\documentclass[a4paper,12pt,authoryear,print,index]{PhDThesisPSnPDF}

\input{Preamble/preamble}

\title{Towards Improved Variational Inference for Deep Bayesian Models}

\author{Sebastian William Ober}

\dept{Department of Engineering}

\university{University of Cambridge}
\crest{\includegraphics[width=0.2\textwidth]{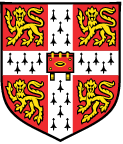}}

\degreetitle{Doctor of Philosophy}

\college{Christ's College}

\degreedate{December 2022} 

\subject{LaTeX} \keywords{{LaTeX} {PhD Thesis} {Engineering} {University of
Cambridge}}

\ifdefineAbstract
\fi

\ifdefineChapter
\fi

\begin{document}

\frontmatter

\maketitle


\begin{declaration}

This thesis is the result of my own work and includes nothing which is the outcome of work done in collaboration except as declared in the Preface and specified in the text.
I further state that no substantial part of my thesis has already been submitted, or, is being concurrently submitted for any such degree, diploma or other qualification at the University of Cambridge or any other University or similar institution except as declared in the Preface and specified in the text. 
This dissertation contains fewer than
65,000 words including appendices, bibliography, footnotes, tables and equations and has fewer than 150 figures.


\end{declaration}

\begin{abstract}
Deep learning has revolutionized the last decade, being at the forefront of extraordinary advances in a wide range of tasks including computer vision, natural language processing, and reinforcement learning, to name but a few.
However, it is well-known that deep models trained via maximum likelihood estimation tend to be overconfident and give poorly-calibrated predictions.
Bayesian deep learning attempts to address this by placing priors on the model parameters, which are then combined with a likelihood to perform posterior inference.
Unfortunately, for deep models, the true posterior is intractable, forcing the user to resort to approximations.

In this thesis, we explore the use of variational inference as an approximation, as it is unique in simultaneously approximating the posterior and providing a lower bound to the marginal likelihood.
If tight enough, this lower bound can be used to optimize hyperparameters and to facilitate model selection.
However, this capacity has rarely been used to its full extent for Bayesian neural networks, likely because the approximate posteriors typically used in practice can lack the flexibility to effectively bound the marginal likelihood.
We therefore explore three aspects of Bayesian learning for deep models.
First, we begin our investigation by asking whether it is necessary to perform inference over as many parameters as possible, or whether it is reasonable to treat many of them as hyperparameters that we optimize with respect to the marginal likelihood.
This would introduce significant computational savings; however, we observe that this can lead to pathological behavior and severe overfitting, suggesting that it is better to be as ``fully Bayesian'' as possible.
We continue our thesis by proposing a variational posterior that provides a unified view of inference in Bayesian neural networks and deep Gaussian processes, which we show is flexible enough to take advantage of added prior hyperparameters.
Finally, we demonstrate how variational inference can be improved in certain deep Gaussian process models by analytically removing symmetries from the posterior, and performing inference on Gram matrices instead of features.
While we do not directly investigate the use of our improvements for model selection, we hope that our contributions will provide a stepping stone to fully realize the promises of variational inference in the future.

\end{abstract}


\begin{acknowledgements}

I would like to start by thanking my supervisor, Carl Edward Rasmussen, for his support and advice over the past years. 
Carl's careful approach to research combined with his deep expertise in probabilistic modeling taught me more than I could have ever imagined.
Perhaps more importantly, his patient encouragement as I navigated the challenges of doing a PhD kept me focused on the important things.
Beyond Carl, I was fortunate to be surrounded by incredible researchers in the Machine Learning Group here at Cambridge.
I would particularly like to single out David R Burt, Andrew YK Foong, Vidhi Lalchand, Ross M Clarke, and Adri{\`a} Garriga-Alonso for being a constant source of learning and friendship in my time there.
I would also like to thank my advisor, Sumeet Singh, and those in his group, for welcoming me into their reading group in my first couple of years: this provided me with a valuable complementary perspective on statistical machine learning.
I am additionally grateful for the invaluable feedback, discussion, and time provided by my examiners, Marc Deisenroth and Carl Henrik Ek, whose insights greatly improved this thesis.
Finally, I would like to thank those who provided feedback on this thesis during its preparation: Samuel Power, David R Burt, and Samuel Duffield.
Of course, any remaining errors or inaccuracies are my own.

This thesis would not have existed in its current form without the many amazing collaborators I have been fortunate enough to work with during my PhD: Laurence Aitchison, Ben Anson, Artem Artemev, David R Burt, Vincent Fortuin, Adri{\`a} Garriga-Alonso, Martin J{\o}rgensen, Edward Milsom, Henry B Moss, Victor Picheny, Pola Schw{\"o}bel, Mark van der Wilk, and Adam X Yang.
Two of these, Mark van der Wilk and Laurence Aitchison, were remarkably generous in their time and support, patiently meeting with me weekly for much of my PhD as I was slowly improving my knowledge and skills.
I learned a huge amount from both of them, both on the technical side and on how to approach research, and I hope we will continue to have many productive collaborations.
In addition to these collaborations, I was fortunate to be able to spend time at Secondmind towards the end of my PhD.
This provided me an invaluable insight into the use of machine learning in industry and also surrounded me with incredible researchers in probabilistic modeling and Bayesian optimization.
In particular, in addition to those mentioned above, I would like to thank Sofia Ceppi, Nicolas Durrande, Vincent Dutordoir, Stratis Markou, Hrvoje Stoji{\'c}, Alan Saul, Fergus Simpson, and Louis Tiao for being a wealth of both knowledge and fun during my time there.

This PhD was funded by the Gates Cambridge Trust. 
I am grateful to them not only for the financial support, but for also providing a community of Scholars dedicated to making the world a better place, and enabling this community to flourish.

On a more personal note, I am deeply grateful for the friends who have supported me and with whom I have shared many amazing memories.
There have been countless such friends throughout my PhD, and as there are far too many to list them all, in addition to those listed above I will have to limit myself to singling out Imran Ahmed, Joey Belleza, Clara-Ann Cheng, Luke Cockerton, Catherine Darlison, the ``FH Crew,'' Gui Freitas, Gabriel Gallardo, Mayeule Huard, Genny Kilburn-Smith, David Losson, Sarah Maple, Keir Martland, Conner McCain, Katy McCulloch, Daniel Miller, Catherine Newman, Andrew Paverd, Christian Raroque, Carlos Rodr{\'i}guez Otero, Andrei Smid, Konrad Suchodolski, Sean Tan, Prakash Thanikachalam, Damian Walsh, Anna Whitehead, and Hania Wyciszczok.
Many of these friends were made at Fisher House, a wonderful and spiritually enriching community that has been tirelessly supported by its amazing Chaplains and pastoral assistants over the years: Sr Ann Swailes OP, Fr Philip Moller SJ, Fr Chase Pepper CSC, Fr Robert Verrill OP, Paul Norris, Matteo Baccaglini, Fr Matthew Gummess O Carm, Fr Paul Keane, and Msgr Mark Langham.
The last of these particularly holds a special place in my heart; may he rest in peace.
I would also like to thank the community of musicians at Cambridge, who welcomed me as one of their own -- I could never have imagined before coming that I would have the opportunity to play music at such a high level.

Finally, and most importantly, I could not have done this PhD without the loving and constant support of my family, and in particular my parents, Raimund and Sally, and my brother, Alexander.



\end{acknowledgements}

\begin{relationship}

Some of the chapters of this thesis are based on work previously published at the following venues:
\begin{itemize}
    \item Chapter 3 expands on ``The promises and pitfalls of deep kernel learning,'' coauthored by \textbf{Sebastian W. Ober}, Carl Edward Rasmussen, and Mark van der Wilk, and published at the Thirty-seventh Conference on Uncertainty in Artificial Intelligence (UAI), 2021.
    \item Chapter 4 expands on ``Global inducing point variational posteriors for Bayesian neural networks and deep Gaussian processes,'', coauthored by \textbf{Sebastian W. Ober} and Laurence Aitchison, and published at the Thirty-seventh International Conference on Machine Learning (ICML), 2021.
    \item Chapter 5 expands on ``A variational approximate posterior for the deep Wishart process,'' coauthored by \textbf{Sebastian W. Ober} and Laurence Aitchison, and published at the Thirty-fifth Conference on Neural Information Processing Systems (NeurIPS), 2021.
\end{itemize}
In addition to these previously published works, Sec.~\ref{sec:dwp:improving} has been recently extended and published at the Thirty-ninth Conference on Uncertainty in Artificial Intelligence (UAI), 2023, as ``An improved variational approximate posterior for the deep Wishart process,'' coauthored by \textbf{Sebastian W. Ober}, Ben Anson, Edward Milsom, and Laurence Aitchison.

\end{relationship}


\tableofcontents

\listoffigures

\listoftables


\printnomenclature

\mainmatter


\chapter{Introduction}  
\label{introduction}

\ifpdf
    \graphicspath{{Introduction/Figs/Raster/}{Introduction/Figs/PDF/}{Introduction/Figs/}}
\else
    \graphicspath{{Introduction/Figs/Vector/}{Introduction/Figs/}}
\fi

Deep learning has shown remarkable success in a wide array of tasks, ranging from image classification to natural language processing.
However, obtaining reliable uncertainty estimates for these models' predictions is difficult.
Bayesian inference, one of the cornerstones of probabilistic machine learning that promises to address this, is intractable in these models, forcing the user to resort to approximations.
In this thesis, we focus on improving variational inference for deep models.

Traditionally, the approximate posteriors used in variational inference for deep models have often sacrificed flexibility in favor of a low computational cost.
More complex approximate posteriors do exist, but they are typically ``brute force'' in nature: they do not take the natural structure of the models into account.
In this thesis, we explore how we can achieve tractable yet flexible approximate inference in deep models by taking advantage of these structures.
In doing so, we try to answer the following questions:
\begin{itemize}
    \item Is it necessary to perform inference over all the model parameters? (Chapter~\ref{sec:dkl})
    \item Can we design approximate posteriors that are both tractable and flexible by taking the natural structure of the models into account? (Chapter~\ref{sec:gi})
    \item Is it possible to transform a deep model to remove symmetries in the true posterior that make variational inference difficult? (Chapter~\ref{sec:dwp})
\end{itemize}

We begin our thesis by first broadly discussing what we would like a good model to do, before introducing the concepts behind probabilistic modeling, which we use to motivate our focus on variational inference.

\section{What do we want from a model?}
In this thesis, we focus on supervised learning tasks where we have a dataset of $N$ input-output pairs $\Data \coloneqq \{(\x_n), (\y_n)\}_{n=1}^N$.
This data could either represent continuous outputs in a regression context (e.g., a stock price or the binding affinity of a new drug candidate), or categorical outputs in the case of classification (e.g., whether an image is of a cat or a dog).
The task of supervised learning is to create a model from this data that will be useful in predicting for unseen data.
At its simplest level, this will involve learning a set of model parameters, $\w$, that we hope will best describe the data.

In the course of acquiring data and modeling, we will face different types of uncertainty.
For instance, it is likely that our acquired data reflects some source of noise --- for instance, measurement error in regression tasks, or labeling error in classification.
Moreover, it is possible that we will not have enough data for the model to understand what prediction to make at every unseen test input.
Ideally, we would like our modeling process to reflect these sources of uncertainty, so that the user can know when to trust a model's predictions, and what action to take based off those predictions.

Following the above examples, we can describe uncertainty as falling under one of two types \citep{kendall2017uncertainties}.
The first of these, \emph{aleatoric uncertainty}, refers to uncertainty that cannot be reduced by collecting more data.
For instance, measurement noise will be present regardless of how much data we collect.
On the other hand, \emph{epistemic uncertainty} is the uncertainty that a model should have about its own parameters, and thereby its predictions.
As opposed to aleatoric uncertainty, we can reduce epistemic uncertainty by adding more data --- in particular where the model is already unsure about its predictions.
Ideally, we would like a method to be able to reason about both types of uncertainty, and to be able to distinguish between the two.
For instance, a good sense of epistemic uncertainty is essential for a model's use in downstream decision making such as active learning, as well as for safety-critical tasks, such as autonomous driving: if the visual recognition model is unsure about what it is seeing, the vehicle should act differently than if the model is sure.

So far, we have implicitly assumed that we know which model to use.
However, this is rarely the case in practice.
Instead, practitioners often have to propose a model that they hope will have the capacity to model the data well, without providing erroneous predictions further away from the data.
Moreover, models will often have a number of hyperparameters to tune.
Ideally, in addition to being able to quantify different types of uncertainty, we would like to have a framework that can choose between models and hyperparameters.

Motivated by these two desirable attributes for modeling, we now turn to illustrating how they can be achieved through concrete examples.

\section{Probabilistic modeling}
\label{sec:intro:probmodel}
\begin{figure}[t]
    \centering
    \begin{subfigure}[b]{0.3\textwidth}
        \includegraphics[width=\textwidth]{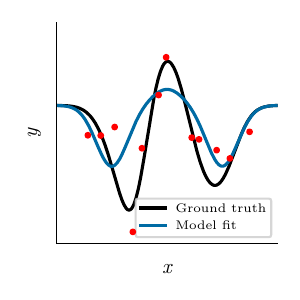}
        \caption{$W = 3$}
        \label{fig:intro:ml_3}
    \end{subfigure}
    \begin{subfigure}[b]{0.3\textwidth}
        \includegraphics[width=\textwidth]{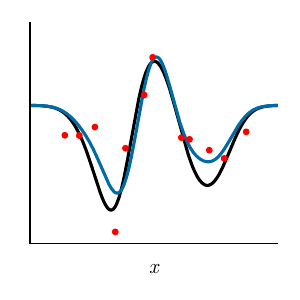}
        \caption{$W = 5$}
        \label{fig:intro:ml_5}
    \end{subfigure}
    \begin{subfigure}[b]{0.3\textwidth}
        \includegraphics[width=\textwidth]{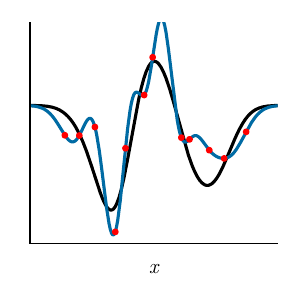}
        \caption{$W = 12$}
        \label{fig:intro:ml_12}
    \end{subfigure}
    \caption{Maximum likelihood fits for three models with different numbers $W$ of ``Gaussian bump'' features. The simplest model (a) cannot effectively model the data, whereas the most complicated model (c) overfits to noise. The best model is therefore a model with intermediate complexity (b).}
    \label{fig:intro:max_likelihood}
\end{figure}

The approach to these problems in this thesis will be to use probabilistic modeling, and in particular a Bayesian perspective.
In order to motivate this approach, we consider a simple toy model, featurized linear regression.
This exposition will give us the opportunity to consider the problems of uncertainty quantification and model selection, eventually arriving at inference with Gaussian processes, while also introducing the notation that we will use throughout.
Consider the model
\begin{align*}
    y_n = \transpose{\w}\feat{\x_n} + \epsilon_n, \quad \epsilon_n \sim \N{0, \sigma^2},
\end{align*}
where $\x_n \in \reals^D$ is an input, $y_n \in \reals$ is its corresponding output, $\feat{\cdot}: \reals^D \rightarrow \reals^W$ is a mapping from the input space to a feature space, and $\w \in \reals^W$ is a vector of weights.\footnote{Note that we have not explicitly included a bias term in the model; instead, as is common we will assume throughout the thesis that any biases have been subsumed into $\w$ using an appropriate modification to $\feat{\cdot}$.}
This model therefore assumes that the observations $y_n$ are noisy realizations of the linear function $f(\cdot) = \transpose{w}\feat{\cdot}$, where the noise has variance $\sigma^2$.
We assume we are given a dataset of $N$ input-output pairs $\Data \coloneqq \{(\x_n), (y_n)\}_{n=1}^N$, which we equivalently write as $\b{\X, \y}$ where $\X \in \reals^{N\times D}$ and $\y \in \reals^{N}$ are formed by stacking the inputs and outputs appropriately.

The simplest approach to learn the weights $\w$ is to perform maximum likelihood estimation, which chooses $\w^*$ as 
\begin{align*}
    \w^* &= \argmax_{\w \in \reals^W}\; \pc{\y\,}{\,\X, \w} \\
    &= \argmax_{\w \in \reals^W}\; \Nc{\y\,}{\,\FeatMat\w,\, \sigma^2\I_N} \\
    &= \argmin_{\w \in \reals^W}\; \norm{\y - \FeatMat\w}_2^2 \\
    &= \b{\transpose{\FeatMat}\FeatMat}^{-1}\transpose{\FeatMat}\y,
\end{align*}
where we have defined $\FeatMat = \Feat{\X} \coloneqq \transpose{[\feat{\x_1},\, \dots,\, \feat{\x_N}]} \in \reals^{N \times W}$ to be the \emph{design matrix}, and where $\I_N \in \reals^{N \times N}$ is the identity matrix.

One natural question that arises when considering this model is how many features to use.
We consider this question in Fig.~\ref{fig:intro:max_likelihood}, where we plot fits using the maximum likelihood estimator for models with a varying number of ``Gaussian bump'' features:
\begin{align*}
    \feat{\x}_i = \exp\b{-(\x - \mathbf{c}_i)^2/l^2},
\end{align*} 
where we spread the centers $\mathbf{c}_i$ evenly across the domain.
From these plots, we make a few observations.
First, the maximum likelihood estimator does not protect against overfitting: the model in Fig.~\ref{fig:intro:ml_12} fits to the noise in the data and provides extreme predictions outside of the data.
Second, the maximum likelihood estimator does not adequately address uncertainty in its predictions.
Whereas the maximum likelihood approach allows us to estimate the noise variance (i.e., aleatoric uncertainty), it does not provide a sense of uncertainty in its parameters and predictions (i.e., epistemic uncertainty).

Our final observation from Fig.~\ref{fig:intro:max_likelihood} is that it is not possible to choose between the models by only looking at the training data: indeed, the model with the best fit to the training data in terms of prediction error (Fig.~\ref{fig:intro:ml_12}) suffers the most from overfitting.
On the other hand, choosing a model with too few features results in a poor fit as well (Fig.~\ref{fig:intro:ml_3}).
Therefore, the best model for this data is a model with an intermediate complexity.
However, the maximum likelihood estimator does not give us a straightforward way of selecting this model, at least not without looking at the performance on held-out validation or test data.
The Bayesian paradigm attempts to address the above deficiencies by simultaneously seeking a good fit for the data, while also protecting against overfitting through a coherent representation of uncertainty in a way that allows for model selection.\footnote{While other paradigms, e.g.,  frequentist methods such as conformal prediction \citep{angelopoulos2022gentle}, can represent epistemic uncertainty, they typically rely on validation or calibration sets, and may not directly allow for model selection.}

\subsection{Bayesian modeling}

In Bayesian modeling, instead of obtaining a point estimate of our parameters $\w$, we wish to infer the \emph{posterior} density of the parameters, $\pc{\w}{\Data}$, which directly accounts for epistemic uncertainty in the model.
We achieve this by proposing a \emph{prior} distribution $\p{\w}$, which encodes our beliefs about what the parameters should be before seeing any data.
Using the likelihood $\pc{\Data}{\w}$, which defines the generative model, we can obtain the posterior by using Bayes' rule:
\begin{align*}
    \pc{\w}{\Data} = \frac{\pc{\Data}{\w}\p{\w}}{\p{\Data}}.
\end{align*}
Here, $\p{\Data}$ is known as the \emph{marginal likelihood} or \emph{model evidence}, and is given by integrating the likelihood with respect to the prior,
\begin{align*}
    \p{\Data} = \int \pc{\Data}{\w} \p{\w} d\w.
\end{align*}
As we shall see, this term is crucial for the Bayesian paradigm: it is greatest when the prior lines up well with the data, a fact that will allow us to choose between models.
We can additionally use the Bayesian posterior to make predictions at a test point $\x_*$, by integrating over the posterior:
\begin{align*}
    \pc{y_*}{\x_*, \Data} = \int \pc{y_*}{\x_*, \w} \pc{\w}{\Data} d\w.
\end{align*}

Returning to the example of featurized linear regression, we propose a prior $\N{\0,\, \alpha^2 \I_W}$ for our parameters. 
It is possible to show \citep[see e.g., Sec.~7.6 of][]{murphy2012machine} that the posterior is given by 
\begin{align*}
    \pc{\w\,}{\,\X, \y} &= \Nc{\w}{\m, \S}\propto \p{\w} \pc{\y\,}{\,\X, \w}, \\
    \m &= \frac{1}{\sigma^2} \S \transpose{\FeatMat}\y, \\
    \S &= \b{\frac{1}{\alpha^2}\I_W + \frac{1}{\sigma^2}\transpose{\FeatMat}\FeatMat}^{-1}, 
\end{align*}
with predictions
\begin{align*}
    \pc{y_*}{\x_*, \X, \y} = \Nc{y_*}{\transpose{\m}\feat{\x_*}, \transpose{\feat{\x_*}}\S\feat{\x_*} + \sigma^2}.
\end{align*}
Finally, we can also compute the marginal likelihood in closed form, giving
\begin{align*}
    \pc{\y}{\X} = \Nc{\y}{\0,\, \alpha^2\FeatMat\transpose{\FeatMat} + \sigma^2 \I_N}.
\end{align*}
Let us briefly consider the logarithm of the marginal likelihood (i.e., the LML):
\begin{align}
    \label{eq:intro:lml-blr}
    \log \pc{\y}{\X} = -\frac{N}{2}\log 2\pi - \underbrace{\frac{1}{2}\transpose{\y}\b{\alpha^2\FeatMat\transpose{\FeatMat} + \sigma^2 \I_N}^{-1}\y}_\text{(a)} - \underbrace{\frac{1}{2}\log \abs{\alpha^2\FeatMat\transpose{\FeatMat} + \sigma^2 \I_N}}_\text{(b)}.
\end{align}
We see that the LML is comprised of three terms, the first being constant with respect to the data.
The second, which we have denoted (a), encourages the prior to conform to the data.
The last term, (b), punishes priors over functions that are too complicated for putting too much mass on functions that do not explain the data well.
This decomposition has led term (a) to be referred to as the ``data fit'' term, whereas term (b) has often been referred to as the ``complexity penalty'' \citep{rasmussen2006gaussian}.
If the LMLs of different models are available, we can therefore compare the LMLs to determine which model is best, as it should trade off data fit and complexity \citep{mackay1995probable,rasmussen2000occam}.
Moreover, to find a suitable model we can optimize any \emph{hyperparameters}, i.e., parameters of the prior and likelihood, with respect to the LML, in a scheme referred to as \emph{type-II maximum likelihood}, or simply \emph{maximum marginal likelihood}.\footnote{We note that type-II maximum likelihood is a form of model selection, as two models with different hyperparameter values are technically different models according to the Bayesian framework.}

\begin{figure}[t]
    \centering
    \begin{subfigure}[b]{0.3\textwidth}
        \includegraphics[width=\textwidth]{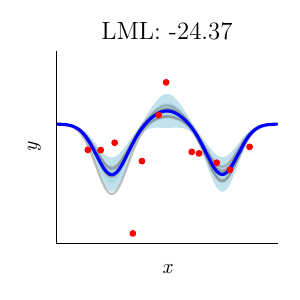}
        \caption{$W = 3$}
    \end{subfigure}
    \begin{subfigure}[b]{0.3\textwidth}
        \includegraphics[width=\textwidth]{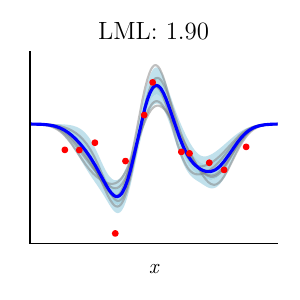}
        \caption{$W = 5$}
    \end{subfigure}
    \begin{subfigure}[b]{0.3\textwidth}
        \includegraphics[width=\textwidth]{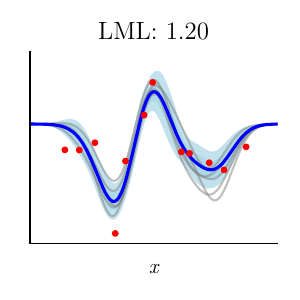}
        \caption{$W = 12$}
    \end{subfigure}
    \caption{Plots of the posterior predictives and posterior samples (gray) for the three models, along with their log marginal likelihoods (LMLs). For the posterior predictives, we plot the mean functions (blue line), with the shaded regions corresponding to one and two standard deviations. The intermediate model (b) has the best LML, as the model with the fewest features (a) cannot effectively model the data, and the model with the most features (c) is penalized for having too much complexity.}
    \label{fig:intro:exact_blr}
\end{figure}

In Fig.~\ref{fig:intro:exact_blr}, we plot the posterior predictives (i.e., the distributions of $f(\cdot) = \transpose{\w}\feat{\cdot}$) for probabilistic models corresponding to the deterministic models from Fig.~\ref{fig:intro:max_likelihood}, displaying their log marginal likelihoods as well.
In these figures, we have optimized the prior variance $\alpha^2$ with respect to the LML, with the noise variance $\sigma^2$ fixed to its true value.
We observe that the plots now show a sense of epistemic uncertainty, and that the best LML is given by the simplest model that explains the data well: the LML balances model fit and complexity, leading to a \emph{Bayesian Occam's razor} \citep{mackay1995probable, rasmussen2000occam}.

\subsection{Gaussian processes}
\label{sec:intro:gps}
In the previous section, we have considered Bayesian linear regression, which performs inference over a finite set of weights.
However, one may argue that for many problems it is more natural to perform inference over \emph{functions}.
For instance, we may have intuition that a function should be either smooth or rough, which could be difficult to encode in terms of features.
Perhaps more importantly, performing inference with a finite number of features limits the capacity of the model, which can be seen clearly in Fig.~\ref{fig:intro:exact_blr}.
While the epistemic uncertainty in these figures is reasonable in the region of the data, the models do not have the capacity to have higher uncertainty outside of the data.
However, if we used an infinite number of features, this would be possible.
In this section, we briefly describe how we can do so using \emph{Gaussian processes (GPs)}, which will be one of the fundamental building blocks for the work in this thesis.
For a more in-depth introduction to Gaussian processes, we refer the reader to \citet{rasmussen2006gaussian}.

A Gaussian process is defined as a potentially uncountable collection of random variables indexed by inputs $\x_1,\, \x_2,\, \ldots \, \in \mathcal{X}$, any finite number of which are distributed according to a Gaussian distribution. 
A GP is uniquely defined by a mean function $m(\cdot): \mathcal{X} \rightarrow \reals$ and a positive semi-definite covariance function, or kernel, $k(\cdot, \cdot): \mathcal{X} \times \mathcal{X} \rightarrow \reals$.
We will denote such a GP by $\mathcal{GP}(m, k)$.
Returning to the linear model of the previous section, we see that the prior over weights, $\w \sim \N{\0, \alpha^2 \I_W}$, implies a Gaussian process prior over function values with mean and covariance
\begin{align*}
    m(\x) &= \E[\w]{\transpose{\w}\feat{\x}} = 0, \\
    k(\x, \x') &= \E[\w]{(\transpose{\w}\feat{\x} - m(\x))(\transpose{\w}\feat{\x'} - m(\x'))} = \alpha^2\transpose{\feat{\x}}\feat{\x'},
\end{align*}
so that the kernel function is an inner product of the feature maps of $\x$ and $\x'$.
Therefore, Bayesian linear regression (with a Gaussian prior and likelihood) can be viewed as a particular type of Gaussian process model.

However, whereas Bayesian linear regression is limited to finite feature maps $\feat{\cdot}$, Gaussian processes allow us to consider \emph{infinite} feature maps defined by the kernel function $k$.
For instance, the popular automatic relevance determination (ARD) squared-exponential (SE) kernel, $k_\text{SE}(\cdot, \cdot): \reals^D \times \reals^D \rightarrow \reals$,
\begin{align}
\label{eq:intro:se-kernel}
    k_\text{SE}(\x, \x') = \sigma_f^2 \exp \b{-\frac{1}{2}\sum_{d=1}^D \frac{(x_d - x_d')^2}{l_d^2}},
\end{align}
with signal variance $\sigma_f^2$ and lengthscales $\{l_d\}_{d=1}^D$, can be obtained by considering the limit of an infinite number of Gaussian basis functions \citep[\originalS 4.2.1]{rasmussen2006gaussian}.
Making use of the algebraic properties of Gaussians, we can directly perform inference using an infinite number of features at a finite computational cost.
Consider the model
\begin{align*}
    y_n = f(\x_n) + \epsilon_n, \quad \epsilon_n \sim \N{0, \sigma^2},
\end{align*}
with a GP prior $f \sim \mathcal{GP}(0, k)$.\footnote{Throughout the rest of the thesis, unless otherwise noted we will assume a zero mean function, as this typically will not lose generality.}
Given a dataset $\Data = (\X, \y)$ as above, we can define the Gram matrix $\K_{\X, \X} \in \reals^{N \times N}$, where we have $(\K_{\X, \X})_{ij} = k(\x_i, \x_j)$ for $1 \leq i, j \leq N$.
To obtain predictions of the function values $\f_* \in \reals^{N_*}$ at a set of points $\X_* \in \reals^{N_* \times D}$, conditioning using properties of Gaussians gives
\begin{align*}
    \pc{\f_*}{\X_*, \X, \y} &= \Nc{\f_*}{\hat{\m}_*, \hat{\S}_*}, \\
    \hat{\m}_* &= \K_{\X_*, \X}\b{\K_{\X, \X} + \sigma^2 \I_N}^{-1}\y, \\
    \hat{\S}_* &= \K_{\X_*, \X_*} - \K_{\X_*, \X}\b{\K_{\X, \X} + \sigma^2 \I_N}^{-1}\K_{\X, \X_*},
\end{align*}
where we have defined $\K_{\X_*, \X_*}$, $\K_{\X_*, \X}$, and $\K_{\X, \X_*}$ analogously to $\K_{\X, \X}$.

\begin{figure}[t]
    \centering
    \begin{subfigure}[b]{0.49\textwidth}
        \includegraphics[width=\textwidth]{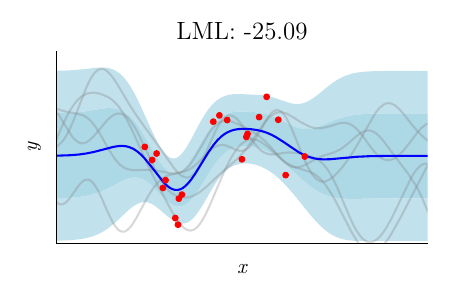}
        \caption{Hyperparameters left at initialization}
    \end{subfigure}
    \begin{subfigure}[b]{0.49\textwidth}
        \includegraphics[width=\textwidth]{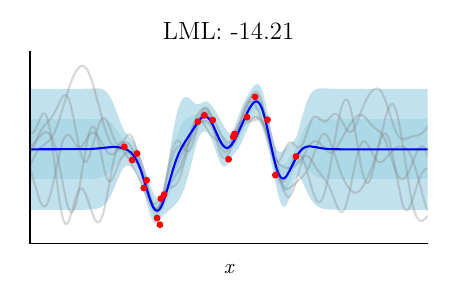}
        \caption{Trained hyperparameters}
    \end{subfigure}
    \caption{Plots of the posterior predictives and samples for GP models with squared exponential kernels, trained on data subsampled from the toy example given in \citet{snelson2006sparse}, along with their log marginal likelihoods. The model in (a) has been left at its initial hyperparameter values, whereas the model in (b) is the result of learning the hyperparameters according to the log marginal likelihood.}
    \label{fig:intro:exact_gpr}
\end{figure}

We can also obtain the marginal likelihood by integrating over the GP prior:
\begin{align*}
    \pc{\y}{\X} &= \Nc{\y}{\0, \K_{\X, \X} + \sigma^2 \I_N}, \\
    \log \pc{\y}{\X} &= -\frac{N}{2} \log 2\pi - \frac{1}{2}\transpose{\y}\b{\K_{\X, \X} + \sigma^2\I_N}^{-1}\y - \frac{1}{2}\log \abs{\K_{\X, \X} + \sigma^2\I_N}.
\end{align*}
We see again that the LML can be divided into three terms, with analogous data fit and complexity terms to those in Eq.~\ref{eq:intro:lml-blr}.
Indeed, the LMLs for the two models are identical if we replace $\alpha^2\FeatMat\transpose{\FeatMat}$ from Eq.~\ref{eq:intro:lml-blr} with $\K_{\X, \X}$, emphasizing again that the two approaches to Bayesian linear regression as described are equivalent.
Therefore, we can again attempt to use the LML for model selection and hyperparameter tuning.
For example, a GP with the squared exponential kernel from Eq.~\ref{eq:intro:se-kernel} will have $\theta = \{\sigma_f, \{l_d\}_{d=1}^D, \sigma\}$ as hyperparameters, which are typically optimized with respect to the LML with a gradient-based optimizer such as L-BFGS \citep{nocedal1980updating, liu1989limited}.
We plot an example of the posterior predictive for a GP with trained hyperparameters and hyperparameters left at initialization, along with the LMLs of the two models, in Fig.~\ref{fig:intro:exact_gpr}.
Importantly, we see that both models represent the uncertainty outside of the data regions well, demonstrating the benefit of using an infinite number of features.
We also observe that the posterior predictive of the GP with trained hyperparameters provides a better fit to the data as well as better uncertainty quantification, showing the benefit provided by using the LML for model selection.

So far, we have explored Bayesian linear regression and Gaussian process models to motivate our use of the Bayesian paradigm to avoid overfitting, provide (epistemic)\footnote{As the focus of this thesis is primarily on epistemic uncertainty quantification, from here onwards we typically use ``uncertainty'' to refer to epistemic uncertainty, unless otherwise noted.} uncertainty estimates, and select models.
However, there are a two primary issues that we have not addressed.
First, we can only obtain exact posteriors, posterior predictives, and marginal likelihoods for models with conjugate priors and likelihoods.
For other models, such as classification models or Bayesian neural networks, we have to use approximations, as the marginal likelihood is intractable to compute exactly.
Second, the computational cost of performing exact inference, particularly for GPs, can be prohibitive for large datasets.
For Bayesian linear regression, the computational cost is $\mathcal{O}(NW + W^3)$ (noting that the marginal likelihood can be rewritten using Woodbury's identity), whereas GP regression requires $\mathcal{O}(N^3)$ computational complexity and $\mathcal{O}(N^2)$ memory.
For large $N$, we may want to avoid this cost for both models; in particular, for large datasets, we may wish to train with stochastic minibatching for both methods.
These issues can be sidestepped by \emph{variational inference (VI)}, which we explore next.

\section{A refresher on variational inference}

In variational inference \citep{jordan1999introduction, blei2017variational}, we attempt to approximate the intractable true posterior of a model, $\pc{\cdot}{\Data}$, with an approximate posterior $\q{\cdot}$.
For now, we assume the model to be parametric with parameters $\w$, so that we are searching for a good approximation $\q{\w} \approx \pc{\w}{\Data}$.
Typically, $\q{\w}$ is a member of a variational family $\mathcal{Q}$, which is often parameterized by a set of variational parameters $\phi$, so that we can write $\q{\w} = \qsub[\phi]{\w}$.
The goal of variational inference is to turn inference into an optimization problem, which is achieved by attempting to minimize the reverse Kullback-Leibler (KL) divergence between the true posterior and its approximation, with respect to the variational parameters:
\begin{align}
    \label{eq:intro:kl-min}\qsub[*]{\w} &= \argmin_{\q{\w} \in \mathcal{Q}}\; \KL{\q{\w}}{\pc{\w}{\Data}}, \\
    \label{eq:intro:kl-def}\KL{\q{\w}}{\pc{\w}{\Data}} &= \int \q{\w} \log \frac{\q{\w}}{\pc{\w}{\Data}} d\w,
\end{align}
which then reduces to
\begin{align*}
    \phi_* = \argmin_{\phi} \; \KL{\qsub[\phi]{\w}}{\pc{\w}{\Data}}.
\end{align*}
Since we do not typically have access to the true posterior in scenarios where we wish to use VI, computing the required KL divergence directly is not possible.
However, let us consider the log marginal likelihood:
\begin{align}
    \log \p{\Data} &= \log \int \p{\Data, \w} d\w \label{eq:intro:elbo-lml} \\
    &= \log \int \p{\Data, \w} \frac{\q{\w}}{\q{\w}} d\w \nonumber \\
    &= \log \E[\q{\w}]{\frac{\p{\Data, \w}}{\q{\w}}} \nonumber \\
    &\geq \E[\q{\w}]{\log \frac{\p{\Data, \w}}{\q{\w}}} \nonumber \\
    &= \log \p{\Data} - \KL{\q{\w}}{\pc{\w}{\Data}} \label{eq:intro:elbo-intractable} \\
    &= \E[\q{\w}]{\log \p{\Data}} - \E[\q{\w}]{\log \q{\w} - \log \pc{\w}{\Data}} \nonumber \\
    &= \E[\q{\w}]{\log \pc{\Data}{\w}} - \E[\q{\w}]{\log \q{\w} - \log \p{\w}} \nonumber \\
    &= \underbrace{\E[\q{\w}]{\log \pc{\Data}{\w}}}_{\text{(a)}} - \underbrace{\KL{\q{\w}}{\p{\w}}}_\text{(b)} \eqqcolon \mathcal{L} \label{eq:intro:elbo-tractable},
\end{align}
where the inequality is due to applying Jensen's inequality, and the subsequent lines are given by applications of Bayes' rule.
We have been able to transform expressions containing two intractable terms, the marginal likelihood and true posterior (Eqs.~\ref{eq:intro:elbo-lml} and~\ref{eq:intro:elbo-intractable}), into an expression $\mathcal{L}$ containing terms that (we assume) we have access to: the prior, the likelihood, and the approximate posterior (Eq.~\ref{eq:intro:elbo-tractable}).
The expected log-likelihood (a) can typically be computed by quadrature or Monte Carlo estimation, whereas the KL term (b) can either be computed in closed form (for simple models) or again estimated via Monte Carlo estimation.
Furthermore, we see that the resulting $\mathcal{L}$ is a lower bound to the marginal likelihood, leading to the name \emph{evidence lower bound (ELBO)}.
Importantly, from Eq.~\ref{eq:intro:elbo-intractable} we see that the tightness of the bound is given exactly by the KL divergence between approximate and true posteriors --- the quantity we wish to minimize.
Finally, we note that for likelihoods that factorize across datapoints, i.e.,
\begin{align*}
    \pc{\Data}{\w} = \prod_{n=1}^N \pc{\Data_{n}}{\w},
\end{align*}
we can perform minibatching to obtain an unbiased estimate of the ELBO: constructing a minibatch of the data $\Tilde{\Data} \subset \Data$ with $\abs{\tilde{\Data}} = \tilde{N}$, we obtain
\begin{align*}
    \tilde{\mathcal{L}} = \frac{N}{\tilde{N}}\sum_{n=1}^{\tilde{N}} \E[\q{\w}]{\log \pc{\tilde{\Data}_n}{\w}} - \KL{\q{\w}}{\p{\w}}.
\end{align*}

\begin{figure}[t]
    \centering
    \begin{subfigure}[b]{0.49\textwidth}
        \includegraphics[width=\textwidth]{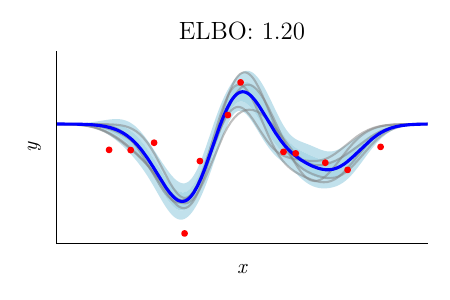}
        \caption{Full-covariance approximate posterior}
    \end{subfigure}
        \begin{subfigure}[b]{0.49\textwidth}
        \includegraphics[width=\textwidth]{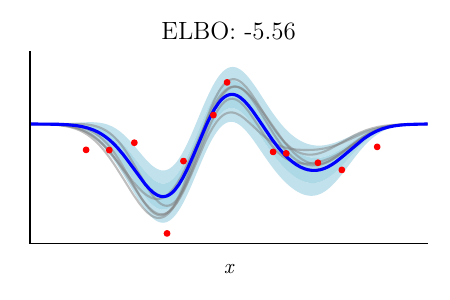}
        \caption{Mean-field approximate posterior}
    \end{subfigure}
    \caption{Demonstration of predictive posteriors for VI for a BLR model ($W=12$) with two approximate posteriors: (a) a full-covariance Gaussian approximate posterior, and (b) a mean-field Gaussian approximate posterior. The full-covariance posterior is able to recover the true posterior as well as the true prior standard deviation $\alpha=0.155$, whereas the mean-field approximate posterior cannot, with a worse ELBO and biased $\alpha=0.092$. }
    \label{fig:intro:blr_vi}
\end{figure}

Because the KL divergence to the true posterior appears directly in the ELBO (Eq.~\ref{eq:intro:elbo-intractable}), our original intractable minimization problem, Eq.~\ref{eq:intro:kl-min}, can now be formulated as the problem of maximizing the ELBO with respect to the variational parameters.
Moreover, as the ELBO is guaranteed to be a lower bound to the log marginal likelihood, we can consider using it as a proxy to the marginal likelihood for both model selection and to optimize any model hyperparameters $\theta$.\footnote{The fact that the ELBO is a lower bound is essential: if it were not, we could risk diverging from the LML when we optimize the hyperparameters.} 
In the latter case, we end up with the joint optimization problem
\begin{align*}
    \phi_*, \theta_* = \argmax_{\phi, \theta} \; \mathbb{E}_{\qsub[\phi]{\w}} \left[\log \pc{\Data}{\w, \theta}\right] - \KL{\qsub[\phi]{\w}}{\pc{\w}{\theta}}.
\end{align*}
While we can use this objective to optimize model hyperparameters, the bound is in general not uniformly tight for different hyperparameter values.
This will result in bias being introduced when estimating hyperparameters, meaning that the optimal hyperparameters given by the ELBO will likely not coincide with the optimal hyperparameters according to the marginal likelihood. 
Moreover, the magnitude of the bias, and hence the accuracy of our hyperparameter optimization, depends on the flexibility of the approximating family --- as can be seen from Eq.~\ref{eq:intro:elbo-intractable}, if our approximating family contains the true posterior, we will recover the LML and thus be able to optimize hyperparameters without bias \citep{turner2011two}.
We make this discussion concrete by illustrating this briefly for our Bayesian linear regression example in Fig.~\ref{fig:intro:blr_vi}.
In this figure, we consider variational inference with two approximating families: 
\begin{enumerate}
    \item the family of full-covariance normal distributions, $\qsub[\phi]{\w} = \Nc{\w}{\m, \S}$, so that $\phi = \{\m, \S\}$ with $\S \in S_{++}$, and
    \item the family of fully-factorized, or mean-field normal distributions, $\qsub[\phi]{\w} = \Nc{\w}{\m, \mathrm{Diag}(\ms^2)}$ so that $\phi = \{\m, \ms^2\}$, and where by $\mathrm{Diag}(\cdot)$ we denote a function that returns a diagonal matrix with elements given by the argument.
\end{enumerate}
We observe that the first family, which contains the true posterior, is able to provide sensible error bars and hyperparameters, whereas for the second family, both the uncertainty quantification and hyperparameter estimates suffer, albeit only slightly.
Nevertheless, this example serves to show that the flexibility of the variational family determines both the quality of the uncertainty estimates and hyperparameters.
For variational inference to be successful, we therefore require an approximating family that is flexible enough to capture the necessary structure in the posterior while retaining computational tractability.
When this is the case, the ELBO will be close to the LML, allowing for reliable hyperparameter selection, which in turn will allow the ELBO to be used as a proxy for the LML for the purposes of model selection.

\subsection{Variational inference for Gaussian processes}
We have seen how variational inference provides a means of simultaneously addressing the intractability of exact Bayesian inference while allowing for model selection, achieving both through the ELBO.
However, our exposition has focused on parametric models.
Here, we briefly review the works of \citet{hensman2013gaussian, hensman2015scalable, titsias2009variational}, which show how to perform variational inference in Gaussian processes. 

For Gaussian processes, recall that our primary aim is to reduce the $\mathcal{O}(N^3)$ computational and $\mathcal{O}(N^2)$ memory complexities of exact GP inference, which are prohibitive for large datasets.
Additionally, exact GP inference is only possible for regression with conjugate likelihoods --- we would like to have a method that works for non-conjugate likelihoods, allowing us to perform classification through e.g., Bernoulli or softmax likelihoods.
Both can be achieved by introducing a set of $M < N$ \emph{inducing variables}, $\u \in \reals^M$, which are values of the latent function $f$ indexed at a set of inducing inputs, $\Z = \{\z\}_{i=1}^M, \z_i \in \mathcal{X}$.\footnote{For clarity of exposition, we assume that the inducing points are distinct from the datapoints.}
VI in GPs relies on using these pseudo-datapoints to construct a sparse approximation to the kernel matrix $\K_{\X, \X}$ to allow faster computation: hence this form of VI for GPs is often referred to as \emph{sparse variational inference}.
Intuitively, the inducing points are trained so that they ``compress'' the $N$ datapoints as effectively as possible into $M$ points.

Instead of minimizing the KL divergence for parameters, we now wish to minimize the KL divergence between two stochastic processes: an approximate process over functions, $\q{f}$, and the true posterior stochastic process, $\pc{f}{\y}$.\footnote{For notational clarity, we from this point drop the conditioning on $\X$, as it is implied.}
This approach was formalized in \citet{matthews2016sparse}; for simplicity, we provide an informal and non-rigorous argument based on their work.
We therefore refer the interested reader to their work for a rigorous treatment of the argument, which addresses technical details such as the meaning of a KL divergence between stochastic processes.
Using the notation $\f = f(\X)$, we begin by writing $f = \{\f, \u, f_{\neq \f, \u}\}$ to divide up the latent function $f$ into its values at the data, inducing inputs, and everywhere else, respectively. 
We then continue by decomposing the true and approximate posteriors following this division:
\begin{align*}
    \pc{f}{\y} &= \pc{\f_{\neq \f, \u}, \f, \u}{\y} \\
    &= \pc{f_{\neq \f, \u}}{\f, \u, \y}\pc{\f}{\u, \y}\pc{\u}{\y} \\
    &= \pc{f_{\neq \f, \u}}{\f, \u}\pc{\f}{\u, \y}\pc{\u}{\y}, \text{ and}\\
    \q{f} &= \q{\f_{\neq \f, \u}, \f, \u} \\
    &= \qc{f_{\neq \f, \u}}{\f, \u}\qc{\f}{\u}\q{\u},
\end{align*}
where we condition implicitly on the datapoints $\X$ and inducing points $\Z$ as necessary.

We now turn to choosing $\q{f}$ to minimize the KL divergence between the approximate and true posterior stochastic processes, $\KL{\q{f}}{\pc{f}{\y}}$.
Using the chain rule of the KL divergence,
\begin{align*}
    \KL{\q{f}}{\pc{f}{\y}} &= \KL{\qc{f_{\neq \f, \u}}{\f, \u}}{\pc{f_{\neq \f, \u}}{\f, \u}} + \KL{\q{\f, \u}}{\pc{\f, \u}{\y}},
\end{align*}
we see that we should choose $\qc{f_{\neq \f, \u}}{\f, \u} = \pc{f_{\neq \f, \u}}{\f, \u}$ to minimize the first term.
We are then left with the choice of $\q{\f, \u} = \qc{\f}{\u}\q{\u}$ to minimize the second term.
For the former of these terms, we choose $\qc{\f}{\u} = \pc{\f}{\u}$, which is a simplifying assumption that assumes that $\u$ will be a sufficient statistic for $\f$.
In reality, this will only be true for $\Z = \X$, so that $M = N$ \citep{titsias2009variational}.
However, this assumption allows us to reduce the computational complexity from $\mathcal{O}(N^3)$ to $\mathcal{O}(NM^2)$.
For the second term, we choose $\q{\u} = \Nc{\u}{\m, \S}$.
We  are now ready to derive our ELBO:
\begin{align*}
    \mathcal{L} &= \log \p{\y} - \KL{\q{f}}{\pc{f}{\y}} \\
    &= \E[\q{f}]{\log \frac{\p{\y}\cancel{\pc{f_{\neq \f, \u}}{\f, \u}}\pc{\f}{\u, \y}\pc{\u}{\y}}{\cancel{\pc{f_{\neq \f, \u}}{\f, \u}}\pc{\f}{\u}\q{\u}}} \\
    &= \E[\q{\f}]{\log \frac{\p{\y, \f, \u}}{\pc{\f}{\u}\q{\u}}} \\
    &= \E[\q{\f}]{\log \frac{\pc{\y}{\f}\cancel{\pc{\f}{\u}}\p{\u}}{\cancel{\pc{\f}{\u}}\q{\u}}} \\
    &= \E[\q{\f}]{\log \pc{\y}{\f}} - \KL{\q{\u}}{\p{\u}}.
\end{align*}
This ELBO can be computed in $\mathcal{O}(NM^2)$, and can again be minibatched for factorizing likelihoods, resulting in $\mathcal{O}(\tilde{N}M^2 + M^3)$ complexity per evaluation.
As with the linear regression model, we use the ELBO to optimize the variational parameters, which in this case are $\phi = \{\Z, \m, \S\}$, and any model hyperparameters $\theta$.
We can provide predictions of the latent function value at test points using
\begin{align*}
    \q{\f_*} &= \int \pc{\f_*}{\u} \q{\u} d\u = \Nc{\f_*}{\hat{\m}_*, \hat{\S}_*}, \\
    \hat{\m}_* &= \K_{\X_*, \Z}\K_{\Z, \Z}^{-1}\m, \\
    \hat{\S}_* &= \K_{\X_*, \X_*} - \K_{\X_*, \Z}\K_{\Z, \Z}^{-1}\b{\K_{\Z, \Z} - \S}\K_{\Z, \Z}^{-1}\K_{\Z, \X_*}.
\end{align*}
Finally, for regression with a Gaussian likelihood, \citet{titsias2009variational} showed that we can obtain the optimal $\q{\u}$ in closed form, so that
\begin{align*}
    \m &= \frac{1}{\sigma^2}\K_{\Z, \Z}\b{\K_{\Z, \Z} + \frac{1}{\sigma^2}\K_{\Z, \X}\K_{\X, \Z}}^{-1}\K_{\Z, \X}\,\y, \\
    \S &= \K_{\Z, \Z}\b{\K_{\Z, \Z} + \frac{1}{\sigma^2}\K_{\Z, \X}\K_{\X, \Z}}^{-1}\K_{\Z, \Z}.
\end{align*}
Plugging this optimal posterior into the ELBO, it is possible to derive a ``collapsed'' version of the bound as
\begin{align*}
    \mathcal{L} = \log \Nc{\y}{\0,\, \K_{\X, \Z}\K_{\Z, \Z}^{-1}\K_{\Z, \X} + \sigma^2\I_N} - \frac{1}{2\sigma^2}\mathrm{tr}\b{\K_{\X, \X} - \K_{\X, \Z}\K_{\Z, \Z}^{-1}\K_{\Z, \X}}.
\end{align*}
While this version of the bound can no longer be minibatched, it is often easier to train to a good optimum, as it has fewer (optimization) parameters and is suitable for use with quasi-Newton optimizers such as L-BFGS.

\begin{figure}[t]
    \centering
    \begin{subfigure}[b]{0.49\textwidth}
        \includegraphics[width=\textwidth]{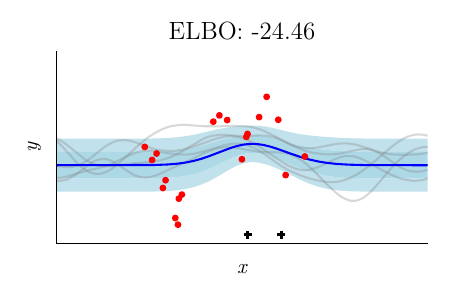}
        \caption{$M = 2$}
    \end{subfigure}
    \begin{subfigure}[b]{0.49\textwidth}
        \includegraphics[width=\textwidth]{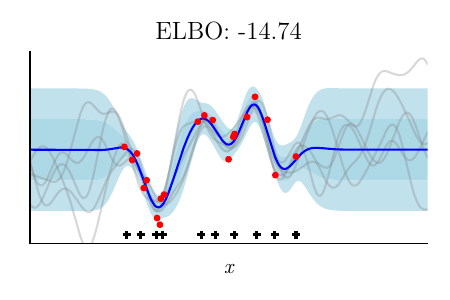}
        \caption{$M = 10$}
    \end{subfigure}
    \caption{Illustration of sparse variational inference for GPs with a squared exponential kernel. The approximate posterior formed by taking only 2 inducing points (a) neither models the data well, nor provides sensible hyperparameters: for instance, the lengthscale is chosen as $\ell_1 = 1.100$, as compared to $\ell_1 = 0.415$ given by the LML (cf. Fig.~\ref{fig:intro:exact_gpr}). By contrast, 10 inducing points (b) provide a better model of the data, obtain a much better ELBO (cf. LML from Fig.~\ref{fig:intro:exact_gpr} of -14.21), and a much better lengthscale of $\ell_1 = 0.526$.}
    \label{fig:intro:sgpr}
\end{figure}

We demonstrate sparse variational inference in Fig.~\ref{fig:intro:sgpr}, using the collapsed bound of \citet{titsias2009variational} for training.
We see that the variational posterior improves as more inducing points are used.
This behavior is to be expected, as adding an inducing point can only improve the ELBO \citep{bauer2016understanding}.
Moreover, we again see (and this time in a more pronounced manner) that more flexibility in the approximate posterior leads directly to improved hyperparameter estimates, as there is less bias in the hyperparameter loss landscape.
Thus, we can perform model selection with sparse GP models, but only reliably when we have a sufficient number of inducing points.
Sparse methods for GPs are therefore most successful when there is some level of redundancy in the data.

\section{Thesis overview}

In the previous sections, we have seen that
\begin{enumerate}
    \item the Bayesian approach provides a principled way of obtaining uncertainty estimates for a model's predictions,
    \item model selection can be performed using the marginal likelihood, and
    \item variational inference can be used to achieve both of these, given sufficiently flexible approximate posteriors.
\end{enumerate}
The rest of the thesis is devoted to showing how variational inference for deep models can be improved to facilitate these promises entailed by the Bayesian framework.
In Chapter~\ref{sec:back}, we introduce the deep models we will consider in this thesis: Bayesian neural networks (BNNs) and deep Gaussian processes (DGPs).
For both, we discuss prior work on inference in these models, including non-variational approaches, and highlight some of the challenges that need to be overcome to make variational inference successful.
Following this background material, we turn our focus to our main contributions.

In Chapter~\ref{sec:dkl}, we consider whether it is necessary to marginalize over \emph{all} the parameters in a Bayesian model, or whether we can treat them as hyperparameters to be optimized with respect to the marginal likelihood.
In doing so, we explore some of the limitations of the marginal likelihood, which are important to understand if we wish to use it for hyperparameter optimization and model selection.
We also consider whether the same limitations apply to the ELBO.
Chapters~\ref{sec:gi} and~\ref{sec:dwp} are devoted to demonstrating improved variational inference in BNNs and DGPs.
We begin in Chapter~\ref{sec:gi} by demonstrating how BNNs and DGPs can be treated as equivalent models for the purpose of inference, and provide a tractable approximating family for both that provides correlations between all layers.
In the BNN case, we also introduce new priors, and show that our improved posterior is sufficiently flexible to take advantage of them, whereas less flexible posteriors struggle.
In Chapter~\ref{sec:dwp}, we show how variational inference in certain DGPs can be improved by analytically removing symmetries in the true posterior, leading to higher ELBOs and better predictive performance.
While we do not explicitly consider the use of variational inference for model selection in these chapters, we believe our work on improving approximate posteriors will lay the groundwork to make this possible.
We conclude our thesis by summarizing our contributions and outlining directions for future work in Chapter~\ref{sec:discuss}.


\chapter[Background]{Background: Deep Bayesian modeling}  
\label{sec:back}

\ifpdf
    \graphicspath{{Background/Figs/Raster/}{Background/Figs/PDF/}{Background/Figs/}}
\else
    \graphicspath{{Background/Figs/Vector/}{Background/Figs/}}
\fi

In the previous chapter, we introduced probabilistic modeling from a Bayesian perspective for shallow models, with a focus on variational inference and its capabilities.
We now turn to providing the background on deep models necessary for the rest of thesis, which will also give us an opportunity to further expound the notation we use.
In doing so, we again motivate our focus on variational inference as a unified framework for both uncertainty quantification and model selection.

\section{Neural networks}
\label{sec:back:nns}
Neural networks have shown remarkable success in a wide variety of tasks in the past decade, ranging from image classification \citep{krizhevsky2012imagenet, he2016deep} to natural language processing \citep{sutskever2014sequence,devlin2019bert} and even molecular property prediction \citep{schutt2017schnet,anderson2019cormorant}.
This success has been driven by the notion that a model should learn features from the data \citep{krizhevsky2009learning,lecun2015deep}.
This is in contrast to the models we considered in the previous chapter, where the features were fixed for both Bayesian linear regression and Gaussian processes.
In these cases, we were only able to tune hyperparameters such as prior and likelihood variances as well as lengthscales.
While neural networks have been shown to learn features effectively, they do not model uncertainty in a principled manner -- indeed, they are generally known to be overconfident in their predictions \citep{guo2017calibration}.
This is a crucial limitation for many tasks, as a sense of the model's uncertainty may be critical for downstream decision making.

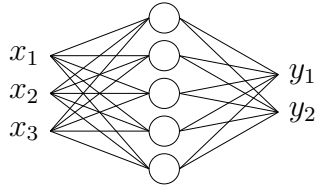
\begin{figure}[t]
\centering
    \centering
    \begin{tikzpicture}
    \draw[very thin] (0, 0) circle (0.2);
    \draw[very thin] (0, 0.5) circle (0.2);
    \draw[very thin] (0, -0.5) circle (0.2);
    \draw[very thin] (0, 1) circle (0.2);
    \draw[very thin] (0, -1) circle (0.2);
    \draw[very thin] (0.2, 0) -- (1.5, 0.25) node[anchor=west] {$y_1$};
    \draw[very thin] (0.2, 0) -- (1.5, -0.25) node[anchor=west] {$y_2$}; 
    \draw[very thin] (0.2, 0.5) -- (1.5, 0.25);
    \draw[very thin] (0.2, 0.5) -- (1.5, -0.25);
    \draw[very thin] (0.2, 1) -- (1.5, 0.25);
    \draw[very thin] (0.2, 1) -- (1.5, -0.25);
    \draw[very thin] (0.2, -0.5) -- (1.5, 0.25);
    \draw[very thin] (0.2, -0.5) -- (1.5, -0.25);
    \draw[very thin] (0.2, -1) -- (1.5, 0.25);
    \draw[very thin] (0.2, -1) -- (1.5, -0.25);
    \draw[very thin] (-0.2, 0) -- (-1.5, 0.5) node[anchor=east] {$x_1$};
    \draw[very thin] (-0.2, 0) -- (-1.5, 0) node[anchor=east] {$x_2$};
    \draw[very thin] (-0.2, 0) -- (-1.5, -0.5) node[anchor=east] {$x_3$};
    \draw[very thin] (-0.2, 0.5) -- (-1.5, 0.5);
    \draw[very thin] (-0.2, 0.5) -- (-1.5, 0);
    \draw[very thin] (-0.2, 0.5) -- (-1.5, -0.5);
    \draw[very thin] (-0.2, -0.5) -- (-1.5, 0.5);
    \draw[very thin] (-0.2, -0.5) -- (-1.5, 0);
    \draw[very thin] (-0.2, -0.5) -- (-1.5, -0.5);
    \draw[very thin] (-0.2, 1) -- (-1.5, 0.5);
    \draw[very thin] (-0.2, 1) -- (-1.5, 0);
    \draw[very thin] (-0.2, 1) -- (-1.5, -0.5);
    \draw[very thin] (-0.2, -1) -- (-1.5, 0.5);
    \draw[very thin] (-0.2, -1) -- (-1.5, 0);
    \draw[very thin] (-0.2, -1) -- (-1.5, -0.5);
    \end{tikzpicture}
    \caption{A single (hidden) layer fully-connected neural network with three inputs $x_1, x_2, \textrm{ and } x_3$, two outputs $y_1 \textrm{ and } y_2$, and one hidden layer with five hidden units.}
    \label{fig:back:fc_net}
\end{figure}

The simplest form of neural network is a feedforward, fully-connected network, which alternates between linear layers and ``neurons'', or \emph{hidden units}, elementwise nonlinear functions $\psi(\cdot): \reals \rightarrow \reals$ also known as \emph{activation functions}.
We refer to the nonlinearities between the linear layers as \emph{hidden layers}, the input to these layers as \emph{pre-activations} or \emph{activities}, and the outputs as \emph{post-activations} or simply \emph{activations}.
We mathematically describe a general fully-connected neural network with $L$ (hidden) layers as follows.
Consider input data $\X \in \reals^{N\times D}$ where $D = \nu_0$; we propagate this input through the network to outputs $\F_{L+1} \in \reals^{N\times \nu_{L+1}}$ via
\begin{align}
    \nonumber
    \F_1 &= \X \W_1, \\ 
    \label{eq:back:fc_net}
    \F_\ell &= \psi\b{\F_{\ell-1}} \W_\ell \quad \textrm{ for } \ell \in \cb{2,\dotsc,L+1},
\end{align}
where we have weights $\slpo{\W_\ell}$, with $\W_\ell \in \reals^{\nu_{\ell-1}\times \nu_\ell}$.
Note that we have again absorbed any bias terms into the weight matrices.
We present an example of a simple single (hidden) layer neural network in Fig.~\ref{fig:back:fc_net}.

By contrasting the equations for a neural networks with those for linear regression presented in Sec.~\ref{sec:intro:probmodel}, we can better understand what it means for a neural network to ``learn features.''
Inspecting the final layer of the network, we see that $\F_{L+1} = \psi\b{\F_{L}}\W_{L+1}$, which we can equivalently write as $\F_{L+1} = \Featsub{\slp{\W_\ell}}{\X}\W_{L+1}$. 
We see that $\Featsub{\slp{\W_\ell}}{\cdot}$ plays the role of the feature map that we saw earlier; however, in this case we do not provide a fixed feature map.
Rather, the features are learned by the weights in the previous layers, $\slp{\W_\ell}$.

While fully-connected networks are useful in regression problems with little obvious structure in the input data, for problems with more structure performance can usually be improved by taking advantage of this specific structure with appropriate modifications to the model's architecture.
This idea is perhaps most prominent in convolutional neural networks \citep[CNNs;][]{lecun1998gradient, krizhevsky2012imagenet}, which make use of the two-dimensional structure present in images.
As the name suggests, the basic building block of CNNs is the convolutional layer, which we now consider in its simplest form.
Consider an image input $\x$ with $C_\text{in}$ channels, height $H_\text{in}$, and width $W_\text{in}$, i.e., $\x \in \reals^{C_\text{in} \times H_\text{in} \times W_\text{in}}$.
Then, given a set of \emph{convolutional filters} $\W \in \reals^{C_\text{out}\times C_\text{in} \times k \times k}$, the output of the convolutional layer, $\Y \in \reals^{C_\text{out}\times H_\text{out} \times W_\text{out}}$, is given by
\begin{align}
    \label{eq:back:conv}
    \Y_{c', :, :} = \sum_{c=1}^{C_\text{in}} \X_{c, :, :} \ast \W_{c', c, :, :},
\end{align}
where we use $\ast$ to denote a two-dimensional discrete convolution,\footnote{Strictly speaking, the operation that is usually implemented is the cross-correlation operation. However, the two are functionally equivalent as the convolutional filters are learned.} and `$:$' denotes that we take all elements along the dimension.\footnote{We note that, as before, it is also possible to include a bias term, which we have omitted for simplicity.}
This basic layer can be made more flexible by considering different convolutional strides, paddings, and filter dilations, although we do not consider these in depth.

\begin{figure}[t]
\centering
\definecolor{darkorange}{RGB}{255,127,14}
\definecolor{forestgreen}{RGB}{20,100,44}

    \begin{tikzpicture}
    
    \fill[forestgreen!20] (0, 1) rectangle ++ (1, 1);
    \fill[darkorange!20] (1, 0.5) rectangle ++ (1, 1);
    \draw[step=0.5] (0, 0) grid (2, 2);
    
    \newcounter{mycount}
    \setcounter{mycount}{`a}
    \foreach \y in {+1.75,+1.25,0.75,0.25}
    \foreach \x in {0.25,0.75,1.25,1.75}
    \node at (\x,\y) {\char\value{mycount}\addtocounter{mycount}{1}};
    
    \node at (2.5, 1) (conv) {$\ast$};
    
    \draw[step=0.5] (3-0.001, 0.5) grid (4, 1.5);
    
    \setcounter{mycount}{`w}
    \foreach \y in {+1.25,+0.75}
    \foreach \x in {3.25,3.75}
    \node at (\x,\y) {\char\value{mycount}\addtocounter{mycount}{1}};
    
    \node at (4.5, 1) (equal) {$=$};
    
    \fill[forestgreen!20] (5, 1.25) rectangle ++ (0.5, 0.5);
    \fill[darkorange!20] (6, 0.75) rectangle ++ (0.5, 0.5);
    \draw[step=0.5,shift={(0., 0.25)}] (5.0-0.001, 0) grid (6.5, 1.5);
    
    \fill[forestgreen!20] (7.0, 1.1) rectangle ++ (0.5, 0.5);
    \node[anchor=west, align=center] at (7.5, 1.35) (result1) {\small $= \text{aw} + \text{bx} + \text{ey} + \text{fz}$};
    
    \fill[darkorange!20] (7.0, 0.4) rectangle ++ (0.5, 0.5);
    \node[anchor=west, align=center] at (7.5, 0.65) (result1) {\small $= \text{gw} + \text{hx} + \text{ky} + \text{lz}$};
    
    \end{tikzpicture}
\caption{A pictorial representation of a convolution operation with a $4\times 4$ image and $2\times 2$ kernel, resulting in a $3\times 3$ image.}
\label{fig:back:convolution}
\end{figure}
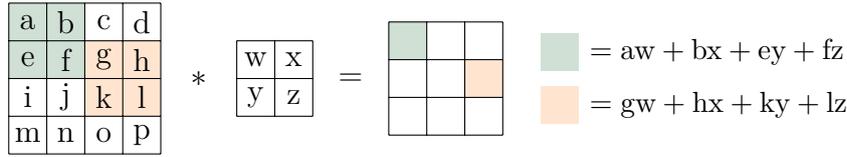

To make a convolutional network, we alternate convolutional layers with activation functions and pooling operations, the latter of which reduce the size of the images. 
The pooling operations help the network to move from identifying low-level features, such as edges, to higher level features such as shapes.
More recently, CNNs have also included normalization layers such as Batch Normalization \citep[BN;][]{ioffe2015batch}, which have been found to be empirically useful.
After being processed through the convolutional layers, the image is flattened and pushed through fully connected layers to form the output.
In Fig.~\ref{fig:back:convolution}, we present a diagram of a simple convolution operation as might be used in a convolutional network.

\subsection{Bayesian neural networks}
\label{sec:back:bnns}
The most popular learning paradigm for neural networks has far and away been maximum likelihood estimation, potentially with some added regularization.
However, despite the success of deterministic neural networks in a wide range of tasks, they are not well-suited for safety-critical tasks such as autonomous driving, as they do not automatically represent uncertainty in their beliefs (i.e., epistemic uncertainty).
Even more problematic for safety, their predictions tend to be poorly calibrated and overconfident \citep{guo2017calibration}: when predicting with a classification likelihood that outputs class probabilities, the predicted class probabilities are often significantly higher than the model's accuracy.
The lack of uncertainty in their predictions additionally makes it difficult to use neural networks in tasks which make explicit use of a model's uncertainty, such as Bayesian optimization.
Finally, as in the case of featurized linear regression, it is difficult to perform model selection without referring to the model's performance on a held-out dataset, which is both expensive to evaluate on and can be wasteful of potentially valuable data.

We therefore follow a Bayesian approach, wherein we apply a prior $\p{\slpo{\W_\ell}}$ on our weights, resulting in a \emph{Bayesian neural network (BNN)}.
We restrict the priors we consider to be Gaussians that factorize over both layers and the outputs within layers.
Returning to the notation of a fully-connected model for simplicity, these priors can be written as
\begin{align}
    \nonumber
    \p{\slpo{\W_\ell}} &= \prodlayers \;\p{\W_\ell} \\
    \label{eq:back:pw}
    &= \prodlayers \prodln \;\Nc{\w_\lambda^\ell}{\0, \frac{1}{\nu_{\ell - 1}}\mS_{\ell}}.
\end{align}
In many prior works, $\mS_\ell$ is set to $\nu_{\ell-1}\I_{\nu_{\ell-1}}$, so that each weight has an independent $\N{0, 1}$ prior distribution.
In Chapter~\ref{sec:gi}, we consider other forms of the prior covariance.
After choosing a likelihood for our model, $\pc{\Y}{\X, \slpo{\W_\ell}} = \pc{\Y}{\F_{L+1}}$, our desired object is again the posterior,
\begin{align*}
    \pc{\slpo{\W_\ell}}{\X, \Y} &= \frac{\pc{\Y}{\F_{\L+1}}\p{\slpo{\W_{L+1}}}}{\pc{\Y}{\X}}.
\end{align*}
Unfortunately, even for the simplest networks the marginal likelihood $\pc{\Y}{\X}$ is intractable.
Therefore, we must resort to approximate inference.
Since the inception of BNNs, countless approaches for approximate Bayesian inference have been proposed.
Here, we highlight some of the dominant approaches, and summarize their strengths and weaknesses.

\subsection{Approximate inference in BNNs}

One of the prominent early approaches for BNN inference was Markov chain Monte Carlo (MCMC). 
This was first explored in depth by \citet{neal1995bayesian}, who used Hamiltonian Monte Carlo (HMC) \citep{duane1987hybrid,neal2011mcmc}, which is often referred to as the ``gold standard'' for approximate Bayesian inference.
By constructing a Markov chain in parameter space whose stationary distribution is the true posterior, MCMC methods are able to guarantee (under some mild regularity conditions) that the samples generated are asymptotically reflective of the true posterior \citep{brooks2011handbook}.
Despite this guarantee, in practice it is very difficult to tell whether the Markov chain has actually converged to the true target distribution.
Moreover, exact implementations of MCMC methods, particularly gradient-based methods such as HMC, require multiple evaluations of the network over the entire dataset to obtain a single sample.
This is prohibitive for large models with large datasets, which has led to increasing interest in stochastic gradient MCMC (SGMCMC) methods \citep{welling2011bayesian,chen2014stochastic,ma2015complete}, which subsample the dataset to obtain stochastic gradients and ignore the costly Metropolis accept/reject step.
For instance, \citet{zhang2020cyclical} showed that SGMCMC could be used to scale up inference to the ImageNet dataset \citep{imagenet}.
However, despite recent progress in improving the accuracy of SGMCMC \citep{garriga2021exact}, SGMCMC does not have the theoretical convergence guarantees of MCMC \citep{johndrow2020no, betancourt2015fundamental}, and both have parameters that are difficult to tune.
Finally, and perhaps most crucially, MCMC methods do not have a straightforward mechanism to perform model comparison.

Another strategy for BNN inference that has received increasing recent interest is Laplace's method.
Originally proposed for BNNs by \citet{denker1990transforming,mackay1992practical}, the Laplace method first finds a \emph{maximum a posteriori (MAP)} estimate of the  model and uses the local Hessian to fit a Gaussian approximation to the true posterior.
As computing the full Hessian is intractable for today's large-scale networks, recent efforts have focused on ways of approximating this computation, showing success in uncertainty quantification on large architectures  \citep{ritter2018scalable,daxberger2021laplace}.
The Laplace approximation further provides an estimate of the LML, which has been shown to work well for model selection \citep{immer2021scalable}, signaling that the marginal likelihood is indeed desirable for deep models.
However, the Laplace approximation has a few limitations.
First, the approximation is restricted to Gaussian posteriors, and so cannot capture heavy tails or multimodality, both of which may be present in the true posterior.
Second, it is difficult to know how good of an approximation the Laplace approximation is, particularly with respect to the LML estimate.
More specifically, there is a possibility that the Laplace approximation provides an overestimate of the true LML for certain hyperparameter values, meaning that the LML estimate can diverge from the true LML when optimizing hyperparameters.
This could be dangerous, as it means that a worse Laplace approximation to the LML may appear better than a good approximation, as the worse approximation could overestimate the LML more.

In another line of work, some approaches of non-Bayesian origin have recently attracted interpretations as approximations to the true Bayesian posterior.
The two most prominent of these are \emph{Monte Carlo dropout} \citep{Gal2015DropoutB,kingma2015variational} and \emph{deep ensembles} \citep{lakshminarayanan2017simple}.
The first uses dropout \citep{srivastava2014dropout} at test time to sample sets of weights, which can be viewed as sampling from a variational posterior.
However, the variational posterior implied by dropout has an infinite KL divergence to the true posterior and does not concentrate with more data, making it difficult to interpret according to the standard variational framework \citep{hron2018dropout,osband2018randomized}.
Indeed, \citet{osband2018randomized} argue that the lack of posterior concentration, which should occur in Bayesian methods, makes MC dropout unsuitable for downstream decision making tasks such as reinforcement learning.
Moreover, MC dropout suffers from significant pathological behavior in its predictive posterior, for instance its inability to model ``in-between'' uncertainty, the epistemic uncertainty that should be present in regions between clusters of datapoints \citep{foong2019between, foong2019pathologies}.
This makes is unsuitable for tasks such as adversarial robustness and active learning.

Deep ensembles, on the other hand, train multiple sets of weights for the same model, relying on random initialization and the multimodality of the loss surface to obtain different fits.
The different fits can then be used analogously to samples from a posterior distribution to obtain predictive uncertainty.
Deep ensembles have been shown to be remarkably effective at both in-distribution performance and uncertainty-related tasks such as robustness to dataset shift \citep{ovadia2019trust}.
Interestingly, while \citet{lakshminarayanan2017simple} were clear that they did not view deep ensembles as a truly Bayesian approach, later work \citep[e.g.,][]{wilson2020bayesian} has reinterpreted it as such.
Nevertheless, neither MC dropout nor deep ensembles directly minimize a divergence to the true posterior, nor is it possible to obtain the true posterior with these methods in an infinite limit.
Finally, they also do not provide LML estimates, limiting their usefulness when it comes to model selection.

The last major approach to inference in BNNs has been variational inference, which is where we focus, as it provides a framework that allows for simultaneous uncertainty quantification and model selection, allowing us to fully realize the potential of the Bayesian paradigm.
Mean-field variational inference (MFVI), where the approximate posterior is a fully-factorized Gaussian, has generally been the most popular form of VI in BNNs \citep{hinton1993keeping,graves2011practical, blundell2015weight}, due to its ease of implementation and computational simplicity.
Nevertheless, although a mean-field posterior can potentially be expressive \citep{farquhar2020liberty}, it has been shown that they can struggle with ``in-between'' uncertainty previously mentioned \citep{foong2019between, foong2019pathologies, yao2019quality}. 
Even more worryingly, wide BNNs fit using MFVI can ignore the data completely \citep{coker2022wide}.
It has also been observed that the ELBO for MFVI is too loose to be used for prior hyperparameter selection, thereby limiting its usefulness for model selection \citep{blundell2015weight,bui2021biases}.
As a result of these issues, there has been ample work on providing structured yet computationally efficient approximate posteriors \citep[e.g.,][]{louizos2016structured, louizos2017multiplicative, krueger2017bayesian, dusenberry2020efficient, tomczak2020efficient}.
However, incorporating the necessary correlation structure into tractable approximate posteriors is difficult, particularly between layers.
This has led to interest in other VI-like objectives and divergences \citep[e.g.,][]{hernandez2015probabilistic} as well as function-space approaches \citep{sun2018functional, ma2019variational}.
However, these approaches often lose the nice properties of VI which we hope to take advantage of, such as providing a lower bound to the LML. 
The function-space approaches are often also theoretically difficult to justify \citep{burt2020understanding}.
In Chapter~\ref{sec:gi}, we propose a method for naturally incorporating correlation structure which provides correlation between all weights of a neural network, and that can be successfully used for model selection.

\subsection{Tempered posteriors and the cold posterior effect}
\label{sec:back:temp}
From a purist Bayesian perspective, one of the most interesting phenomena surrounding modern BNNs is that the vast majority of practical work on BNNs has necessitated some form of posterior tempering to obtain satisfactory results. 
For MCMC, this involves sharpening the posterior by targeting 
\begin{align*}
    \mathrm{p}^{*}\bc{\slpo{\W_\ell}}{\X, \Y} &= \pc{\slpo{\W_\ell}}{\X, \Y}^{1/T},
\end{align*}
where $T < 1$ is known as the \emph{temperature}, resulting in a \emph{cold posterior}.
For VI, the ELBO is modified by introducing a factor $\lambda <  1$, which is distinct from $T$:\footnote{Indeed, the only direct relationship occurs when the true posterior is targeted with $T = \lambda = 1$.}
\begin{align*}
    \mathcal{L}^* &= \log \E[\q{\slpo{\W_\ell}}]{\log \pc{\Y}{\X, \slpo{\W_\ell}}} - \lambda \KL*{\q{\slpo{\W_\ell}}}{\p{\slpo{\W_\ell}}}.
\end{align*}
Unfortunately, this modified ELBO no longer provides a valid bound on the log marginal likelihood.
Therefore, from the Bayesian point of view, tempering is not ideal, as it clouds the Bayesian perspective, both by targeting an object different from the true posterior and limiting the potential for model selection.

Despite this, it has been noted in many pieces of prior work that tempering can dramatically improve performance \citep{wenzel2020good, noci2021disentangling}.
While this is not ideal, we would argue that perhaps it should not be too surprising.
There is no guarantee that the true Bayesian posterior will provide optimal performance when the model (which can be either the architecture or the prior) is mis-specified, as it almost certainly will be for BNNs.
Moreover, the biases introduced by approximate inference can mean that even if the model is well-specified, the approximate posterior for that model may not result in optimal performance.
For instance, \citet{izmailov2021bayesian} argue that the cold posterior effect disappears when models are used without data augmentation, as data augmentation will lead to more data than is accounted for in a na\"{i}ve application of Bayes' rule. 
In order to support this hypothesis, they impressively use long runs of computationally expensive (exact) HMC.
Alternatively, \citet{aitchison2021statistical} argues that data curation is responsible for this effect, as data curation is not accounted for in the modeling assumptions.
Regardless of the exact mechanism for the cold posterior effect, we believe that improved approximations and the consequential improvement in model specification through a tighter ELBO will help resolve whether the cold posterior effect results from poor model specification or whether there is another issue at play.
In the interests of retaining a clean Bayesian interpretation, we will endeavor to avoid tempering or data augmentation, as these practices can muddy the waters conceptually.

\section{Deep Gaussian processes}
\label{sec:back:dgps}
As we did with the Bayesian linear regression problem in Sec.~\ref{sec:intro:gps}, we can consider an alternative view of Bayesian neural networks.
By combining the equations for propagating the features and the form of the prior (Eqs.~\ref{eq:back:fc_net} and~\ref{eq:back:pw}), we see that the distribution of the activations $\F_\ell$ conditioned on those of the previous layer, $\F_{\ell-1}$, is given by
\begin{align*}
    \pc{\F_\ell}{\F_{\ell-1}} &= \prodln \Nc{\f_{\lambda}^\ell}{\0, \Kf{\F_{\ell-1}}}, \\
    \Kf{\F_{\ell-1}} &= k_\mathrm{f}\b{\F_{\ell-1}}= \frac{1}{\nu_{\ell-1}}\psi\b{\F_{\ell-1}}\mS_\ell\transpose{\psi\b{\F_{\ell-1}}},
\end{align*}
where we recall that $\frac{1}{\nu_{\ell-1}}\mS_\ell$ is the prior weight covariance.
This relationship holds for the first layer as well, provided that we define $\psi\b{\F_0} = \X$.
Therefore, a BNN with a Gaussian prior can be viewed as a prior on activations that is conditionally Gaussian, when conditioning on the previous layer's activations.
Such a model is known as a \emph{deep Gaussian process}, with this relationship first being noted by \citet{Gal2015DropoutB}.
However, this view of BNNs has, to the best of our knowledge, had scarce usage in the literature \citep[with][as exceptions]{pleiss2021limitations, dutordoir2021deep, louizos2016structured,aitchison2019bigger}, most likely because the covariance is degenerate for $N$ larger than the width of the layer, which will often be the case.
Nevertheless, in Chapter~\ref{sec:gi}, we make use of this view of BNNs to propose a variational posterior that provides a unified view of inference for both BNNs and deep Gaussian processes.

Formally, a deep Gaussian process \citep[DGP;][]{damianou2013deep} can be defined as 
\begin{align*}
    f_\lambda^\ell &\sim \mathcal{GP}\b{m_\ell, k_\ell}, \quad \lambda \in \{1, \ldots, \nu_\ell\}, \;\ell \in \{1, \ldots, L+1\}, \\
    \pc{\F_\ell}{\F_{\ell-1}, f_\ell} &= \prodln \Nc{\f_\lambda^\ell}{f_\lambda^\ell\b{\F_{\ell-1}}, \sigma_\ell^2},
\end{align*}
along with a suitable likelihood $\pc{\Y}{\F_{L + 1}}$.\footnote{When describing a DGP model, we diverge slightly from how we described BNNs and refer to the number of \emph{GP layers}, and not the number of ``hidden layers.'' For instance, whereas a 3-layer BNN would refer to a model with 3 layers of hidden units and 4 linear layers, a 3-layer DGP indicates that it has 3 GP layers and 2 ``hidden layers.''}
As a minor difference to BNNs, we define $\F_0 = \X$.
We also note that in the BNN case we have set the noise in each layer $\sigma_\ell^2$ to zero, so that the activations coincide with the intermediate function values.
In general, for DGPs we will learn the values of $\sigma_\ell^2$ in each layer (although we will also set it to zero for the last layer, where the likelihood will handle the noise in the data).

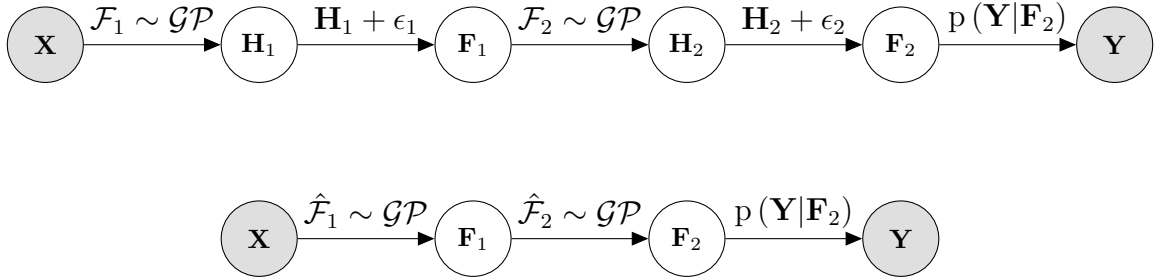
\begin{figure}[!tp]
\centering
    \begin{tikzpicture}[latent/.append style={minimum size=1cm},
    obs/.append style={minimum size=1cm}]
    \node[obs] (x) {$\X$};
    \node[latent, right=of x, xshift=0.8cm] (h1) {$\H_1$};
    \node[latent, right=of h1, xshift=0.8cm] (f1) {$\F_1$};
    \node[latent, right=of f1, xshift=0.8cm] (h2) {$\H_2$};
    \node[latent, right=of h2, xshift=0.8cm] (f2) {$\F_2$};
    \node[obs, right=of f2, xshift=0.8cm] (y) {$\Y$};
    \path (x) edge [->, >={triangle 45}] node[above] {$\mathcal{F}_1 \sim \mathcal{GP}$} (h1);
    \path (h1) edge [->, >={triangle 45}] node[above] {$\H_1 +\epsilon_1$} (f1);
    \path (f1) edge [->, >={triangle 45}] node[above] {$\mathcal{F}_2 \sim \mathcal{GP}$} (h2);
    \path (h2) edge [->, >={triangle 45}] node[above] {$\H_2 + \epsilon_2$} (f2);
    \path (f2) edge [->, >={triangle 45}] node[above] {$\pc{\Y}{\F_2}$} (y);
    \node[below=of x] {};
    \end{tikzpicture}

        \begin{tikzpicture}[latent/.append style={minimum size=1cm},
    obs/.append style={minimum size=1cm}]
    \node[obs] (x) {$\X$};
    \node[latent, right=of x,xshift=0.8cm] (f1) {$\F_1$};
    \node[latent, right=of f1,xshift=0.8cm] (f2) {$\F_2$};
    \node[obs, right=of f2,xshift=0.8cm] (y) {$\Y$};
    \path (x) edge [->, >={triangle 45}] node[above] {$\hat{\mathcal{F}}_1 \sim \mathcal{GP}$} (f1);
    \path (f1) edge [->, >={triangle 45}] node[above] {$\hat{\mathcal{F}}_2 \sim \mathcal{GP}$} (f2);
    \path (f2) edge [->, >={triangle 45}] node[above] {$\pc{\Y}{\F_2}$} (y);
    \end{tikzpicture}

    \caption{
      Comparison of the graphical models for two approaches to DGPs. The top model depicts a DGP model as originally proposed by \citet{damianou2013deep}, and as used in \citet{dai2016variational}. In this model, inference is performed on both $\cb{\H_\ell}_\ell$ and $\cb{\F_\ell}_\ell$, where $\F_\ell$ is a noisy realization of $\H_\ell$, with the noise $\epsilon_\ell \sim \N{0, \sigma_\ell^2}$. The bottom model depicts the DGP model as used in \citet{hensman2014nested} and \citet{salimbeni2017doubly}.
      In this model, inference is only performed on $\cb{\F_\ell}_\ell$, with the noise $\epsilon_\ell$ in the original model being incorporated into the kernel of the layer's GP, which we now denote by $\hat{\mathcal{F}}_\ell$.
      \label{fig:back:dgp-compare}
    }
\end{figure}

Exact inference in DGPs is again intractable and is usually handled with VI, although there have been a few exceptions \citep{bui2016deep, havasi2018inference, lu2020interpretable}.
\citet{damianou2013deep}, who proposed DGPs, perform inference on both the underlying process and activations using latent variables, which allows for an approximate posterior that can model heteroscedasticity.
We depict this in the top graphical model of Fig.~\ref{fig:back:dgp-compare}, where $\H_\ell = \mathcal{F}_\ell(\F_{\ell-1})$ are the function values of the underlying process $\mathcal{F}_\ell$ at the previous layer's outputs.
Hence, $\F_\ell = \H_\ell + \epsilon_\ell$, where $\epsilon \sim \N{0, \sigma_\ell^2}$, and inference is performed over both $\cb{\H_\ell}_\ell$ and $\cb{\F_\ell}_\ell$ by using a latent variable-based approach.
\citet{damianou2013deep} show that they can use their resulting lower bound for model selection; however, their inference method is expensive for large datasets and results in an ELBO that cannot be minibatched.
\citet{dai2016variational} improves on this by amortizing the inference over the latent variables \citep{rezende2014stochastic,kingma2013auto}, which allows for distributed computation across datapoints; however, the ELBO still cannot be minibatched.
Unfortunately, in our experience \citep[and as reported by][]{bui2018efficient}, these models that rely on latent variable inference are difficult to train, with different initializations of the variational parameters often resulting in wildly different fits.

Therefore, more recent works remove the need for latent variables by incorporating the layer-wise noise $\sigma_\ell^2$ directly into the kernel and performing inference only on the resulting process.
We depict this approach to DGPs in the bottom graphical model of Fig.~\ref{fig:back:dgp-compare}.
We observe that we have removed the function values $\cb{\H_\ell}_\ell$ and perform inference only on the activations $\cb{\F_\ell}_\ell$, incorporating the noise $\epsilon_\ell$ directly into the GP layers, now denoted by $\hat{\mathcal{F}_\ell}$ \citep[note that as we do not return to the model of][we drop the $\hat{\cdot}$ for the remainder of this work]{damianou2013deep}.
Using this modification, \citet{hensman2014nested} build on the work of \citet{hensman2013gaussian} to provide an inducing point posterior which factorizes across layers, resulting in an ELBO that can be minibatched.
In order to do so, they had to restrict the form of the posterior so that predictions are almost deterministic conditioned on the inducing variables, with only added $\sigma_\ell^2$ noise \citep[similar to the projected latent variables/deterministic training conditional method of][]{seeger2003fast}.
To allow for a probabilistic treatment of the training and testing data, \citet{salimbeni2017doubly} take a doubly-stochastic approach, where instead of analytically computing integrals, they satisfy themselves with Monte Carlo estimates using reparameterized variational inference (see App.~\ref{app:reparam}).
This allows them to fully extend the minibatched approach of \citet{hensman2013gaussian, hensman2015scalable} to the deep case, so that predictions within a layer use the prior conditional $\pc{\F_\ell}{\U_\ell}$ for predicting, where $\U_\ell$ are the inducing variables in that layer. 
While this approach has been shown remarkably successful, the removal of the latent variables comes at the cost of removing the ability for the model to handle heteroscedastic noise; this was later resolved in \citet{salimbeni2019deep} by combining the approach of \citet{dutordoir2018gaussian} with an importance-weighted ELBO \citep{burda2016importance}.

As the approach of \citet{salimbeni2017doubly} is central to our contributions in Chapters~\ref{sec:gi} and~\ref{sec:dwp}, we briefly outline it here. Incorporating the noise into the kernel for each layer, we can rewrite the model as
\begin{align*}
    \pc{\F_\ell}{\F_{\ell-1}} &= \prodln \Nc{\f_\lambda^\ell}{\m_\ell\b{\F_{\ell-1}}, \Kfl{\F_{\ell-1}}},
\end{align*}
where we have defined $\b{\Kfl{\F_{\ell-1}}}_{ij} = k_\ell\b{\F_{\ell-1,i}, \F_{\ell-1, j}}$.
As in the GP case, we augment the model with inducing variables $\U_\ell \in \reals^{M_\ell \times \nu_\ell}$, which are defined at every layer and correspond to the latent function values of $f_\ell$ at the inducing locations $\Z_{\ell-1} \in \reals^{M \times \nu_{\ell-1}}$.
This results in the augmented prior model
\begin{align*}
    \pc{\begin{pmatrix}
    \F_\ell \\
    \U_\ell
    \end{pmatrix}
    }{\F_{\ell-1}, \Z_{\ell-1}} &= \prodln
    \Nc{\begin{pmatrix}
    \f_\lambda^\ell \\
    \u_\lambda^\ell
    \end{pmatrix}}{
    \m_\ell\b{\begin{pmatrix}
    \F_{\ell-1} \\
    \Z_{\ell-1}
    \end{pmatrix}},
    \Kfl{\begin{pmatrix}
    \F_{\ell-1} \\
    \Z_{\ell-1}
    \end{pmatrix}
    }
    }.
\end{align*}
Taking inspiration from the Gaussian process literature \citep{titsias2009variational, hensman2013gaussian}, we then propose an approximate posterior
\begin{align*}
    \q{\slpo{\F_\ell, \U_\ell}} &= \prodlayers \qc{\F_\ell, \U_\ell}{\F_{\ell-1}, \Z_{\ell-1}} \\
    &= \prodlayers \pc{\F_\ell}{\U_\ell, \F_{\ell-1}, \Z_{\ell-1}}\qc{\U_\ell}{\Z_{\ell-1}},
\end{align*}
so that the conditional posterior over the activations given the inducing points is the same as the prior conditional.
As with the case for GPs, this allows us to cancel the matching terms in the ELBO, resulting in
\begin{align*}
    \mathcal{L} &= \E[\q{\F_{L+1}}]{\log \pc{\Y}{\F_{L+1}}} - \sum_{\ell=1}^{L+1} \KL{\qc{\U_\ell}{\Z_{\ell-1}}}{\pc{\U_\ell}{\Z_{\ell-1}}}.
\end{align*}
The expected log likelihood term cannot be computed in closed form, but we can obtain an unbiased sample-based estimate by sampling from the posterior.
Note that this involves sampling the final-layer activations, which involves propagating the training data through the model using the approximate posterior.
While this may seem expensive, \citet{salimbeni2017doubly} showed that for factorizing likelihoods, we do not need joint samples for each datapoint --- uncorrelated samples suffice.
Moreover, the expected log likelihood term is amenable to minibatching, resulting in two sources of stochasticity in the estimate of the ELBO, hence the term \emph{doubly stochastic}.

For the form of $\qc{\U_\ell}{\Z_{\ell-1}}$, \citet{salimbeni2017doubly} again take inspiration from the GP literature and choose 
\begin{align*}
    \qc{\U_\ell}{\Z_{\ell-1}} &= \prodln \Nc{\u_\lambda^\ell}{\m_\lambda^\ell, \S_\lambda^\ell}.
\end{align*}
Unlike the shallow GP case, there is no limit in which a Gaussian posterior over the inducing variables will be optimal, even for regression.
However, it does have computational benefits.
First, the KL divergences can be computed in closed form with a Gaussian posterior.
Second, the inducing variables can be marginalized out analytically, meaning that the activations can be sampled without first having to sample inducing variables.
This will reduce the variance of the estimate of the ELBO.
Computing the ELBO with this choice of posterior requires $\mathcal{O}(L\tilde{N}M^2\nu + LM^3\nu)$ computational cost (assuming constant width $\nu$), where $\tilde{N}$ is the minibatch size.

Finally, we note that \citet{salimbeni2017doubly} use an identity mean function, $m_\ell(x) = x$, in all the layers except the last.\footnote{For wide ($\nu > 30$) models, they use a more complicated mean function involving a singular value decomposition of the input data. However, we do not consider models that wide.}
This choice, equivalent to residual connections in neural networks \citep{he2016deep}, was made in response to \citet{duvenaud2014avoiding}, who found that DGPs with zero mean functions can exhibit pathological behavior.

\section{Symmetries in deep models}\label{sec:back:symm}

One of the notable features of deep models is that they inherently have many symmetries and non-identifiabilities in their parameter space.
For instance, permuting the neurons in the hidden layer of a neural network, i.e., $\hat{\W}_\ell = \W_\ell \mP$ and $\hat{\W}_{\ell+1} = \transpose{\mP}\W_{\ell+1}$, where $\mP$ is a permutation matrix, will have no effect on the output of the model.
As most priors we consider will put equal density on each permutation of the weights, the resulting posterior will have multiple modes in weight space that represent the same posterior over functions.
The number of modes resulting from permutation symmetries will be determined by the width of the hidden layers, as each layer will contribute $\nu_\ell !$ modes.
Even for a relatively small neural network, this will result in an enormous number of modes, making inference difficult for any method.

Depending on the activation function used, neural networks may also have scaling symmetries.
For instance, for the popular rectified linear unit \citep[ReLU;][]{nair2010rectified} activation function, defined by $\mathrm{ReLU}\b{x} = \max \b{0, x}$, for any non-zero $\alpha$, we have that $\mathrm{ReLU}\b{x} = \alpha \mathrm{ReLU}\b{x/\alpha}$.
Therefore, scaling the weights into a hidden unit by $\alpha$, and scaling the corresponding output weights by $1/\alpha$ will lead to the same overall network output.
This property will lead to a posterior that is an (uncountably) infinite mixture of modes, each mode corresponding to a particular scaling of the weights that will be weighted by the prior density.

DGPs have similar symmetries in their posteriors.
Analogously to BNNs, we can consider $\hat{\F}_\ell = \F_\ell \mQ$, where $\mQ$ is an orthogonal matrix.
When the kernel function for the next layer is isotropic, this transform will not change its output, as orthogonal matrix transformations preserve distance.
Therefore, the true posterior for a zero-mean DGP with isotropic kernels will contain rotational symmetries, whereby the set $\{\F_\ell \mQ : \mQ \in O(\nu_\ell) \}$ will have the same posterior density, where $O(\nu_\ell)$ refers to the orthogonal group in dimension $\nu_\ell$.
A similar observation holds for ARD kernels, but where the matrices being considered scale the features to match the individual lengthscales (so that the scaled distance is preserved).
When non-zero mean functions are considered, it is more difficult to understand what symmetries exist, although we still believe that symmetries will play an important role in the true posterior.

Although these symmetries can make inference difficult for any method, we will now briefly explore the effect we may expect these symmetries to have in VI, as VI will be the focus of our thesis.
We first note that most approximate posteriors in the literature are unimodal, meaning that they will not be able to explicitly capture these symmetries present in the true posterior.
However, if the approximate posterior is capable of capturing one of the true posterior modes perfectly, it should not need to capture the other modes caused by the model's symmetries, as a single mode entirely contains the information of all the equivalent modes.
Here, VI has another advantage: the reverse KL that it targets, $\KL{\mathrm{q}}{\mathrm{p}}$, results in ``mode-seeking'' behavior which encourages a unimodal approximate posterior to focus on a single mode, rather than interpolating between modes \citep{minka2005divergence}.
However, if two modes are sufficiently close together such that their tails overlap, it is still possible for VI to place a significant amount of mass on an area where there should be none \citep{moore2016symmetrized,pourzanjani2017improving}.
We now illustrate this with a simple toy example.

\begin{figure}[t]
    \centering
    \begin{subfigure}[b]{\textwidth}
        \includegraphics[width=\textwidth]{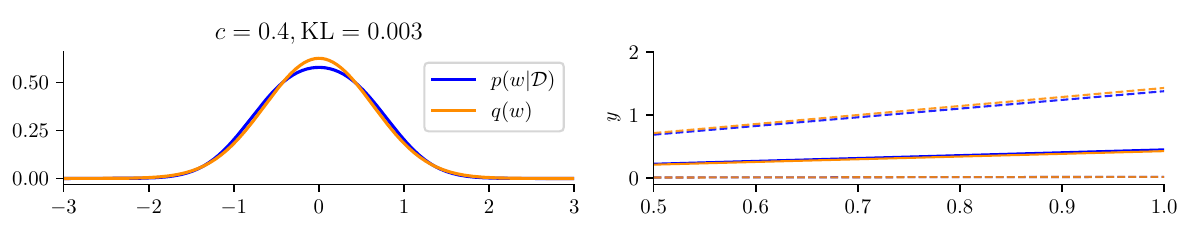}
    \end{subfigure}
    \begin{subfigure}[b]{\textwidth}
        \includegraphics[width=\textwidth]{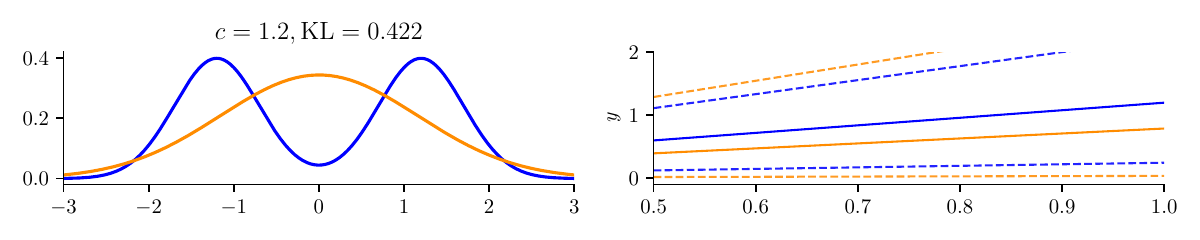}
    \end{subfigure}
    \begin{subfigure}[b]{\textwidth}
        \includegraphics[width=\textwidth]{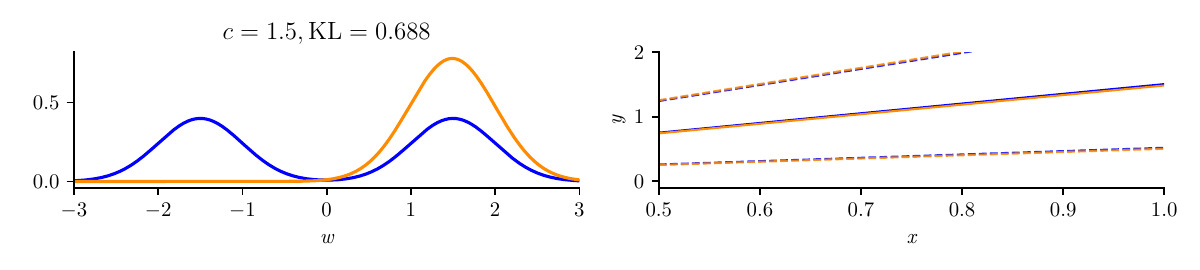}
    \end{subfigure}
    \caption{Fits of a Gaussian approximate posterior $\q{w}$ to true posteriors of the model described in Eqs.~\ref{eq:back:symm-model}--\ref{eq:back:symm-post}, with varying $c$ and $\sigma_\text{post}=0.5$. On the left, we plot the densities resulting from the KL minimization, with the resulting KL indicated in the title. On the right, we plot the resulting posterior predictives for the models, with the solid lines representing the median prediction and the dashed lines representing 95\% confidence intervals. For $c \in \{0.4, 1.5\}$ the approximate posterior predictives are nearly indistinguishable from the true posterior predictives, whereas for $c = 1.2$ there is significant deviation due to $\q{w}$ placing significant mass on low-mass regions of the true posterior.}
    \label{fig:back:symm_vi}
\end{figure}

Consider the simple one-dimensional model
\begin{equation}\label{eq:back:symm-model}
    y = \lvert w \rvert x + \epsilon, \quad \epsilon \sim \N{0, \sigma^2},
\end{equation}
where the absolute value endows the model with a simple symmetry.
With a suitable prior whose density is symmetric about $w = 0$, we can imagine that the posterior is given by (or at the very least well-approximated by)
\begin{equation}\label{eq:back:symm-post}
    \pc{w}{\Data} = \frac{1}{2}\Nc{w}{c, \sigma_\text{post}^2} + \frac{1}{2}\Nc{w}{-c, \sigma_\text{post}^2},
\end{equation}
i.e., a symmetric mixture of two Gaussians.
We wish to understand how a unimodal approximate posterior, $\q{w} = \Nc{w}{m, s^2}$, will behave when trained on this model with VI.

We plot the approximate and true posteriors and posterior predictives for various values of $c$ in Fig.~\ref{fig:back:symm_vi}.
We also show their corresponding KL divergences.
When $c$ is small, the true posterior is unimodal and can be well-approximated by the unimodal $q$.
For intermediate $c$, however, the tails of the mixture posterior overlap significantly and the approximate posterior places a significant amount of its mass on a region where there is little mass in the true posterior.
The effect of this misplacement can be seen in the predictive posterior, which does a poor job of matching the true posterior's predictions.
Finally, for large enough $c$ we observe that the modes of the true posterior are sufficiently well-separated that the KL penalty for ignoring one mode entirely is better than the penalty for putting significant mass where there is none.
In this case, the approximate posterior can model the remaining mode well and thus the true posterior predictive as well.
However, the approximate posterior receives a $\log 2$ penalty for ignoring half the mass of the true posterior.

We have therefore seen that a unimodal approximate posterior can indeed approximate the posterior predictive of a model with symmetries thanks to VI's mode-seeking behavior. 
However, good performance requires that the modes are well-separated, so that the KL penalty for ignoring a mode is less than the penalty for putting mass on low-density areas.
From the definition of the KL divergence (Eq. \ref{eq:intro:kl-def}) it is straightforward to see that the penalty for modeling only one mode out of $N$ equivalent modes is $\log N$, which will therefore be the level of the threshold between interpolating $N$ modes versus modeling only one.
In Chapter~\ref{sec:dwp}, we will demonstrate how we can avoid some of these symmetries entirely by avoiding inference on features, instead focusing entirely on \emph{Gram matrices}.

\section{How should we evaluate VI in deep models?}

One crucial question that we have not yet addressed is how to evaluate variational inference in deep models: what does it mean for us to say that one method is better than another?
Unfortunately, the intractability of exact inference in deep models makes it difficult to understand when one approximation is better than another.
Much of the literature \citep[e.g.,][]{blundell2015weight,louizos2016structured,dusenberry2020efficient} places an emphasis on predictive performance.
For instance, for regression, metrics such as negative log predictive density (NLPD)/test log likelihood (LL) and root mean squared error (RMSE) on held-out test data are often reported.
In classification, test accuracy as well as calibration metrics such as expected calibration error \citep[ECE;][]{naeini2015obtaining, guo2017calibration} and out-of-distribution uncertainty are viewed as important for a method to perform well on.
The algorithms may also be evaluated on uncertainty-aware tasks such as reinforcement learning \citep{blundell2015weight,Gal2015DropoutB} or continual learning \citep{farquhar2020radial}.

Whilst predictive performance is certainly important, it may be misleading when it comes to evaluating an approximate inference algorithm.
This is because poor-quality approximate inference can lead to better results in terms of predictive performance if the model is bad.
For instance, an approximate posterior that ignores the model's uncertainty may perform better on predictive metrics when the prior variance is too large as compared to an approximate posterior that takes the prior uncertainty into account properly.
Fundamentally, under model misspecification, which will typically be the case, it is difficult to determine whether improvements in predictive performance are truly indicative of a better approximation.

One might be tempted to compare the ELBOs of different approximate posteriors to determine whether one approximate posterior is better than another, as a better ELBO will imply that the KL to the true posterior is smaller (assuming equal hyperparameters for the models).
While this is true, the true posterior in the \emph{weight space} of BNNs (or equivalently the activation space for DGPs) is not our primary object of inferential interest.
Rather, we care about the posterior over functions, as this is what will be used to make predictions.
Unfortunately, the weight-space and function-space KL divergences between approximate and true posterior do not have a straightforward relationship for deep models, as there is no one-to-one mapping between the weight space and function space posteriors \citep{burt2020understanding}.

Indeed, we can see this by referring back to the toy example introduced in the previous section.
There, the worst-performing posterior predictive has a parameter-space KL divergence that is in-between the parameter-space KL divergences for the two best-performing cases.
Therefore, if one approximate posterior has a better ELBO than another in the weight space, it is practically impossible to say whether its performance will be better in function space.
This observation has led to recent interest in function-space VI methods \citep{sun2018functional, ma2019variational}, which attempt to directly approximate the posterior predictive, $\pc{f}{\Data}$, where $f$ is the function implied by the parameters of the model.
However, these methods rely on implicit inference, which can have theoretical limitations \citep{mcallester2020formal,burt2020understanding} and do not perform well on simple problems where we can evaluate the true function-space KL divergence \citep{burt2020understanding}.

Despite the limitations of the weight-space ELBO as a metric with which to evaluate VI, the above discussion does point to a way by which we can assess VI algorithms, at least to a limited extent.
We are guaranteed by the data processing inequality \citep[Thm. 7.4 in][]{polyanskiy2022information} that when the KL divergence to the true posterior is zero in weight space, it is also zero in function space.
This suggests that it is not entirely unreasonable to hope that a sufficiently higher weight-space ELBO would lead to a higher function-space ELBO.
However, while there is no guarantee of this, our toy example suggests that we may expect the predictive posterior to suffer in its predictions in cases where the weight-space and function-space KLs are substantially mismatched.
Therefore, we can reasonably hope that an approximate posterior that yields both a better ELBO and better predictive metrics will be the better approximate posterior.
This is especially true if we allow for model or hyperparameter selection: a good approximate posterior should be able to select a better model according to these metrics than a poor one.


Finally, we note that it is popular to evaluate deep Bayesian models by their uncertainty on out-of-distribution data (OOD) and their performance on corrupted data, following the work of \citet{ovadia2019trust}.
While these objectives are certainly important for reliable Bayesian deep learning, we view them as orthogonal targets to the evaluation of VI itself in deep models.
Indeed, there is no reason to expect that the true posterior will be any better at these tasks than an approximate one, even if the model's hyperparameters are well-chosen.
Furthermore, while the marginal likelihood will tend to prefer models that have better uncertainty quantification on out-of-distribution or extrapolated data, as far as we are aware there is no guarantee of this.
Therefore, while evaluating VI for our deep models, we will primarily focus on the following questions:
\begin{enumerate}
    \item How high is the ELBO? and,
    \item How well does the model perform on in-distribution predictive tasks?
\end{enumerate}

\section{Discussion}
In this chapter, we have introduced the background on deep Bayesian models necessary for the remainder of the thesis.
We started by reviewing neural networks and their Bayesian counterparts, discussing different approaches to Bayesian inference in these models.
In doing so, we motivated the variational approach, which we believe has advantages over other approaches: namely, its promise of a combined framework for both uncertainty quantification and model selection.
We will investigate Bayesian inference for neural networks further in Chapters~\ref{sec:dkl} and~\ref{sec:gi}.
In the first of these chapters, we investigate the need for performing inference over all model parameters as opposed to a subset thereof, the latter of which could yield major computational benefits.
In the second, we develop a flexible inducing point variational inference scheme for BNNs.

We continued by discussing DGPs, models that attempt to give GPs the representational learning power of BNNs.
We described different approaches for variational inference in these models, focusing in particular on the work of \citet{salimbeni2017doubly}.
The tie between BNNs and DGPs noted at the beginning of Sec.~\ref{sec:back:dgps} will be crucial in extending the variational approximate posterior we develop for BNNs in Chapter~\ref{sec:gi} to DGPs.
We subsequently discussed the detrimental effect that model symmetries can have in variational inference, describing as well the types of symmetries that deep models are prone to have.
This discussion is the main motivation for Chapter~\ref{sec:dwp}, which attempts to remove the rotational symmetries in DGP models by reframing them as models acting on Gram matrices, as opposed to features.
Finally, we concluded this chapter by discussing how we assess variational inference in deep models in the remainder of this work.


\chapter{Exploring the limitations of partial Bayesian inference in deep models}\label{sec:dkl}

\ifpdf
    \graphicspath{{DKL/Figs/Raster/}{DKL/Figs/PDF/}{DKL/Figs/}}
\else
    \graphicspath{{DKL/Figs/Vector/}{DKL/Figs/}}
\fi

In the previous chapters, we have motivated the use of variational inference for deep models.
However, full variational inference over the potentially millions (if not billions) of parameters in modern neural network architectures is costly.
Ideally, we would like to avoid the need to perform Bayesian inference over this many parameters.
Moreover, BNNs typically require multiple forward passes of the same data for predictions.
Deep kernel learning (DKL) techniques attempt to simplify Bayesian inference for neural networks by replacing the last layer of a neural network with a Gaussian process, and performing inference only on this last layer \citep{calandra2016manifold, wilson2016deep, wilson2016stochastic}.
In performing this style of last-layer only inference, these techniques aim to combine the representational power of neural networks with the reliable uncertainty estimates of Gaussian processes. 
The neural network parameters are trained by treating them as hyperparameters of the Gaussian process and therefore optimizing them with respect to the marginal likelihood.

One crucial aspect of these models is an expectation that, because they are treated as Gaussian process models optimized using the marginal likelihood, they are protected from overfitting. 
We identify situations where this is not the case. 
We explore this behavior, explain its origins and consider how it applies to real datasets. 
Through careful experimentation on the UCI, CIFAR-10, and the UTKFace datasets, we find that the overfitting from using maximum marginal likelihood in overparameterized models (where the overparameterization is with respect to the model hyperparameters), in which the model is ``somewhat Bayesian'', can in certain scenarios be worse than that from not being Bayesian at all. 
We explain how and when DKL can still be successful by investigating optimization dynamics. 
We also find that the failures of DKL can be rectified by a fully Bayesian treatment, which leads to the desired performance improvements over standard neural networks and Gaussian processes.

This chapter is based on joint work with Carl Edward Rasmussen and Mark van der Wilk that was published in \citep*{ober2021promises}.

    
    
    
    


\section{What is deep kernel learning?}
\label{sec:dkl:background}

Gaussian processes and deep neural networks have often been seen to have complementary properties.
On the one hand, Gaussian processes are interpretable models with reliable uncertainty estimates.
However, most popular kernels can only adjust a degree of smoothing, rather than learn sophisticated representations from the data that might aid predictions.
This greatly limits the applicability of GPs to high-dimensional and structured data such as images.
On the other hand, deep neural networks are known to learn powerful representations which are then used to make predictions on unseen test inputs. 
While deterministic neural networks have achieved state-of-the-art performance throughout supervised learning and beyond, they suffer from overconfident predictions \citep{guo2017calibration}, and do not provide reliable uncertainty estimates, as we have already discussed.

It is natural, therefore, to try to combine the uncertainty-representation advantages of GPs with the representation-learning advantages of neural networks, and thus obtain the ``best of both worlds.'' 
Ideally, such an approach would achieve the desiderata of a Bayesian model: training without overfitting, good uncertainty representation, and the ability to learn hyperparameters without using a validation set.
In this chapter, we focus on a line of work that seeks to achieve these desiderata under the name \textit{deep kernel learning} \citep[DKL;][]{calandra2016manifold, wilson2016deep, wilson2016stochastic}.
These works use a neural network to map inputs to points in an intermediate feature space, which is then used as the input space for a GP.
The network parameters can be treated as hyperparameters of the kernel, and thus are optimized with respect to the (log) marginal likelihood, as in standard GP inference.
This leads to an end-to-end training scheme that results in a model that hopefully benefits from the representational power of neural networks while also enjoying the benefits of reliable uncertainty estimation from the GP.
Moreover, as the feature extraction done by the neural network is deterministic, inference only requires one forward pass of the neural net, unlike fully Bayesian BNNs.

In mathematical notation, deep kernel learning maps inputs $\x_n \in \reals^D$ to intermediate values $\v_n \in \reals^Q$ through a neural network $g_\phi(\cdot)$ parameterized by weights and biases $\phi$. 
These intermediate values are then used as inputs to a \emph{base kernel}, which we will choose to be a squared exponential, resulting in the effective kernel $k_\text{DKL}(\x, \x') = k_\text{SE}(g_\phi(\x), g_\phi(\x'))$.
This kernel can then be used as part of a standard Gaussian process regression model: as described in Sec.~\ref{sec:intro:gps},
\begin{equation}
    y_n = f(\x_n) + \epsilon_n, \quad \epsilon_n \sim \N{0, \sigma^2},  \label{eq:dkl:gp-regression}
\end{equation}
where $f\sim \mathcal{GP}(0, k_\text{DKL})$, and we denote its hyperparameters by $\theta \coloneqq \{\phi, \sigma, \sigma_f, \{l_q\}_{q=1}^Q\}$, where $\sigma_f$ is the signal variance of the squared exponential kernel and $\{l_q\}_{q=1}^Q$ its lengthscales.

To the best of our knowledge, \citet{salakhutdinov2007using} were the first to train a DKL-type model, using the marginal likelihood as a training objective to fine tune neural network weights after pretraining using deep belief networks.
\citet{lazaro-gredilla2010marginalized} proposed using mixtures of single-hidden-layer neural networks with last-layer Bayesian linear regression, where the input weights are trained by maximizing the marginal likelihood.
This approach is equivalent to a mixture of multiple ``neural linear'' models \citep{riquelme2018deep, ober2019benchmarking}, where the base kernel for DKL is simply a linear kernel.
\citet{calandra2016manifold} were the first to propose fully end-to-end training of a model with a nonlinear kernel, terming the resulting model the \emph{manifold Gaussian process}, which in an abuse of terminology we will refer to simply as a standard DKL model.\footnote{We note that \citet{wilson2016deep}, which first used the name ``deep kernel learning'', improved scalability on the manifold GP model using KISS-GP \citep{wilson2015kernel}. However, we use ``DKL'' when exact GP inference is used, as this seems to have become standard terminology for all models which transform the inputs to a standard GP using a deterministic neural network.}

Straightforward DKL with exact inference suffers from two major drawbacks. 
First, the $\mathcal{O}(N^3)$ computational cost of GPs causes poor scalability in the number of data.
\citet{wilson2016deep} attempt to address this by using KISS-GP \citep{wilson2015kernel}, and coined the term ``deep kernel learning'' to describe the result.
Second, as discussed in Sec.~\ref{sec:intro:gps}, exact inference is only possible for Gaussian likelihoods, and therefore approximate techniques must be used for classification.
To address both the scalability and non-Gaussian likelihood issues, \citet{wilson2016stochastic} extend the work of \citet{wilson2016deep} by using stochastic variational inference \citep[SVI, ][]{hoffman2013stochastic}.
This model uses a neural network feature extractor, followed by $Q$ independent, single-output GPs with squared exponential kernels, where each GP acts on a single output of the feature extractor.
The outputs of the GPs are then mixed using a matrix $\A$ to create the final output of the model.
As each GP component acts on a single-dimensional input, by placing inducing points on a grid the model can be easily scaled to large datasets using Toeplitz and circulant structure, following the work of \citet{wilson2015thoughts}.

However, we choose to follow \citet{bradshaw2017adversarial} in using a slightly different SVKDL model.
In their model, the neural network maps into a single (potentially multi-output) squared exponential kernel GP, which can act on arbitrary input dimension $Q$ and which maps directly to the output of the model.
In doing so, the model of \citet{bradshaw2017adversarial} returns to the original DKL model.
This difference to the work of \citet{wilson2016stochastic} is facilitated by the use of the stochastic variational inference for GPs \citep{hensman2015scalable} described in Sec.~\ref{sec:intro:gps}, which does not share the practical limits on input dimensionality (for GPs) of \citet{wilson2015thoughts}.
In this case, we optimize the network weights jointly with the hyperparameters by maximizing the ELBO as opposed to the true LML:
\begin{align}
\label{eq:dkl:ELBO}
    \mathcal{L} = \E[q(\F)]{\log \pc{\Y}{\F}} - \KL{\q{\U}}{\p{\U}}.
\end{align}
Note that we allow the model to have multiple outputs as we are usually interested in multiple classes for classification.
In this case, we assume that the kernel is shared between outputs of the GP; however, this assumption could be relaxed if desired.

DKL-related approaches have been increasingly popular for their purported ability to combine the representational power of neural networks with the uncertainty estimates of Gaussian processes.
Indeed, they have been used successfully in wide-ranging downstream tasks such as transfer testing and adversarial robustness \citep{bradshaw2017adversarial}, Bayesian optimization \citep{snoek2015scalable}, reinforcement learning \citep{riquelme2018deep}, and causal prediction for personalized medicine \citep{van2021improving}.

We investigate to what extent DKL is actually able to achieve flexibility and good uncertainty, and what makes it successful in practice: for DKL to be useful from a Bayesian perspective, a higher marginal likelihood should lead to better performance.
In particular, it is often claimed that optimizing the marginal likelihood will automatically calibrate the complexity of the model, preventing overfitting.
For instance, \citet{wilson2016deep} states ``the information capacity of our model grows with the amount of available data, but its complexity is automatically calibrated through the marginal likelihood of the Gaussian process, without the need for regularization or cross-validation.''
This claim is based on the common decomposition of the log marginal likelihood into ``data fit'' and ``complexity penalty'' terms that we observed in~\ref{sec:intro:gps}, which leads to the belief that a better marginal likelihood will result in better test performance.

This is generally true when selecting a small number of hyperparameters. However, in models like DKL with many hyperparameters, we show that marginal likelihood training can encourage overfitting that is \emph{worse} than that from a standard, deterministic neural network.
This is because the marginal likelihood tries to correlate \emph{all} the datapoints, rather than just those for which correlations will be important.
As most standard GP models typically only have a few hyperparameters, this sort of overfitting is not usually an issue, but when many hyperparameters are involved, as in DKL, they can give the model the flexibility to overfit in this way.
As such, our work has implications for all GP methods which use highly parameterized kernels, as well as methods that optimize more than a handful of model parameters according to the marginal likelihood or ELBO.

In this work, we make the following claims:
\begin{itemize}
\itemsep0em
    \item Using the marginal likelihood can lead to overfitting for DKL models.
    \item This overfitting can be worse than the overfitting observed using standard maximum likelihood approaches for neural networks.
    \item The marginal likelihood overfits by overcorrelating the datapoints, as it tries to correlate all the data, not just the points that should be correlated.
    \item Stochastic minibatching can mitigate this overfitting, and helps DKL to work in practice.
    \item A fully Bayesian treatment of deep kernel learning can avoid overfitting and obtain the benefits of both neural networks and Gaussian processes.
\end{itemize}

We note that some works have discussed that overfitting can be an issue for Gaussian processes trained with the marginal likelihood \citep{rasmussen2006gaussian,cawley2010over,lalchand2020approximate}.
Moreover, both \citet{lazaro-gredilla2010marginalized} and \citet{calandra2016manifold} mention that overfitting can be an issue in the context of DKL-related models when there are a large number of network weights.
\citet{tran2019calibrating} showed that these models can also be poorly calibrated, and proposed Monte Carlo dropout \citep{Gal2015DropoutB} to perform approximate Bayesian inference over the neural network weights in the model to fix this.
Additionally, \citet{ober2019benchmarking} showed that it is difficult to get the neural linear model to perform well for regression without considerable hyperparameter tuning, and that fully Bayesian approaches for BNNs often require much less tuning to obtain comparable results.
Recent approaches \citep[e.g., ][]{liu2020simple, van2021improving} carefully regularize the neural network to mitigate these issues, but do still require tuning some hyperparameters on a validation set.
Adding to these works, we explain the undesirable behavior that DKL methods can exhibit, and the mechanism with which the marginal likelihood overfits.

\section{Behavior in a toy problem}
\label{sec:dkl:toy}
\begin{figure*}
    \centering
    \begin{subfigure}[b]{0.49\textwidth}
    \centering
    \centerline{\includegraphics[width=\textwidth]{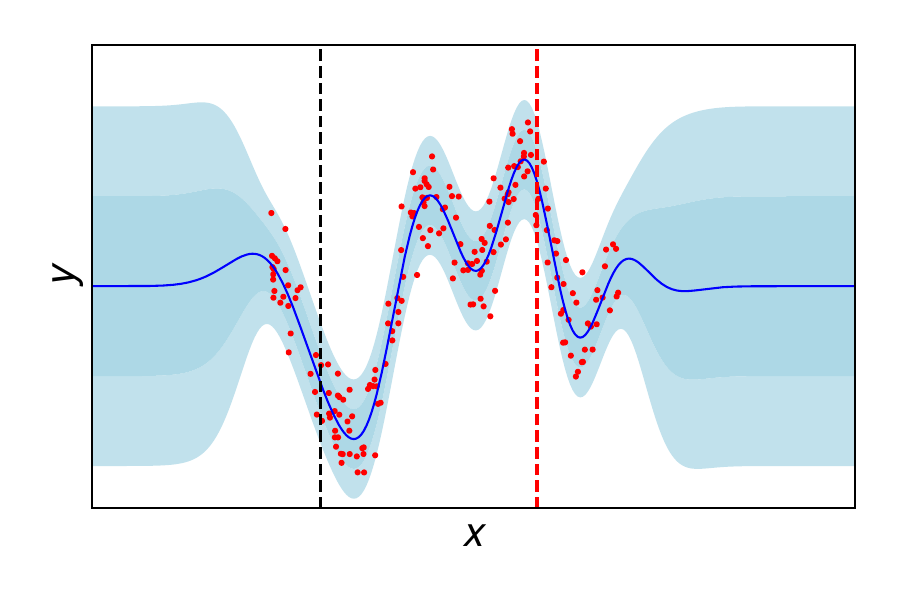}}
    \caption{SE kernel}
    \label{fig:dkl:toy_SE_fit}
    
    \end{subfigure}
    \hfill
    \begin{subfigure}[b]{0.49\textwidth}
    \centering
    \centerline{\includegraphics[width=\textwidth]{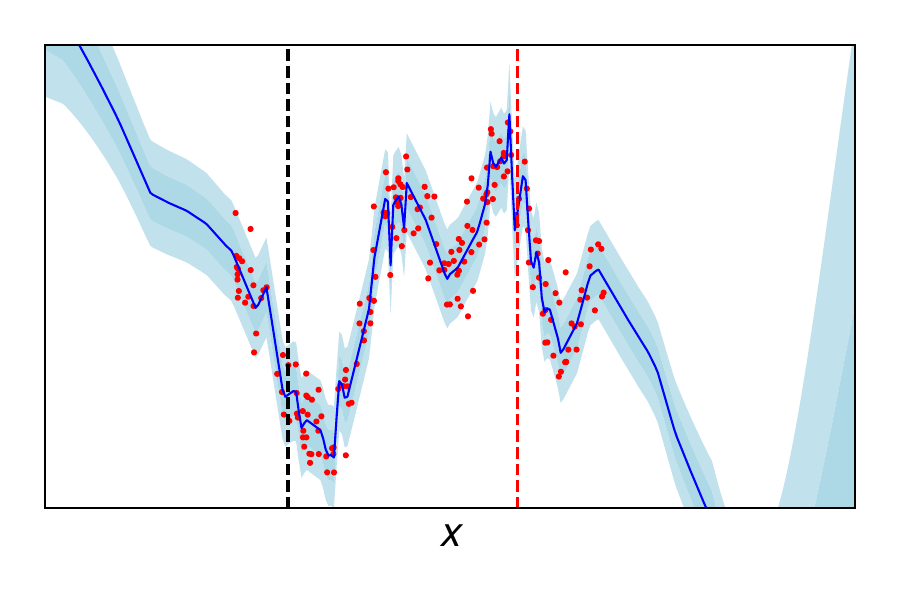}}
    \caption{Exact DKL kernel}
    \label{fig:dkl:toy_DKL_fit}
    
    \end{subfigure}
    \hfill
    \begin{subfigure}[b]{0.49\textwidth}
    \centering
    \centerline{\includegraphics[width=\textwidth]{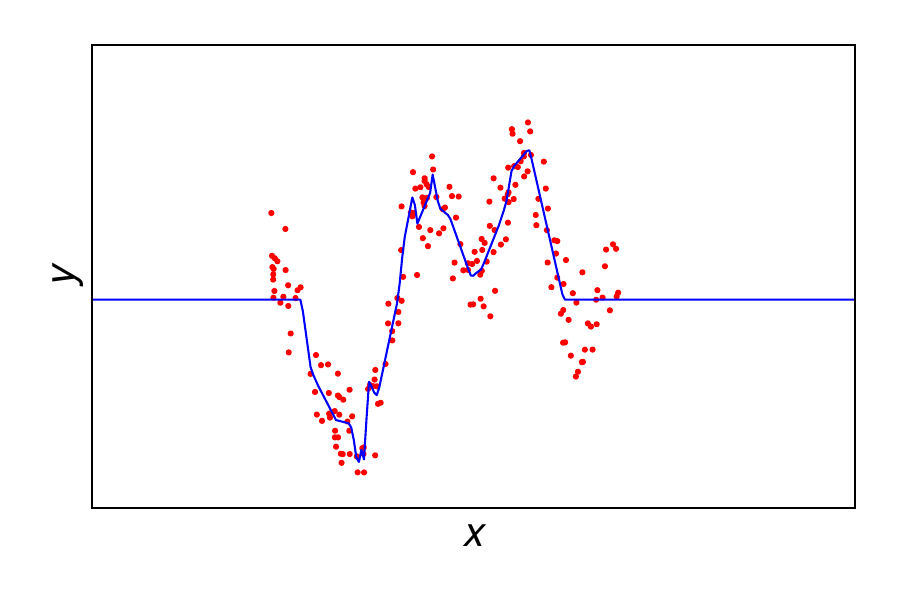}}
    \caption{Neural network fit}
    \label{fig:dkl:toy_NN_fit}
    
    \end{subfigure}
    \hfill
    \begin{subfigure}[b]{0.49\textwidth}
    \centering
    \centerline{\includegraphics[width=\textwidth]{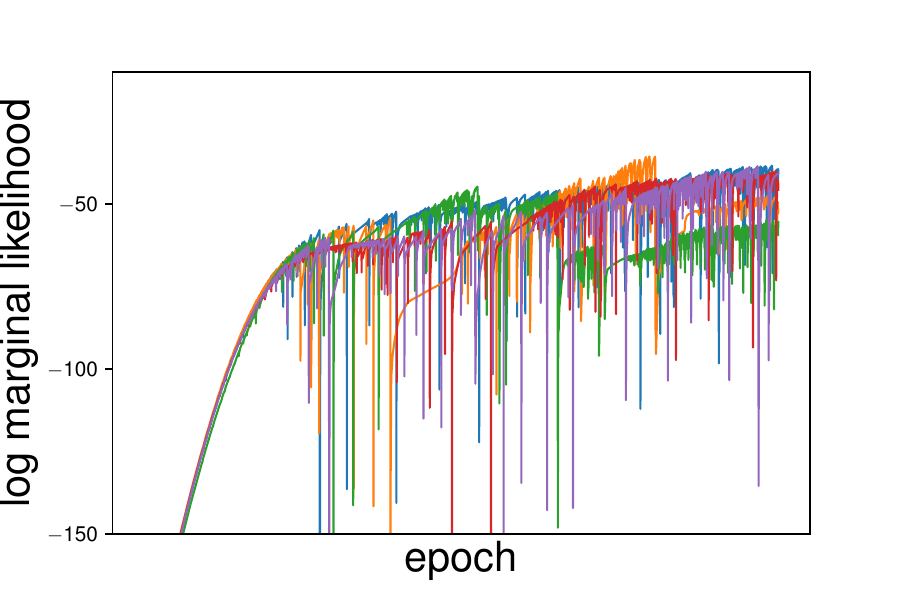}}
    \caption{LML training curves}
    \label{fig:dkl:toy_DKL_traincurves}
    
    \end{subfigure}
    \hfill
    \begin{subfigure}[b]{0.49\textwidth}
    \centering
    \centerline{\includegraphics[width=\textwidth]{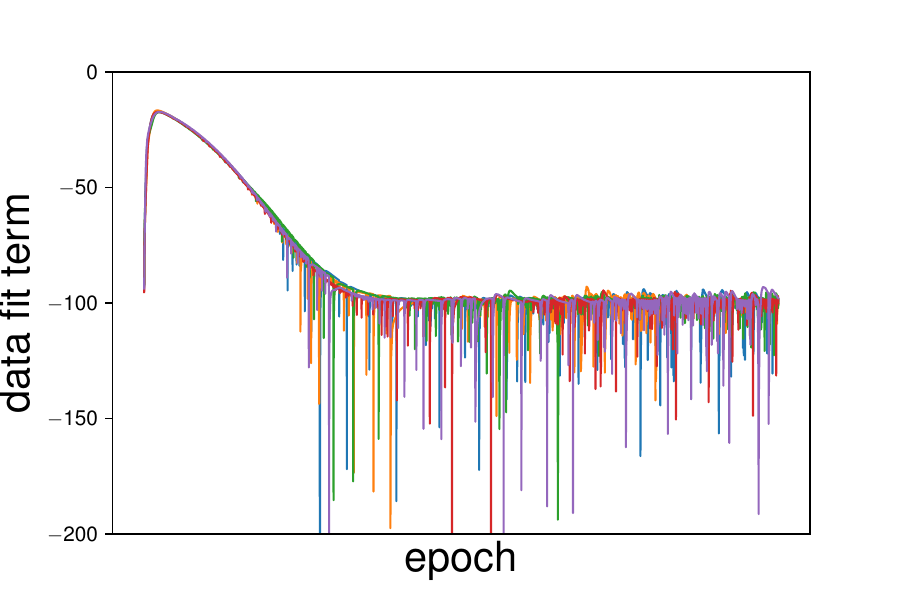}}
    \caption{Data fit}
    \label{fig:dkl:toy_fit_term}
    
    \end{subfigure}
    \hfill
    \begin{subfigure}[b]{0.49\textwidth}
    \centering
    \centerline{\includegraphics[width=\textwidth]{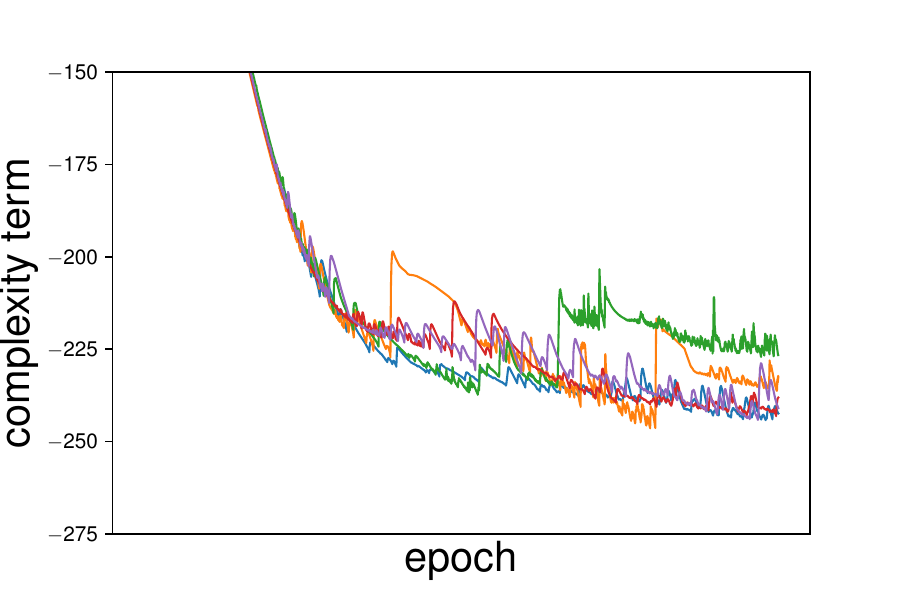}}
    \caption{Complexity penalty}
    \label{fig:dkl:toy_comp_term}
    \end{subfigure}
    
    \caption{Results on toy 1D dataset. Plots (a) and (b) show the predictive posterior for squared exponential (SE) and deep kernel learning (DKL) kernels, respectively. (c) shows the fit given by the neural network analogous to the DKL model. Finally, (d) shows training curves of the log marginal likelihood for 5 different initializations of DKL, with (e) and (f) showing the training curves divided into the LML's data fit and complexity penalty terms, respectively.}
    \label{fig:dkl:toy_problem}

\end{figure*}

To motivate the rest of this chapter, we first consider (exact) DKL in the full toy 1D regression problem from \citet{snelson2006sparse}, with 200 datapoints.
We consider DKL using a two hidden-layer fully-connected ReLU network with layer widths $[100, 50]$ as the feature extractor, letting $Q = 2$ with a squared exponential kernel for the GP.\footnote{We note that this is a smaller feature extractor than that proposed for a dataset of this size in \citet{wilson2016deep}.}
We describe the architecture and experimental details in more detail in App.~\ref{app:dkl:exp-deets}.

We plot the predictive posteriors of both a baseline GP with an SE kernel, and DKL in Figures~\ref{fig:dkl:toy_SE_fit} and \ref{fig:dkl:toy_DKL_fit}, respectively.
We observe that DKL suffers from poor behavior: the fit is very jagged and extrapolates wildly outside the training data.
On the other hand, the fit given by the SE kernel is smooth and fits the data well without any signs of overfitting.
We therefore make the following observation:
\begin{rem}\label{rem:overfitting-exists}
DKL models can be susceptible to overfitting, suggesting that the ``complexity penalty'' of the marginal likelihood may not always prevent overfitting.
\end{rem}

We next compare to the fit given by the deterministic neural network which uses the same feature extractor as the DKL model, so that both models have the same depth.
To ensure a fair comparison, we retain the same training procedure, learning rates, full batch training, and number of optimization steps, so that we only change the model and training loss (from the LML to mean squared error).
We display the fit in Fig.~\ref{fig:dkl:toy_NN_fit}, which shows a nicer fit than the DKL fit of Fig.~\ref{fig:dkl:toy_DKL_fit}: while there is some evidence of overfitting, it is less than that of DKL.
This leads us to our second observation:
\begin{rem}\label{rem:worse-overfitting}
DKL can exhibit \emph{worse} overfitting than a standard neural network trained using maximum likelihood.
\end{rem}

We next plot training curves from five different runs of DKL in Fig.~\ref{fig:dkl:toy_DKL_traincurves}.
From these, we observe that training is very unstable, with many significant spikes in the marginal likelihood objective.
We found that reducing the learning rate does improve stability, but only slightly (App.~\ref{app:dkl:add_exp}).
We also observe that runs often end up settling in different locations with different final values of the log marginal likelihood. 
We plot different fits from different initializations in Fig.~\ref{fig:dkl:app-toy} in the Appendix, showing that these different local minima give very different fits with different generalization properties.

In general, this behavior is concerning: one would hope that adding a Bayesian layer to a deterministic network would improve performance, as introducing Bayesian principles is often touted as a method to reduce overfitting \citep[e.g., ][]{osawa2019practical}. 
However, this toy problem illustrates that performance can even deteriorate upon augmenting the model with a Bayesian output layer.
As this finding is seemingly at conflict with most of the literature, which has found that DKL or variations thereof can be useful, we devote the rest of this chapter to understanding when and why this pathology arises, including for real datasets.

\section{Understanding the pathology}
\label{sec:dkl:understanding}
\subsection{Regression}
To help understand the observed pathological behavior, we first look at the curves of the ``data fit'' and ``complexity penalties'' for five different initializations on the toy dataset. 
We present these curves in Fig.~\ref{fig:dkl:toy_fit_term} and \ref{fig:dkl:toy_comp_term}. We note that each of the data fit curves largely stabilize around -100 nats, so that the complexity terms seem to account for most of the differences in the final marginal likelihood (Fig.~\ref{fig:dkl:toy_DKL_traincurves}). 
This behavior is explained by the following proposition, which states that the data fit term becomes uninteresting for any GPs with learnable signal variance trained on the marginal likelihood.

\begin{restatable}[]{prop}{lmlprop}\label{prop:lml}
Consider the GP regression model as described in Eq.~\ref{eq:dkl:gp-regression}. 
Then, for any valid kernel function that can be written in the form $k(\x, \x') = \sigma_f^2 \hat{k}(\x, \x')$, where $\sigma_f^2$ is a learnable hyperparameter along with learnable noise $\sigma^2$ (and any other kernel hyperparameters), we have that the ``data fit'' term will equal $-N/2$ (where $N$ is the number of datapoints) at the optimum of the marginal likelihood.
\end{restatable}

The proof is simply achieved by differentiation with respect to $\sigma_f^2$:
\begin{proof}
We reparameterize $\sigma^2 = \hat{\sigma}^2\sigma_f^2$. Then, writing $\K_{\X, \X} + \sigma_n^2 \I_N = \sigma_f^2(\hat{\K}_{\X, \X} + \hat{\sigma}^2 \I_N)$, the result follows by differentiating the log marginal likelihood with respect to $\sigma_f^2$:
\begin{align*}
    \frac{d}{d\sigma_f^2} \log \p{\y} &= \frac{d}{d\sigma_f^2} \left(-\frac{N}{2}\log \sigma_f^2 -\frac{1}{2}\log |\hat{\K}_{\X, \X}+\hat{\sigma}^2 \I_N| - \frac{1}{2\sigma_f^2}\transpose{\y}(\hat{\K}_{\X, \X} + \hat{\sigma}^2 \I_N)^{-1}\y\right)\\
    &= -\frac{N}{2\sigma_f^2} + \frac{1}{2\sigma_f^4}\transpose{\y}(\hat{\K}_{\X, \X}+\hat{\sigma}^2 \I_N)^{-1}\y.
\end{align*}
Setting the derivative equal to zero gives:
\begin{equation*}
    \sigma_f^2 = \frac{1}{N}\transpose{\y}(\hat{\K}_{\X, \X} + \hat{\sigma}^2 \I_N)^{-1}\y.
\end{equation*}
Substituting this into the data fit term gives the desired result.
\end{proof}

\begin{figure*}
    \centering
    \begin{subfigure}[b]{0.49\textwidth}
    \centering
    \centerline{\includegraphics[width=\textwidth]{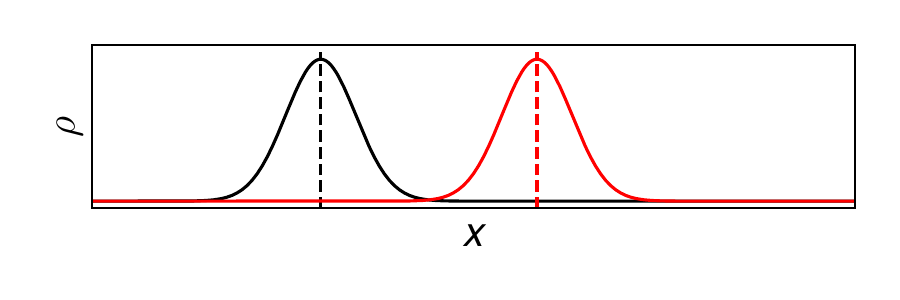}}
    \caption{SE kernel}
    \label{fig:dkl:toy_SE_corr}
    
    \end{subfigure}
    \hfill
    \begin{subfigure}[b]{0.49\textwidth}
    \centering
    \centerline{\includegraphics[width=\textwidth]{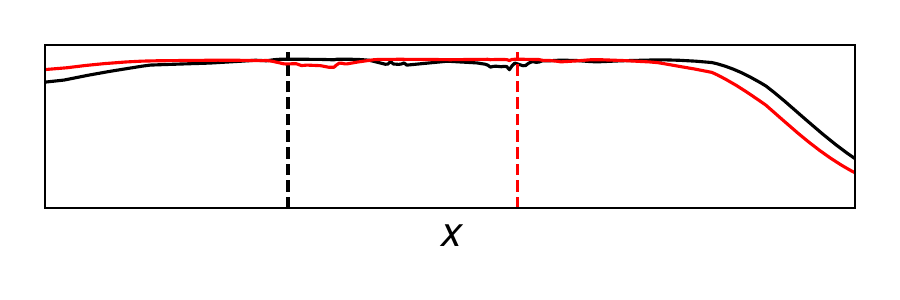}}
    \caption{Exact DKL kernel}
    \label{fig:dkl:toy_DKL_corr}
    \end{subfigure}
    
    \caption{Correlation functions $\rho_{x'}(x) = k(x, x')/\sigma_f^2$ at two points $x'$ given by the vertical dashed lines in Fig.~\ref{fig:dkl:toy_problem}.}
    \label{fig:dkl:toy_problem_corr}

\end{figure*}

We note that this result was essentially proven by \citet{moore2016fast}, although they did not consider the last step of substituting the result into the data fit term.
Instead, they used the result as a means of analytically solving for the optimal signal variance to reduce the number of parameters and hence speed up optimization.
This result is far-reaching, as the use of a learnable signal variance $\sigma_f^2$ is almost universal.
This proposition therefore implies that after training the complexity penalty is responsible for any difference in marginal likelihood for GPs with different kernels.
Recall that the complexity penalty is given by
\begin{equation}
    \label{eq:dkl:complexity}
    \frac{1}{2}\log |\K_{\X, \X} + \sigma^2 \I_N| = \frac{N}{2}\log \sigma_f^2 + \frac{1}{2}\log |\hat{\K}_{\X, \X} + \hat{\sigma}^2 \I_N|.
\end{equation}
Maximizing the marginal likelihood encourages this term to be minimized, which can be done in at least two ways: minimizing $\sigma_f$, or minimizing the $\log |\hat{\K}_{\X, \X} + \hat{\sigma}^2 \I_N|$ term. 
However, there is little freedom in minimizing $\sigma_f$, because that would compromise the data fit. 
Therefore, the main mechanism for minimizing the complexity penalty would be through minimizing the second term. 
One way of doing this is to correlate the input points as much as possible: if there are enough degrees of freedom in the kernel, it is possible to ``hack'' the Gram matrix so that it can do this while minimizing the impact on the data fit term. 
We see this by looking at the correlation plots for the previous SE and DKL fits: in Fig.~\ref{fig:dkl:toy_problem_corr} we have plotted correlation functions $\rho_{x'}(x) = k(x, x')/\sigma_f^2$ at the two points $x'$ given by the vertical dashed lines from Fig.~\ref{fig:dkl:toy_problem}. 
We see that while Fig.~\ref{fig:dkl:toy_SE_corr} shows the expected Gaussian bump for the SE kernel, Fig.~\ref{fig:dkl:toy_DKL_corr} shows near-unity correlation functions for all values. 
Furthermore, in Appendix~\ref{app:dkl:add_exp} we show empirically that for fits that do not show as much correlation, the final marginal likelihood is worse (with the overfitting arguably less pronounced), suggesting that increasing the correlation is indeed the main mechanism by which the model increases its marginal likelihood.
We note that one mechanism of correlating all the datapoints has concurrently been explored and termed ``feature collapse'' \citep{van2021improving}, where the neural network feature extractor learns to collapse all the datapoints onto a low-dimensional surface.
We summarize our findings in the remark:

\begin{rem}\label{rem:correlation}
The complexity penalty encourages high correlation between different points. 
Overparameterizing the  covariance function can lead to pathological results, as it allows all points to be correlated in the prior, not only the points where we would expect correlations to appear.
\end{rem}

\subsection{Classification}\label{sec:dkl:mnist}
\begin{figure}
    \centering
    \includegraphics[width=\textwidth]{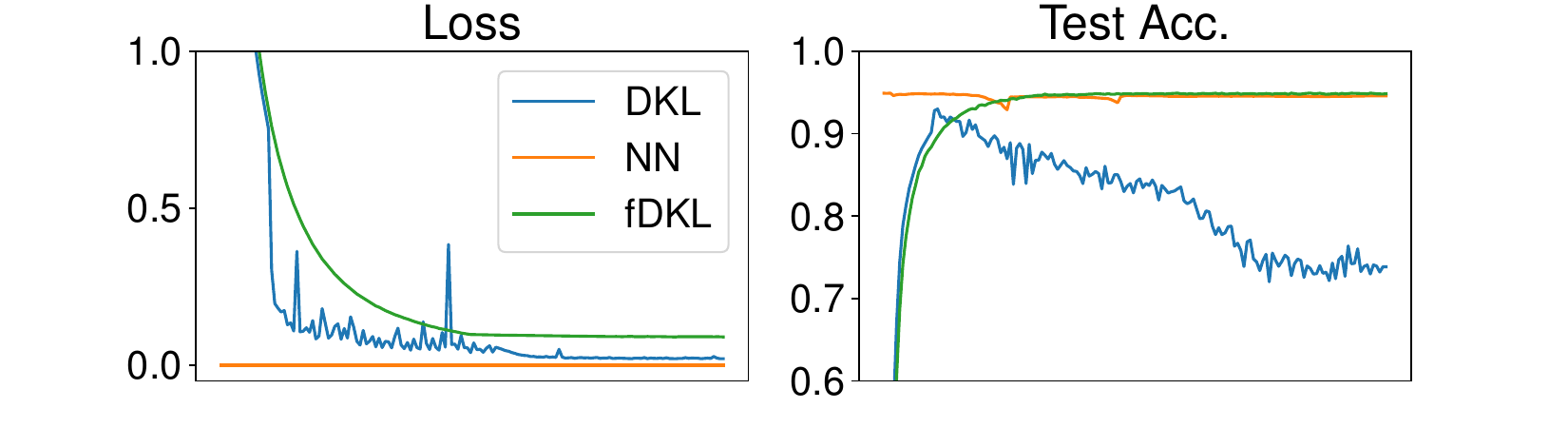}
    \caption{Training curves of both the losses and test accuracy for a 5k subset of MNIST, using a pretrained neural network as starting points for each of DKL, a neural network, and ``fixed network DKL'' (fDKL).}
    \label{fig:dkl:mnist}
\end{figure}
We now briefly consider a simple classification example. 
We compare a neural network (NN) using the usual softmax loss with a DKL model. 
Due to the non-Gaussian likelihood, we use the variational approximation to the marginal likelihood (Eq.\ref{eq:dkl:ELBO}). 
We also compare a DKL model with the NN feature extractor fixed to what is obtained from the normal NN training procedure, which we refer to as ``fixed network DKL'' (fDKL).
All models are initialized with the same pretrained neural network using standard NN training for a fair comparison.
Fig.~\ref{fig:dkl:mnist} shows the training curves for the losses and test accuracies on a subset of 5000 points of MNIST. 
All models are trained with full batches (see App.~\ref{app:dkl:exp-deets} for additional details). 
We observe that the standard NN has a near-zero loss without worsening test accuracy. 
DKL also attains low loss but significantly overfits. 
By contrast, the loss obtained by fDKL is the highest of the three models, but fDKL does not overfit, and achieves the best test accuracy by a small margin.

We can explain the results in a similar way to regression, even though the SVDKL loss is different. 
In this case, the expected log-likelihood measures the data fit, while the KL enforces simplicity of the prior and approximate posterior.
Indeed, the KL contains a $\frac{1}{2}\log|\K_{\Z, \Z}|$ term, which can be viewed analogously to the $\frac{1}{2} \log |\K_{\X, \X} + \sigma^2 \I_N|$ term in the LML.
Since the MNIST classes are well-separated, we expect a near-zero data fit term. 
In the standard NN training loss, there is little encouragement to overfit, since a well-fitted model already achieves a loss close to the global minimum of zero. 
The DKL objective, on the other hand, contains the complexity penalty which can be further reduced by over-correlating points, just as in regression.

We now investigate how these observations relate to real, complex datasets, as well as to the prior literature which has shown that DKL can obtain good results.

\section{DKL for real datasets}
\label{sec:dkl:realdata}
Despite our previous findings, multiple works have shown that DKL methods can perform well in practice \citep[e.g., ][]{wilson2016stochastic, bradshaw2017adversarial}.
We now consider experiments on various datasets and architectures to further investigate the observed pathological behavior and how DKL succeeds.
We provide full experimental details in Appendix~\ref{app:dkl:exp-deets} and additional experimental results in Appendix~\ref{app:dkl:add_exp}

\subsection{DKL for UCI Regression}
\begin{figure*}
    \centering
    \includegraphics[width=\linewidth]{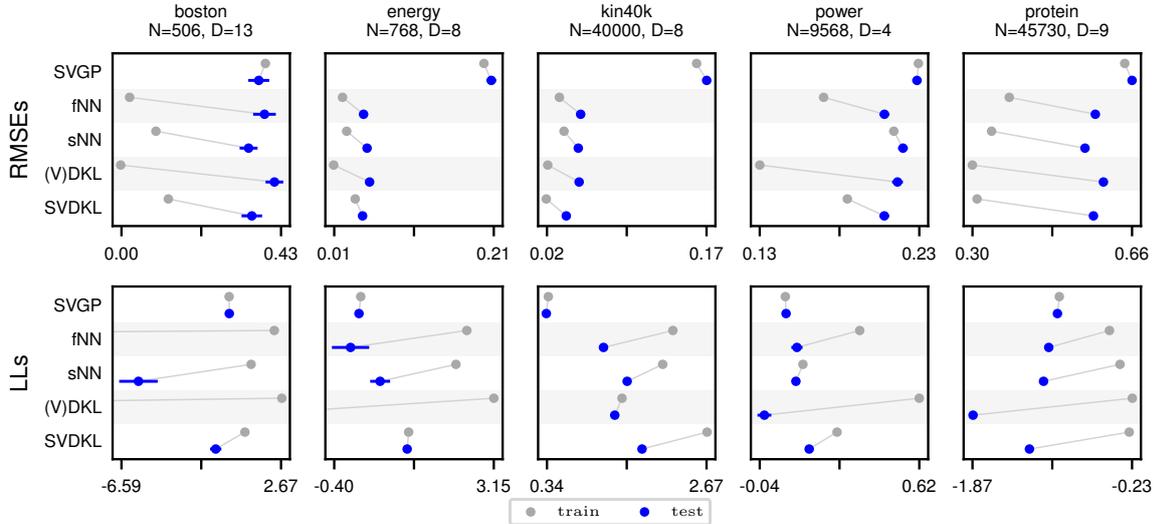}

    \caption{Results for the UCI datasets. We report train and test RMSEs and log likelihoods (LLs) for each method, averaged over the 20 splits. Further left is better for RMSEs; right is better for LLs. Error bars represent one standard error.}
    \label{fig:dkl:uci_plots}

\end{figure*}

We first consider DKL applied to a selection of regression datasets from the UCI repository \citep{Dua:2019}: \textsc{Boston}, \textsc{Energy}, \textsc{Kin40K}, \textsc{Power}, \textsc{Protein}.
These represent a range of different sizes and dimensions: \textsc{Energy}, \textsc{Power}, and \textsc{Protein} were chosen specifically because we expect that they can benefit from the added depth to a GP \citep{salimbeni2017doubly}.

We consider a range of different models, and we report train and test root mean square errors (RMSEs) and log likelihoods (LLs) in Fig.~\ref{fig:dkl:uci_plots}, and tabulate the log marginal likelihoods or ELBOs in Table~\ref{tab:dkl:uci_elbos}.
First, we consider a baseline stochastic variational GP (SVGP) model with an ARD SE kernel.
As this is a GP model with few hyperparameters, we would not expect significant differences between training and testing performances. 
Indeed, looking at Fig.~\ref{fig:dkl:uci_plots}, this is exactly what we observe: the test performance is comparable to, and sometimes even slightly better than, the training performance for both RMSEs and LLs.

We compare to a neural network trained with mean squared error loss and DKL using the same neural network architecture for feature extractor (so that the depths are equal).
We first consider DKL models where we use full-batch training, compared to a neural network with full-batch training, which we refer to as fNN.
As full-batch training for DKL is expensive for larger datasets, for the \textsc{Kin40K}, \textsc{Power}, and \textsc{Protein} we instead use SVDKL trained with 1000 inducing points but full training batches, which we term \emph{variational} DKL (VDKL).
For both methods we use a small weight decay to help reduce overfitting, and we use the same number of gradient steps for each to ensure a fair comparison.
Looking at the results for fNN and (V)DKL in Fig.~\ref{fig:dkl:uci_plots}, we see that both of these methods overfit quite drastically.
This mirrors our observations in Remark~\ref{rem:overfitting-exists} that DKL models can be susceptible to overfitting.
In most cases the overfitting is noticeably worse for (V)DKL than it is for fNN, reflecting our observation in Remark~\ref{rem:worse-overfitting}.
This is particularly concerning for the log likelihoods, as one would hope that the ability of DKL to express epistemic uncertainty through the last-layer GP would give it a major advantage over the neural network, which cannot do so.

\begin{table}
\small
    \centering
    \caption{LMLs/ELBOs per datapoint for UCI datasets. We highlight the best results (taking error bars into account) in bold.}\label{tab:dkl:uci_elbos}
    \begin{tabular}{rccc}
\toprule 
& SVGP & (V)DKL & SVDKL \\ 
\midrule 
\textsc{Boston}& -1.66 $\pm$ 0.06 & $\mathbf{2.47 \pm 0.00}$ & 0.47 $\pm$ 0.01 \\ 
\textsc{Energy}& -0.07 $\pm$ 0.01 & $\mathbf{3.01 \pm 0.02}$ & 1.21 $\pm$ 0.00 \\ 
\textsc{Kin40K}& 0.14 $\pm$ 0.00 & 1.41 $\pm$ 0.00 & $\mathbf{2.62 \pm 0.00}$ \\ 
\textsc{Power}& 0.01 $\pm$ 0.00 & $\mathbf{0.57 \pm 0.00}$ & 0.25 $\pm$ 0.00 \\
\textsc{Protein}& -1.06 $\pm$ 0.00 & -$\mathbf{0.32 \pm 0.01}$ & -0.35 $\pm$ 0.00 \\ 
\bottomrule
\end{tabular}
\end{table}

In practice, however, many approaches for DKL and neural networks alike make use of stochastic minibatching during training.
In fact, it is well-known that minibatch training induces implicit regularization for neural networks that helps generalization \citep{keskar2016large}.
We therefore investigate this for both DKL and neural networks: we refer to the stochastic minibatched network as sNN and compare to SVDKL, using the same batch sizes for both.
Referring again to Fig.~\ref{fig:dkl:uci_plots}, we see that minibatching generally reduces overfitting compared to the full-batch versions, for both model types.
However, the difference between the full batch and stochastic minibatch performances of DKL seem to be greater than the corresponding differences for the standard neural networks, suggesting that the implicit regularization effect is stronger.
The exception to this trend is \textsc{Kin40K}, which appears to be low-noise and simple for a deep model to predict for.
We also note that with the exception of \textsc{Protein}, SVDKL now performs the best of the deep models in terms of log likelihoods, and generally performs better than SVGP.

Finally, we consider Table~\ref{tab:dkl:uci_elbos}, which shows the ELBOs/LMLs for each of the GP methods.
SVGP has by far the worse ELBOs, whereas (V)DKL generally has by far the best.
It is important to note that the ELBOs for SVDKL are worse than those for (V)DKL despite its generally better test performance.
This suggests that improving the marginal likelihood for DKL models does not improve test performance, as one would desire for a Bayesian model.
We summarize our findings in the following remark:
\begin{rem}\label{rem:minibatching}
The reason for DKL's successful performance is not an improved marginal likelihood, but rather that stochastic minibatching provides implicit regularization that protects against overfitting with the marginal likelihood.
\end{rem}

Therefore, we observe again that the Bayesian benefits of the marginal likelihood do not apply in the overparameterized regime: indeed, we find that using the marginal likelihood can be worse than not being Bayesian at all.

\subsection{DKL for image datasets}\label{sec:dkl:image-datasets}

We now explore how these findings relate to high-dimensional, highly structured image datasets.
We might expect that the benefits of DKL would be stronger for images than in the previous regression datasets, as the design of kernels for these high-dimensional spaces remains an open question despite numerous recent advances \citep{van2017convolutional, dutordoir2020bayesian}, and neural networks generally perform far better than kernel methods.

We first consider a regression problem using image inputs: an age regression task using the UTKFace dataset \citep{zhifei2017cvpr}.
The dataset consists of 23,708 images of aligned and cropped faces of size $200\times200\times3$.
These images are annotated with age, gender and race --- we focus on predicting age.\footnote{We note that ethical issues arise in the use of a dataset with these attributes. While we do not directly address ethical concerns in this work, we strongly encourage researchers and practitioners to consider the ethical issues inherent in the creation and use of such datasets.} 
We consider models based on a ResNet-18 \citep{he2016deep}: we take the standard ResNet-18 with 10-dimensional output, to which we add a ReLU nonlinearity and then either a linear output layer or an ARD SE GP, corresponding to the baseline neural network and SVDKL, respectively.
We consider different feature widths $Q$ in Appendix~\ref{app:dkl:fd}.
This construction ensures that both models have the same depth, so that any differences in performance cannot be attributed to the fact that the models have different depths.
We consider the baseline neural network (NN) and SVDKL models.
Additionally, as both \citet{wilson2016stochastic} and \citet{bradshaw2017adversarial} use pretraining followed by finetuning for their models, we compare to this as well.
In our implementation, we take the trained baseline NNs, and fixing the neural networks first learn the variational parameters and GP hyperparameters.
We refer to the result as the fixed net SVDKL (fSVDKL) model.
For finetuning, we then train everything jointly for a number of epochs, resulting in the pretrained SVKDL (pSVDKL) model.
Finally, so that any improvement for f/pSVDKL is not just from additional gradient steps, we also further train the neural networks for the same number of epochs, resulting in the pretrained NN (pNN) model.
We average all results over 3 independent runs using a batch size of 100, and we refer the reader to App.~\ref{app:dkl:exp-deets} for full experimental details.

We report ELBOs, train and test RMSEs, and train and test log likelihoods for the normalized data in the top left portion of Table~\ref{tab:dkl:batch} (batch size 100).
We see that SVDKL, the method without pretraining, obtains lower ELBOs than either fSVDKL or pSVDKL, which obtain broadly similar ELBOs.
We suspect that this is because of the difficulty in training large DKL models from scratch, as noted in \citet{bradshaw2017adversarial}; this is also consistent with our earlier observation that training can be very unstable.
We see that each method, except fSVDKL (with the fixed pretrained network), achieves similar train RMSE. 
However, the test RMSEs are significantly worse for each method, with fSVDKL obtaining the best.
Unsurprisingly, the NN models perform poorly in terms of LL, as they are unable to express epistemic uncertainty, with additional training (pNN) worsening the test performance on both metrics.
pSVDKL (SVDKL with pretraining and finetuning) obtains the best test LL of all methods, as well as better test RMSE than the neural networks, showing that SVDKL can yield improvements consistent with the prior literature.
We note, however, that there is still a substantial gap between train and test performance, indicating overfitting in a way consistent with Remark~\ref{rem:overfitting-exists}.

\subsubsection{Increasing the batch size}
From our UCI experiments, we hypothesized that implicit regularization from minibatch noise was key in obtaining good performance for SVDKL (Remark~\ref{rem:minibatching}). 
We therefore consider increasing the batch size from 100 to 200 for the pretrained methods, keeping the pretrained neural networks the same (Table~\ref{tab:dkl:batch}, top right).
We make a few key observations.
First, this leads to a significantly improved ELBO for pSVDKL, which ends up helping the test RMSE.
However, we see that instead of improving the test LL, it becomes significantly worse, whereas the train LL becomes better: clear evidence of overfitting.
Moreover, fSVDKL, where the network is kept fixed, now outperforms pSVDKL, which has a better ELBO.
Finally, we note that the behavior of pNN does not change significantly, in fact slightly improving with increased batch size: this suggests that the implicit regularization from minibatching is stronger for SVDKL than for standard NNs.
All of these observations are consistent with our findings surrounding Remark~\ref{rem:minibatching}, which argues that stochastic minibatching is crucial to the success of DKL methods, and a better marginal likelihood is associated with worse performance.

\subsubsection{Image classification}
Our analysis in Section~\ref{sec:dkl:understanding} only applies directly to regression, although we were able to show that we can find similar trends in a small classification problem (c.f. Sec.~\ref{sec:dkl:mnist}). 
As one of the main successes of current deep learning is in classification, it is therefore natural to wonder whether the trends we have observed also apply to large-scale classification tasks.
We consider CIFAR-10 \citep{krizhevsky2009learning}, a popular dataset of $32\times32\times3$ images belonging to one of 10 classes.
We again consider a modified ResNet-18 model, in which we have ensured that the depths remain the same between NN and DKL models.
We consider training the models with batch sizes of 100 and 500.
We look at ELBOs, accuracies, and LLs, as well as the LL for incorrectly classified test points, which can indicate overconfidence in predicting wrongly.
We also look at expected calibration error \citep[ECE;][]{guo2017calibration}, a popular metric evaluating model calibration: results are shown in the lower portion of Table~\ref{tab:dkl:batch}.
Here, we see that simple SVDKL struggles even more to fit well, indicating the importance of pretraining.
For the batch size 100 experiments, pSVDKL generally performs the best, reflecting the experience of \citet{wilson2016stochastic} and \citet{bradshaw2017adversarial}.
However, we again observe that increasing the batch size hurts pSVDKL, and fSVKDL outperforms it despite worse ELBOs.

\subsection{Data augmentation}
It is common practice with image datasets to perform data augmentation, which effectively increases the size of the training dataset by using modified versions of the images.
Bayesian inference does not strictly permit this, instead requiring that the model be adjusted \citep{vdw2018inv,nabarro2021}.
We briefly consider whether this changes the overfitting behavior we observed, by repeating the same experiments (without plain SVDKL, as it struggles to fit) with random cropping and horizontal flipping augmentations; see Table~\ref{tab:dkl:batch_da}.
Overall, we once again find that increasing the batch size still significantly hurts the performance of pSVDKL: whereas pSVDKL outperforms the fixed-network version for batch size 100, larger batch sizes reverse this, so that finetuning the network according to the ELBO hurts, rather than helps, performance.
Therefore, in this case, using last-layer Bayesian inference is worse than not being Bayesian at all.
These results reflect our findings in the previous remarks that using the marginal likelihood can be worse than using a standard likelihood, and that stochastic minibatching is one of the main reasons that DKL can be successful.

\begin{landscape}
\begin{table*}
\scriptsize
    \centering
    \caption{Results for the UTKFace age regression task and CIFAR-10 classification, without data augmentation. We report means plus/minus one standard error, averaged over three runs.}\label{tab:dkl:batch}
    \begin{tabular}{rcccccccc}
\toprule 
& \multicolumn{5}{c}{Batch size: 100} & \multicolumn{3}{c}{Batch size: 200/500} \\ 
\cmidrule(lr){2-6}
\cmidrule(lr){7-9}
& NN & SVDKL & pNN & fSVDKL & pSVDKL & pNN & fSVDKL & pSVDKL \\ 
\midrule
UTKFace - ELBO & - & 0.92 $\pm$ 0.01 & - & 1.05 $\pm$ 0.30 & 1.03 $\pm$ 0.10 & - & 0.75 $\pm$ 0.34 & 1.43$\pm$0.04 \\ 
Train RMSE & 0.04$\pm$0.00 & 0.04$\pm$0.00 & 0.04$\pm$0.00 & 0.08 $\pm$ 0.03 & 0.04$\pm$0.00 & 0.04 $\pm$ 0.00 & 0.12 $\pm$ 0.03 & 0.04 $\pm$ 0.00 \\ 
Test RMSE & 0.40 $\pm$ 0.00 & 0.40 $\pm$ 0.01 & 0.41 $\pm$ 0.00 & 0.31 $\pm$ 0.07 & 0.38 $\pm$ 0.02 & 0.39 $\pm$ 0.01 & 0.23 $\pm$ 0.07 & 0.34 $\pm$ 0.02 \\ 
Train LL & 1.81 $\pm$ 0.01 & 1.30 $\pm$ 0.01 & 1.83 $\pm$ 0.01 & 1.16 $\pm$ 0.31 & 1.20 $\pm$ 0.08 & 1.83 $\pm$ 0.01 & 0.82 $\pm$ 0.34 & 1.60 $\pm$ 0.03 \\ 
Test LL & -48.73 $\pm$ 1.64 & -6.88 $\pm$ 0.38 & -53.72 $\pm$ 1.71 & -7.55 $\pm$ 3.42 & -4.74 $\pm$ 1.35 & -48.48 $\pm$ 2.07 & -5.36 $\pm$ 4.78 & -10.43 $\pm$ 2.94 \\
\midrule
CIFAR-10 - ELBO & - & -0.76 $\pm$ 0.28 & - & -0.02 $\pm$ 0.00 & -0.00 $\pm$ 0.00 & - & -0.02 $\pm$ 0.00 & -0.00 $\pm$ 0.00 \\ 
Train Acc. & 1.00 $\pm$ 0.00 & 0.76 $\pm$ 0.09 & 1.00 $\pm$ 0.00 & 1.00 $\pm$ 0.00 & 1.00 $\pm$ 0.00 & 1.00 $\pm$ 0.00 & 1.00 $\pm$ 0.00 & 1.00 $\pm$ 0.00 \\ 
Test Acc. & 0.79 $\pm$ 0.00 & 0.63 $\pm$ 0.03 & 0.79 $\pm$ 0.00 & 0.78 $\pm$ 0.00 & 0.79 $\pm$ 0.00 & 0.79 $\pm$ 0.00 & 0.79 $\pm$ 0.00 & 0.79 $\pm$ 0.00 \\ 
Train LL & -0.00 $\pm$ 0.00 & -0.71 $\pm$ 0.28 & -0.00 $\pm$ 0.00 & -0.01 $\pm$ 0.00 & -0.00 $\pm$ 0.00 & -0.00 $\pm$ 0.00 & -0.00 $\pm$ 0.00 & -0.00 $\pm$ 0.00 \\ 
Test LL & -2.05 $\pm$ 0.03 & -1.37 $\pm$ 0.10 & -2.30 $\pm$ 0.11 & -1.14 $\pm$ 0.00 & -1.13 $\pm$ 0.01 & -2.88 $\pm$ 0.04 & -1.07 $\pm$ 0.01 & -1.45 $\pm$ 0.00 \\ 
Inc. Test LL & -8.87 $\pm$ 0.10 & -3.38 $\pm$ 0.77 & -9.48 $\pm$ 0.30 & -5.10 $\pm$ 0.01 & -5.24 $\pm$ 0.05 & -10.77 $\pm$ 0.07 & -4.73 $\pm$ 0.03 & -6.63 $\pm$ 0.03 \\ 
ECE & 0.18 $\pm$ 0.00 & 0.10 $\pm$ 0.05 & 0.19 $\pm$ 0.00 & 0.14 $\pm$ 0.00 & 0.15 $\pm$ 0.00 & 0.19 $\pm$ 0.00 & 0.13 $\pm$ 0.00 & 0.15 $\pm$ 0.00 \\ 
\bottomrule 
    \end{tabular}
\end{table*}
\begin{table*}
\scriptsize
    \centering
    \caption{Results for the image datasets with data augmentation. We report means $\pm1$ standard error, averaged over 3 runs.}\label{tab:dkl:batch_da}
    \begin{tabular}{rcccccccc}
\toprule 
& \multicolumn{4}{c}{Batch size: 100} & \multicolumn{3}{c}{Batch size: 200 (UTKFace) / 500 (CIFAR-10)} \\ 
\cmidrule(lr){2-5}
\cmidrule(lr){6-8}
& NN & pNN & fSVDKL & pSVDKL & pNN & fSVDKL & pSVDKL \\ 
\midrule 
UTKFace - ELBO & - & - & 0.16 $\pm$ 0.03 & 0.14 $\pm$ 0.03 & - & 0.12 $\pm$ 0.06 & 0.45 $\pm$ 0.03 \\ 
Train RMSE & 0.19 $\pm$ 0.01 & 0.18 $\pm$ 0.00 & 0.19 $\pm$ 0.00 & 0.17 $\pm$ 0.01 & 0.13 $\pm$ 0.00 & 0.20 $\pm$ 0.01 & 0.12 $\pm$ 0.01 \\ 
Test RMSE & 0.36 $\pm$ 0.00 & 0.36 $\pm$ 0.00 & 0.36 $\pm$ 0.00 & 0.35 $\pm$ 0.00 & 0.35 $\pm$ 0.00 & 0.31 $\pm$ 0.04 & 0.35 $\pm$ 0.01 \\ 
Train LL & 0.25 $\pm$ 0.03 & 0.31 $\pm$ 0.01 & 0.25 $\pm$ 0.03 & 0.30 $\pm$ 0.03 & 0.65 $\pm$ 0.02 & 0.20 $\pm$ 0.06 & 0.63 $\pm$ 0.04 \\ 
Test LL & -1.03 $\pm$ 0.07 & -1.22 $\pm$ 0.05 & -0.92 $\pm$ 0.07 & -0.76 $\pm$ 0.03 & -2.72 $\pm$ 0.21 & -0.63 $\pm$ 0.30 & -1.55 $\pm$ 0.17 \\ 
\midrule
CIFAR-10 - ELBO & - & - & -0.07 $\pm$ 0.00 & -0.03 $\pm$ 0.00 & - & -0.06 $\pm$ 0.01 & -0.01 $\pm$ 0.00 \\ 
Train Acc. & 0.98 $\pm$ 0.00 & 0.99 $\pm$ 0.00 & 0.99 $\pm$ 0.00 & 0.99 $\pm$ 0.00 & 1.00 $\pm$ 0.00 & 0.98 $\pm$ 0.00 & 1.00 $\pm$ 0.00 \\ 
Test Acc. & 0.86 $\pm$ 0.00 & 0.86 $\pm$ 0.00 & 0.86 $\pm$ 0.00 & 0.86 $\pm$ 0.00 & 0.87 $\pm$ 0.00 & 0.86 $\pm$ 0.00 & 0.86 $\pm$ 0.00 \\ 
Train LL & -0.05 $\pm$ 0.00 & -0.02 $\pm$ 0.00 & -0.05 $\pm$ 0.00 & -0.03 $\pm$ 0.00 & -0.01 $\pm$ 0.00 & -0.05 $\pm$ 0.01 & -0.01 $\pm$ 0.00 \\ 
Test LL & -0.70 $\pm$ 0.01 & -0.90 $\pm$ 0.00 & -0.68 $\pm$ 0.00 & -0.64 $\pm$ 0.00 & -1.38 $\pm$ 0.03 & -0.67 $\pm$ 0.02 & -0.84 $\pm$ 0.00 \\ 
Inc. Test LL & -4.83 $\pm$ 0.12 & -6.31 $\pm$ 0.00 & -4.65 $\pm$ 0.00 & -4.58 $\pm$ 0.00 & -8.97 $\pm$ 0.07 & -4.66 $\pm$ 0.13 & -6.06 $\pm$ 0.01 \\ 
ECE & 0.09 $\pm$ 0.00 & 0.11 $\pm$ 0.00 & 0.09 $\pm$ 0.00 & 0.09 $\pm$ 0.00 & 0.12 $\pm$ 0.00 & 0.09 $\pm$ 0.00 & 0.11 $\pm$ 0.00 \\ 
\bottomrule 
    \end{tabular}
\end{table*}
\end{landscape}

\section{Addressing the pathology}\label{sec:dkl:fix}
We have seen that the empirical Bayesian approach to overparameterized GP kernels can lead to pathological behavior.
In particular, we have shown that methods that rely on the marginal likelihood to optimize a large number of hyperparameters can overfit, and that learning is unstable.
While minibatching can help mitigate these issues, the overall performance is sensitive to the batch size, leading to a separate hyperparameter to tune.
It is therefore natural to wonder whether we can address this by using a fully Bayesian approach, which has been shown to improve the predictive uncertainty of GP models \citep{lalchand2020approximate}.
Indeed, \citet{tran2019calibrating} showed that using Monte Carlo dropout to perform approximate Bayesian inference over the network parameters in DKL can improve calibration.

\begin{figure}[t]
    \centering
    \begin{subfigure}[b]{0.49\textwidth}
    \centering
    \centerline{\includegraphics[width=\textwidth]{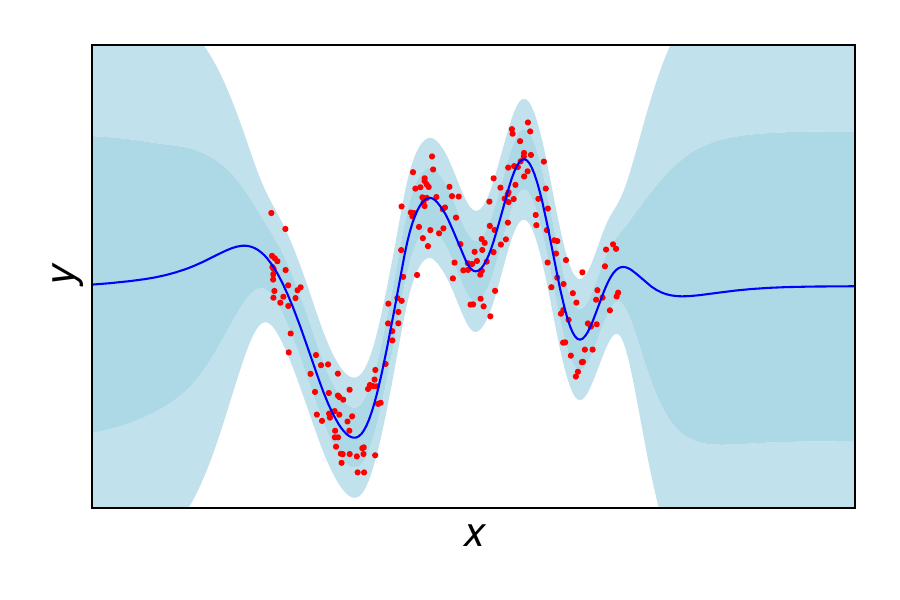}}
    \caption{Original dataset}
    \label{fig:dkl:toy_hmc_200}
    
    \end{subfigure}
    \hfill
    \begin{subfigure}[b]{0.49\textwidth}
    \centering
    \centerline{\includegraphics[width=\textwidth]{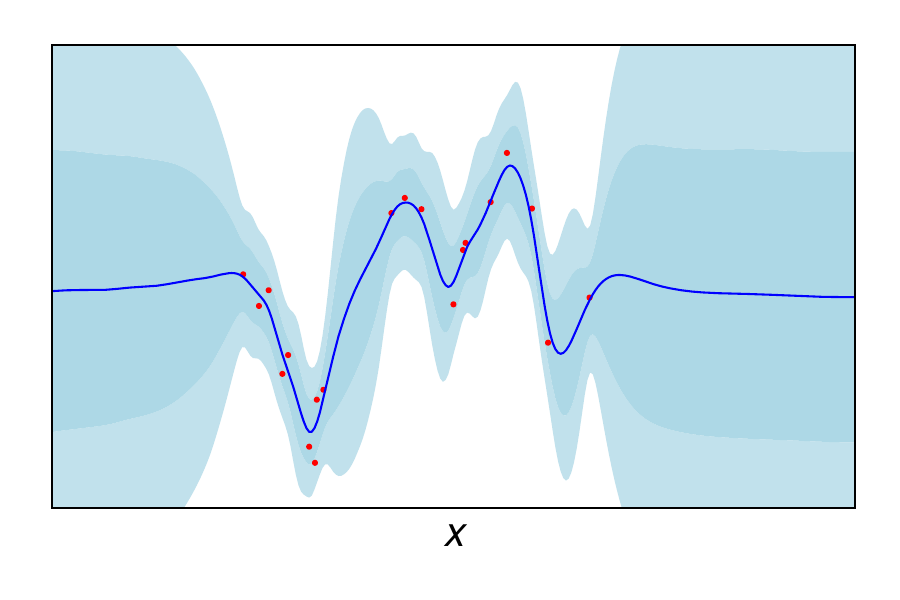}}
    \caption{Subsampled dataset}
    \label{fig:dkl:toy_hmc_20}
    
    \end{subfigure}
    \caption{Posteriors for fully Bayesian DKL using HMC.}
    \label{fig:dkl:hmc}
\end{figure}

We test this hypothesis using sampling methods.
We first consider the 1D toy problem, using HMC \citep{neal2011mcmc} to sample the neural network weights along with the other GP hyperparameters, using the marginal likelihood as the potential.
We plot the resulting posterior in Fig.~\ref{fig:dkl:toy_hmc_200}, and see that this completely resolves the problems observed earlier. 
In fact, the uncertainty in the outer regions is even greater than that given by the standard SE fit in Fig.~\ref{fig:dkl:toy_SE_fit}, while still concentrating where there is data, indicating potentially improved epistemic uncertainty quantification.
We additionally consider a subsampled version of the dataset in Fig.~\ref{fig:dkl:toy_hmc_20}.
There is still no overfitting despite the small dataset size: for a comparison to the baseline SE kernel and DKL, see Fig.~\ref{fig:dkl:app-toy} in Appendix~\ref{app:dkl:add_exp}.

Unfortunately, HMC in its standard form does not scale to the larger datasets considered in Sec.~\ref{sec:dkl:image-datasets}, due to the necessity of calculating gradients over the entire dataset and the calculation of the acceptance probability.
Therefore, we consider stochastic gradient Langevin dynamics \citep[SGLD;][]{welling2011bayesian}, which allows us to use minibatches.
We note that SGLD has relatively little additional training cost compared to SGD, as it simply injects scaled Gaussian noise into the gradients; the main cost is in memory and at test time.
While we do not necessarily expect that this will be as accurate to the true posterior as HMC \citep[see e.g., ][]{johndrow2020no}, we hope that it will give insights into what the performance of a fully Bayesian approach would be.
We select a batch size of 100, and give test results for the NN and SVDKL models for both UTKFace and CIFAR-10 without data augmentation in Table~\ref{tab:dkl:sgld}.
We see that for both datasets, the additional uncertainty significantly helps the NN models.
The improvement is significant for SVDKL for the UTKFace dataset, and while not so significant for CIFAR-10, we still observe slight improvements in log likelihoods and ECE at the expense of slightly lower test accuracy.
Moreover, the fully Bayesian SVDKL outperforms the Bayesian NN in nearly every metric, and significantly so for the uncertainty-related metrics.
In fact, for CIFAR-10, the original version of SVDKL (i.e. pSVDKL) outperforms the Bayesian NN for the uncertainty metrics, even for the larger batch size experiments.
Therefore, we arrive at our final remark:
\begin{rem}\label{rem:fully-Bayesian}
A fully Bayesian approach to deep kernel learning can prevent overfitting and obtain the benefits of both neural networks and Gaussian processes.
\end{rem}%

\begin{table}
\small
    \centering
    \caption{Results for the image datasets with SGLD. We highlight the best results (taking error bars into account) in bold.}\label{tab:dkl:sgld}
    \begin{tabular}{rcc}
\toprule 
& NN & SVDKL \\ 
\midrule 
UTKFace - Test RMSE & 0.16 $\pm$ 0.00 & 0.16 $\pm$ 0.00 \\
Test LL & 0.39 $\pm$ 0.04 & 0.42 $\pm~$ 0.03 \\
\midrule
CIFAR-10 - Test Acc. & $\mathbf{0.79 \pm 0.00}$ & 0.78 $\pm$ 0.00 \\
Test LL & -1.89 $\pm$ 0.02 & -$\mathbf{1.11 \pm 0.02}$ \\
Inc. Test LL & -8.78 $\pm$ 0.11 & -$\mathbf{4.94 \pm 0.10}$ \\
ECE & 0.18 $\pm$ 0.00 & $\mathbf{0.13 \pm 0.00}$ \\
\bottomrule 
    \end{tabular}
\end{table}

\section{Concurrent \& subsequent related work}
We now briefly summarize and comment on concurrent and subsequent work related to the contents of this chapter.
Deep kernel learning techniques have enjoyed increasing attention and success in the past few years.
\citet{schwoebel2021last} attempted to use DKL techniques to learn data augmentations by following the approach introduced by \citet{vdw2018inv}.
However, they were only able to show limited success, as the co-adaptation of features and learned invariances, required to learn the latter, also allows the model to overfit.
In another line of work, \citet{liu2020simple} introduce the  \emph{spectral normalized Gaussian process (SNGP)}, which uses spectral normalization \citep{miyato2018spectral} to enforce ``distance awareness'' through a neural network's layers, so that points that are far away from training data remain far away as they are propagated through the feature extractor.
In order to make the GP layer scalable, they use a random Fourier feature representation of the GP \citep{rahimi2007random} with a Laplace approximate posterior.
\citet{fortuin2021bayesian} modify this by introducing a latent variable to the SNGP model to explicitly account for aleatoric uncertainty in the model's predictions.
Improving on this line of work, \citet{van2021improving} introduce spectral normalization for the batch normalization layers, and argue that the RFF approximation should be replaced by a standard SVGP.
Moreover, they identify \emph{feature collapse} as a key mechanism for the overfitting observed in DKL models without spectral normalization, wherein the feature extractor collapses all the input data onto a low-dimensional manifold.

Combining this observation with ours, we can understand that one potential mechanism that DKL models achieve over-correlation of its inputs is by feature collapse: if all the data points are forced to lie next to each other in feature space, they will be highly correlated.
It then becomes clear that a test point that is distant from the data manifold in input space will then be mapped to be next to data in feature space, causing overfitting.
Spectral normalization techniques therefore mitigate this failure mode of DKL models by enforcing appropriate distance between data points in the feature space, which will make it harder for the marginal likelihood to correlate all the data points.
This is as opposed to more classical regularization techniques for neural networks, which are largely ineffectual on DKL models.
For instance, with standard $l_2$ regularization, the neural network will simply shrink its weights, resulting in a shrunken feature space. 
Unfortunately, the GP can straightforwardly compensate for this by shrinking its lengthscales appropriately.
It would therefore be interesting to understand in more depth how spectral normalization constrains the marginal likelihood to avoid pathological behavior, and whether failure modes in addition to feature collapse still remain.

Finally, \citet{lotfi2022bayesian} discuss the marginal likelihood in the context of model selection and generalization, using DKL as one of many examples.
They argue that the marginal likelihood can lead to both under- and over-fitting, and is therefore not a good proxy for either generalization or model selection.
Instead, they propose the use of the \emph{conditional LML (CLML)}, in which the LML is computed conditioned on a subset of the data, which they argue is more robust to model mis-specification.
They illustrate this in part by performing experiments comparing the performance of DKL models optimized using either the LML or the CLML.
They argue that the CLML can mitigate the tendency of the LML to underfit in the context of DKL.
This may be puzzling, given the contents of this chapter.
However, on closer inspection, they only optimize their models for 100 gradient steps of the Adam optimizer \citep{kingma2014adam}.
In our experience, this number of steps is nowhere near sufficient to adequately optimize these models, and so is unlikely to indicate any significant differences in the optima of the CLML compared to those of the LML.

Considering the broader picture, it is true that the marginal likelihood does not directly provide the model with the best test performance.
However, we are not aware of a probabilistic interpretation of the CLML that would directly lend itself to being more useful than the LML for model selection.
By contrast, the LML has a simple interpretation from the Bayesian perspective.
Moreover, the CLML introduces an additional parameter that needs to be tuned: the size of the dataset to condition on.
Therefore, we would generally agree with David MacKay's argument when discussing failure of the Bayesian method to find models that generalize well \citep[][emphasis added]{mackay1992practical}:
\begin{displayquote}
A failure indicates one of two things, and in either case we are able to learn and improve: either numerical inaccuracies in the evaluation of the probabilities caused the failure, or \emph{else the alternative that were offered to Bayes were a poor selection}, ill-matched to the real world (for example, using inappriopriate regularizers). When such a failure is detected, \emph{it prompts us to examine our models and try to discover the implicit assumptions in the model that the data did not agree with}; alternative models can be tried until one is found that makes the data more probable.
\end{displayquote}
Indeed, one of the failure modes of the LML identified by \citet{lotfi2022bayesian} is due to the prior placing significant mass on regions of the parameter space which correspond to unlikely posterior models.
In this case, we would argue that it would be better to identify such settings and improve the prior to avoid this, as \citet{mackay1992practical} demonstrates.

\section{Conclusions}
In this chapter, we have explored the performance of DKL in different regimes. 
We have shown that, while DKL models can achieve good performance, this is mostly because of implicit regularization due to stochastic minibatching rather than a better marginal likelihood.
This stochastic regularization appears to be stronger than that for plain neural networks.
Moreover, we have shown that when this stochastic regularization is limited, the performance can be worse than that of standard neural networks, with more overfitting and unstable training.
This is surprising, because DKL models are ``more Bayesian'' than deterministic neural networks, and so one might expect that they would be less prone to overfitting.
However, we have shown that for highly parameterized models, the marginal likelihood tries to correlate all the datapoints rather than those that should be correlated: therefore, a higher marginal likelihood does not improve performance.
This means that when the number of hyperparameters is large, the marginal likelihood cannot be relied upon for model selection as it often is, just as the standard maximum likelihood training loss cannot be used for model selection.
Finally, we showed that a fully Bayesian approach to the neural network hyperparameters can overcome this limitation and improve the performance over the less Bayesian approach, fully showing the advantages of DKL models. 
While we have restricted our analysis to DKL models, there is no reason that the conclusions we have arrived at should be limited only to this class of models: therefore, we would argue that wherever possible, all the parameters of a model should be integrated over via Bayes' rule, and the number of hyperparameters should be limited to as few as possible.

\chapter[Global inducing points]{Improving variational inference in deep Bayesian models with global inducing points}\label{sec:gi}

\ifpdf
    \graphicspath{{GI/Figs/Raster/}{GI/Figs/PDF/}{GI/Figs/}}
\else
    \graphicspath{{GI/Figs/Vector/}{GI/Figs/}}
\fi

In the previous chapter, we argued that we should integrate over as many of the parameters in a model as possible, rather than treating them as learnable.
Motivated by our discussion in Chapters~\ref{introduction} and~\ref{sec:back}, we explore the variational approach for Bayesian neural networks and deep Gaussian processes.
We begin by discussing prior attempts at variational inference in these models, and argue that they typically omit crucial between-layer correlations in their approximate posteriors.
By contrast, to build our own approximate posterior, we first consider the optimal approximate posterior over the top-layer weights in a Bayesian neural network for regression, and show that it exhibits strong dependencies on the lower-layer weights. 
We adapt this result to develop a correlated approximate posterior over the weights at all layers in a Bayesian neural network.
Using the equivalence between BNNs and DGPs discussed in Sec.~\ref{sec:back:dgps}, we extend this approach to deep Gaussian processes, providing a unified approach to variational inference in the two model classes. 
Our approximate posterior uses learned ``global'' inducing points, which are defined only at the input layer and propagated through the network to obtain inducing inputs at subsequent layers. 
By contrast, the standard ``local'' inducing point methods from the deep Gaussian process literature optimise a separate set of inducing inputs at every layer, and thus do not model correlations across layers. 
Our method gives state-of-the-art performance for a variational fully Bayesian method, without data augmentation or tempering, on CIFAR-10 of 86.7\%, which is comparable to SGMCMC without tempering but with data augmentation \citep[88\% in ][]{wenzel2020good}.

This chapter is based on joint work with Laurence Aitchison that was published in \citep*{ober2020global}.

\section{Motivation}
As we have seen, VI in Bayesian neural networks requires the user to specify a family of approximate posteriors over the weights, with the classical approach using Gaussian distributions that are independent across not only layers but individual weights \citep{hinton1993keeping,graves2011practical,blundell2015weight}.
Later work has considered more complex approximate posteriors, for instance using a Matrix-Normal distribution as the approximate posterior for a full weight matrix \citep{louizos2016structured} and hierarchical variational inference \citep{louizos2017multiplicative, dusenberry2020efficient}.
By contrast, DGPs use an approximate posterior defined over functions: the standard approach is to specify the inputs and outputs at a finite number of ``inducing'' points \citep{damianou2013deep,salimbeni2017doubly}.

Critically, these classical BNN and DGP approaches define approximate posteriors over functions that are independent across layers. 
An approximate posterior that factorises across layers is problematic, because what matters for a deep model is the overall input-output transformation for the full model, not the input-output transformation for individual layers.
This raises the question of what family of approximate posteriors should be used to capture correlations across layers.
One approach for BNNs would be to introduce a flexible ``hypernetwork,'' a separate neural network used to generate the weights \citep{krueger2017bayesian,pawlowski2017implicit}.
However, this type of brute-force approach is likely to be suboptimal, as it does not sufficiently exploit the rich structure in the underlying neural network.
Finally, most VI approaches in the literature rely on data augmentation and/or posterior tempering for larger networks, both of which cloud the Bayesian perspective and make the resulting ELBO difficult to interpret as a true lower bound to the marginal likelihood (cf. Sec.~\ref{sec:back:temp}).

For guidance, we consider the optimal approximate posterior over the top-layer units in a deep network for regression, conditioned on the lower layers.
This optimal approximate posterior is given by propagating the training inputs through lower layers to compute the top-layer representation, then using Bayesian linear regression to map from the top-layer representation to the outputs.
Inspired by this result, we use Bayesian linear regression to define a generic family of approximate posteriors for BNNs.
In particular, we introduce learned ``pseudo-outputs'' at every layer, and compute the posterior over the weights by performing linear regression from the inputs (propagated from lower layers) onto the pseudo-outputs.
We reduce the burden of working with many training inputs by summarising the posterior using a small number of ``inducing'' points.
Using the BNN-DGP correspondence described in Sec.~\ref{sec:back:dgps}, our approach can be extended to DGPs. 
We explore connections to the inducing point GP literature, showing that inference in the two classes of models can be unified.

Concretely, our contributions are:
\begin{itemize}
    \item We propose an approximate posterior for BNNs based on Bayesian linear regression that naturally induces correlations between layers (Sec.~\ref{sec:gi:full-posterior}).
    \item We provide an efficient implementation of this posterior for convolutional layers (Sec.~\ref{sec:gi:efficient_conv}).
    \item We introduce new BNN priors that allow for more flexibility with inferred hyperparameters (Sec.~\ref{sec:gi:prior}).
    \item We show how our approximate posterior can be naturally extended to DGPs, resulting in a unified approach for inference in BNNs and DGPs (Sec.~\ref{sec:gi:dgp-ext}).
\end{itemize}

\section{Methods}\label{sec:gi:methods}
To motivate our approximate posterior, we first consider the optimal top-layer posterior for a fully-connected Bayesian neural network in the regression case.
We follow the notation outlined in Sec.~\ref{sec:back:nns}, where we have lower-layer weights $\slp{\W_\ell}$, $\W_\ell\in\reals^{\nu_{\ell-1}\times \nu_\ell}$, and output weights $\W_{L+1}\in\Reals^{\nu_L \times \nu_{L+1}}$.
For input data $\X \in \reals^{N \times \nu_0}$, these weights imply activities $\F_\ell \in \reals^{N \times \nu_\ell}$ for $\nu \in \cb{1, \dotsc, L+1}$ according to Eq.~\ref{eq:back:fc_net}.
Defining a likelihood $\pc{\Y}{\F_{L+1}}$ along with a prior $\p{\slpo{\W_\ell}}$ that factorizes across layers and output units (c.f. Sec.~\ref{sec:back:bnns}), we consider a generic approximate posterior $\q{\slpo{\W_\ell}}$.
We fit this approximate posterior by maximizing the ELBO,
\begin{align}
  \label{eq:gi:bnn-elbo}
  \mathcal{L} = \E[\mathrm{q}]{\log \pc{\Y}{\X, \slpo{\W_\ell}} + \log \p{\slpo{\W_\ell}} - \log \q{\slpo{\W_\ell}}}.
\end{align}

To build intuition about how to parameterise $\q{\slpo{\W_\ell}}$, we consider the optimal $\qc{\W_{L+1}}{\slp{\W_\ell}}$ for any given $\q{\slp{\W_\ell}}$, i.e., the optimal top-layer posterior conditioned on the lower layers.
We begin by simplifying the ELBO by incorporating terms that do not depend on $\W_{L+1}$ into $c\b{\slp{\W_\ell}}$, starting by combining the prior and likelihood:
\begin{align*}
  \mathcal{L} = \E[\mathrm{q}]{\log \pc{\Y, \slpo{\W}}{\X} - \log \q{\slpo{\W}}}.
\end{align*}
Then, splitting out $\W_{L+1}$ gives
\begin{multline*}
  \mathcal{L} = \Eb[\mathrm{q}]\big[\log \pc{\Y, \slp{\W}}{\X} + \log \pc{\W_{L+1}}{\Y, \X, \slp{\W_\ell}}\\ - \log \qc{\W_{L+1}}{\slp{\W_\ell}} - \log \q{\slp{\W_\ell}}\big].
\end{multline*}
As we are interested in the optimal $\qc{\W_{L+1}}{\slp{\W_\ell}}$ for any setting of $\slp{\W_\ell}$, we can collect $\pc{\Y, \slp{\W}}{\X}$ and $\q{\slp{\W_\ell}}$, which do not have any dependence on $\W_{L+1}$, into  a single term $c\b{\slp{\W_\ell}}$.
We therefore obtain
\begin{align*}
  \mathcal{L} = \E[\mathrm{q}]{\log \pc{\W_{L+1}}{\Y, \X, \slp{\W_\ell}} - \log \qc{\W_{L+1}}{\slp{\W_\ell}} + c\b{\slp{\W_\ell}}}.
\end{align*}
By using the law of total expectation to split out the expectations,
\begin{multline*}
  \mathcal{L} = \Eb[{\q{\slp{\W}}}]\bigg[\Eb[{\qc{\W_{L+1}}{\slp{\W_\ell}}}]\big[\log \pc{\W_{L+1}}{\Y, \X, \slp{\W_\ell}} \\- \log \qc{\W_{L+1}}{\slp{\W_\ell}}\big] + c\b{\slp{\W}}\bigg].
\end{multline*}
Finally, noticing that the inner expectation is a KL divergence leads us to the expression,
\begin{multline}
\label{eq:gi:last-layer-elbo}
  \mathcal{L} = \Eb[{\q{\slp{\W_\ell}}}]\big[-\KL*{\qc{\W_{L+1}}{\slp{\W_\ell}}}{\pc{\W_{L+1}}{\Y, \X, \slp{\W_\ell}}}\\
  + c\b{\slp{\W_\ell}}\big].
\end{multline}

Thus, the optimal approximate posterior is the true last-layer posterior conditioned on the previous layers' weights,
\begin{align}
  \nonumber
  \qc{\W_{L+1}}{\slp{\W_\ell}} &= \pc{\W_{L+1}}{\Y, \X, \slp{\W_\ell}}\\ 
  \label{eq:gi:WLp1:W}
  &\propto \pc{\Y}{\W_{L+1}, \F_L} \p{\W_{L+1}},
\end{align}
where the final proportionality comes by applying Bayes' theorem and exploiting the model's conditional independencies.
For regression with a Gaussian likelihood,
\begin{align}
  \label{eq:gi:def:reg}
  \pc{\Y}{\W_{L+1}, \F_L} = \prodlnlp \Nc{\y_\lambda}{\psi\b{\F_L} \w^{L+1}_\lambda, \La^{-1}_{L+1}},
\end{align}
where $\y_\lambda$ is the value of a single output channel for all training inputs, and $\La_{L+1}$ is a precision matrix, which for convenience we assume to be shared across outputs. 
Thus, the posterior is given in closed form by Bayesian linear regression \citep{rasmussen2006gaussian}:
\begin{align}
    \label{eq:gi:blr-refresher}
    \qc{\W_{L+1}}{\slp{\W_\ell}} = \prodlnlp \Nc{\w^{L+1}_\lambda}{\S \transpose{\psi\b{\F_L}} \La_{L+1} \y_\lambda, \S},
\end{align}
where
\begin{align*}
    \S = \b{\nu_L \Sm_{L+1}^{-1} + \transpose{\psi\b{\F_L}}\La_{L+1}\psi\b{\F_L}}^{-1}.
\end{align*}
While this result may be neither particularly novel nor surprising, it neatly highlights our motivation for the rest of the paper.
In particular, it shows that for regression, we can always obtain the optimal conditional top-layer posterior, which to the best of our knowledge has not been used before in BNN inference.
Moreover, doing top-layer Bayesian linear regression based on the propagated features from the previous layers naturally introduces correlations between layers.

\subsection{Defining the full approximate posterior with global inducing points and pseudo-outputs} \label{sec:gi:full-posterior}
We adapt the optimal top-layer approximate posterior above to give a scalable approximate posterior over the weights at all layers.
To avoid propagating all training inputs through the network, which is intractable for large datasets, we instead propagate $M$ \textit{global} inducing locations, $\U_0 \in \Reals^{M\times \nu_0}$,
\begin{align}
  \nonumber
  \U_1 &= \U_0 \W_1, \\ 
  \label{eq:gi:bnn_pseudo_prop}
  \U_\ell &= \psi\b{\U_{\ell-1}}\W_\ell \quad \text{ for } \ell = 2, \dots, L+1.
\end{align}
Next, the optimal posterior requires outputs, $\Y$.
However, no outputs are available at inducing locations for the output layer, let alone for intermediate layers.
We thus introduce (learnable) variational parameters to mimic the form of the optimal posterior. 
In particular, we use the product of the prior over weights and an ``inducing-likelihood'',
$\N{\v_\lambda^\ell; \u_\lambda^\ell, \La_\ell^{-1}}$, representing noisy ``pseudo-outputs'' of the outputs of the linear layer at the inducing locations, $\u_\lambda^\ell = \psi\b{\U_{\ell-1}} \w_\lambda^\ell$.
Substituting $\u_\lambda^\ell$ into the inducing-likelihood, the approximate posterior becomes
\begin{align}
\nonumber
    \qc{\W_\ell}{\cb{\W_{\ell'}}_{\ell' = 1}^{\ell - 1}} &\propto \prodln \Nc{\v_\lambda^\ell}{\psi\b{\U_{\ell-1}}\w_\lambda^\ell,\, \La_\ell^{-1}}\p{\w_\lambda^\ell},\\
\nonumber
    \qc{\W_\ell}{\cb{\W_{\ell'}}_{\ell' = 1}^{\ell - 1}} &= \prodln \Nc{\w_\lambda^\ell}{\S^\w_\ell \transpose{\psi\b{\U_{\ell-1}}}\La_\ell \v_\lambda^\ell,\, \S_\ell^\w}, \\
    \label{eq:gi:Qw|v}
    \S_\ell^\w &= \b{\nu_{\ell-1}\Sm_\ell^{-1} + \transpose{\psi\b{\U_{\ell-1}}} \La_\ell \psi\b{\U_{\ell-1}}}^{-1},
\end{align}
where $\v_\lambda^\ell$ and $\La_\ell$ are variational parameters.

Summarizing our notation, we have: $\u_\lambda^\ell, \v_\lambda^\ell \in\Reals^{M}$, so that $\U_{\ell-1}\in\reals^{M\times \nu_{\ell-1}}$ and $\V_{\ell} \in \reals^{M \times \nu_\ell}$ are formed by stacking these vectors, and $\w_\lambda^\ell \in\reals^{N_{\ell-1}}$, with $\Sm_\ell, \S^\w_\ell\in\reals^{\nu_{\ell-1}\times \nu_{\ell-1}}$ and $\La_\ell\in\reals^{M \times M}$.
Therefore, our full approximate posterior factorizes as
\begin{align*}
  \q{\slpo{\W_\ell}} &= \prod_{\ell=1}^{L+1} \qc{\W_\ell}{\cb{\W_{\ell'}}_{\ell'=1}^{\ell-1}}.
\end{align*}
Substituting this approximate posterior and the factorised prior into the ELBO (Eq.~\ref{eq:gi:bnn-elbo}), the full ELBO can be written as
\begin{align*}
  \mathcal{L} = \E[\q{\slpo{\W}}]{\log \p{\Y, |\X, \slpo{\W}}+ \sum_{\ell=1}^{L+1} \log \frac{\p{\W_\ell}}{\q{\W_\ell| \cb{\W_{\ell'}}_{\ell'=1}^{\ell'-1}}}},
\end{align*}
where $\p{\W_\ell}$ is given by Eq.~\eqref{eq:back:pw} and $\qc{\W_\ell}{\cb{\W_{\ell'}}_{\ell'=1}^{\ell'-1}}$ is given by Eq.~\eqref{eq:gi:Qw|v}.
The forms of the ELBO and approximate posterior suggest a sequential procedure to evaluate and subsequently optimize it: we alternate between sampling the weights using Eq.~\eqref{eq:gi:Qw|v} and propagating the data and inducing points (Eq.~\ref{eq:back:fc_net} and Eq.~\ref{eq:gi:bnn_pseudo_prop};  see Alg.~\ref{algo}).
In summary, the parameters of the approximate posterior are the global inducing inputs, $\U_0$, and the pseudo-outputs and precisions at all layers, $\slpo{\V_\ell, \La_\ell}$.
As each factor $\qc{\W_\ell}{\cb{\W_{\ell'}}_{\ell'=1}^{\ell-1}}$ is Gaussian, these parameters can be optimised using standard reparameterized variational inference \citep{kingma2013auto,rezende2014stochastic} in combination with the Adam optimiser \citep{kingma2014adam} (see Appendix~\ref{app:reparam} for a description of the reparaterization trick).
Importantly, by placing inducing inputs on the training data (i.e.\ $\U_0 = \X$), and setting $\v_\lambda^\ell = \y_\lambda$, this approximate posterior matches the optimal top-layer posterior (Eq.~\ref{eq:gi:WLp1:W}).
Finally, we note that while this posterior is conditionally Gaussian, the full posterior over all $\slpo{\W_\ell}$ is non-Gaussian, which might have benefits over a full-covariance Gaussian approximate posterior, which is intractable for larger networks.
\begin{algorithm}[tb]
\caption{Global inducing points for neural networks}
\label{algo}
\begin{algorithmic}
  \STATE \textbf{Parameters:} $\U_0$, $\slpo{\V_\ell, \La_\ell}$.
  \STATE \textbf{Neural network inputs:} $\F_0$
  \STATE \textbf{Neural network outputs:} $\F_{L+1}$
  \STATE $\mathcal{L} \leftarrow 0$
  \FOR{$\ell$ {\bfseries in} $\{1,\dotsc,L+1\}$}
  \STATE \textcolor{gray}{Compute the mean and cov.\ for weights at this layer}
  \STATE $\S^\w_\ell = \b{\nu_{\ell-1} \Sm^{-1}_\ell + \transpose{\psi\b{\U_{\ell-1}}} \La_\ell \psi\b{\U_{\ell-1}}}^{-1}$
  \STATE $\M_\ell = \S^\w_\ell \transpose{\psi\b{\U_{\ell-1}}} \La_\ell \V_\ell$
  \STATE \textcolor{gray}{Sample the weights and compute the ELBO}
  \STATE $\W_\ell \sim \N{\M_\ell, \S^\w_\ell} = \qc{\W_\ell}{\cb{\W_{\ell'}}_{\ell'=1}^{\ell-1}}$
  \STATE $\mathcal{L} \leftarrow \mathcal{L} + \log \p{\W_\ell} - \log \Nc{\W_\ell}{\M_\ell, \S^\w_\ell}$
  \STATE \textcolor{gray}{Propagate the inputs and inducing points using sampled weights,}
  \STATE $\U_\ell = \psi\b{\U_{\ell-1}} \W_\ell$
  \STATE $\F_\ell = \psi\b{\F_{\ell-1}} \W_\ell$
  \ENDFOR
  \STATE $\mathcal{L} \leftarrow \mathcal{L} + \log \pc{\Y}{\F_{L+1}}$
\end{algorithmic}
\end{algorithm}

\subsection{Efficient convolutional Bayesian linear regression}
\label{sec:gi:efficient_conv}
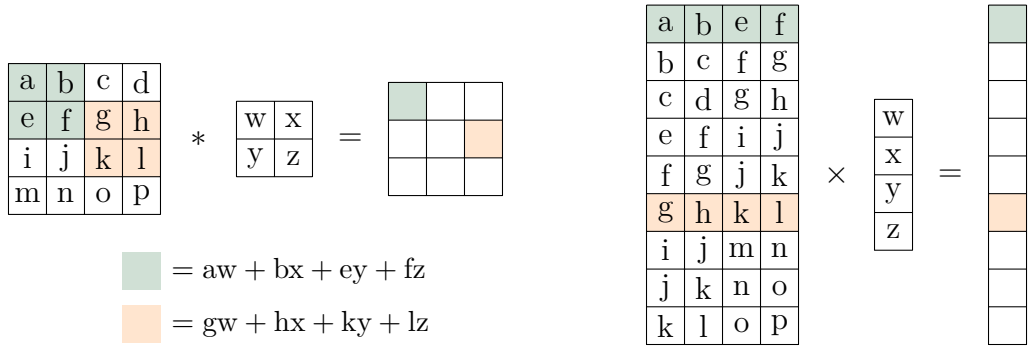
\begin{figure}[t]
\centering
\begin{subfigure}[b]{0.49\textwidth}
\centering
\definecolor{darkorange}{RGB}{255,127,14}
\definecolor{forestgreen}{RGB}{20,100,44}

    \begin{tikzpicture}
    
    \fill[forestgreen!20] (0, 1) rectangle ++ (1, 1);
    \fill[darkorange!20] (1, 0.5) rectangle ++ (1, 1);
    \draw[step=0.5] (0, 0) grid (2, 2);
    
    \newcounter{mycountgi}
    \setcounter{mycountgi}{`a}
    \foreach \y in {+1.75,+1.25,0.75,0.25}
    \foreach \x in {0.25,0.75,1.25,1.75}
    \node at (\x,\y) {\char\value{mycountgi}\addtocounter{mycountgi}{1}};
    
    \node at (2.5, 1) (conv) {$\ast$};
    
    \draw[step=0.5] (3-0.001, 0.5) grid (4, 1.5);
    
    \setcounter{mycountgi}{`w}
    \foreach \y in {+1.25,+0.75}
    \foreach \x in {3.25,3.75}
    \node at (\x,\y) {\char\value{mycountgi}\addtocounter{mycountgi}{1}};
    
    \node at (4.5, 1) (equal) {$=$};
    
    \fill[forestgreen!20] (5, 1.25) rectangle ++ (0.5, 0.5);
    \fill[darkorange!20] (6, 0.75) rectangle ++ (0.5, 0.5);
    \draw[step=0.5,shift={(0., 0.25)}] (5.0-0.001, 0) grid (6.5, 1.5);
    
    \fill[forestgreen!20] (1.5, -1) rectangle ++ (0.5, 0.5);
    \node[anchor=west, align=center] at (2, -.75) (result1) {\small $= \text{aw} + \text{bx} + \text{ey} + \text{fz}$};
    
    \fill[darkorange!20] (1.5, -1.7) rectangle ++ (0.5, 0.5);
    \node[anchor=west, align=center] at (2, -1.45) (result1) {\small $= \text{gw} + \text{hx} + \text{ky} + \text{lz}$};
    
    \end{tikzpicture}
    \caption{The standard convolutional representation.}
\end{subfigure}
\begin{subfigure}[b]{0.49\textwidth}
\centering
\definecolor{darkorange}{RGB}{255,127,14}
\definecolor{forestgreen}{RGB}{20,100,44}

    \begin{tikzpicture}
    
    \fill[forestgreen!20] (0, 4) rectangle ++ (2, 0.5);
    \fill[darkorange!20] (0, 1.5) rectangle ++ (2, 0.5);
    \draw[step=0.5] (0, 0) grid (2, 4.5);
    
    \node at (0.25,4.25) {a};
    \node at (0.75,4.25) {b};
    \node at (1.25,4.25) {e};
    \node at (1.75,4.25) {f};
    \node at (0.25,3.75) {b};
    \node at (0.75,3.75) {c};
    \node at (1.25,3.75) {f};
    \node at (1.75,3.75) {g};
    \node at (0.25,3.25) {c};
    \node at (0.75,3.25) {d};
    \node at (1.25,3.25) {g};
    \node at (1.75,3.25) {h};
    \node at (0.25,2.75) {e};
    \node at (0.75,2.75) {f};
    \node at (1.25,2.75) {i};
    \node at (1.75,2.75) {j};
    \node at (0.25,2.25) {f};
    \node at (0.75,2.25) {g};
    \node at (1.25,2.25) {j};
    \node at (1.75,2.25) {k};
    \node at (0.25,1.75) {g};
    \node at (0.75,1.75) {h};
    \node at (1.25,1.75) {k};
    \node at (1.75,1.75) {l};
    \node at (0.25,1.25) {i};
    \node at (0.75,1.25) {j};
    \node at (1.25,1.25) {m};
    \node at (1.75,1.25) {n};
    \node at (0.25,0.75) {j};
    \node at (0.75,0.75) {k};
    \node at (1.25,0.75) {n};
    \node at (1.75,0.75) {o};
    \node at (0.25,0.25) {k};
    \node at (0.75,0.25) {l};
    \node at (1.25,0.25) {o};
    \node at (1.75,0.25) {p};
    
    \node at (2.5, 2.25) (times) {$\times$};
    
    \draw[step=0.5,shift={(0., 0.25)}] (3-0.001, 1-0.001) grid (3.5, 3);
    
    \setcounter{mycountgi}{`w}
    \foreach \y in {3, 2.5, 2, 1.5}
    \node at (3.25,\y) {\char\value{mycountgi}\addtocounter{mycountgi}{1}};
    
    \node at (4, 2.25) (equal) {$=$};
    
    \fill[forestgreen!20] (4.5, 4) rectangle ++ (0.5, 0.5);
    \fill[darkorange!20] (4.5, 1.5) rectangle ++ (0.5, 0.5);
    \draw[step=0.5] (4.5-0.001, 0) grid (5, 4.5);
    
    
    
    \end{tikzpicture}
    \caption{The linear regression equivalent.}
\end{subfigure}
\caption{Two representations of the standard convolution operation used in CNNs. In (a), we show the standard representation in its convolutional form, where we have a $4\times 4$ image and $2\times 2$ kernel, resulting in a $3\times 3$ image. In (b), we show how the same operation can be expressed as linear regression, where we have extracted and flattened the $2\times 2$ patches in the image and flattened the kernel into a weight vector.}
\label{fig:gi:convolution}
\end{figure}

The previous sections were valid for a fully connected network.
The extension to convolutional networks is straightforward in principle: we transform the convolution into a matrix multiplication by treating each patch as a separate input feature vector, flattening the spatial and channel dimensions together into a single vector, a process which we illustrate in Fig.~\ref{fig:gi:convolution}.
Thus, the feature vectors have length \verb!in_channels! $\times$ \verb!kernel_width! $\times$ \verb!kernel_height!, and the matrix $\U_\ell$ contains \verb!patches_per_image! $\times$ \verb!minibatch! patches.
Likewise, we now have inducing outputs, $\v_\lambda^\ell$, at each location in all the inducing images, so this again has length \verb!patches_per_image! $\times$ \verb!minibatch!.
After explicitly extracting the patches, we can straightforwardly apply standard Bayesian linear regression.

However, explicitly extracting image patches is very memory intensive in a DNN.
If we consider a standard convolution with a $3\times3$ convolutional kernel, then there is a $3\times 3$ patch centered at each pixel in the input image, meaning a factor of $9$ increase in memory consumption.
Instead, we note that computing the matrices required for linear regression, $\transpose{\psi\b{\U_{\ell-1}}} \La_\ell\psi\b{\U_{\ell-1}}$ and $\transpose{\psi\b{\U_{\ell-1}}} \La_\ell \V_\ell$, does not require explicit extraction of image patches.
Instead, these matrices can be computed by taking the autocorrelation of the image/feature map, i.e., a convolution operation where we treat the image/feature map, as \textit{both} the inputs and the weights.
We now describe this process in detail for 1-dimensional convolutions, with the 2-dimensional case being a straightforward extension.

Explicitly expanding Eq.~\ref{eq:back:conv}, the standard form for a convolution with input features $\X$, output features $\Y$, and convolutional weights/kernel $\W$ is
\begin{align}
    \Y_{i, c', :} &= \sum_{c} \X_{i, c, :} \ast \W_{c', c, :},
\intertext{where each element is given by}
  Y_{i,c',u} &= \sum_{c,\delta} X_{i,c,\b{u+\delta}} W_{c',c,\delta}.
\end{align}
Here, $i$ indexes images, $c$ and $c'$ index channels, $u$ indexes the location within the image, and $\delta$ indexes the location within the convolutional patch.
Later, we will swap the identity of the ``patch location'' and the ``image location'' and to facilitate this, we define them both to be centered on zero,
\begin{align}
  u &\in \cb{-(S-1)/2,\dotsc,(S-1)/2}, & \delta \in \cb{-(K-1)/2,\dotsc, (K-1)/2},
\end{align}
where $S$ is the size of an image and $K$ is the size of a patch, such that, for example for a size 3 kernel, $\delta\in\cb{-1,0,1}$.

Following the illustration in Fig.~\ref{fig:gi:convolution}, we now express the above convolution as linear regression by forming a new input, $\X'$, by cutting out each image patch:
\begin{align}
  X'_{i,c,u,\delta} &= X_{i,c,\b{u +\delta}},
\end{align}
leading to
\begin{align}
  Y_{i,c',u} &= \sum_{c,\delta} X'_{i,c,u,\delta} W_{c',c,\delta}.
\end{align}
We then proceed by combining both $i$ and $u$ into a single index, combining $\delta$ and $c$, followed by permuting indices so that $\Y$ is indexed by (the combined index) $iu$ and $c'$, $\X'$ is indexed by $iu$ and $c\delta$, and $\W$ is indexed by $c\delta$ and $c'$.
We then obtain the simple linear regression,
\begin{align}
  \Y &=  \X' \W.
\end{align}
This means that we can directly apply the approximate posterior we derived for the fully-connected case in Eq.~\eqref{eq:gi:Qw|v} to the convolutional case. To allow for this, we take
\begin{align}
  \X' &= \La_\ell^{1/2} \psi\b{\U_{\ell-1}}, & \Y &= \La_\ell^{1/2} \V_\ell.
\end{align}

For linear regression (Eq.~\ref{eq:gi:Qw|v}), we first need to compute
\begin{align}
  \b{\transpose{\psi\b{\U_{\ell-1}}} \La_\ell\V_\ell}_{c \delta, c'} &= \b{\transpose{\X'} \Y}_{c \delta, c'} = \sum_{i u} X'_{i u, c \delta} Y_{i u, c'}.\\
  \intertext{Rewriting this in terms of the original $\X$ (i.e., without explicitly cutting out image patches), we obtain}
  \b{\transpose{\X'} \Y}_{c \delta, c'} &= \sum_{i u} X_{i,c,\b{u+\delta}} Y_{iu, c'}.
\end{align}
This can be directly viewed as the convolution of $\X$ and $\Y$, where we treat $\Y$ as the ``convolutional weights,'' $u$ as the location within the now very large (size $S$) ``convolutional patch,'' and $\delta$ as the location in the resulting output.
Once we realise that the computation is a spatial convolution, it is possible to fit it into standard convolution functions provided by deep learning frameworks.

Next, we need to compute
\begin{align}
  \b{\transpose{\psi\b{\U_{\ell-1}}}\La_\ell\psi\b{\U_{\ell-1}}}_{c \delta, c' \delta'} &= \b{\transpose{\X'} \X'}_{c \delta, c'\delta'} = \sum_{i u} X'_{i u, c\delta} X'_{i u, c'\delta'}.
  \intertext{Again, rewriting this in terms of $\X$ (i.e., without explicitly cutting out image patches), we obtain,}
  \b{\transpose{\X'} \X'}_{c \delta, c' \delta'} &= \sum_{i u} X_{i, c, \b{u+\delta}} X_{i, c', \b{u+\delta'}}.
\end{align}
To treat this as a convolution, we first need exact translational invariance, which can be achieved by using circular boundary conditions.
Note that circular boundary conditions are not typically used in neural networks for images, and we therefore only use circular boundary conditions to define the approximate posterior over weights. 
The variational framework does not restrict us to also using circular boundary conditions within our feedforward network, and as such, we use standard zero-padding.
With exact translational invariance, we can write this expression directly as a convolution,
\begin{align}
  \b{\transpose{\X'} \X'}_{c \delta, c' \delta'} &= \sum_{i u} X_{i, c, u} X_{i, c', \b{u+\delta'-\delta}},
  \intertext{where}
  \b{\delta'-\delta} &\in \cb{-(K-1), \dotsc, (K-1)}.
\end{align}
I.e., for a size $3$ kernel, $\b{\delta'-\delta} \in \cb{-2, -1, 0, 1, 2}$,
where we treat $\X_{i, c, u}$ as the ``convolutional weights,'' $u$ as the location within the ``convolutional patch,'' and $\delta'-\delta$ as the location in the resulting output ``feature map.''

We have therefore succeeded in rewriting the terms needed for convolutional Bayesian linear regression purely in terms of convolutions, which offers considerable benefits in terms of memory consumption.
In particular, the output matrices are usually quite small -- the number of channels is typically $32$ or $64$, and the number of locations within a patch is typically $9$, giving a very manageable total size that is typically smaller than $1000 \times 1000$.

\subsection{Priors}
\label{sec:gi:prior}
To ensure we make the most of our BNNs, and to illustrate the importance of prior choice, we investigate the use of four priors of varying flexibility in this chapter. 
We are careful to ensure that all hyperparameters in these priors have a prior and approximate posterior, which is necessary to ensure that ELBOs are comparable across models.

First, we consider a Gaussian prior with fixed scale, NealPrior, so named because it is necessary to obtain meaningful results when considering infinite networks \citep{neal1996priors},
\begin{align*}
  \Sm_\ell &= \I.
\end{align*}
We note that this prior bears strong similarities to the ``He'' initialisation \citep{he2015delving}.\footnote{Recall that we include a factor of $\frac{1}{\nu_{\ell-1}}$ scaling of the covariance in our prior definition, Eq.~\ref{eq:back:pw}.}
NealPrior is defined so as to ensure that the variance of the prior activations do not grow exponentially as they are propagated through the network.
We compare this to the standard $\mathcal{N}(0, 1)$ (StandardPrior), which causes the activations to increase exponentially as they propagate through network layers (see Eq.~\ref{eq:back:pw}):
\begin{align*}
  \Sm_\ell &= \nu_{\ell-1} \I.
\end{align*}
Next, we consider ScalePrior, which defines a prior and approximate posterior over the scale,
\begin{gather*}
  \Sm_\ell = \tfrac{1}{s_\ell} \I,\\
  \p{s_\ell} = \text{Gamma}\b{s_\ell;\, 2,\, 2},\\
  \q{s_\ell} = \text{Gamma}\b{s_\ell;\, 2+\alpha_\ell,\, 2+\beta_\ell},
\end{gather*}
where here we parameterise the Gamma distribution with the shape and rate parameters, and $\alpha_\ell$ and $\beta_\ell$ are non-negative learned parameters of the approximate posterior over $s_\ell$.
Finally, we consider SpatialIWPrior, which allows for spatial correlations in the weights, i.e., it allows for prior correlations between elements of the individual convolutional kernels \citep[see e.g.,][for a similar, but more restrictive spatial prior over weights]{fortuin2021bayesian}.
In particular, we take the covariance to be the Kronecker product of an identity matrix over channel dimensions, and a Wishart-distributed matrix, $\L^{-1}_\ell$, over the spatial dimensions,
\begin{align}
  \nonumber
  \Sm_\ell &= \I \otimes \L^{-1}_\ell,\\
  \nonumber
  \p{\L_\ell} &= \mathcal{W}^{-1}\b{\L_\ell; \b{\nu_{\ell{-}1}{+}1} \mathrlap{\I} \phantom{\I + \mathbf{\Psi}}, \nu_{\ell{-}1}{+}1},\\ 
  \label{eq:gi:IWPrior}
  \q{\L_\ell} &= \mathcal{W}^{-1}\b{\L_\ell; \b{\nu_{\ell{-}1}{+}1} \I + \mathbf{\Psi}, \nu_{\ell{-}1}{+}1{+}\delta},
\end{align}
where $\mathcal{W}^{-1}$ is the inverse Wishart distribution, and the non-negative real number, $\delta$, and the positive definite matrix, $\mathbf{\Psi}$, are learned parameters of the approximate posterior (see Appendix~\ref{app:jac} for details on the inverse Wishart distribution).

\subsection{Extension to DGPs} \label{sec:gi:dgp-ext}
We showed in Sec.~\ref{sec:back:dgps} that BNNs with Gaussian priors can be viewed as a special case of DGPs with a particular choice of kernel.
In this section, we show that by following similar reasoning, we can derive a DGP approximate posterior that is equivalent to our approximate posterior for BNNs.
For ease of exposition, we restrict our attention to DGPs with zero function prior means, although this assumption can be relaxed straightforwardly.

Following Sec.~\ref{sec:back:dgps}, our BNN approximate posterior defines a prior over inducing outputs given by
\begin{align}
  \label{eq:gi:U}
  \pc{\U_\ell}{\U_{\ell-1}} &= \prodln \Nc{\u_\lambda^\ell}{\0, \Kfl{\U_{\ell-1}}},
\end{align}
where $\Kfl{\U_{\ell-1}}$ is the Gram matrix implied by the BNN prior (Eq.~\ref{eq:back:pw}), i.e.,
\begin{align*}
    \Kfl{\U_{\ell-1}} = \frac{1}{\nu_{\ell-1}} \psi\b{\F_{\ell-1}} \Sm_\ell \transpose{\psi\b{\F_{\ell-1}}}.
\end{align*}
We now consider the posterior over $\slpo{\U_\ell}$ that is implied by our BNN posterior.
Recall that $\u_\lambda^\ell = \psi\b{\U_{\ell-1}}\w_\lambda^\ell$, and that $\qc{\w_\lambda^{\ell}}{\cb{\W_{\ell'}}_{\ell'=1}^{\ell-1}}$ only depends on $\cb{\W_{\ell'}}_{\ell'=1}^{\ell-1}$ through $\U_{\ell-1}$.
Given that $\qc{\W_\ell}{\cb{\W_{\ell'}}_{\ell'=1}^{\ell-1}}$ is Gaussian, we can therefore transform the BNN weight posterior into a conditionally Gaussian posterior for $\U_\ell$ given $\U_{\ell-1}$:
\begin{align*}
  \qc{\U_\ell}{\U_{\ell-1}} &= \prodln \Nc{\u_\lambda^\ell}{\psi\b{\U_{\ell-1}}\S^\w_\ell \transpose{\psi\b{\U_{\ell-1}}}\La_\ell \v_\lambda^\ell,\, \psi\b{\U_{\ell-1}}\S_\ell^\w\transpose{\psi\b{\U_{\ell-1}}}}.
\end{align*}
By applying the Woodbury identity and straightforward linear algebraic manipulations, while substituting $\Kfl{\U_{\ell-1}}$ as defined above where possible, we find that this simplifies to
\begin{align}
  \nonumber
  \qc{\U_\ell}{\U_{\ell-1}} &= \prodln \Nc{\u_\lambda^\ell}{\S^\u_\ell \La_\ell \v_\lambda^\ell, \S^\u_\ell}, \\
  \label{eq:gi:U|V}
  \S^\u_\ell &= \b{\Kfl{\U_{\ell-1}}^{-1} + \La_\ell}^{-1}.
\end{align}

Unfortunately, this approximate posterior cannot be used directly in the BNN case, as the degeneracy of the kernel function makes it non-trivial to sample $\F_\ell$ from $\U_\ell$.
However, for general DGPs with non-degenerate kernels, by combining Eq.~\ref{eq:gi:U|V} with the GP conditional prior (c.f. Sec.~\ref{sec:back:dgps}), we now have an approximate posterior that straightforwardly encodes dependencies between layers:
\begin{align*}
    \q{\slpo{\F_\ell, \U_\ell}} &= \prodlayers \qc{\F_\ell, \U_\ell}{\F_{\ell-1}, \U_{\ell-1}} \\
    &= \prodlayers \pc{\F_{\ell}}{\U_{\ell}, \F_{\ell-1}, \U_{\ell-1}}\qc{\U_\ell}{\U_{\ell-1}}.
\end{align*}
After canceling out the prior conditional terms, the full ELBO takes the form
\begin{align}
  \mathcal{L} = \E[\qc{\slpo{\F_\ell, \U_\ell}}{\X, \U_0}]{\log \pc{\Y}{\F_{L+1}} + \sum_{\ell=1}^{L+1} \log \frac{\pc{\U_\ell}{\U_{\ell-1}}}{\qc{\U_\ell}{\U_{\ell-1}}}},
\end{align}
which we optimize with respect to any prior hyperparameters and the variational parameters: the global inducing inputs, $\U_0$, and the pseudo-outputs and precisions at all layers, $\slpo{\V_\ell, \La_\ell}$, which can be optimized using reparameterized variational inference \citep[][see App.~\ref{app:reparam}]{kingma2013auto,rezende2014stochastic}.
Moreover, in App.~\ref{sec:gi:app:gpmotivation}, we show how the same approximate posterior can be arrived at by directly considering the optimal top-layer conditional posterior for DGPs, just as we did for BNNs in Sec.~\ref{sec:gi:methods}.
Algorithm~\ref{algo:dgp} describes the resulting algorithm for DGPs.

\begin{algorithm}[tb]
\caption{Global inducing points for deep Gaussian processes}
\label{algo:dgp}
\begin{algorithmic}
  \STATE \textbf{Parameters:} $\U_0$, $\slpo{\V_\ell, \La_\ell}$.
  \STATE \textbf{Neural network inputs:} $\F_0$
  \STATE \textbf{Neural network outputs:} $\F_{L+1}$
  \STATE $\mathcal{L} \leftarrow 0$
  \FOR{$\ell$ {\bfseries in} $\{1,\dotsc,L+1\}$}
    \STATE \textcolor{gray}{Compute the mean and covariance over the inducing outputs at this layer}
    \STATE $\S^\u_\ell = \b{\Kfl{\U_{\ell-1}}^{-1} + \La_\ell}^{-1}$
    \STATE $\M_\ell = \S^\u_\ell \La_\ell \V_\ell$
    \STATE \textcolor{gray}{Sample the inducing outputs and compute the ELBO}
    \STATE $\U_\ell \sim \N{\M_\ell, \S^\u_\ell} = \qc{\U_\ell}{\U_{\ell-1}}$
    \STATE $\mathcal{L} \leftarrow \mathcal{L} + \log \qc{\U_\ell}{\U_{\ell-1}} - \log \Nc{\U_\ell}{\M_\ell, \S^\u_\ell}$
    \STATE \textcolor{gray}{Propagate the inputs using the sampled inducing outputs,}
    \STATE $\F_\ell \sim \pc{\F_\ell}{\U_\ell, \F_{\ell-1}, \U_{\ell-1}}$
  \ENDFOR
  \STATE $\mathcal{L} \leftarrow \mathcal{L} + \log \pc{\Y}{\F_{L+1}}$
\end{algorithmic}
\end{algorithm}


\subsubsection{Comparing our deep GP approximate posterior to previous work}
\label{sec:gi:graph-compare}
\begin{figure}[!tp]
    \centering
    \begin{tikzpicture}[latent/.append style={minimum size=1cm}]
    \node[latent] (fl) {$\F_\ell$};
    \node[latent, right=of fl] (flp) {$\F_{\ell+1}$};
    \node[latent, left=of fl] (flm) {$\F_{\ell-1}$};
    \node[latent, above=of fl] (fnl) {$\Fn_\ell$};
    \node[latent, above=of fnl] (ul) {$\U_\ell$};
    \node[latent, right=of fnl] (fnlp) {$\Fn_{\ell+1}$};
    \node[latent, right=of ul] (ulp) {$\U_{\ell+1}$};
    \node[latent, left=of fnl] (fnlm) {$\Fn_{\ell-1}$};
    \node[latent, left=of ul] (ulm) {$\U_{\ell-1}$}; 
    \edge {ulm} {fnlm};
    \edge {ul} {fnl};
    \edge {ulp} {fnlp};
    \edge {fnl} {fl};
    \edge {fnlm} {flm};
    \edge {fnlp} {flp};
    \edge {flm} {fl};
    \edge {fl} {flp};
    \draw[dashed, ->] (-3.5cm, 0cm) -- (-2.5cm, 0cm);
    \draw[dashed, ->] (2.5cm, 0cm) -- (3.5cm, 0cm);
    \node[above left=of ulm] {\textbf{A}};
    \end{tikzpicture}

    \begin{tikzpicture}[latent/.append style={minimum size=1cm}]
    \node[latent] (fl) {$\F_\ell$};
    \node[latent, right=of fl] (flp) {$\F_{\ell+1}$};
    \node[latent, left=of fl] (flm) {$\F_{\ell-1}$};
    \node[latent, above=of fl] (fnl) {$\Fn_\ell$};
    \node[latent, above=of fnl] (ul) {$\U_\ell$};
    \node[latent, right=of fnl] (fnlp) {$\Fn_{\ell+1}$};
    \node[latent, right=of ul] (ulp) {$\U_{\ell+1}$};
    \node[latent, left=of fnl] (fnlm) {$\Fn_{\ell-1}$};
    \node[latent, left=of ul] (ulm) {$\U_{\ell-1}$}; 
    \edge {ulm} {fnlm};
    \edge {ulm} {fnl};
    \edge {ul} {fnl};
    \edge {ulp} {fnlp};
    \edge {ul} {fnlp};
    \edge {fnl} {fl};
    \edge {fnlp} {flp};
    \edge {fnlm} {flm};
    \edge {flm} {fl};
    \edge {fl} {flp};
    \draw[dashed, ->] (-3.5cm, 0cm) -- (-2.5cm, 0cm);
    \draw[dashed, ->] (2.5cm, 0cm) -- (3.5cm, 0cm);
    \draw[dashed, ->] (-3.5cm, 3.5cm) -- (-2.35cm, 2.35cm);
    \draw[dashed, ->] (2.35cm, 3.65cm) -- (3.5cm, 2.5cm);
    \node[above left=of ulm] {\textbf{B}};
    \end{tikzpicture}

    \begin{tikzpicture}[latent/.append style={minimum size=1cm}]
    \node[latent] (fl) {$\F_\ell$};
    \node[latent, right=of fl] (flp) {$\F_{\ell+1}$};
    \node[latent, left=of fl] (flm) {$\F_{\ell-1}$};
    \node[latent, above=of fl] (fnl) {$\Fn_\ell$};
    \node[latent, above=of fnl] (ul) {$\U_\ell$};
    \node[latent, right=of fnl] (fnlp) {$\Fn_{\ell+1}$};
    \node[latent, right=of ul] (ulp) {$\U_{\ell+1}$};
    \node[latent, left=of fnl] (fnlm) {$\Fn_{\ell-1}$};
    \node[latent, left=of ul] (ulm) {$\U_{\ell-1}$}; 
    \edge {ulm} {fnlm}; 
    \edge {ulm} {fnl};
    \edge {ulm} {ul};
    \edge {ul} {fnl};
    \edge {ulp} {fnlp};
    \edge {ul} {fnlp};
    \edge {ul} {ulp};
    \edge {fnl} {fl};
    \edge {fnlp} {flp};
    \edge {fnlm} {flm};
    \edge {flm} {fl};
    \edge {fl} {flp};
    \draw[dashed, ->] (-3.5cm, 0cm) -- (-2.5cm, 0cm);
    \draw[dashed, ->] (2.5cm, 0cm) -- (3.5cm, 0cm);
    \draw[dashed, ->] (-3.5cm, 3.5cm) -- (-2.35cm, 2.35cm);
    \draw[dashed, ->] (2.35cm, 3.65cm) -- (3.5cm, 2.5cm);
    \draw[dashed, ->] (-3.5cm, 4cm) -- (-2.5cm, 4cm);
    \draw[dashed, ->] (2.5cm, 4cm) -- (3.5cm, 4cm);
    \node[above left=of ulm] {\textbf{C}};
    \end{tikzpicture}
    \caption{
      Comparison of the graphical models for three approaches to inference in deep GPs: \textbf{A)} \citet{salimbeni2017doubly}, \textbf{B)} \citet{ustyuzhaninov2019compositional}, and \textbf{C)} ours. 
      \label{fig:gi:graph-compare}
    }
\end{figure}
The standard approach to inference in deep GPs (c.f. Sec.~\ref{sec:back:dgps}) involves ``local'' inducing points $\Z_{\ell-1}$, defined at every layer, and an approximate posterior over $\slpo{\U_\ell}$ that is factorised over layers,
\begin{align}
  \nonumber
  \pc{\U_\ell}{\Z_{\ell-1}} &= \prodln \Nc{\u_\lambda^\ell}{\0, \Kfl{\Z_{\ell-1}}},\\
  \label{eq:gi:dgpU}
 \qc{\slpo{\U_\ell}}{\slpo{\Z_{\ell-1}}} &= \prodlayers \prodln \N{\u_\lambda^\ell; \m^\ell_\lambda, \S^\ell_\lambda}.
\end{align}
In particular, \citet{salimbeni2017doubly} learn a set of local inducing locations $\slpo{\Z_{\ell-1}}$ and propagate the data through the model using the learned approximate posterior over $\U_\ell$ and the model's prior conditional, $\pc{\F_\ell}{\U_\ell, \F_{\ell-1}, \Z_{\ell-1}}$.
Through the prior conditional, we can think of the approximate posterior over $\slpo{\U_\ell}$ as inducing an approximate posterior over underlying infinite-dimensional processes $\Fn_\ell$ at each layer, which are implicitly used to propagate the data through the network via $\F_\ell = \Fn_\ell(\F_{\ell-1})$.
We show a graphical model summarising the standard approach in Fig.~\ref{fig:gi:graph-compare}A. 
While, as \citet{salimbeni2017doubly} point out, the function values $\slpo{\F_\ell}$ are correlated, the functions $\slpo{\Fn_\ell}$ themselves are independent across layers. 
We note that for BNNs, this is equivalent to having a posterior over weights that factorises across layers: the activities $\slpo{\F_\ell}$ will be correlated even though the weights are not.

One approach to introduce dependencies across layers for the functions would be to introduce the notion of global inducing points, propagating the initial $\U_0$ through the model. 
In fact, \citet{ustyuzhaninov2019compositional} independently proposed this approach to introducing dependencies, using a toy problem to motivate the approach.
They kept the form of the approximate posterior the same as the standard approach (Eq.~\ref{eq:gi:dgpU}).
If our goal is to introduce dependencies across layers, however, this would seem inappropriate, as the form of the approximate posterior implies a level of independence between layers.
We show the corresponding graphical model for this approach in Fig.~\ref{fig:gi:graph-compare}B. 
The graphical model shows that as adjacent functions $\Fn_\ell$ and $\Fn_{\ell+1}$ share the parent node $\U_\ell$, they are in fact dependent. 
However, non-adjacent functions do not share any parent nodes, and so are independent: this can be seen by considering the d-separation criterion \citep{pearl1988probabilistic} for $\Fn_{\ell-1}$ and $\Fn_{\ell+1}$, which have parents $(\U_{\ell-2}, \U_{\ell-1})$ and $(\U_{\ell}, \U_{\ell+1})$ respectively.

Our approach, by contrast, determines the form of the approximate posterior over $\U_\ell$ by performing Bayesian regression using $\U_{\ell-1}$ as input to that layer's GP, where the output data is $\V_\ell$ with precision $\La_\ell$. 
This results in a posterior that depends on the previous layer, $\qc{\U_\ell}{\U_{\ell-1}}$. We show the corresponding graphical model in Fig.~\ref{fig:gi:graph-compare}C. 
From this graphical model it is straightforward to see that our approach results in a posterior over functions that are correlated across all layers.

\subsection{Asymptotic complexity}
\label{sec:gi:complexity}
In the deep GP case, the complexity for global inducing is exactly that of standard inducing point Gaussian processes, i.e., $\mathcal{O}(L\tilde{N}M^2\nu + LM^3\nu)$ where $M$ is the number of inducing points, $\tilde{N}$ is the size of the data batch, $L$ is the depth, and $\nu$ is the width, assumed for simplicity to be constant across layers.
The first term comes from propagating the data through the model, whereas the second term comes from computing and sampling the posterior over $\U_\ell$ based on the inducing points (e.g., inverting the covariance).

In the fully-connected BNN case, we have three terms, $\mathcal{O}(L\nu^3 + LM \nu^2 + L\tilde{N} \nu^2)$. 
The first term arises from taking the inverse of the covariance matrix in Eq.~\eqref{eq:gi:Qw|v}, but is also the complexity e.g., for propagating the inducing points from one layer to the next (Eq.~\ref{eq:gi:bnn_pseudo_prop}).
The second term comes from computing that covariance in Eq.~\eqref{eq:gi:Qw|v}, by taking the product of input features with themselves.
The final term comes from multiplying the training inputs/minibatch by the sampled weights.

\section{Results}
\begin{figure*}
    \centering
    \includegraphics{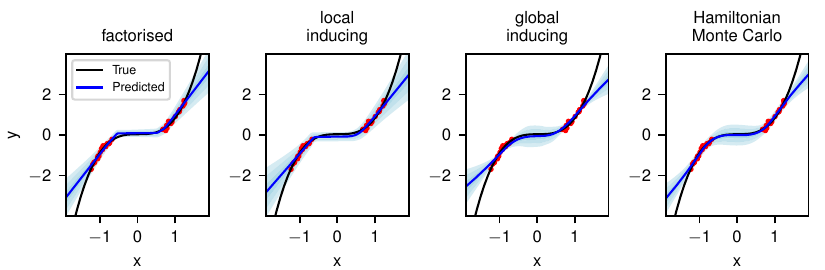}
    \caption{
      Predictive distributions on the toy dataset. Shaded regions represent one standard deviation.
      \label{fig:gi:toy}
    }
\end{figure*}
We describe our experiments and results to assess the performance of global inducing points (`gi') against local inducing points (`li') and the fully factorised/mean field (`fac') approximation family. We additionally consider models where we use one method up to the last layer and another for the last layer, which may have computational advantages; we denote such models `method1 $\rightarrow$ method2'. 

\subsection{Uncertainty in 1D regression}
\begin{figure}[t]
    \centering
    \includegraphics{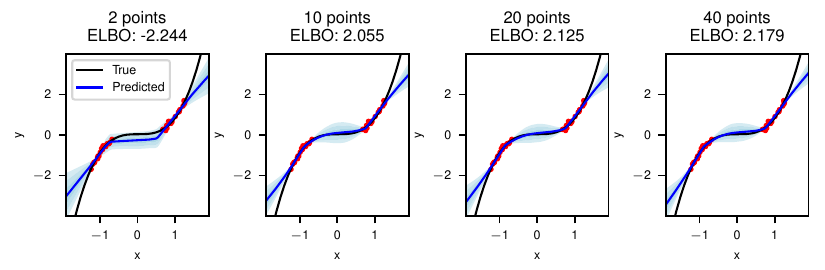}
    \caption{Predictive distributions on the toy dataset as the number of inducing points changes.}
    \label{fig:gi:toy-points}
\end{figure}
\label{sec:gi:toy}
We demonstrate the use of local and global inducing point methods in a toy 1-D regression problem, comparing it with fully factorised VI and Hamiltonian Monte Carlo \citep[HMC;][]{neal2011mcmc}. Following \citet{hernandez2015probabilistic}, we generate 40 input-output pairs $(x, y)$ with the inputs $x$ sampled i.i.d. from $\mathcal{U}([-4, -2]\cup[2, 4])$ and the outputs generated by $y = x^3 + \epsilon$, where $\epsilon \sim \mathcal{N}(0, 3^2)$. We then normalized the inputs and outputs. Note that we have introduced a `gap' in the inputs, following recent work \citep{foong2019between, yao2019quality, foong2019pathologies} that identifies the ability to express `in-between' uncertainty as an important quality of approximate inference algorithms. We evaluated the inference algorithms using fully-connected BNNs with 2 hidden layers of 50 ReLU hidden units, using the NealPrior. For the inducing point methods, we used 100 inducing points per layer.

The predictive distributions for the toy experiment can be seen in Fig.~\ref{fig:gi:toy}. We observe that of the variational methods, the global inducing method produces predictive distributions closest to HMC, with good uncertainty in the gap. Meanwhile, factorised and local inducing fit the training data, but do not produce reasonable error bars, demonstrating an important limitation of methods lacking correlation structure between layers.

\subsubsection{Exploring the effect of the number of inducing points}
\label{sec:gi:numinducing}

We now briefly consider the effect of changing the number of inducing points, $M$, used in global inducing. We consider the same toy problem and plot predictive posteriors obtained with global inducing as the number of inducing points increases from 2 to 40 (noting that in Fig.~\ref{fig:gi:toy} we used 100 inducing points). 

We plot the results of our experiment in Fig.~\ref{fig:gi:toy-points}. While two inducing points are clearly not sufficient, we observe that there is remarkably very little difference between the predictive posteriors for 10 or more inducing points. This observation is reflected in the ELBOs per datapoint (listed above each plot), which show that adding more points beyond 10 gains very little in terms of closeness to the true posterior. 

However, we note that this is a very simple dataset: it consists of only two clusters of close points with a very clear trend. Therefore, we would expect that for more complex datasets more inducing points would be necessary. We leave a full investigation of how many inducing points are required to obtain a suitable approximate posterior, such as that found in \citet{burt2020convergence} for sparse GP regression, to future work.

We provide additional toy experiments looking at the compositional uncertainty \citep{ustyuzhaninov2019compositional} in both BNNs and DGPs for 1D regression in Appendix~\ref{sec:gi:app:compositional}.

\subsection{Depth dependence in deep linear networks}
\begin{figure*}
    \centering
    \includegraphics{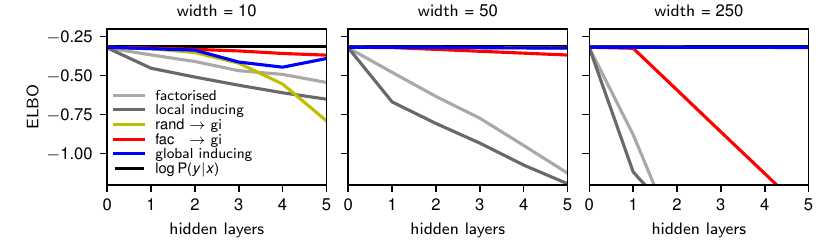}
    \caption{ELBO for different approximate posteriors as we change network depth and width on a dataset generated using a linear Gaussian model.  The rand $\rightarrow$ gi line lies behind the global inducing line in width $=50$ and width $=250$.
    \label{fig:gi:linear}}
\end{figure*}
The lack of correlations between layers might be expected to become more problematic in deeper networks.
To isolate the effect of depth on different approximate posteriors, we considered deep linear networks trained on data generated from a toy linear model: 5 input features were mapped to 1 output feature, where the 1000 training and 100 test inputs are drawn IID from a standard Gaussian, and the true outputs are drawn using a weight-vector drawn IID from a Gaussian with variance $1/5$, and with noise variance of $0.1$.
We can evaluate the model evidence under the true data generating process which forms an upper bound (in expectation) on the model evidence and ELBO for all models.

We found that the ELBO for methods that factorise across layers -- factorised and local inducing -- drops rapidly as networks get deeper and wider (Fig.~\ref{fig:gi:linear}).
This is undesirable behaviour, as we know that wide, deep networks are necessary for good performance on difficult machine learning tasks.
In contrast, we found that methods with global inducing points at the last layer decay much more slowly with depth, and perform better as networks get wider.
Remarkably, global inducing points gave good performance even with lower-layer weights drawn at random from the prior, which is not possible for any method that factorises across layers.
We believe that fac $\rightarrow$ gi performed poorly at width $=250$ due to optimization issues since rand $\rightarrow$ gi achieves better performance despite being a special case of fac $\rightarrow$ gi.

\subsection{Regression benchmark: UCI}\label{sec:gi:uci}
\begin{figure*}
    \centering
    \includegraphics{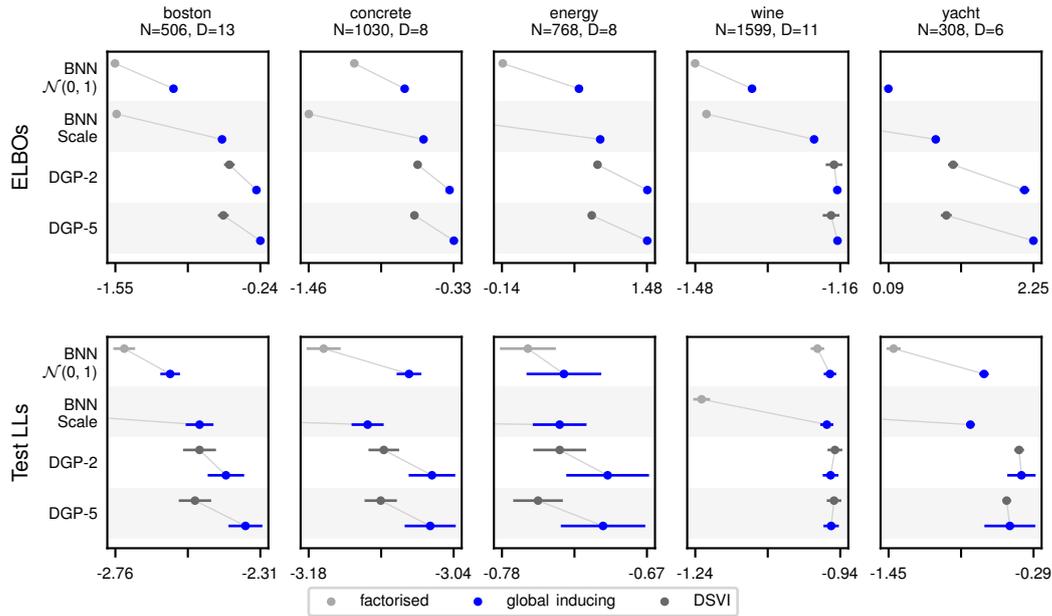}
    \caption{Average test log likelihoods for BNNs on the UCI datasets (in nats). Error bars represent one standard error. Shading represents different priors. We connect the factorised models with the fac $\rightarrow$ gi models with a thin grey line as an aid for easier comparison. Further to the right is better.
    \label{fig:gi:uci_bnn}}
\end{figure*}
We benchmark our methods on the UCI datasets used in \citet{hernandez2015probabilistic}, popular benchmark regression datasets for BNNs and DGPs.
Following the standard approach \citep{Gal2015DropoutB}, each dataset uses 20 train-test `splits' (except for protein with 5 splits) and the inputs and outputs are normalised to have zero mean and unit standard deviation.
We focus on the five smallest datasets, as we expect Bayesian methods to be most relevant in small-data settings (see App.~\ref{sec:gi:app:ucibnn} and\ref{sec:gi:app:ucidgp} for all datasets).
We consider two-layer fully-connected ReLU networks, using the fully factorised and global inducing approximating families, as well as two- and five-layer DGPs with doubly-stochastic variational inference (DSVI) \citep{salimbeni2017doubly} and global inducing.
For the BNNs, we consider the standard $\mathcal{N}(0, 1)$ prior and ScalePrior.

We display ELBOs and average test log likelihoods for the un-normalised data in Fig.~\ref{fig:gi:uci_bnn}, where the dots and error bars represent the means and standard errors over the test splits, respectively.
We observe that global inducing obtains better ELBOs than factorised and DSVI in almost every case, indicating that it does indeed approximate the true posterior better (since the ELBO is the marginal likelihood minus the KL to the posterior).
While this is the case for the ELBOs, this does not always translate to a better test log likelihood due to model misspecification, as we see that occasionally DSVI outperforms global inducing by a very small margin.
The very poor results for factorised on ScalePrior indicate that it has difficulty learning useful prior hyperparameters for prediction \citep[see also][]{blundell2015weight}, which is due to the looseness of its bound to the marginal likelihood.
We provide experimental details, as well as additional results with additional architectures, priors, datasets, and RMSEs, in Appendices~\ref{sec:gi:app:ucibnn} and \ref{sec:gi:app:ucidgp}, for BNNs and DGPs, respectively.

\begin{landscape}
\begin{table*}
  \caption{CIFAR-10 classification accuracy. The first block shows our main results without data augmentation or tempering with SpatialIWPrior, (with ScalePrior in brackets).  
  The next block shows our results with data augmentation and tempering on with a larger ResNet18 with SpatialIWPrior.
  The subsequent block shows comparable past results, from GPs and BNNs.  
  The final block show non-comparable (sampling-based) methods.  Dashes indicate that the figures were either not reported, are not applicable.
  The best results are highlighted in bold.
  The time is reported per epoch with ScalePrior and for MNIST, rather than CIFAR-10 because of a known performance bug in the convolutions required in Sec.~\ref{sec:gi:efficient_conv} with $32 \times 32$ (and above) images \url{https://github.com/pytorch/pytorch/issues/35603}.}
  \label{tab:gi:cifar10}
  \centering
  \begin{tabular}{llcccc}
    \toprule
    & & test log like. & accuracy (\%) & ELBO & time\\
    \midrule
    & factorised & -0.58 (-0.66) & 80.27 (77.65) & -1.06 (-1.12) & 19 s \\
    no tempering or & local inducing & -0.62 (-0.60) & 78.96 (79.46) & -0.84 (-0.88) & 33 s \\
    data augmentation & fac $\rightarrow$ gi & -0.49 (-0.56) & 83.33 (81.72) & -0.91 (-0.96) & 25 s \\
    & \textbf{global inducing} & \textbf{-0.40 (-0.43)} & \textbf{86.70 (85.73)} &  \textbf{-0.68 (-0.75)} & 65 s \\
    \midrule
    with tempering and & factorised & -0.39 & 87.52 & --- & --- \\
    data augmentation & \textbf{fac $\rightarrow$ gi} & \textbf{-0.24} & \textbf{92.41} & --- & --- \\
    \midrule
    & \citet{shi2019sparse} & --- & $80.30\%$ & ---\\
    VI prior work & \citet{li2019enhanced} & --- & $81.40\%$ & ---\\
    & \citet{shridhar2019comprehensive} & --- & $73\%$ & ---\\
    \midrule
    sampling prior work& \citet{wenzel2020good} & $-0.35$ & $88.50\%$ & ---\\
    \bottomrule
  \end{tabular}
\end{table*}
\end{landscape}

\subsection{Convolutional benchmark: CIFAR-10}
For CIFAR-10, we considered a ResNet-inspired model consisting of conv2d-relu-block-avgpool2-block-avgpool2-block-avgpool-linear, where the ResNet blocks consisted of a shortcut connection in parallel with conv2d-relu-conv2d-relu, using 32 channels in all layers.
In all our experiments, we used no data augmentation and 500 inducing points.
Our training scheme (see App.~\ref{sec:gi:app:deets}) ensured that our results did not reflect a `cold posterior' \ref{sec:back:temp}. 
Our results are shown in Table~\ref{tab:gi:cifar10}. 
We achieved remarkable performance of $86.7\%$ predictive accuracy, with global inducing points used for all layers, and with a spatial inverse Wishart prior on the weights.
These results compare very favourably with comparable Bayesian approaches, i.e.\ those without data augmentation or posterior sharpening: past work with deep GPs obtained $80.3\%$ \citep{shi2019sparse}, and work using infinite-width neural networks to define a GP obtained $81.4\%$ accuracy \citep{li2019enhanced}.
Remarkably, with only 500 inducing points we are approaching the accuracy of sampling-based methods \citep{wenzel2020good} (which used data augmentation), which are in principle able to more closely approximate the true posterior.
Furthermore, we see that global inducing performs the best in terms of ELBO (per datapoint) by a wide margin, demonstrating that it gets far closer to the true posterior than the other methods.
We provide additional results on uncertainty calibration and out-of-distribution detection in Appendix~\ref{sec:gi:app:uncertainty}.

Finally, while we have focused this work on achieving good results with a fully principled, Bayesian approach, we briefly consider training a full ResNet-18 \citep{he2016deep} using more popular techniques such as data augmentation and tempering. 
While data augmentation and tempering are typically viewed as clouding the Bayesian perspective, there is work attempting to formalise both within the context of modified probabilistic generative models \citep{aitchison2021statistical,nabarro2021}.
Using a cold posterior with a factor of 20 reduction on the KL term, with horizontal flipping and random cropping, we obtained a test accuracy of $87.52\%$ and a test log likelihood of $-0.39$ nats with a standard fully factorised Gaussian posterior using SpatialIWPrior.
Using fac $\rightarrow$ gi, we obtain a significant improvement of $92.41\%$ test accuracy and a test log likelihood of $-0.24$ nats.
We attempted to train a full global inducing model; however, we encountered difficulties in scaling the method to the large widths of ResNet-18 layers.
We believe this presents an avenue for fruitful future work in scaling global inducing point posteriors.

\section{Related work}
\subsection{Prior related work}

\subsubsection{Bayesian neural networks}
Several works have attempted to scale structured Gaussian approximate posteriors to BNNs.
\citet{ghosh2017model} use matrix-variate Gaussians that factorize across layers, along with a hierarchical prior, to prune network weights while obtaining extra flexibility.
By contrast, \citet{zhang2017noisy} use an information-geometric optimization routine based on natural gradient methods \citep{amari1998natural} to fit a matrix-variate Gaussian to the weights in each layer, allowing for correlations within a layer.
Similarly, \citet{mishkin2018slang} use natural gradients to optimize a low-rank-plus-diagonal Gaussian posterior over \emph{all} the weights in a network; however, they did not demonstrate whether it could scale to large BNNs, restricting their focus to smaller problems.
\citet{farquhar2020radial} and \citet{oh2020radial} decompose the weight matrices into radial and directional components, and sample each independently, leading to correlations between the weight matrices and rows thereof, respectively.
However, both of these still retain factorization across layers.
\citet{tomczak2018neural} considered ensembles of mean-field approximate posteriors, showing that it outperforms standard MFVI.

In perhaps the most relevant work to ours for BNNs, \citet{louizos2016structured} attempted to use pseudo-data along with matrix variate Gaussians to form an approximate posterior for BNNs. 
Their approach factorises across layers, thus missing the important layerwise correlations that we obtain.
Moreover, they encountered an important limitation: the BNN prior implies that $\U_\ell$ is low-rank and it is difficult to design an approximate posterior capturing this constraint.
As such, they were forced to use $M < \nu_\ell$ inducing points, which is particularly problematic in the convolutional, global-inducing case where there are many patches (points) in each inducing image input.
Finally, it is not clear that the ELBO they provide is a valid lower bound on the log marginal likelihood, and while they show connections between BNNs and DGPs, they do not show how their work can be used in DGPs.
Nevertheless, it would be interesting to further consider connections between their work and ours.

A few works have attempted to introduce even more flexibility into their approximate posteriors using neural networks to help parameterize their approximate posteriors.
For instance, \citet{louizos2017multiplicative} use hierarchical variational inference \citep{ranganath2016hierarchical} with normalizing flows \citep{rezende2015variational} to introduce explicit correlations between neurons.
\citet{krueger2017bayesian} take a similar approach, but avoid using hierarchical variational inference by learning deterministic weights which are then scaled by the normalizing flow.
We note that this risks overfitting for the reasons pointed out in the previous chapter.
Finally, \citet{pawlowski2017implicit} use implicit inference to enable a neural network to transform a noise source into a direct approximate posterior over weights, i.e., they use a hypernetwork.
However, to enable this approach to scale to larger networks, they had to introduce layerwise factorization.
Moreover, KL estimation with implicit inference is known to be theoretically difficult for high-dimensional problems \citep{mcallester2020formal}.
Finally, while these approaches may be flexible, they tend to ignore the natural structure of the model, and therefore may miss out on important correlations through their structural assumptions.

Recently, function-space approaches have become popular \citep{ma2019variational, sun2018functional}.
These methods attempt to perform inference on the stochastic processes implied by BNNs.
However, these rely on implicit inference, which as mentioned above has theoretical limitations.
Moreover, these approaches often attempt to minimize the KL divergence of the BNN to a Gaussian process, which is in fact infinite \citep{burt2020understanding}.
By contrast, our global inducing approach can be seen as a compromise between weight-space and function-space inference, whereby we perform tractable inference on the pre-activations.

Finally, we note that some prior work on BNNs reports better perfomance than ours on datasets such as CIFAR-10.
However, to the best of our knowledge, no variational Bayesian method outperforms ours without modifying the BNN model, some form of posterior tempering \citep{wenzel2020good}, or data augmentation \citep{zhang2017noisy,bae2018eigenvalue,osawa2019practical,ashukha2020pitfalls}. 
All of these may cloud the Bayesian perspective, whereas we attempt to retain the full Bayesian perspective as far as possible.

\paragraph{Priors} As part of our contribution is to propose new priors for BNNs, we highlight some priors proposed in other works.
For a more comprehensive overview, we refer the interested reader to \citet{fortuin2022priors}.
As previously mentioned, most prior works propose an independent $\N{0, 1}$ prior over the weights \citep{blundell2015weight, louizos2017multiplicative, zhang2020cyclical}.
In order to use their more expressive structured posteriors, some of the works mentioned above have had to use different priors to ensure that the KLs can be computed.
For instance, \citet{oh2020radial} had to use a modified prior mimicking the structure of its posterior.
It is also interesting to notice that while early work \citep{hinton1993keeping, graves2011practical} proposed priors with learnable hyperparameters, this has largely been abandoned in more recent work, perhaps in part because of the comment by \citet{blundell2015weight} that they were unable to learn prior variances.

A few works have proposed more flexible priors.
\citet{hinton1993keeping} proposed mixture of Gaussian priors with learnable hyperparameters, while \citet{blundell2015weight} propose a scale mixture of two Gaussians, both centered at zero.
Meanwhile, \citet{ghosh2017model} introduce horseshoe priors, which enable pruning and thereby a degree of automatic model selection.
\citet{wu2019deterministic} propose a prior nearly identical to our ScalePrior; however, they do not perform inference over the scale parameter, instead learning it through the marginal likelihood.
On the other hand, \citet{cui2020informative} show how a hierarchical prior such as ScalePrior can be used to induce sparsity.
Finally, \citet{atanov2019deep} propose an implicit prior for convolutional kernels, trained on kernels trained from adjacent tasks; however, we expect this to have difficulties with implicit inference, as well as being costly.

\subsubsection{Deep Gaussian processes}

The most similar prior work for DGPs to ours is \citet{ustyuzhaninov2019compositional}, who propose two methods for introducing correlations between layers.
In the first, they use local inducing points and assume the factorization
\begin{align*}
    \q{\slpo{\U}} = \prodlayers \qc{\U_\ell}{\U_{\ell-1}},
\end{align*}
which results in a block tridiagonal structure in the full precision matrix.
Their second approach is the most similar to ours: they use a global inducing scheme, which they refer to as ``inducing points as inducing locations.''
However, as described in Sec.~\ref{sec:gi:dgp-ext}, they choose a factorizing posterior over $\slpo{\U_\ell}$, which means that the functions for non-adjacent layers are conditionally independent.
By contrast, our approximate posteriors have marginal dependencies across $\U_\ell$ and functions at all layers, and are capable of capturing the optimal top-layer posterior.

Another line of work comes from applying implicit inference through adversarial variational Bayes \citep{mescheder2017adversarial}, seen in \citet{yu2019implicit}.
However, this comes with the difficulties mentioned above with implicit inference: indeed, the authors find that they had to use parameter sharing for the \emph{variational parameters} to avoid overfitting.
We believe this is actually a sign that the inference does not work well, as variational parameters should be protected from overfitting.
Finally, \citet{havasi2018inference} proposed using stochastic gradient Hamiltonian Monte Carlo \citep{chen2014stochastic} for inference in DGPs.
While this is theoretically very expressive, it is also costly, difficult to tune,\footnote{Personal communication with the first author.} and it is difficult to assess both its convergence and how accurate the samples are (cf. Sec.~\ref{sec:back:bnns}). 

\subsection{Concurrent and subsequent related work}
\subsubsection{Bayesian neural networks}
A few works have contemporaneously or since considered hierarchical or structured VI for BNNs.
\citet{dusenberry2020efficient} propose a scheme whereby for each weight matrix $\W$, they perform inference over vectors $\mathbf{r}$ and $\mathbf{s}$, so that $\W = \W' \circ \mathbf{r} \transpose{\mathbf{s}}$, where $\circ$ represents elementwise multiplication, and $\W'$ is learned by maximum likelihood.
They frame this scheme as performing hierarchical variational inference, and they are able to obtain impressive results while scaling their method to ImageNet \citep{imagenet}.
While they do not use tempering, they still use data augmentation.
Moreover, it is likely that this scheme is prone to overfitting, as discussed in the previous chapter; indeed, there may be evidence for this, as they obtain worse results when using a mixture distribution with the correct log likelihood calculation (c.f. Sec. 3.4 of their paper).
However, the overfitting may be mitigated by their choice of variational expectation maximization as a training procedure, which might prevent the co-adaptation of variational parameters and hyperparameters necessary for overfitting \citep[see e.g., ][]{schwoebel2021last}.

Alternatively, \citet{tomczak2020efficient} investigate the use of low-rank-plus-diagonal variational posteriors for large BNNs, where they factorize the approximate posterior over layers and derive a local reparameterization trick \citep{kingma2015variational} for the posterior.
They find that they are able to obtain gains over MFVI if they fix the added diagonal variance term. 
However, when they allow for the diagonal term to be learned, these gains disappear.
We believe that this indicates that the approximate posterior is not sufficiently flexible to achieve good results.
\citet{havasi2021sampling} take an alternative approach to structured VI by proposing a method that locally refines a learned MFVI posterior using auxiliary variables.
While potentially quite flexible, it is not clear how scalable this method is.

A few works have attempted to use inducing point-related methods to improve VI for BNNs.
\citet{ritter2021sparse} extend Matheron's rule for GPs \citep{wilson2020efficiently} to matrix-variate Gaussians, and use this to augment the weight space with a set of ``inducing weights''.
They then show that they can significantly reduce the number of parameters needed for VI by parameterising a posterior over these inducing weights, which encode a sparse representation of the true weights.
This approach can then be used to sparsify the BNN.
While their posterior is currently limited by being factorized across layers, this is a fascinating direction, and it would be very interesting to understand more how it could be related to our method.
\citet{morales2021activation} take a different approach, and attempt to improve BNN models by incorporating GP layers in a BNN, which they perform inference with using local inducing points.
However, they learn the weights deterministically, again risking overfitting.

Finally, \citet{bui2021biases} uses annealed importance sampling \citep{neal2001annealed} to assess the tightness of the ELBO to the LML for multiple posteriors at different hyperparameter settings.
Of the VI methods they tried -- mean-field, full-covariance, importance-weighted VI \citep{burda2016importance}, the thermodynamic variational objective \citep{masrani2019thermodynamic}, and ours - only ours gave ELBOs close the computed LML across different hyperparameter values and thereby exhibit limited bias in its hyperparameter selection.

\paragraph{Priors}
There has been a recent surge in interest in priors for BNNs; we highlight some of the main contributions.
First, \citet{karaletsos2020hierarchical} propose a hierarchical GP prior over network \emph{weights} to obtain more flexible and correlated priors.
In order to use this model, they propose a unique approximation scheme as well.
However, they do not demonstrate that it can scale to large architectures.

\citet{izmailov2021bayesian} use HMC to explore the effect of the prior in large-scale classification.
They argue that the choice of prior has a much more limited effect on the performance of the model than the choice of architecture.
While this may be true, they use filter response normalization layers \citep{singh2020filter}, which likely removes the effect of many of the changes in prior they make.
It would be interesting to see a similar analysis performed in the absence of these layers.

Meanwhile, \citet{tomczak2021collapsed} derive collapsed variational bounds for BNNs, whereby the inference over prior parameters is solved analytically, thus tightening the ELBO.
They show that this can significantly improve the performance of MFVI.
\citet{tran2022all} argue for functional priors that can mimic the inductive biases of GPs \citep[similar to][]{sun2018functional}, and show how to learn a weight prior that mimics GP priors by using Wasserstein distances.

Several works have also investigated the use of priors that encode correlations within convolutional kernels, similar to our SpatialIWPrior.
\citet{fortuin2021bayesian} hypothesize that the choice of prior might be partially responsible for the cold posterior effect (c.f. Sec.~\ref{sec:back:temp}), and present some evidence with correlated priors that this is the case.
\citet{pearce2020structured} also provided an initial investigation into the use of correlated priors in convolutional networks, while \citet{garriga2021correlated} explored the effect similar priors might have in infinitely wide networks, showing that they can be useful.

\subsubsection{Deep Gaussian processes}
Since our work, subsequent research has further developed and exploited the relationship between BNNs and DGPs.
\citet{dutordoir2021deep} demonstrate an equivalence between the neurons in a deterministic neural network and the mean values of inducing variables in local-inducing DGPs.
They use this equivalence to propose a training procedure whereby a neural network is trained to initialize the inducing variables of a DGP, demonstrating improved performance.
This approach can therefore be seen either as a way of introducing uncertainty into a neural network or providing a DGP with a better initialization.
On the theoretical side, \citet{pleiss2021limitations} take inspiration from the infinitely-wide neural network literature \citep{matthews2018gaussian, lee2017deep} and extend these results to DGPs, showing that DGPs effectively become shallow GP models in the wide limit.

A couple of works have also recently proposed new inference procedures for DGPs.
\citet{lindinger2020beyond} propose a joint multivariate Gaussian (with structured covariance) for the approximate posterior for all inducing variables at all layers, within a local inducing approach.
While the structure they introduce does improve scalability over a full covariance Gaussian, this approach still suffers from cubic complexity scaling with depth.
Motivated by a desire to reduce the number of total parameters to be trained in a DGP, \citet{jain2021subset} propose subset-of-data variational inference. 
With this approach, a subset of the data is chosen (e.g., by an approximation to k-means clustering) and fixed as the global inducing inputs.
The approximate posterior is otherwise essentially equivalent to the ``inducing points as inducing locations'' proposed by \citet{ustyuzhaninov2019compositional}.

\section{Conclusions}
We derived optimal top-layer variational approximate posteriors for BNNs and deep GPs, and used them to develop generic, scalable approximate posteriors.
These posteriors make use of \textit{global} inducing points, which are learned only at the bottom layer and are propagated through the network.
This leads to extremely flexible posteriors, which even allow the lower-layer weights to be drawn from the prior.
We showed that these global inducing variational posteriors lead to improved performance with better ELBOs, and state-of-the-art performance for variational BNNs on CIFAR-10.



\chapter[Deep Wishart processes]{Performing inference over Gram matrices with deep Wishart processes}\label{sec:dwp}  
\ifpdf
    \graphicspath{{DWP/Figs/Raster/}{DWP/Figs/PDF/}{DWP/Figs/}}
\else
    \graphicspath{{DWP/Figs/Vector/}{DWP/Figs/}}
\fi

In the previous chapter, we have seen how we can define more flexible scalable approximate posteriors that introduce correlations across layers for both BNNs and DGPs.
We achieved this by taking the structure of these deep models into account, applying the insights from optimal final-layer inference to the entire model.
However, BNNs and DGPs possess many symmetries that most approximate posteriors for variational inference --- including global inducing posteriors --- do not exploit.
As described in Sec.~\ref{sec:back:symm}, these symmetries have the potential to bias approximate posteriors to regions of the space which have little true posterior mass, despite variational inference's mode-seeking tendencies.
In this chapter, we explore how this behavior can be mitigated in deep models, with a focus on DGPs.
We do so by following \citet{aitchison2020deep} in viewing deep Bayesian models from a different viewpoint: instead of viewing them as defining priors over layered features, we argue that in some circumstances, they should be reframed as defining a prior over \emph{Gram matrices}.

This chapter is based on joint work with Laurence Aitchison, published in \citep*{ober2021variational}.
Additionally, Sec~\ref{sec:dwp:improving} has recently been extended and published in \citep*{ober2023improved}, as joint work with Ben Anson, Edward Milsom, and Laurence Aitchison.

\section{Introduction \& Motivation}

The successes of modern deep learning have highlighted that good performance on tasks such as image classification \citep{krizhevsky2012imagenet} requires deep models with lower layers that have the flexibility to learn good representations.
Up until very recently, this was only possible in feature-based methods such as neural networks.
Kernel methods do not have this flexibility because the kernel could be modified only using a few kernel hyperparameters.
However, with the advent of deep kernel processes \citep[DKPs;][]{aitchison2020deep}, we now have deep kernel methods that offer neural network-like flexibility in the kernel.
DKPs introduce this flexibility by taking the kernel from the previous layer, then sampling from a Wishart or inverse Wishart centered on that kernel, followed by a nonlinear transformation.
The sampling and nonlinear transformation steps are repeated multiple times to form a deep architecture.
Remarkably, deep Gaussian processes \citep[DGPs;][]{damianou2013deep,salimbeni2017doubly}, standard Bayesian NNs, infinite-width Bayesian NNs  \citep[neural network Gaussian processes or NNGPs;][]{lee2017deep,matthews2018gaussian,novak2018bayesian,garriga2018deep}, and infinite NNs with finite-width bottlenecks \citep{agrawal2020wide,aitchison2019bigger} can be written as DKPs \citep{aitchison2020deep}.
Indeed, for kernels that can be expressed in terms of operations on Gram matrices, \citet{aitchison2020deep} showed that a particular DKP, the \emph{deep Wishart process (DWP)}, induces a prior over functions which is equivalent to that of a DGP.
In a DGP, the random variables inferred in variational inference are the model's intermediate features, with kernels computed as a function of these features at each layer.
However, in a DWP, there are no features \textit{at all}. 
The only random variables are the positive semi-definite kernel matrices themselves, which are sampled directly from Wishart distributions: the DWP works entirely on the kernel matrices implied by the DGP's features.

\citet{aitchison2020deep} argued that DWPs should have considerable advantages over related feature-based models, because feature-based models have pervasive symmetries in the true posterior, which are difficult to capture in standard variational approximate posteriors.
For instance, as described in Sec.~\ref{sec:back:symm}, it is possible to permute rows and columns of weight matrices in a neural network, such that the activations at a given layer are permuted, but the network's overall input-output function remains the same \citep{mackay1992practical,sussmann1992uniqueness,bishop1995neural}.
These permutations result in network weights with exactly the same probability density under the true posterior, but with very different probability densities under standard variational approximate posteriors, which are generally unimodal.
However, these issues do not arise with DWPs, because all permutations of the hidden units correspond to the same kernel \citep[see Appendix D in][for more details]{aitchison2020deep}.

While \citet{aitchison2020deep} showed the equivalence between DWPs and DGPs, they were not able to perform inference in DWPs, as they were not able to find a sufficiently flexible distribution over positive semi-definite matrices to form the basis of an approximate posterior.
Instead, they were forced to work with a different DKP: the deep \emph{inverse} Wishart processes (DIWPs), which was easier because the inverse Wishart itself forms a suitable approximate posterior.
While the DIWP also avoids using features, it does not correspond directly to an already-established Bayesian model.
Moreover, we show in Sec.~\ref{sec:dwp:diwpcomp} that the inverse Wishart is not particularly well-suited to inducing point approximate posteriors.
With this in mind, enabling accurate inference for the DWP is conceptually important, as it allows for a direct comparison between feature-based and kernel-based inference with \emph{equivalent} models.
In this work, we show how to create a sufficiently flexible approximate posterior for DWPs, thereby enabling us to compare directly to their equivalent DGPs.
In particular, our contributions are:

\begin{itemize}
  \item We develop a new family of flexible distributions over positive semi-definite matrices by generalizing the Bartlett decomposition (Sec.~\ref{sec:dwp:gen_wishart}).
  \item We use this distribution to develop an effective approximate posterior for the deep Wishart process which incorporates dependency across layers (Sec.~\ref{sec:dwp:approx_post}).
  \item We develop a doubly stochastic inducing-point inference scheme for the DWP. While the derivation mostly follows that for deep inverse Wishart processes \citep{aitchison2020deep}, we need to develop a novel scheme for sampling the training/test points conditioned on the inducing points, as the dependence structure of the DWP poses new challenges for this task, in comparison to the DIWP (Sec.~\ref{sec:dwp:dsvi}).
  \item We empirically compare DGP and DWP inference under \emph{equivalent function-space priors} (up to kernel hyperparameters, which we allow to vary between models). This was not possible in \citet{aitchison2020deep} as they only derived an inference scheme for the deep inverse Wishart processes, whose prior is not equivalent to a DGP prior. 
\end{itemize}

We now describe in depth the equivalence between DWPs and previously-proposed deep Bayesian models such as BNNs (with Gaussian priors) and DGPs (Sec.~\ref{sec:dwp:wishdgp}), and give motivation for why we would prefer the DWP over DGPs (Sec.~\ref{sec:dwp:symmetry} and~\ref{sec:dwp:dwpdgpelbo}) and the previously-proposed DIWP (Sec.~\ref{sec:dwp:diwpcomp}).

\subsection{A Wishart formulation of BNNs \& DGPs}\label{sec:dwp:wishdgp}
To help address the problem of performing variational inference in deep models, \citet{aitchison2020deep} reframes BNNs and DGPs as specific instances of \emph{deep kernel processes}, namely \emph{deep Wishart processes}.
In this section, we follow and extend their argument to show how this is possible.
Recall the general DGP model, which defines a prior over features $\slpo{\F_\ell}$ (with $\F_\ell \in \reals^{N \times \nu_\ell}$) as
\begin{align*}
    \pc{\F_\ell}{\F_{\ell-1}} &= \prodln \Nc{\f_\lambda^\ell}{\m_\ell\b{\F_{\ell-1}}, \Kfl{\F_{\ell-1}}},
\end{align*}
where $\m_{\ell}(\cdot)$ is the layer's mean function, $\Kfl{\cdot}$ its covariance function. 
We also define a likelihood $\pc{\Y}{\F_{L+1}}$ for the model.
Following \citet{aitchison2020deep}, we show how this DGP model can be expressed as a \emph{deep Wishart process}. We first consider the $N\times N$ Gram matrices defined as
\begin{align*}
  \G_\ell &= \frac{1}{\nu_\ell} \F_\ell \transpose{\F}_\ell = \frac{1}{\nu_\ell} \sum_{\lambda=1}^{\nu_\ell} \f^\ell_{\lambda} \transpose{\f^\ell_{\lambda}}.
\end{align*}
Under the prior above, it is clear that $\f^\ell_\lambda$ are IID and multivariate-Gaussian distributed conditioned on the features at the previous layer. 
Assuming a zero mean function, $\m_\ell\b{\cdot} = 0$, $\G_\ell$ follows the definition of the \emph{Wishart distribution}:
\begin{restatable}[The Wishart distribution; {\citealp[][Eq. 3.2]{srivastava2003singular}}]{defi}{defwish}
\label{defi:dwp:wish}
Consider a matrix $\F \in \reals^{N \times \nu}$ with columns $\f_\lambda \in \reals^{N} \sim \N{\0, \S}$. Then, the product
\begin{align*}
    \G \coloneqq \F \transpose{\F} = \sum_{\lambda = 1}^\nu \f_\lambda \transpose{\f}_\lambda
\end{align*}
is said to be distributed according the \emph{Wishart distribution} with \emph{scale matrix} $\S$ and $\nu$ \emph{degrees of freedom}.
When the scale matrix is the identity, $\S = \I$, we refer to the distribution as a \emph{standard Wishart}.
We denote this by $\G \sim \Wish{\S, \nu}$, and $\G$ has density
\begin{align}
    \label{eq:dwp:wishdens}
    \p{\G} = \frac{\pi^{\nu (\nt - N)/2}}{2^{\nu N/2}|\S|^{\nu/2} \Gamma_\nt\b{\tfrac{\nu}{2}}} \abs{\G_{:\nt, :\nt}}^{\b{\nu-N-1}/2} \etr \b{-\S^{-1}\G/2},
\end{align}
where we have defined $\nt = \min(\nu, N)$, $\Gamma_\nt\b{\cdot}$ is the multivariate gamma function, and $\etr(\cdot)$ is the exponentiated trace function, $\etr(\cdot) = \exp{\mathrm{tr}(\cdot)}$.
\end{restatable}
We take note of a few things from this definition.
First, it is clear that Wishart random variables are positive semi-definite by definition, and for $\nu \geq N$ will be full rank with probability one.
Moreover, the density is valid for $\nu < N$, where the resulting $\G$ will be low rank: this case defines the \emph{singular} Wishart.
We further note that for $\nu \geq N$, we can straightforwardly extend the Wishart to take any real-valued degrees of freedom $\nu$ with density also given by Eq.~\ref{eq:dwp:wishdens}.
In addition, it is possible to sample from a Wishart distribution, either by sampling $\nu$ Gaussian variables and using the definition above, or, as we shall later see (Sec.~\ref{sec:dwp:bart}), by using the Bartlett decomposition (which is particularly relevant for non-integer $\nu$).
Finally, the Wishart distribution has mean 
\begin{align}
  \label{eq:dwp:mean_wishart}
  \mathbb{E}\sqb{\G} &= \nu \mathbb{E}\sqb{\f_\lambda \transpose{\f}_\lambda} = \nu \S.
\end{align}

From this definition, it is straightforward to see that $\G_\ell$ is indeed Wishart-distributed with degrees of freedom equal to the width of the DGP layer, $\nu$, and scale matrix $\frac{1}{\nu_\ell} \Kfl{\F_{\ell-1}}$.
Therefore, given the features from the previous layer, $\F_{\ell-1}$, it is possible to directly sample the Gram matrix $\G_\ell$, using
\begin{align}
\label{eq:dwp:pG|F}
  \pc{\G_\ell}{\F_{\ell-1}} &= \Wish{\G_\ell; \tfrac{1}{\nu_\ell} \Kfl{\F_{\ell-1}}, \nu_\ell}.
\end{align}
While this shows how to sample $\G_\ell$ when we have $\F_{\ell-1}$, this is not sufficient to define a true deep Wishart process: we would need to find features $\F_{\ell-1}$ consistent with the Gram matrix $\G_{\ell-1}$ that we have already sampled.
To work entirely with Gram matrices, we would need to be able to compute $\Kfl{\F_{\ell-1}}$ directly from $\G_{\ell-1}$, i.e., obtain a function $\mathbf{K}_\ell: \reals^{N \times N} \rightarrow \reals^{N \times N}$ such that
\begin{align*}
\label{eq:dwp:kfl-kl}
    \Kl{\G_{\ell - 1}} = \Kl{\tfrac{1}{\nu_{\ell-1}} \F_{\ell-1} \transpose{\F}_{\ell-1}} = \Kfl{\F_{\ell-1}}.
\end{align*}
Remarkably, this is possible for a large family of practically relevant kernels, particularly those where the influence of the features on the kernel matrix is through a dot product operation (which can be seen by the definition of $\G_{\ell}$) or through the Euclidean distance between features.
For instance, this includes any isotropic kernel as well as the arc-cosine kernel \citep{aitchison2020deep}.
For isotropic kernels, e.g., squared exponential and Mat\'{e}rn kernels,\footnote{We note that this does \emph{not} include automatic relevance determination (ARD) versions of these kernels, as these rely directly on determining the relative importance of \emph{features}, which we do not have access to. However, as we will discuss in Sec.~\ref{sec:dwp:limitations}, we do not believe this to be a strong limitation of our method.} by definition \citep[][p. 80]{rasmussen2006gaussian} we have
\begin{align*}
    k_\ell\b{\F_{\ell-1, i}, \F_{\ell-1, j}} &= k_{\mathrm{dist},\ell}\b{\|\F_{\ell-1, i} - \F_{\ell-1, j}\|},
\end{align*}
for a suitable function $k_{\mathrm{dist},\ell}(\cdot)$.
Manipulating the square of the argument on the right hand side,
\begin{align*}
    \|\F_{\ell-1, i} - \F_{\ell-1, j}\|^2 &= \sum_{\lambda=1}^{\nu_{\ell - 1}} \b{F_{i\lambda}^{\ell-1} - F_{j\lambda}^{\ell-1}}^2 \\
    &= \sum_{\lambda=1}^{\nu_{\ell-1}} \b{F_{i\lambda}^{\ell - 1}}^2 - 2F_{i\lambda}^{\ell-1}F_{j\lambda}^{\ell-1} + \b{F_{j\lambda}^{\ell-1}}^2 \\
    &= \nu_{\ell-1}\b{G_{ii}^{\ell-1} -2G_{ij}^{\ell-1} + G_{jj}^{\ell-1}},
\end{align*}
where the final line is given by the definition of $\G_{\ell-1}$, showing that we can indeed rewrite isotropic kernels in terms of $\G_{\ell}$.

Therefore, by using $\Kl{\cdot}$ instead of $\Kfl{\cdot}$, we can rewrite our model entirely in terms of Gram matrices (up until the last-layer features, as the likelihood depends on features instead of Gram matrices):
\begin{subequations}
\label{eq:dwp:def:deepiw}
\begin{align}
  \pc{\G_\ell}{\G_{\ell-1}} &= \Wish{\G_\ell; \tfrac{1}{\nu_\ell} \Kl{\G_{\ell-1}}, \nu_\ell} & \text{with } \G_0 &= \tfrac{1}{\nu_0} \X \X^T,\\
  \pc{\F_{L+1}}{\G_L} &= \prodlnlp \N{\f_\lambda^{L+1}; \0, \Kl{\G_L}},
\end{align}
\end{subequations}
again with a likelihood $\pc{\Y}{\F_{L+1}}$.
We have thus succeeded in transforming our original deep Gaussian process model into an \emph{equivalent} deep Wishart process model, where we propagate Gram matrices instead of features.
Note that the assumption of a zero mean function in the DGP layers is essential to this equivalence: while a non-zero mean function is theoretically possible, it would require the use of the non-central Wishart distribution for the Gram matrices, which are difficult to use in practice \citep{koev2006efficient}.
Finally, the above derivations can likewise be applied to BNNs up to Eq.~\ref{eq:dwp:pG|F}.
However, in the case of BNNs, $\tfrac{1}{\nu_\ell}\Kfl{\F_{\ell-1}}$ is low rank when $N > \nu_{\ell-1}$, making it difficult to interpret the result, and it is not clear that we can find a $\Kl{\cdot}$ equivalent to $\Kfl{\cdot}$.
Therefore, for the remainder of the chapter, we focus on the DWP-DGP equivalence for full-rank $\tfrac{1}{\nu_\ell}\Kfl{\F_{\ell-1}}$.

\subsection{The DWP formulation captures true-posterior symmetries while DGPs do not}
\label{sec:dwp:symmetry}
We now have two equivalent generative models: one phrased in terms of features, $\F_\ell$, and another phrased in terms of Gram matrices, $\G_\ell$.
We will now show that deep Wishart processes are able to account for the permutation and rotational symmetries that arise in both BNNs and DGPs.
Consider again the transformation of features $\hat{\F}_\ell = \F_\ell \mQ$, where $\mQ$ is an orthogonal matrix.
Since $\mQ \transpose{\mQ} = \I$, the Gram matrix $\G_\ell$ is invariant to these transformations:
\begin{align*}
    \hat{\G}_\ell = \frac{1}{\nu_\ell} \hat{\F}_\ell \transpose{\hat{\F}_\ell} = \frac{1}{\nu_\ell} \F_\ell \mQ \transpose{\mQ} \transpose{\F}_\ell = \frac{1}{\nu_\ell} \F_\ell \transpose{\F}_\ell = \G_\ell.
\end{align*}
As such, DWP approximate posteriors written in terms of $\G_\ell$ implicitly respect these symmetries in the model.

\subsubsection{Global inducing DGP posteriors and symmetries}
While the DWP automatically respects the rotational symmetries in DGPs, it is still in theory possible to construct a feature-based approximate DGP posterior that respects these symmetries.
We first note that a pure Gaussian approximate posterior would have to learn a mean of zero to ensure that the density is invariant to rotations.
As this will clearly not be the case, we focus on investigating whether current DGP posteriors, and in particular the global inducing posterior of the last chapter, respect \emph{any} form of rotational invariance.
We first consider rotating the inducing variables and features going into a layer, $\hat{\U}_{\ell-1} = \U_{\ell-1} \mQ$ and $\hF_{\ell-1} = \F_{\ell-1}\mQ$.
By observing the form of the global inducing approximate posterior in Eq.~\ref{eq:gi:U|V} and since $\Kfl{\cdot}$ is invariant to rotations in the input, we see that the approximate posterior is invariant to this rotation.
Note that this will not be true of local inducing posteriors, as the inducing locations remain fixed at their original locations.\footnote{This insight was much of the motivation for \citet{ustyuzhaninov2019compositional}.}

Now let us consider rotating the output inducing variables and features, so that $\hat{\U}_\ell = \U_\ell \mQ$ and $\hF_\ell = \F_\ell \mQ$.
In this case, the global inducing approximate posterior is not invariant to this symmetry, i.e., it assigns different posterior density to $\hF_\ell$ and $\F_\ell$.
This will also be the case for local inducing posteriors: we expect that this is where the DWP formulation can show its advantages.

\subsection{Equivalent DWP posteriors will have better ELBOs and generalization}\label{sec:dwp:dwpdgpelbo}
We have just shown that the DWP implicitly respects some of the symmetries in deep Bayesian models.
Let us now consider a DGP approximate posterior $\q{\slpo{\F_\ell}}$, and its associated ELBO
\begin{align*}
    \mathcal{L}_\mathrm{DGP} = \log \p{\Y} - \KL*{\q{\slpo{\F_\ell}}}{\pc{\slpo{\F_\ell}}{\Data}}.
\end{align*}
Taking this posterior, and transforming it into a posterior over Gram matrices $\q{\slpo{\G_\ell}}$ by applying $\G_\ell = \frac{1}{\nu_\ell} \F_\ell \transpose{\F}_\ell$ results in the ELBO
\begin{align*}
    \mathcal{L}_\mathrm{DWP} = \log \p{\Y} - \KL*{\q{\slpo{\G_\ell}}}{\pc{\slpo{\G_\ell}}{\Data}}.
\end{align*}
Importantly, as we can write the Gram matrices $\G_\ell$ as a function of the features $\F_\ell$, we can apply the data processing inequality:
\begin{restatable}[see e.g., Thm. 7.4 in \citealp{polyanskiy2022information}]{prop}{dataprocprop}\label{prop:data-proc}
Let $g(\cdot)$ be a function. Given probability densities $\mathrm{p}_x$ and $\mathrm{q}_x$ on $x$, and defining $y = g(x)$, we have that $\KL{\mathrm{q}_y(y)}{\mathrm{p}_y(y)} \leq \KL{\mathrm{q}_x(x)}{\mathrm{p}_x(x)}$, where $\mathrm{p}_y$ and $\mathrm{q}_y$ are the densities on $y$ implied by $\mathrm{p}_x$, $\mathrm{q}_x$, and $g(\cdot)$ (i.e., the densities according to pushforward measures).
\end{restatable}
This guarantees that the KL between the approximate posterior over features and the true posterior will upper bound the KL for Gram matrices, meaning that $\mathcal{L}_\mathrm{DWP} \geq \mathcal{L}_\mathrm{DGP}$.
Moreover, equality is only achieved if $\q{\slpo{\F_\ell}}$ gives equal density to \emph{all} $\F_\ell$ that result in the same $\G_\ell$, for which invariance to rotation is a necessary condition.\footnote{In essence, the data processing inequality tells us that we lose information by transforming from $\F_\ell$ to $\G_\ell$. Therefore, equality is achieved when we have the same amount of information in $\F_\ell$ and $\G_\ell$.}
In other words, the higher ELBO for the implied posterior $\q{\slpo{\G_\ell}}$ is due to the fact that we have implicitly encoded invariance to the model's symmetries.
For a deeper theoretical understanding of this result we refer the reader to \citet{burt2020understanding}, as this is one of the main motivations for the recent interest in function-space inference \citep{sun2018functional, ma2019variational}.
Moreover, this fact can also be used to derive better PAC-Bayes generalization bounds \citep[cf. Section 6.1.3 of][]{alquier2021userfriendly}.

This analysis relied on the assumption that $\q{\{\G_\ell\}_\ell}$ would be the distribution implied by $\q{\{\F_\ell\}_\ell}$. 
In practice, this will not be the case, as we cannot compute this density analytically.
Moreover, we wish to deal entirely in Gram matrices, which requires that we specify approximate posteriors directly for Gram matrices.
However, given a sufficiently flexible posterior over Gram matrices, we should still observe these advantages in practice.

\subsection{Comparing the DWP to the DIWP}\label{sec:dwp:diwpcomp}
While \citet{aitchison2020deep} showed the above equivalence between deep Wishart and deep Gaussian processes, they did not demonstrate how to design a practical approximate posterior for DWPs.
Instead, they focused on the deep \emph{inverse} Wishart process, a similar construction where instead $\G_\ell$ are conditionally \emph{inverse} Wishart given $\G_{\ell-1}$ (see App.~\ref{app:jac} for details about the inverse Wishart distribution).
As the deep inverse Wishart process does not correspond to any prior feature-based deep model that we are aware of, it is difficult to determine whether the promised gains in performance will actually materialize.

Moreover, the inverse Wishart has properties that make it unsuited for inducing point inference.
In particular, inverse Wishart approximate posteriors as used in \citet{aitchison2020deep} can become \emph{worse} as inducing points are added.
Consider an inducing point-based approximate posterior $\mathrm{q}_{1}\b{\U, *, \F} = \mathrm{q}_{1}\b{\U}\pc{*}{\U}\pc{\F}{\U, *}$, where $\U$ represents the inducing points, $*$ represents a point outside of $\U$, and $\F$ represents any set of additional training or test points.
As in \citet{aitchison2020deep}, we have followed the standard GP literature approach of defining an approximate posterior $\q{\U}$ over inducing points and using the conditional prior $\pc{\cdot}{\U}$ to predict.
Now consider adding $*$ to our set of inducing points, so that $\mathrm{q}_{2}\b{\U, *, \F} = \mathrm{q}_{2}\b{\U, *}\pc{\F}{\U, *} = \mathrm{q}_{2}\b{\U}\mathrm{q}_{2}\bc{*}{\U}\pc{\F}{\U, *}$.
To ensure that adding this inducing point does not worsen the ELBO, we should be able to set $\mathrm{q}_2\b{\U, *} = \mathrm{q}_1\b{\U, *}$, i.e., adding an inducing point should only increase the flexibility of the approximate posterior.
We will now show that this is not possible for the inverse Wishart, whereas it is possible for the Wishart.

Consider an inverse Wishart prior model similar to that proposed in \citet[][cf. Eq. 31]{aitchison2020deep}:
\begin{align*}
    \G &= \begin{pmatrix}
      \G_{\u\u} & \g_{\u *} \\
      \transpose{\g}_{\u *} & g_{**}
    \end{pmatrix}
    \sim
    \iWish{\S,
    \nu + M + 1
    }, \\
    \S &= \begin{pmatrix}
      \S_{\u\u} & \boldsymbol{\sigma}_{\u *} \\
      \transpose{\boldsymbol{\sigma}}_{\u *} & \sigma_{**}
    \end{pmatrix},
\end{align*}
where $M$ is the number of inducing points $\U$ and $\nu$ is a degrees of freedom parameter. 
According to this prior, the prior conditional $\pc{*}{\U} = \pc{g_{**}, \g_{\u *}}{\G_{\u\u}}$ can be found via the following \citep[cf. App.~I in ][]{aitchison2020deep}:
\begin{align}
\nonumber
    g_{**\cdot \u}| \G_{\u \u}, \S &\sim \mathrm{InverseGamma}\b{\alpha = \tfrac{1}{2}(\nu + M + 1),\, \beta = \tfrac{1}{2}\sigma_{**\cdot \u}}, \\
    \label{eq:dwp:iwcond}
    \G_{\u \u}^{-1} \g_{\u *} | g_{**\cdot \u}, \G_{\u\u}, \S &\sim \N{\S_{\u\u}^{-1}\boldsymbol{\sigma}_{\u *},\, g_{**\cdot \u}\S_{\u\u}^{-1}},
\end{align}
where we have defined $g_{** \cdot \u}$ and $\sigma_{** \cdot \u}$ as the Schur complements
\begin{align*}
    g_{**\cdot \u} &= g_{**} - \transpose{\g}_{\u *} \G_{\u\u}^{-1} \g_{\u *}, \\
    \sigma_{**\cdot \u} &= \sigma_{**} - \transpose{\boldsymbol{\sigma}}_{\u *}\S_{\u\u}^{-1}\boldsymbol{\sigma}_{\u *}.
\end{align*}
Similar to \citet{aitchison2020deep}, we now consider an approximate posterior 
\begin{align*}
    \q{\G} &= \iWish{\G; \boldsymbol{\Psi}, \nu + M + 1}, \\
    \boldsymbol{\Psi} &= \begin{pmatrix}
      \boldsymbol{\Psi}_{\u\u} & \x \\
      \transpose{\x} & a
    \end{pmatrix},
\end{align*}
where $\x$ and $a$ are learned.
Inspecting the conditional $\mathrm{q}_2\bc{*}{\U}$, which follows the form of Eq.~\ref{eq:dwp:iwcond}, we obtain the following system of equations:
\begin{align*}
    \boldsymbol{\Psi}_{\u\u}^{-1} \x &= \S_{\u\u}^{-1}\boldsymbol{\sigma}_{\u *} \implies \x = \boldsymbol{\Psi}_{\u\u} \S_{\u\u}^{-1} \boldsymbol{\sigma}_{\u *}, \\
    a - \transpose{\x}\boldsymbol{\Psi}_{\u\u}^{-1}\x &= \sigma_{**\cdot \u} \implies a = \sigma_{**} - \transpose{\boldsymbol{\sigma}}_{\u *}\S_{\u\u}^{-1}\b{\S_{\u\u} - \boldsymbol{\Psi}_{\u\u}}\S_{\u\u}^{-1}\boldsymbol{\sigma}_{\u *}, \\
    g_{** \cdot \u}\boldsymbol{\Psi}_{\u\u}^{-1} &= g_{** \cdot \u}\S_{\u\u}^{-1},
\end{align*}
the last of which cannot be resolved, as it would imply $\mathrm{q}_1\b{\U} \neq \mathrm{q}_2 \b{\U}$.
Intuitively, this is because the inverse Wishart requires that the degrees of freedom increase as points are added, resulting in a contracting approximate posterior.

We now consider a similar setup for the Wishart distribution, and show that it is possible to set $\qc{*}{\U} = \pc{*}{\U}$.
For the Wishart, we consider a similar setup for the prior, where we make the features explicit:
\begin{align*}
    \begin{pmatrix}
      \U \\
      *
    \end{pmatrix}
    \begin{pmatrix}
      \transpose{\U} & \transpose{*}
    \end{pmatrix}
    = \G &=
    \begin{pmatrix}
      \G_{\u\u} & \g_{\u *} \\
      \transpose{\g}_{\u *} & g_{**}
    \end{pmatrix}
    \sim
    \Wish{\begin{pmatrix}
        \S_{\u\u} & \boldsymbol{\sigma}_{\u *} \\
        \transpose{\boldsymbol{\sigma}}_{\u *} & \sigma_{**}
    \end{pmatrix}, \nu
    }.
\end{align*}
\citet{eaton2007wishart} shows us that the conditional prior
\begin{align*}
    \pc{*}{\U} &= \MN{\transpose{\boldsymbol{\sigma}}_{\u *}\S_{\u\u}^{-1}\U, \sigma_{**\cdot \u}, \I},
\end{align*}
where $\sigma_{** \cdot \I}$ is defined as above.
Defining a Wishart approximate posterior as
\begin{align*}
    \q{\G} = \Wish{\G; \begin{pmatrix}
      \boldsymbol{\Psi}_{\u\u} & \x \\
      \transpose{\x} & a
    \end{pmatrix}, \nu},
\end{align*}
and comparing the prior conditional $\pc{*}{\U}$ with the implied posterior conditional $\qc{*}{\U}$, we obtain the following system of equations:
\begin{align*}
    \transpose{\x}\boldsymbol{\Psi}_{\u\u}^{-1} &= \transpose{\boldsymbol{\sigma}}_{\u *}\S_{\u\u}^{-1} \implies \x = \boldsymbol{\Psi}_{\u\u} \S_{\u\u}^{-1} \boldsymbol{\sigma}_{\u *}, \\
    a - \transpose{\x}\boldsymbol{\Psi}_{\u\u}^{-1}\x &= \sigma_{**\cdot \u} \implies a = \sigma_{**} - \transpose{\boldsymbol{\sigma}}_{\u *}\S_{\u\u}^{-1}\b{\S_{\u\u} - \boldsymbol{\Psi}_{\u\u}}\S_{\u\u}^{-1}\boldsymbol{\sigma}_{\u *}.
\end{align*}
Therefore, for the Wishart we are able to ensure that adding inducing points will not worsen the ELBO.

We now turn to the challenge of developing an approximate posterior for the deep Wishart process.

\section{Methods}

As \citet{aitchison2020deep} explained, it is difficult to define sufficiently flexible approximate posteriors for the deep Wishart process (cf. App. E therein).
In particular, as the number of datapoints $N$ can be greater than the width of the model $\nu$, the sampled Gram matrices $\G_\ell$ will be low-rank with rank $\nu$ (almost surely).
Therefore, we will need a flexible approximate posterior over rank $\nu$ Gram matrices.
An obvious first choice would be to simply choose the Wishart distribution with degrees of freedom $\nu$, so that the rank of the posterior and prior are matched (which is necessary for the KL divergence to be finite).
However, for fixed degrees of freedom, the Wishart variance
\begin{align*}
  \Var \sqb{G_{ij}} &= \nu \b{\Sigma_{ij}^2 + \Sigma_{ii} \Sigma_{jj}}
\end{align*}
cannot be specified independently of the mean (Eq.~\ref{eq:dwp:mean_wishart}), which is essential for a variational approximate posterior that can flexibly capture potentially narrow true posteriors.
An alternative approach would be to work with a non-central Wishart, which is defined by taking $\F$ from the definition of the Wishart to have non-zero mean, i.e., to sample $\F \in \reals^{N \times \nu}$ with columns $\f_\lambda \in \reals^N \sim \N{\m, \S}$ for some learned $\m$.
However, as previously mentioned, the non-central Wishart is difficult to use in practice, as it has a probability density function whose current cost of evaluation renders it unsuitable for repeated evaluation in the inner loop of a modern machine learning algorithm \citep{koev2006efficient}.
Instead, we develop a new class of generalized singular Wishart distribution, which modifies the Wishart to give independent control over the mean and variance of sampled matrices.
We achieve this by using a modified \emph{Bartlett decomposition}, which we now describe.

\subsection{The Bartlett decomposition}\label{sec:dwp:bart}
The Bartlett decomposition \citep{bartlett1933on} allows full-rank standard Wishart random variables (i.e., with identity scale matrix, $\S = \I$) to be sampled efficiently, particularly for large values of $\nu$ and non-integer $\nu$.
Instead of sampling $\F$, the Bartlett decomposition samples the Cholesky decomposition of $\G$, which we denote by $\T$.
According to the Bartlett decomposition, 
\begin{subequations}
  \label{eq:dwp:PA}
  \begin{align}
    \label{eq:dwp:PA_ondiag}
    \p{T_{jj}^2} &= \text{Gamma}\b{T_{jj}^2;\,\alpha{=}\tfrac{\nu-j+1}{2}, \beta{=}\tfrac{1}{2}},\\
    \label{eq:dwp:PA_offdiag}
    \p{T_{j > k}} &= \Nc{T_{jk}}{\,0, 1}. 
  \end{align}
\end{subequations}
For Wishart distributions with non-identity scale matrices, we can sample $\G$ by using the Cholesky decomposition $\L$ of $\S = \L \transpose{\L}$, so that $\G = \L \T \transpose{\T} \transpose{\L}$.

\subsection{Generalizing the singular Wishart distribution}
\label{sec:dwp:gen_wishart}
We now present our generalizations of the singular Wishart distribution.
First, we need to generalize the Bartlett construction to potentially singular matrices (i.e., those for which $\nu < N$).
In the singular case, the Bartlett factor is given by
\begin{subequations}
\begin{align}
  \label{eq:dwp:T}
  \T &= \begin{pmatrix}
    T_{11}    & \dotsm & 0          \\ 
    \vdots    & \ddots & \vdots     \\ 
    T_{\nu 1} & \dotsm & T_{\nu \nu} \\ 
    \vdots    & \ddots & \vdots     \\ 
    T_{N 1}   & \dotsm & T_{N \nu}
  \end{pmatrix},\\
  \p{T_{jj}^2} &= \text{Gamma}\b{T_{jj}^2;\,\tfrac{\nu-j+1}{2}, \tfrac{1}{2}}, &
  \p{T_{i > j}} &= \Nc{T_{ij}}{0, 1}. 
\end{align}
\end{subequations}
Recalling that $\G = \L \T \transpose{\T} \transpose{\L}$ and by applying the results of Appendices~\ref{sec:jac:J_LLT} and~\ref{sec:jac:J_LA}, we have that
\begin{align*}
  \p{\G} &= \b{\prod_{j=1}^N \frac{1}{L_{jj}^{\min(j,\nu)}}} \prod_{j=1}^\nt \frac{\text{Gamma}\b{T_{jj}^2; \tfrac{\nu-j+1}{2}, \tfrac{1}{2}}}{T_{jj}^{N-j} L_{jj}^{N-j+1}} \prod_{i=j+1}^N \Nc{T_{ij}}{0, 1}, 
\end{align*}
where we recall that $\nt = \min(\nu, N)$.
In Appendix~\ref{sec:jac:singular_proof} we prove that this corresponds to the known singular Wishart density (Eq.~\ref{eq:dwp:wishdens}). 
Equipped with the singular Bartlett, we can now develop our generalization of the Wishart distribution: 

\begin{defi}[The generalized singular Wishart distribution]\label{def:gen-wishart}
The generalized singular Wishart, $\gWish{\G; \S, \nu, \boldsymbol{\alpha}, \boldsymbol{\beta}, \boldsymbol{\mu}, \boldsymbol{\sigma}}$, is a distribution over positive semi-definite $N\times N$ matrices $\G$, with positive definite scale matrix $\S = \L \transpose{\L} \in \mathbb{R}^{N\times N}$, a positive, integer-valued degrees of freedom parameter $\nu$, and Bartlett-generalizing parameters $\boldsymbol{\alpha}, \boldsymbol{\beta}, \boldsymbol{\mu}, \boldsymbol{\sigma}$. These latter parameters modify the Bartlett decomposition as follows:
\begin{subequations}
\begin{align*}
  \q{T_{jj}^2} &= \textup{Gamma}\b{T_{jj}^2;\,\alpha_j, \beta_j} && \text{for } j \leq \nu,\\
  \q{T_{i > j}} &= \Nc{T_{ij}}{\,\mu_{ij}, \sigma_{ij}^2} && \text{for } j \leq \nu. 
\end{align*}
\end{subequations}
This implies a distribution over $\G = \L \T \transpose{\T} \transpose{\L}$ with density
\begin{align*}
  \q{\G} &= \b{\prod_{j=1}^N \frac{1}{L_{jj}^{\min(j,\nu)}}} \prod_{j=1}^{\nt} \frac{\textup{Gamma}\b{T_{jj}^2;\, \alpha_j, \beta_j}}{T_{jj}^{N-j} L_{jj}^{N-j+1}} \prod_{i=j+1}^N \N{T_{ij}; \mu_{ij}, \sigma_{ij}^2}. 
\end{align*}
\end{defi}
Note that we use $\q{\cdot}$ to reflect the fact that we will use the generalized singular Wishart as the basis for our approximate posterior.
The density is derived using the same transformations and Jacobians as for the singular Wishart above.

\subsection{The full approximate posterior}
\label{sec:dwp:approx_post}
To determine the scale matrix in our generalized singular Wishart, we follow a global inducing point approach as in the previous chapter, which will enable us to obtain between-layer correlations.
However, unlike the case of BNNs and DGPs as well as the deep inverse Wishart process in \citet{aitchison2020deep}, it is not possible to obtain an optimal last-layer posterior for the deep Wishart process. 
Therefore, we choose a form that mimics the form of these posteriors, allowing for similar across-layer dependencies: 
\begin{align}
\label{eq:dwp:def:ap}
  \qc{\G_\ell}{\G_{\ell-1}} &= \gWish{\G_\ell; (1-q_\ell) \tfrac{1}{\nu_\ell} \Kl{\G_{\ell-1}} + q_\ell \V_\ell \transpose{\V}_\ell,\, \nu_\ell,\, \boldsymbol{\alpha}_\ell,\, \boldsymbol{\beta}_\ell,\, \boldsymbol{\mu}_\ell,\, \boldsymbol{\sigma}_\ell},
\end{align}
where the approximate posterior parameters are $\sl{\V_\ell, \boldsymbol{\alpha}_\ell, \boldsymbol{\beta}_\ell, \boldsymbol{\mu}_\ell, \boldsymbol{\sigma}_\ell, q_\ell}$, where $0 < q_\ell < 1$ is a scalar, and $\V_\ell\in \reals^{N \times N}$.
Here, the learnable parameter $q_\ell$ allows us to trade off the influence on the scale matrix from the prior and the learned covariance $\V_\ell \transpose{\V}_\ell$, which allows for similar across-layer dependencies as in the previous chapter and \citet{aitchison2020deep}.
$\nu_\ell$ is fixed, as it determines the width of the layer; the remaining parameters are simply the parameters from the Bartlett generalization.

\subsection{Doubly-stochastic inducing point variational inference in deep Wishart processes}
\label{sec:dwp:dsvi}
While we could directly use the above posterior in our DWP, it would be prohibitively expensive for larger datasets due to the $\mathcal{O}(N^3)$ cost of the matrix factorizations required.
For efficient inference in high-dimensional problems, we therefore take inspiration from the DGP literature by developing doubly-stochastic inducing point posteriors.
We follow the global inducing scheme laid out in the previous chapter, which \citet{aitchison2020deep} also use, and define a set of learnable global inducing inputs $\X_\text{i} \in \reals^{M \times \nu_0}$.
We begin by extending and decomposing all variables into inducing and training (or test) points, to be able to propagate inducing points and data points simultaneously. 
For instance, we extend and decompose $\X$ into $\X_\text{i}\in\reals^{M \times \nu_0}$ and $\X_\text{t}\in\reals^{N\times \nu_0}$, where $M$ is the number of inducing points, and $N$ is the number of training/testing points.
This leads to
\begin{align*}
  \X &= \begin{pmatrix}
    \X_\text{i}\\
    \X_\text{t}
  \end{pmatrix}, &
  \F_{L+1} &= \begin{pmatrix}
    \F^{L+1}_\text{i}\\
    \F^{L+1}_\text{t}
  \end{pmatrix}, &
  \G_\ell &=\begin{pmatrix}
    \G_\text{ii}^\ell & \G_\text{it}^\ell\\
    \G_\text{ti}^\ell & \G_\text{tt}^\ell
  \end{pmatrix}, &
\end{align*}
where e.g., $\G_\text{ii}^\ell$ is $M \times M$ and $\G_\text{it}^\ell$ is $M \times N$.
The full ELBO, including variables for all the inducing and training points, is
\begin{align}
  \label{eq:dwp:def:elbo}
  \mathcal{L} &= \E[\mathrm{q}]{\log \pc{\Y}{\F^{L+1}_\text{t}} + \log \frac{\pc{\slt{\G_\ell}, \F_{L+1}}{\X}}{\qc{\slt{\G_\ell}, \F_{L+1}}{\X}}},
\end{align}
where the expectation is taken over $\qc{\slt{\G_\ell},\F_{L+1}}{\X}$.
The prior is given by combining all terms in Eq.~\eqref{eq:dwp:def:deepiw} for both inducing and train/test inputs,
\begin{align*}
  \pc{\slt{\G_\ell},\F_{L+1}}{\X} = \sqb{\prod_{\ell=1}^{L} \pc{\G_\ell}{\G_{\ell-1}}} \pc{\F_{L+1}}{\G_L},
\end{align*}
where the dependence on $\X$ enters on the right because $\G_0 = \tfrac{1}{\nu_0} \X \transpose{\X}$.
Taking inspiration from the Gaussian process literature, we factorize the full approximate posterior as the product of an approximate posterior over inducing points and the conditional prior for train/test points,
\begin{multline}
  \label{eq:dwp:Q_inducing_fac}
  \qc{\slt{\G_\ell},\F_{L+1}}{\X} =\\
  \qc{\slt{\G^\ell_\text{ii}},\F_\text{i}^{L+1}}{\X_\text{i}} \pc{\slt{\G^\ell_\text{it}},\slt{\G^\ell_\text{tt}},\F_\text{t}^{L+1}}{\slt{\G^\ell_\text{ii}},\F_\text{i}^{L+1}, \X}.
\end{multline}
We can write the prior in the same way, so that
\begin{multline}
  \label{eq:dwp:P_inducing_fac}
  \pc{\slt{\G_\ell},\F_{L+1}}{\X} =\\
  \pc{\slt{\G^\ell_\text{ii}},\F_\text{i}^{L+1}}{\X_\text{i}} \pc{\slt{\G^\ell_\text{it}},\slt{\G^\ell_\text{tt}},\F_\text{t}^{L+1}}{\slt{\G^\ell_\text{ii}},\F_\text{i}^{L+1}, \X}.
\end{multline}
For now, we put aside discussion of the conditional prior until later, in Eq.~\eqref{eq:dwp:condP}. 
The approximate posterior and prior over inducing matrices $\G_\text{ii}^\ell$ and last-layer features $\F_\text{i}^{L+1}$ are given by combining terms in Eq.~\eqref{eq:dwp:def:deepiw} and Eq.~\eqref{eq:dwp:def:ap}, so that
\begin{align}
  \pc{\slt{\G^\ell_\text{ii}},\F_\text{i}^{L+1}}{\X_\text{i}} &= \sqb{\prod_{\ell=1}^{L} \pc{\G^\ell_\text{ii}}{\G^{\ell-1}_\text{ii}}} \pc{\F_\text{i}^{L+1}}{\G_\text{ii}^L},\\
  \label{eq:dwp:Q_inducing}
  \qc{\slt{\G^\ell_\text{ii}},\F_\text{i}^{L+1}}{\X_\text{i}} &= \sqb{\prod_{\ell=1}^{L} \qc{\G^\ell_\text{ii}}{\G^{\ell-1}_\text{ii}}} \qc{\F_\text{i}^{L+1}}{\G_\text{ii}^L}.
\end{align}
For the approximate posterior terms over inducing point Gram matrices, $\qc{\G_\text{ii}^\ell}{\G_\text{ii}^{\ell-1}}$, we use our generalized singular Wishart posterior defined in Sec.~\ref{sec:dwp:approx_post}, whereas for the approximate posterior over features $\qc{\F_\text{i}^{L+1}}{\G_\text{ii}^L}$, we apply the global inducing posterior from the previous chapter.
Substituting Eqs.~(\ref{eq:dwp:Q_inducing_fac}--\ref{eq:dwp:Q_inducing}) into the ELBO (Eq.~\ref{eq:dwp:def:elbo}), the conditional prior cancels and we obtain
\begin{align*}
  \mathcal{L} &= \E[\mathrm{q}]{\log \pc{\Y}{\F^{L+1}_\text{t}} + \log \frac{\sqb{\tprod_{\ell=1}^{L} \qc{\G^\ell_\text{ii}}{\G^{\ell-1}_\text{ii}}} \qc{\F_\text{i}^{L+1}}{\G_\text{ii}^L}}{\sqb{\tprod_{\ell=1}^{L} \pc{\G^\ell_\text{ii}}{\G^{\ell-1}_\text{ii}}} \pc{\F_\text{i}^{L+1}}{\G_\text{ii}^L}}}.
\end{align*}
The first term is a summation across train datapoints (for likelihoods that factorize across datapoints), and the second term depends only on the inducing points. 
Therefore, as in \citep{salimbeni2017doubly}, we can compute unbiased estimates of the expectation by taking only a minibatch of datapoints. 
We also never need to compute the density of the conditional prior in Eq.~\eqref{eq:dwp:P_inducing_fac}.
Rather, we only need to be able to sample from it.
We now inspect this term more closely:
\begin{multline}
  \label{eq:dwp:condP}
  \pc{\slt{\G_\text{ti}^\ell, \G_\text{tt}^\ell},\F_\text{t}^{L+1}}{\slt{\G^\ell_\text{ii}},\F_\text{i}^{L+1}, \X} =\\ 
  \pc{\F^{L+1}_\text{t}}{\F^{L+1}_\text{i}, \G_{L}} \prod_{\ell=1}^{L} \pc{\G^\ell_\text{ti}, \G^\ell_\text{tt}}{\G^\ell_\text{ii}, \G_{\ell-1}}.
\end{multline}
The first term, $\pc{\F^{L+1}_\text{t}}{\F^{L+1}_\text{i}, \G_{L}}$, is a multivariate Gaussian, and can be evaluated using methods from the GP literature \citep{rasmussen2006gaussian}.
The second distribution is more difficult to sample from.
To address this issue, we refer back to the feature-based representation, where we define scaled imagined features $\tF_\ell$ so that
\begin{align}
  \label{eq:dwp:FFT}
  \tF_\ell \transpose{\tF}_\ell = \G_\ell &\sim \Wish{\S_\ell, \nu_\ell},
\end{align}
with $\S_\ell = \frac{1}{\nu_\ell}\Kl{\G_{\ell-1}}$, and
\begin{align*}
  \tF_\ell &= \begin{pmatrix}
    \tF^\ell_\text{i} \\ \tF^\ell_\text{t}
  \end{pmatrix}, &
   \S_\ell &= \begin{pmatrix}
    \S_\text{ii}^\ell & \transpose{\S^\ell_\text{ti}}\\
    \S_\text{ti}^\ell & \S_\text{tt}^\ell
  \end{pmatrix},
\end{align*}
where $\tF_\ell \in \reals^{(M + N)\times \nu_\ell}$, $\tF_\text{i}\in\reals^{M\times \nu_\ell}$ and $\tF_\text{t}\in\reals^{N\times \nu_\ell}$.
Our goal is to sample $\G_\text{it}^\ell$ and $\G_\text{tt}^\ell$ given $\G_\text{ii}^\ell$.
Our approach is to note that $\tF_\text{t}$ conditioned on $\tF_\text{i}$ is given by a matrix normal \citep[][page 310]{eaton2007wishart}:
\begin{align}
  \label{eq:dwp:Ft|Fi}
  \pc{\tF^\ell_\text{t}}{\tF^\ell_\text{i}} &= \MN{\S^\ell_\text{ti} \b{\S^\ell_\text{ii}}^{-1} \tF^\ell_\text{i}, \,\S^\ell_{\text{tt}\cdot \text{i}}, \,\I},
\end{align}
where
\begin{align*}
  \S^\ell_{\text{tt}\cdot \text{i}} &= \S^\ell_\text{tt} - \S^\ell_\text{ti} \b{\S^\ell_\text{ii}}^{-1} \transpose{\S^\ell_\text{ti}}.
\end{align*}
Note that we sample each test/train point independently, in which case $N=1$ and $\S_{22 \cdot 1}$ is scalar.

Returning to $\G_\ell$, which includes $\G^\ell_\text{it}$ and $\G^\ell_\text{tt}$, we have
\begin{align*}
  \G_\ell &=\begin{pmatrix}
    \G_\text{ii}^\ell & \G_\text{it}^\ell\\
    \G_\text{ti}^\ell & \G_\text{tt}^\ell
  \end{pmatrix} = 
  \begin{pmatrix}
    \tF_\text{i}^\ell \transpose{\mbox{$\tF_\text{i}^\ell$}} & \tF_\text{i}^\ell\transpose{\mbox{$\tF_\text{t}^\ell$}} \\
    \tF_\text{t}^\ell \transpose{\mbox{$\tF_\text{i}^\ell$}} & \tF_\text{t}^\ell\transpose{\mbox{$\tF_\text{t}^\ell$}} 
  \end{pmatrix} 
  = \tF_\ell \transpose{\tF}_\ell.
\end{align*}
We emphasize that these features are \emph{imagined}, and do not represent real features as in a DGP.
Therefore, for $\tF_\text{i}^\ell$ we are free to use any value, provided that $\G^\ell_\text{ii} = \tF_\text{i}^\ell \transpose{\mbox{$\tF_\text{i}^\ell$}}$.
Under this condition, the resulting distribution over $\G_\ell$ arising from Eq.~\eqref{eq:dwp:FFT} is independent of the specific choice of $\tF_\text{i}^\ell$ (we confirm this in App.~\ref{sec:dwp:app:F}).
Recall that to sample $\G_\text{ii}^\ell$ in our approximate posterior, we explicitly sample its potentially low-rank Cholesky factor, $\L_\ell \T_\ell$. 
We can therefore directly use
\begin{align*}
  \tF_\text{i}^\ell &= \L_\ell \T_\ell 
\end{align*}
However, this only works if $\nu \leq M$, in which case $\L_\ell \T_\ell \in \reals^{M \times \nu_\ell}$.
In the unusual case where we have fewer inducing points than degrees of freedom, $M<\nu$, then $\L_\ell \T_\ell \in \reals^{M \times M}$, so we need to pad to achieve the required size of $M \times \nu_\ell$ to use Eq.~\ref{eq:dwp:Ft|Fi}:
\begin{align*}
  \tF_\text{i}^\ell &= \begin{pmatrix}
    \L_\ell \T_\ell & \0
  \end{pmatrix}.
\end{align*}

Finally, note that we can optimise all the variational parameters using standard reparameterized variational inference \citep{kingma2013auto,rezende2014stochastic}.
We provide an algorithm in Alg.~\ref{alg}.\footnote{For clarity, we ignore the kernel hyperparameters and variational parameters required for the last layer global inducing posterior in the algorithm and focus on the Wishart process variational parameters.}

\begin{algorithm}[tb]
\caption{Computing predictions/ELBO for one batch}
\label{alg}
\begin{algorithmic}
  \STATE {\bfseries parameters:} $\sl{\V_\ell, q_\ell, \boldsymbol{\alpha}_\ell, \boldsymbol{\beta_\ell}, \boldsymbol{\mu}_\ell, \boldsymbol{\sigma}_\ell}, \X_\text{i}$.
  \STATE {\bfseries Inputs:} $\X_\text{t}$; {\bfseries Targets:} $\Y$
  \STATE \textcolor{gray}{combine inducing and test/train inputs}
  \STATE $\X = \begin{pmatrix} \X_\text{i} \\ \X_\text{t} \end{pmatrix}$ 
  \STATE \textcolor{gray}{sample first Gram matrix}
  \STATE $\G_0 = \tfrac{1}{\nu_0} \X \transpose{\X}$
  \FOR{$\ell$ {\bfseries in} $\{1,\dotsc,L\}$}
    \STATE \textcolor{gray}{sample inducing Gram matrix and its Cholesky, $\L_\ell \T_\ell$ and update ELBO}
    \STATE $\L_\ell \T_\ell \transpose{\T}_\ell \transpose{\L}_\ell = \G_\text{ii}^\ell \sim \qc{\G_\text{ii}^\ell}{\G_\text{ii}^{\ell-1}}$ 
    \STATE $\mathcal{L} \leftarrow \mathcal{L} + \log \pc{\G_\text{ii}^\ell}{\G_\text{ii}^{\ell-1}} - \log \qc{\G_\text{ii}^\ell}{\G_\text{ii}^{\ell-1}}$ 
    \STATE \textcolor{gray}{sample full Gram matrix from conditional prior}
    \STATE $\S_\ell = \tfrac{1}{\nu_\ell} \Kl{\G_{\ell-1}}$
    \STATE $\S^\ell_{\text{tt}\cdot \text{i}} = \S^\ell_\text{tt} - \S^\ell_\text{ti} \b{\S^\ell_\text{ii}}^{-1} \transpose{\S^\ell_\text{ti}}$
    \STATE $\tF_\text{i}^\ell = \L_\ell \T_\ell$
    \STATE $\tF^\ell_\text{t} \sim \MN{\S^\ell_\text{ti} \b{\S^\ell_\text{ii}}^{-1} \tF^\ell_\text{i}, \S^\ell_{\text{tt}\cdot \text{i}}, \I}$
    \STATE $\G_\ell = 
    \begin{pmatrix}
      \G_\text{ii}^\ell & \tF^\ell_\text{i} \transpose{\mbox{$\tF^\ell_\text{t}$}}\\
      \tF^\ell_\text{t} \transpose{\mbox{$\tF^\ell_\text{i}$}} & \tF^\ell_\text{t} \transpose{\mbox{$\tF^\ell_\text{t}$}}
    \end{pmatrix}$
  \ENDFOR
  \STATE \textcolor{gray}{sample GP inducing outputs and update ELBO}
  \STATE $\F_\text{i}^{L+1} \sim \qc{\F_\text{i}^{L+1}}{\G^L_\text{ii}}$
  \STATE $\mathcal{L} \leftarrow \mathcal{L} + \log \pc{\F_\text{i}^{L+1}}{\G_\text{ii}^L} - \log \qc{\F_\text{i}^{L+1}}{\G_\text{ii}^L}$ 
  \STATE \textcolor{gray}{sample GP predictions conditioned on inducing points}
  \STATE $\F_\text{t}^{L+1} \sim \qc{\F_\text{t}^{L+1}}{\G^L, \F_\text{i}^{L+1}}$
  \STATE \textcolor{gray}{add likelihood to ELBO}
  \STATE $\mathcal{L} \leftarrow \mathcal{L} + \log \pc{\Y}{\F_\text{t}^{L+1}}$
\end{algorithmic}
\end{algorithm}

\subsection{Asymptotic complexity}
\label{sec:dwp:complexity}
Recalling that $\nu_\ell$ is the width of the $\ell$th layer, $M$ the number of inducing points, and $N$ the number of train or test points, the computational complexity of one DWP layer is given by $\mathcal{O}(M^3 + N M^2)$. 
This is a decrease of a factor of $\nu_{\ell}$ over the complexity for standard DGP inference, such as doubly stochastic variational inference \citep{salimbeni2017doubly}, which has complexity $O(\nu_{\ell}(M^3 + N M^2))$. 
The difference arises from the fact that in a DGP, $\nu_{\ell}$ Gaussian processes are sampled in each layer, whereas for a DWP we sample a single Gram matrix.

\section{Experimental Results}
We focus on comparing our DWP approximate posterior to those obtained using equivalent DGPs, i.e., those where the depth $L+1$ and width $\nu$ are the same.
To ensure equivalent models, we used a DGP model with a zero mean function, i.e., without the skip connections used by \citet{salimbeni2017doubly}).
In doing so, we found that local inducing posteriors such as the DSVI posterior from \citet{salimbeni2017doubly} were incapable of obtaining non-trivial results.
Therefore, to obtain a sensible comparison, we compare only to the global inducing posterior we proposed in the last chapter.
We first look at the cubic toy example from the previous chapter (Sec.~\ref{sec:gi:toy}), before moving to experiments on the UCI datasets from \citet{Gal2015DropoutB}.

\subsection{Visualizing the features}\label{sec:dwp:toy}
\begin{figure}
    \centering
    \includegraphics[width=0.8\textwidth]{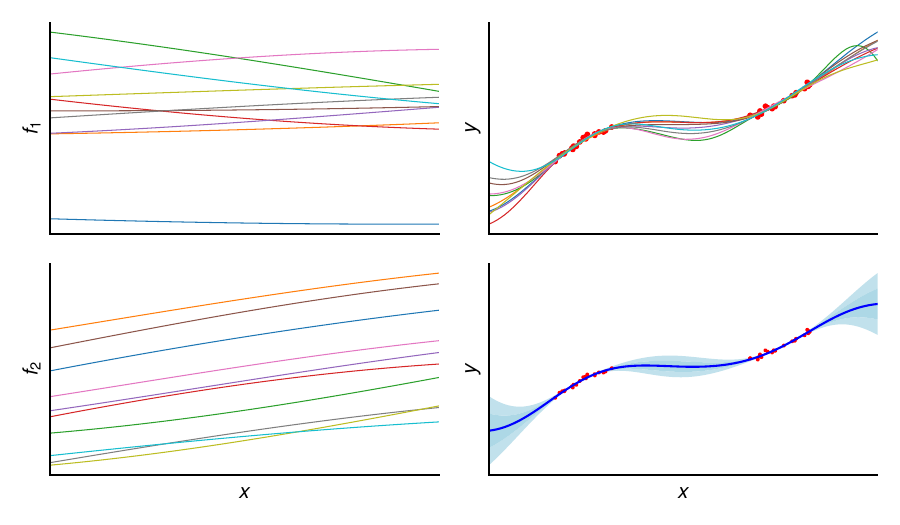}
    \caption{Features from a 2-layer DGP posterior with intermediate width 2: feature samples $f_1$ (first layer, first output; top left), $f_2$ (first layer, second output; bottom left), posterior samples (top right), posterior predictive (bottom right).}
    \label{fig:dwp:dgp_features}
\end{figure}

\begin{figure}
    \centering
    \includegraphics[width=0.8\textwidth]{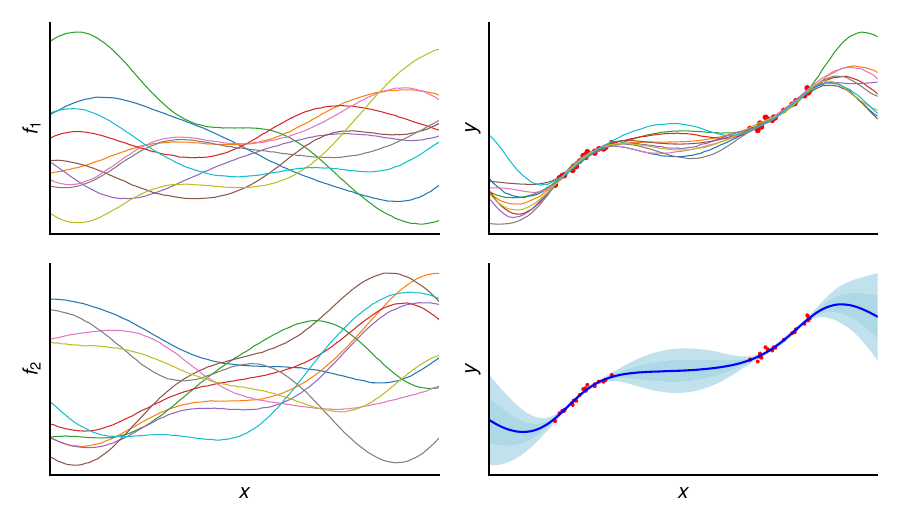}
    \caption{Features from a 2-layer DWP posterior with intermediate width 2: feature samples $f_1$ (first layer, first output; top left), $f_2$ (first layer, second output; bottom left), posterior samples (top right), posterior predictive (bottom right).}
    \label{fig:dwp:dwp_features}
\end{figure}

In this section, we compare intermediate-layer features for trained 2-layer, width-2 DWP and DGP models with squared exponential kernels on the cubic 1-dimensional toy example. 
We plot the intermediate samples for the DGP in Figure~\ref{fig:dwp:dgp_features}, and imagined features obtained from the sampled Gram matrices for the DWP in Figure~\ref{fig:dwp:dwp_features}.
We observe that the features learned by the DWP are both more interesting, and more varied. 
This allows for a greater predictive uncertainty in the posterior away from the data.
These improved characteristics are due to the increased ability of the DWP to capture true-posterior symmetries, which allow the DWP to learn a smaller lengthscale, leading to more interesting features.

\subsection{Experiments on the UCI datasets}
\newpage
\begin{table}[ht]
  \caption{ 
    ELBOs, test log-likelihoods, and test root mean square error for UCI datasets from \citep{Gal2015DropoutB} for five-layer models, comparing our DWP with equivalent DGP models.
    Better results are highlighted; see Appendix~\ref{sec:dwp:tables} for other depths and additional information.
    Note that for the test LLs and RMSEs we do not take the error bars into account for highlighting better results. 
    This is because we found most of the variance in these to be due to the splitting of the datasets, as opposed to the models having varying relative performance each time they are fit. 
    We therefore found that the ``better'' model according to the particular metric mean would generally perform better consistently across splits. 
    For the ELBOs, we do take the error bars into account, however, as the train set is large enough for the variance due to the splits to be small.
  }
  \label{tab:dwp:uci_comb}
  \centering
  \begin{tabular}{lrcc}
    \toprule
& dataset & DGP & DWP\\
\midrule 
& \textsc{\textsc{Boston}} & -0.45 $\pm$ 0.00 & \textbf{-0.37 $\pm$ 0.01} \\
& \textsc{Concrete} & -0.50 $\pm$ 0.00 & \textbf{-0.49 $\pm$ 0.00} \\
& \textsc{Energy} & 1.38 $\pm$ 0.00 & \textbf{1.40 $\pm$ 0.00} \\
& \textsc{Kin8nm} & -0.14 $\pm$ 0.00 & -0.14 $\pm$ 0.00 \\
ELBO & \textsc{Naval} & \textbf{3.92 $\pm$ 0.04} & 3.59 $\pm$ 0.12 \\
& \textsc{Power} & \textbf{0.03 $\pm$ 0.00} & 0.02 $\pm$ 0.00 \\
& \textsc{Protein} & \textbf{-1.00 $\pm$ 0.00} & -1.01 $\pm$ 0.00 \\
& \textsc{Wine} & -1.19 $\pm$ 0.00 & -1.19 $\pm$ 0.00 \\
& \textsc{Yacht} & 1.46 $\pm$ 0.02 & \textbf{1.59 $\pm$ 0.02} \\
\midrule
& \textsc{\textsc{Boston}} & -2.43 $\pm$ 0.04 & \textbf{-2.38 $\pm$ 0.04} \\
& \textsc{Concrete} & -3.13 $\pm$ 0.02 & -3.13 $\pm$ 0.02 \\
& \textsc{Energy} & -0.71 $\pm$ 0.03 & -0.71 $\pm$ 0.03 \\
& \textsc{Kin8nm} & 1.38 $\pm$ 0.00 & \textbf{1.40 $\pm$ 0.01} \\
LL & \textsc{Naval} & \textbf{8.28 $\pm$ 0.04} & 8.17 $\pm$ 0.07 \\
& \textsc{Power} & -2.78 $\pm$ 0.01 & \textbf{-2.77 $\pm$ 0.01} \\
& \textsc{Protein} & -2.73 $\pm$ 0.01 & \textbf{-2.72 $\pm$ 0.01} \\
& \textsc{Wine} & -0.96 $\pm$ 0.01 & -0.96 $\pm$ 0.01 \\
& \textsc{Yacht} & -0.73 $\pm$ 0.07 & \textbf{-0.58 $\pm$ 0.06} \\
\midrule
& \textsc{\textsc{Boston}} & \textbf{2.81 $\pm$ 0.14} & 2.82 $\pm$ 0.17 \\
& \textsc{Concrete} & \textbf{5.49 $\pm$ 0.10} & 5.53 $\pm$ 0.10 \\
& \textsc{Energy} & 0.49 $\pm$ 0.01 & \textbf{0.48 $\pm$ 0.01} \\
& \textsc{Kin8nm} & 0.06 $\pm$ 0.01 & 0.06 $\pm$ 0.01 \\
RMSE & \textsc{Naval} & 0.00 $\pm$ 0.00 & 0.00 $\pm$ 0.00 \\
& \textsc{Power} & 3.88 $\pm$ 0.04 & \textbf{3.84 $\pm$ 0.04} \\
& \textsc{Protein} & 3.77 $\pm$ 0.02 & \textbf{3.76 $\pm$ 0.02} \\
& \textsc{Wine} & 0.63 $\pm$ 0.01 & 0.63 $\pm$ 0.01 \\
& \textsc{Yacht} & 0.57 $\pm$ 0.05 & \textbf{0.50 $\pm$ 0.04} \\
\bottomrule
  \end{tabular}
\end{table}
For more quantitative experiments, we trained a DWP and a DGP with the exact same generative model with squared exponential kernels on UCI datasets.
We trained both models for 20,000 gradient steps using the Adam optimizer \citep{kingma2014adam}; we detail the exact experimental setup in Appendix~\ref{app:dwp:exp-details}.
We report ELBOs, test log likelihoods (LLs), and test root mean square error (RMSE) for depth 5 in Table~\ref{tab:dwp:uci_comb}; we report other depths and quote the relevant results using the DIWP from \citet{aitchison2020deep} in Appendix~\ref{sec:dwp:tables}.
We found that the DWP sometimes outperformed the DGP model, evident for instance if we look at the ELBOs and smaller datasets (\textsc{Boston}, \textsc{Concrete}, \textsc{Energy}, and \textsc{Yacht}).
On larger datasets, the benefits often disappear, which we hypthoesize is because uncertainty modelling is less important for good performance.
On the predictive metrics, the DWP was usually comparable or better on LLs while not as compelling on RMSEs, again suggesting that the DWP may perform better when it comes to uncertainty quantification.

\subsection{Runtimes \& training curves}
\begin{figure}
     \centering
     \includegraphics[width=\textwidth]{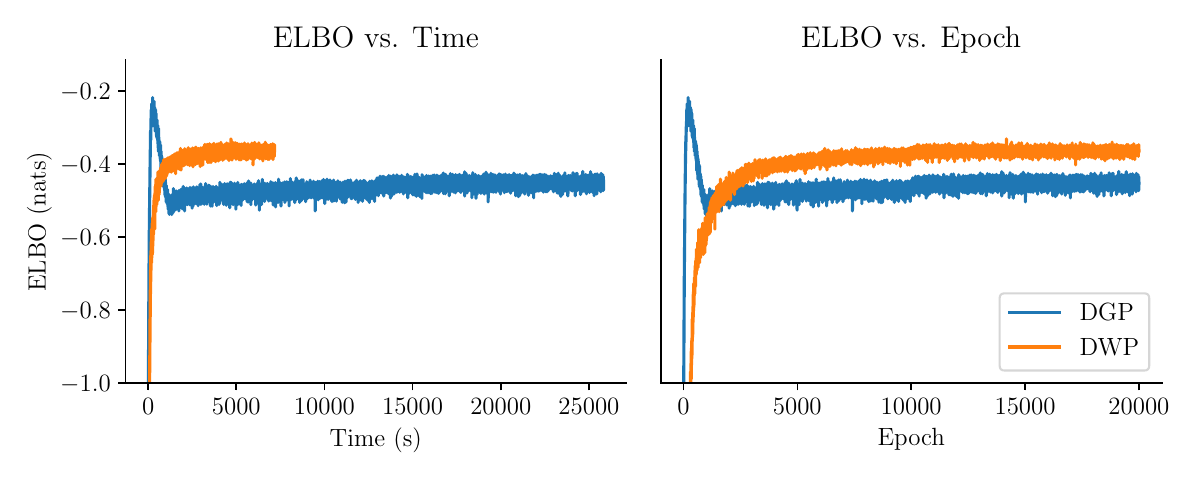}
        \caption{ELBO versus time and epoch for five-layer models on one split of \textsc{Boston}}
        \label{fig:dwp:runtime}
\end{figure}

In Sec.~\ref{sec:dwp:complexity}, we showed that DWPs have a lower computational complexity than DGPs, because of the need for DGPs to sample $\nu_\ell$ features in each layer, whereas DWPs only need to sample one Gram matrix.
Here, we briefly discuss the runtimes of our implementations.
We show a plot of the training curves for one split of the \textsc{Boston} dataset with a 5-layer DGP and DWP in Fig.~\ref{fig:dwp:runtime}, plotted against both runtime and epoch.
From these plots, we make two observations. 
First, the DWP trains much more quickly than the DGP in terms of runtime.
However, it seems to require slightly more epochs than the DGP to converge (note that the spike at the start of the DGP curve is an artifact of the tempering scheme we use).
In Appendix~\ref{sec:dwp:tables}, we provide a table of time per epoch, which shows that we obtain faster runtime for \textsc{Protein} and for shallower models, although the gains are slightly more modest due to the models being shallower and the fact that we run \textsc{Protein} on a GPU, as opposed to a CPU for \textsc{Boston}.

\section{Related work}
\subsection{Work prior to ours}
The DWP prior was introduced by \citet{aitchison2020deep}.
However, as already discussed, they were not able to do variational inference with the DWP because they did not have a sufficiently flexible approximate posterior over positive semi-definite matrices.
Instead, they were forced to work with a deep \textit{inverse} Wishart process, which is easier because the inverse Wishart itself is a suitable approximate posterior.
Here, we give a flexible generalized Wishart distribution over positive semi-definite matrices which is suitable for use as a variational approximate posterior in the DWP.
As the deep Wishart process prior is equivalent to a DGP prior, we were able to directly compare DGP and DWP inference in models with the exact same prior.
Such a comparison with equivalent priors was not possible in \citet{aitchison2020deep}, because the deep \textit{inverse} Wishart process priors studied therein are induce distinct function space priors to DGPs.

There is an alternative line of work using \textit{generalized} Wishart processes \citep{wilson2010generalised}, as opposed to our \textit{deep} Wishart processes.
A generalized Wishart process specifies a distribution over an infinite number of finite-dimensional Wishart-distributed matrices.
These matrices might represent, e.g., the noise covariance in a dynamical system, in which case there might be an infinite number of such matrices, one for each time or location in the state-space \citep{wilson2010generalised,heaukulani2019scalable,jorgensen2020stochastic}.
In contrast, the Wishart process \citep{dawid1981some,bru1991wishart} describes finite dimensional marginals of a single, potentially infinite dimensional matrix. 
In our context, we stack (non-generalized) Wishart processes to form a deep Wishart process.
Importantly, these generalized Wishart priors do not have the flexibility to capture a DGP prior because the underlying features at all locations are jointly multivariate Gaussian \citep[Sec.~4 in][]{wilson2010generalised} and therefore lack the required nonlinearities between layers.
Further, not only do the underlying stochastic processes (deep vs generalized Wishart process) differ, inference is also radically different.
In particular, work on the generalized Wishart performs inference on the underlying multivariate Gaussian feature vectors (e.g., Eq.\ 15--18 in \citealt{wilson2010generalised}, Eq.~12 in \citealt{heaukulani2019scalable}, and Eq.~24 in \citealt{jorgensen2020stochastic}).
Unfortunately, variational approximate posteriors defined over multivariate Gaussian feature vectors fail to capture symmetries in the true posterior (Sec.~\ref{sec:dwp:symmetry}).
In contrast, we define approximate posteriors directly over the symmetric positive semi-definite Gram matrices themselves, which required us to develop new, more flexible distributions over these matrices.

To the best of our knowledge, there has been only one prior work not yet mentioned that attempts to resolve the issues of symmetries in variational inference for deep Bayesian models.
Namely, \citet{pourzanjani2017improving} attempt to resolve both permutation symmetries and scaling symmetries in Bayesian neural networks by enforcing an ordering on the neurons and enforcing a single scale on the weight matrices.
These involve a change to the prior over weights which will have no effect on the prior over functions.
However, it is not immediately clear how one would extend their approach to resolve the rotational symmetries in DGPs that we focus on.
Nevertheless, this is a promising line of work that should be investigated further.

\subsection{Concurrent and subsequent work}
To the best of our knowledge, there has only been a limited amount of concurrent and subsequent work related to the topics we have discussed.
\citet{popescu2022tproc} propose an augmented inverse Wishart distribution which does not suffer as badly from the issues noted in Sec.~\ref{sec:dwp:diwpcomp}, where adding an inducing point can hurt the approximate posterior.
However, they only propose a fix which would be relevant for one size of $\U$: i.e., adding inducing points in more than one batch would still suffer from the issues we noted therein.
In an attempt to further understand the effect of symmetries on variational inference in deep models, \citet{kurle2022detrimental} identify \emph{translation} invariance as being important in neural networks.
They provide theoretical analysis on the effect of this invariance on MFVI for shallow models, relating it to the work done by \citet{coker2022wide}, and briefly apply those findings to deep BNNs.
While they do not provide a practical algorithm for mitigating effects of translation invariance, we believe this is an interesting line for future work.

\section{Limitations}\label{sec:dwp:limitations}
There are a few limitations of our work.
First, it is only possible to derive equivalent DWPs for certain kernels - namely, those where we can skip the feature representation and work entirely in Gram matrices.
While this holds for a large range of kernels, such as isotropic kernels and the arc-cosine kernel, it does not hold for some common kernels such as automatic relevance determination (ARD) kernels.
However, we note that we are able to use any kernel for the first layer, and that in practice we did not find that ARD kernels in intermediate layers significantly improved performance, as all the features have a shared prior. 
Moreover, we were not yet able to apply our work to BNNs, as the singularity of the input Gram matrix to the Wishart layers as well as the lack of an equivalent kernel function on Gram matrices makes this difficult.

The second main limitation is that it is not currently possible to incorporate modifications to the basic DGP model such as skip connections \citep{duvenaud2014avoiding, salimbeni2017doubly}, which are often important to good empirical performance in practice.
As we have noted, such modifications would require the use of the non-central Wishart distribution, which is difficult to evaluate in the inner loop of a deep learning algorithm. 
While we can obtain decent results without such modifications by using global inducing posteriors, they often significantly help the performance of the models.\footnote{For instance, the reader may be interested to compare the results of this chapter to those of Sec.~\ref{sec:gi:uci}, where ResNet-like skip connections as in \citet{salimbeni2017doubly} were used.}
Therefore, if good test performance is the main desired modeling outcome, it is advisable to still use a more standard DGP formulation.
Nevertheless, there are perhaps some scenarios where a DWP formulation may be preferable, for instance in Bayesian optimization, due to its improved uncertainty quantification.
However, we leave an investigation of this to future work.

Finally, the generalization of our work to more complex architectures such as convolutional models is non-trivial, and will also be left to future work.
Finally, it seems that the performance is not as competitive for larger datasets, where uncertainty representation is of more limited use, and that perhaps the predictive flexibility of the DWP is not as good as the DGP (considering the generally less compelling RMSE).
We now spend the remainder of the chapter exploring a couple of approaches to improve the performance of the deep Wishart process.

\subsection{Improving the generalized singular Wishart}\label{sec:dwp:improving}
Inspecting the form of the generalized singular Wishart we have proposed more closely, we make a few observations.
First, the individual terms in the generalized Bartlett are independent.
Moreover, the constraint that it be lower-triangular introduces an asymmetry in how different datapoints are treated with respect to its columns: some terms are distributed according to a gamma distribution whereas others are normal.
This contrasts with the explicit feature-based representation of a Wishart, whereby we sample $\f_\lambda$ columns according to multivariate Gaussians.
Note that this is not a problem for the non-generalized Wishart, as we are able to prove that the standard Bartlett decomposition results in the same density.
However, for our generalized Bartlett, this might be problematic, as we cannot prove that it is equivalent to, e.g., a non-central Wishart.
More specifically, we might expect the true posterior distribution over Gram matrices to have a certain \emph{directionality}, in that the implied features might have more uncertainty in some directions of the feature-space than others.
Therefore, we hypothesize that it may be beneficial to apply rotations and/or stretching of the columns of our generalized Bartlett.
Thus we propose the \emph{A-generalized singular Wishart} distribution, which is defined by applying a learned matrix $\A$ to the features:
\begin{defi}[The A-generalized singular Wishart distribution]
The A-generalized singular Wishart, $\text{A-}\gWish{\G; \S, \nu, \boldsymbol{\alpha}, \boldsymbol{\beta}, \boldsymbol{\mu}, \boldsymbol{\sigma}, \A}$ is a distribution over positive semi-definite $N \times N$ matrices $\G$, with positive definite scale matrix $\S = \L \transpose{\L} \in \mathbb{R}^{N\times N}$, a positive, integer-valued degrees of freedom parameter $\nu$, an invertible matrix $\A\in\reals^{N \times N}$, and Bartlett-generalizing parameters $\boldsymbol{\alpha}, \boldsymbol{\beta}, \boldsymbol{\mu}, \boldsymbol{\sigma}$. These latter parameters modify the Bartlett decomposition as follows:
\begin{subequations}
\begin{align*}
  \q{T_{jj}^2} &= \textup{Gamma}\b{T_{jj}^2;\,\alpha_j, \beta_j} && \text{for } j \leq \nu,\\
  \q{T_{i > j}} &= \Nc{T_{ij}}{\,\mu_{ij}, \sigma_{ij}^2} && \text{for } j \leq \nu. 
\end{align*}
\end{subequations}
We use $\A$ to rotate and stretch the sampled Bartlett columns, so that $\G = \L \A \T \transpose{(\L\A\T)}$.
This implies a distribution over $\G$ with density
\begin{multline*}
        \q{\G} = \b{\prod_{j=1}^N \frac{1}{L_{jj}^{\min \b{j, \nu}}}}\frac{\abs{\D_{:\nt, :\nt}}^{(\nu - N - 1)/2}}{|\A|^\nu\abs{\C_{:\nt, :\nt}}^{(\nu - N - 1)/2}} \times \\
    \prod_{j=1}^\nt \frac{\textup{Gamma}\b{T_{jj}^2;\,\alpha_j, \beta_j}}{T_{jj}^{N-j}L_{jj}^{N-j+1}}\prod_{i=j+1}^N \Nc{T_{ij}}{\mu_{ij}, \sigma_{ij}^2},
\end{multline*}
where we have defined $\C = \T\transpose{\T}$ and $\D = \A\C\transpose{\A}$.
\end{defi}

The addition of the $\A$ matrix to the generalized Wishart ensures that we can stretch and rotate the implicit features in the generalized Wishart. 
However, recall that we are interested in distributions over features, and not simply deterministic features.
While the addition of $\A$ would be sufficient to span the entire space of \emph{deterministic} features, the \emph{probabilistic} features we have may not be sufficiently flexible to adequately span the space of distributions over features.
We can attempt to further improve the flexibility of our generalized singular Wishart by introducing a learnable lower-triangular matrix $\B$ to right-multiply the Bartlett matrix, which may result in a more flexible basis of probabilistic features:
\begin{defi}[The AB-generalized singular Wishart distribution]
The AB-generalized singular Wishart, $\text{AB-}\gWish{\G; \S, \nu, \boldsymbol{\alpha}, \boldsymbol{\beta}, \boldsymbol{\mu}, \boldsymbol{\sigma}, \A, \B}$ is a distribution over positive semi-definite $N \times N$ matrices $\G$, with positive definite scale matrix $\S = \L \transpose{\L} \in \mathbb{R}^{N\times N}$, a positive, integer-valued degrees of freedom parameter $\nu$, an invertible matrix $\A\in\reals^{N \times N}$, an (invertible) lower-triangular matrix $\B \in \reals^{\nu \times \nu}$, and Bartlett-generalizing parameters $\boldsymbol{\alpha}, \boldsymbol{\beta}, \boldsymbol{\mu}, \boldsymbol{\sigma}$. These latter parameters modify the Bartlett decomposition as follows:
\begin{subequations}
\begin{align*}
  \q{T_{jj}^2} &= \textup{Gamma}\b{T_{jj}^2;\,\alpha_j, \beta_j} && \text{for } j \leq \nu,\\
  \q{T_{i > j}} &= \Nc{T_{ij}}{\mu_{ij}, \sigma_{ij}^2} && \text{for } j \leq \nu. 
\end{align*}
\end{subequations}
We use $\A$ to rotate and stretch the sampled Bartlett columns, and $\B$ to mix the Bartlett columns, so that $\G = \L \A \T \B \transpose{(\L\A\T\B)}$.
This implies a distribution over $\G$ with density
\begin{multline*}
        \q{\G} = \b{\prod_{j=1}^N \frac{1}{L_{jj}^{\min \b{j, \nu}}}}\frac{\abs{\D_{:\nt, :\nt}}^{(\nu - N - 1)/2}}{|\A|^\nu\abs{\C_{:\nt, :\nt}}^{(\nu - N - 1)/2}} \times \\
    \prod_{j=1}^\nt \frac{\textup{Gamma}\b{T_{jj}^2;\,\alpha_j, \beta_j}}{T_{jj}^{N-j}L_{jj}^{N-j+1}B_{jj}^{2(N-j+1)}}\prod_{i=j+1}^N \Nc{T_{ij}}{\mu_{ij}, \sigma_{ij}^2},
\end{multline*}
where we have defined $\C = \T\B\transpose{\b{\T\B}}$ and $\D = \A\C\transpose{\A}$.
\end{defi}
By introducing this matrix $\B$, we are mixing the columns of $\T$ in the hope that this will yield more flexibility in the implied features to allow the resulting AB-generalized Wishart to span a larger space of distributions than the A-generalized Wishart.

We provide the details on how to derive these densities in App.~\ref{app:jac}.
In practice, we parameterize $\A$ using an LU factorization, which makes the resulting density easier to compute.
\subsection{Results}
\begin{table}[!ht]
  \caption{
    ELBOs, test log-likelihoods, and test root mean square error for UCI datasets from \citep{Gal2015DropoutB} for five-layer models, comparing our different approximate posteriors.
    Better results are highlighted; see Appendix~\ref{sec:dwp:tables} for other depths and additional information.
    Note that for the test LLs and RMSEs we do not take the error bars into account for highlighting better results. 
    This is because we found most of the variance in these to be due to the splitting of the datasets, as opposed to the models having varying relative performance each time they are fit. 
    We therefore found that the ``better'' model according to the particular metric mean would generally perform better consistently across splits. 
    For the ELBOs, we do take the error bars into account, however, as the train set is large enough for the variance due to the splits to be small.
  }
  \label{tab:dwp:uci_genwish}
  \centering
  \begin{tabular}{lrccc}
    \toprule
& dataset & DWP & DWP-A & DWP-AB \\
\midrule 
& \textsc{\textsc{Boston}} & -0.37 $\pm$ 0.01 & -0.36 $\pm$ 0.00 & -0.36 $\pm$ 0.00 \\
& \textsc{Concrete} & -0.49 $\pm$ 0.00 & \textbf{-0.45 $\pm$ 0.00} & \textbf{-0.45 $\pm$ 0.00} \\
& \textsc{Energy} & 1.40 $\pm$ 0.00 & \textbf{1.42 $\pm$ 0.00} & 1.41 $\pm$ 0.00 \\
& \textsc{Kin8nm} & -0.14 $\pm$ 0.00 & \textbf{-0.11 $\pm$ 0.00} & \textbf{-0.11 $\pm$ 0.00} \\
ELBO & \textsc{Naval} & 3.59 $\pm$ 0.12 & \textbf{3.97 $\pm$ 0.02} & 3.63 $\pm$ 0.22 \\
& \textsc{Power} & 0.02 $\pm$ 0.00 & \textbf{0.03 $\pm$ 0.00} & \textbf{0.03 $\pm$ 0.00} \\
& \textsc{Protein} & -1.01 $\pm$ 0.00 & \textbf{-1.00 $\pm$ 0.00} & \textbf{-1.00 $\pm$ 0.00} \\
& \textsc{Wine} & -1.19 $\pm$ 0.00 & -1.19 $\pm$ 0.00 & -1.19 $\pm$ 0.00 \\
& \textsc{Yacht} & 1.59 $\pm$ 0.02 & \textbf{1.79 $\pm$ 0.02} & \textbf{1.79 $\pm$ 0.02} \\
\midrule
& \textsc{\textsc{Boston}} & \textbf{-2.38 $\pm$ 0.04} & -2.39 $\pm$ 0.05 & \textbf{-2.38 $\pm$ 0.04} \\
& \textsc{Concrete} & -3.13 $\pm$ 0.02 & \textbf{-3.07 $\pm$ 0.02} & -3.08 $\pm$ 0.02 \\
& \textsc{Energy} & -0.71 $\pm$ 0.03 & \textbf{-0.70 $\pm$ 0.03} & \textbf{-0.70 $\pm$ 0.03} \\
& \textsc{Kin8nm} & 1.40 $\pm$ 0.01 & \textbf{1.41 $\pm$ 0.01} & \textbf{1.41 $\pm$ 0.01} \\
LL & \textsc{Naval} & 8.17 $\pm$ 0.07 & \textbf{8.40 $\pm$ 0.02} & 8.10 $\pm$ 0.19 \\
& \textsc{Power} & -2.77 $\pm$ 0.01 & \textbf{-2.76 $\pm$ 0.01} & \textbf{-2.76 $\pm$ 0.01} \\
& \textsc{Protein} & -2.72 $\pm$ 0.01 & -2.71 $\pm$ 0.01 & \textbf{-2.70 $\pm$ 0.00} \\
& \textsc{Wine} & -0.96 $\pm$ 0.01 & -0.96 $\pm$ 0.01 & -0.96 $\pm$ 0.01 \\
& \textsc{Yacht} & -0.58 $\pm$ 0.06 & -0.22 $\pm$ 0.09 & \textbf{-0.18 $\pm$ 0.07} \\
\midrule
& \textsc{\textsc{Boston}} & 2.82 $\pm$ 0.17 & \textbf{2.77 $\pm$ 0.16} & 2.81 $\pm$ 0.17 \\
& \textsc{Concrete} & 5.53 $\pm$ 0.10 & 5.26 $\pm$ 0.11 & \textbf{5.24 $\pm$ 0.11} \\
& \textsc{Energy} & 0.48 $\pm$ 0.01 & 0.48 $\pm$ 0.01 & 0.48 $\pm$ 0.01 \\
& \textsc{Kin8nm} & 0.06 $\pm$ 0.01 & 0.06 $\pm$ 0.00 & 0.06 $\pm$ 0.00  \\
RMSE & \textsc{Naval} & 0.00 $\pm$ 0.00 & 0.00 $\pm$ 0.00 & 0.00 $\pm$ 0.00 \\
& \textsc{Power} & 3.84 $\pm$ 0.04 & \textbf{3.80 $\pm$ 0.04} & \textbf{3.80 $\pm$ 0.04} \\
& \textsc{Protein} & 3.76 $\pm$ 0.02 & 3.73 $\pm$ 0.02 &\textbf{3.70 $\pm$ 0.01} \\
& \textsc{Wine} & 0.63 $\pm$ 0.01 & 0.63 $\pm$ 0.01 & 0.63 $\pm$ 0.01 \\
& \textsc{Yacht} & 0.50 $\pm$ 0.04 & \textbf{0.37 $\pm$ 0.03} & 0.38 $\pm$ 0.03 \\
\bottomrule
  \end{tabular}
\end{table}
We compare our improved generalized singular Wishart distributions using the same setup on the UCI datasets as before.
We keep the form of the approximate posterior the same as before (cf. Sections~\ref{sec:dwp:approx_post} and~\ref{sec:dwp:dsvi}), only changing the generalized singular Wisharts used to the A- and AB-generalizations.
We demonstrate the results on five-layer architectures in Table~\ref{tab:dwp:uci_genwish}, where DWP denotes the original posterior, and DWP-A and DWP-AB refer to the A- and AB-generalizations, respectively.
The full results, which compare to the DGP model as well as on additional depths, are tabulated in App.~\ref{sec:dwp:tables}.

These results show near-universal improvement on all metrics (excepting those datasets where all methods seem to perform similarly) over the original generalized singular Wishart and hence over the DGP as well.
However, there is no clear winner between the A- and AB-generalizations, showing that it is perhaps unnecessary to combine the sampled Bartlett features.
Finally, Table~\ref{tab:dwp:runtime} shows that the A- and AB-generalizations incur negligible time cost over the original, suggesting that there is little reason not to use these new generalizations.

\section{Conclusions}
In this chapter, we have discussed how to practically perform inference in deep Wishart processes instead of their equivalent deep Gaussian processes.
We argued that performing inference on Gram matrices rather than features is advantageous, as it removes the rotational symmetries present in the true posterior.
We achieved this in practice by proposing a range of generalizations to the singular Wishart distribution, and demonstrating how to use these to perform doubly stochastic inducing point inference.
We showed that for a sufficiently flexible parameterization of the approximate posterior, we can improve over state-of-the-art inference in deep Gaussian processes in both performance and computational complexity.
However, even with our most flexible parameterizations, we do not match the performance of DGPs with skip connections (cf. Sec.~\ref{sec:gi:uci} and Tables~\ref{tab:gi:uci_dsvi2_lls}--\ref{tab:gi:uci_dsvi2_elbos}).
Nevertheless, we believe this demonstration shows the promise of deep kernel methods over feature-based methods, and we believe paves the way for exciting future research.

\chapter[Discussion]{Discussion}\label{sec:discuss}  

In this thesis, we have attempted to explore various aspects of variational inference for Bayesian deep learning.
We have focused our attention on three topics for deep models: the use of the marginal likelihood, improving approximate posteriors, and accounting for symmetries.
We now summarize our contributions concretely, before turning to future directions we would like to explore.
In order to facilitate this, we recall our motivation for using variational inference.

First, variational inference provides a natural means of trading off flexibility and compute: if we desire, we can always sacrifice compute to improve our approximation by increasing the flexibility of the approximate posterior.
This is in contrast to other approximations such as the Laplace approximation, which is limited to Gaussian approximations.
Second, by minimizing the KL divergence to the true posterior, variational inference simultaneously provides a lower bound to the log marginal likelihood.
For most other methods, it cannot be guaranteed that a marginal likelihood estimate is a lower bound, if it is even possible to obtain an estimate straightforwardly.\footnote{For instance, with standard MCMC methods, substantial algorithmic modifications are typically required to obtain estimates of the LML.}
For variational inference, this means that we can confidently maximize the evidence lower bound with respect to model hyperparameters without worrying that we are deviating from the behavior of the marginal likelihood, whereas for other methods we might be increasing the estimate away from the marginal likelihood.
This opens up the possibility for straightforward model selection using the ELBO.
While discussing our contributions and future directions, we therefore keep these two points in mind, asking 1) how have we exploited the ability of variational inference to trade off between flexibility and computation, and 2) how have we demonstrated the promise of the ELBO as a lower bound to the log marginal likelihood?

\section{Summary of contributions}
We now summarize our contributions in this thesis.
\begin{itemize}
    \item \textbf{Limitations of the marginal likelihood:} We began our contributions with Chapter~\ref{sec:dkl}, where we addressed an important question: what limitations might the marginal likelihood have? We investigated this question in the context of deep kernel learning. We found that when given a large number of hyperparameters, the marginal likelihood can be susceptible to overfitting. Surprisingly, we found that this overfitting can be worse than that exhibited by deterministic training of e.g., neural networks, casting doubt on the promise that the complexity penalty of the marginal likelihood will effectively regulate overfitting. By inspecting the data fit and complexity penalties of the LML more closely, we found that the overfitting is due to over-correlation of the data points. We showed that these limitations also applied to the ELBO when considering larger datasets and classification tasks. However, as DKL is a well-established technique, we argued that in practice its success may be due to stochastic minibatching, which provides implicit regularization of the ELBO/LML. Finally, we found that a Bayesian treatment of the large number of hyperparameters can mitigate this overfitting, which therefore should encourage us to be as ``fully Bayesian'' as possible.
    \item \textbf{Improved posteriors for BNNs and DGPs:} We continued our thesis in Chapter~\ref{sec:gi} by attempting to formulate a more effective structured approximate posterior for BNNs and DGPs. In particular, we argued that correlations between layers in an approximate posterior were essential for good performance, and so we proposed an approximate posterior, the global inducing point posterior, that induces correlations between all layers while also providing a unified view of BNNs and DGPs for inference. We further showed how to derive a memory-efficient version suitable for convolutional neural networks, and introduced new, more flexible priors. Experimentally, we showed that our approximate posterior gave better ELBOs, which usually resulted in better test performance. We also showed that the improved tightness of the ELBO to the marginal likelihood allowed our global inducing posteriors to take advantage of the additional flexibility in our learned priors, whereas the less flexible posteriors we tested against were not. Combining these contributions, we were able to obtain a state-of-the-art accuracy of 86.7\% for the CIFAR-10 dataset, without using either data augmentation or tempering.
    \item \textbf{Inference in a Gram matrix formulation of DGPs:} Finally, in Chapter~\ref{sec:dwp} we showed how even tighter ELBOs could be obtained for DGPs by analytically integrating out their rotational symmetries. This results in a model called the deep Wishart process, which for inference necessitated sufficiently flexible distributions over singular positive semi-definite matrices and an inducing point inference scheme. We demonstrated how to do both by introducing three generalizations of the singular Wishart distribution by using a generalized Bartlett decomposition, which could then be used as the basis of a global inducing point posterior. We then numerically demonstrated that we were able to obtain better ELBOs than equivalent global inducing DGPs, which in turn led to better test performance.
\end{itemize} 

In summary, our contributions have shed light on whether the LML or ELBO can be used reliably for deep models, and we have demonstrated that we can significantly improve the flexibility of tractable variational inference in deep models, thereby providing strides towards answering our first question.
However, while we have shown that our approximate posteriors do lead to better ELBOs that can facilitate better hyperparameter selection and predictive performance, we have not fully demonstrated the promise of the ELBO for model selection, leaving our second question at least partly unanswered.
We bear this in mind while turning to discuss future directions.

\section{Future directions}
We now discuss future directions of research that we hope will fully realize the promises of variational inference for Bayesian deep learning.
\begin{itemize}
    \item \textbf{The marginal likelihood:} Whereas past work \citep[e.g.,][]{mackay1992practical} has strongly advocated the use of the marginal likelihood as a metric for model selection, we have identified some of its limitations, and more recent work \citep{lotfi2022bayesian} has called this into question entirely. It would therefore be essential to further develop our understanding of the marginal likelihood as an indicator for model selection. Several questions here would be interesting. For instance, how many hyperparameters can the marginal likelihood handle before overfitting becomes an issue, as it is often impractical to integrate over \emph{all} hyperparameters? Could we develop more theoretical intuition for this? Finally, and most relevantly, can we theoretically quantify how well the marginal likelihood correlates with test performance or generalization?
    \item \textbf{Improved posteriors:} Our approximate posteriors have focused entirely on unimodal approximate posteriors. However, there is evidence in the literature that suggests that unimodal approximate posteriors may have poor performance \citep[e.g.,][]{lotfi2022bayesian, wilson2020bayesian} relative to multimodal posteriors, which may represent a much wider array of functions. One solution would be to use a mixture distribution, which was investigated to a limited extent in \citet{tomczak2018neural} for MFVI. However, as constructing a mixture over all the modes in a BNN or DGP is infeasible, the most pressing research question becomes: how can we decide which modes are the most important to include? Moreover, is it possible to reach a feasible point such that adding additional modes to a mixture posterior doesn't improve the performance? 
    \item \textbf{Improving the applicability of global inducing points and deep kernel processes:} While global inducing point posteriors tractably introduce correlations between all layers, they have some limitations which make them difficult to scale for non-fully-connected architectures. While we addressed these in part for CNNs, we are still effectively extracting far more patches than should be needed. This problem is exacerbated for deep convolutional Gaussian processes \citep{dutordoir2020bayesian, blomqvist2018deep}, where we cannot use the tricks developed in Sec.~\ref{sec:gi:efficient_conv} for CNNs. To fully resolve these issues, a framework for global \emph{interdomain} inducing points would need to be developed \citep{lazaro-gredilla2009inter}. Further to this, while we have successfully introduced an effective approximate posterior for deep Wishart processes, we are still severely limited in terms of the prior model architecture. For this line of work to truly be successful, we would require the level of flexibility we already have with BNNs and DGPs, for instance allowing for skip connections/non-zero mean functions and architectural elements such as convolutions.
    \item \textbf{Model selection:} One of the key promises of the variational framework is to be able to use model selection when the approximating family is flexible enough. While we have shown this to a limited extent with the priors we introduced in Chapter~\ref{sec:gi}, and \citet{bui2021biases} showed that global inducing posteriors obtain tight ELBOs across a range of hyperparameter values, it remains to be seen whether our approximate posteriors are flexible enough to allow for selection between differing architectures. In particular, can the ELBO successfully determine a model's depth and width? How can we obtain reliable (even if computationally expensive) estimates of the LML to compare our ELBOs to? If so, can we describe the biases introduced by the ELBO as compared to the LML for different architectures?
    \item \textbf{Software:} Finally, for the practitioner to be willing to use the techniques we have developed, it will be necessary to develop reliable and easy-to-use software. Indeed, the introduction of packages such as PyTorch \citep{paszke2019pytorch} and TensorFlow \citep{tensorflow2015-whitepaper} have been extraordinarily influential in increasing the spread and popularity of deep learning techniques. Our aim, therefore, would be to create a package that allows the user to build their own variational Bayesian models, including GPs, BNNs, DGPs, and DKPs, within a unified framework, as effortlessly as with current deep learning packages.
\end{itemize}

While there is clearly much work to be done to fully realize the promises of variational inference in deep Bayesian learning, we are hopeful that there is a future where the problems of uncertainty quantification and model selection can be jointly solved using this unique framework.


\begin{spacing}{0.9}


\bibliographystyle{plainnat} 
\cleardoublepage
\bibliography{References/references} 



\end{spacing}


\begin{appendices} 

\chapter{Derivations of Jacobians and Wishart Densities} \label{app:jac}

In this Appendix, we provide further background on Wishart and inverse Wishart distributions, and provide derivations of the densities for our generalized singular Wishart distributions.

\section{Background}
We begin by providing additional background on the Wishart, the inverse Wishart, and the Bartlett decomposition.

\subsection{The Wishart distribution}

The Wishart, $\Wish{\S, \nu}$, is a distribution over positive semi-definite $N \times N$ matrices, $\G$, with positive definite scale parameter $\S\in\mathbb{R}^{N\times N}$ and a positive, integer-valued degrees-of-freedom parameter, $\nu$.
The Wishart distribution is defined by taking $\nu$ vectors $\f_\lambda \in \mathbf{R}^N$ sampled from a zero-mean Gaussian with covariance $\S$.
These vectors can be generated from standard Gaussian vectors, $\boldsymbol{\xi}_\lambda$, by transforming them with the Cholesky $\L$ of the scale parameter, $\S = \L \transpose{\L}$:
\begin{align*}
  \L \boldsymbol{\xi}_\lambda &= \f_\lambda \sim \N{\0, \S} \quad\quad \text{ where } \quad\quad \boldsymbol{\xi}_\lambda \sim \N{\0, \I}.
\end{align*}
Both $\f_\lambda$ and $\boldsymbol{\xi}_\lambda$ can be stacked to form $N \times \nu$ matrices, $\F$ and $\mathbf{\Xi}$,
\begin{align*}
  \F &= \b{\f_1 \quad \f_2 \quad \dotsm \quad \f_\nu}, &
  \mathbf{\Xi} &= \b{\boldsymbol{\xi}_1 \quad \boldsymbol{\xi}_2 \quad \dotsm \quad \boldsymbol{\xi}_\nu}.
\end{align*}
Wishart samples are defined by taking the sum of the outer products of the sampled $\f_\lambda$ vectors, which can be written as a matrix multiplication:
\begin{align}
  \label{eq:jac:def:wishart}
  \sum_{\lambda=1}^\nu \f_\lambda \transpose{\f}_\lambda = \F \transpose{\F} &= \L \mathbf{\Xi} \transpose{\mathbf{\Xi}} \transpose{\L} =  \L \H \transpose{\L} = \G \sim \Wish{\S, \nu},
  \intertext{where $\H=\mathbf{\Xi} \transpose{\mathbf{\Xi}}$ is a sample from a \emph{standard Wishart} (i.e., one with an identity scale parameter),}
  \label{eq:jac:def:standard_wishart}
  \sum_{\lambda=1}^\nu \boldsymbol{\xi}_\lambda \transpose{\boldsymbol{\xi}}_\lambda &= \mathbf{\Xi} \transpose{\mathbf{\Xi}} = \H \sim \Wish{\I, \nu}.
\end{align}
Note that therefore the Wishart has mean
\begin{align}
  \label{eq:jac:mean_wishart}
  \mathbb{E}\sqb{\G} &= \nu \mathbb{E}\sqb{\f_\lambda \transpose{\f}_\lambda} = \nu \S.
\end{align}
Summarizing these equations, we obtain the definition for the Wishart distribution:

\defwish*
This definition encapsulates both singular and non-singular Wisharts.
As in the main text, we note that for the case of $\nu \geq N$, we can extend the definition of the Wishart to include non-integer $\nu$, as the density still applies in that case.

\subsection{The inverse Wishart distribution}
For $\nu \geq N$, it can be shown that samples from the Wishart distribution are almost surely full rank.
Therefore, we can sensibly consider the distribution of the inverse of a Wishart-distributed random variable:

\begin{defi}[The inverse Wishart distribution; {\citealp[][, Sec. 4.5.1]{murphy2012machine}}]
\label{defi:jac:iwish}
Consider a matrix $\G \in \reals^{N \times N}$, where we have that $\G^{-1} \sim \Wish{\S^{-1}, \nu}$ with $\nu \geq N$.
Then $\G$ is said to be distributed according to the \emph{inverse Wishart distribution}, denoted $\G \sim \iWish{\S, \nu}$, and has density
\begin{align*}
    \p{\G} = \frac{|\S|^{\nu/2}}{2^{\nu N/2}\Gamma_{N}\b{\tfrac{\nu}{2}}}|\G|^{-(\nu + N + 1)/2}\etr \b{-\S\G^{-1}/2},
\end{align*}
where $\Gamma_N\b{\cdot}$ is the multivariate gamma function.
\end{defi}
It is possible to show that the inverse Wishart distribution has mean
\begin{align*}
    \mathbb{E}\sqb{\G} = \frac{\S}{\nu - N - 1}
\end{align*}
for $\nu > N + 1$.

\subsection{The Bartlett decomposition}
The Bartlett decomposition \citep{bartlett1933on} allows full-rank standard Wishart random variables, $\G \sim \Wish{\I_N, \nu}$ with $\nu \geq N$, to be sampled efficiently, particularly for large or non-integer $\nu$.
Instead of sampling $\F$, the Bartlett decomposition samples the Cholesky decomposition of $\G$, which we denote $\T$ (so that $\G = \T \transpose{\T}$).
According to the Bartlett decomposition, 
\begin{align*}
\T &= \begin{pmatrix}
  T_{11} & \dotsm & 0 \\
  \vdots & \ddots & \vdots \\
  T_{N1} & \dotsm & T_{NN}
\end{pmatrix},\\
\p{T_{jj}^2} &= \text{Gamma}\b{T_{jj}^2;\, \alpha{=}\tfrac{\nu-j+1}{2}, \beta{=}\tfrac{1}{2}},\\
\p{T_{j > k}} &= \Nc{T_{jk}}{\,0, 1}. 
\end{align*}
In words, the square of the on-diagonal elements of the Bartlett factor are gamma-distributed (with the shape-rate parameterization), whereas the off-diagonals are i.i.d. standard Gaussian.
For Wishart distributions with non-identity scale matrices, we can sample $\G$ by using the Cholesky decomposition $\L$ of $\S = \L \transpose{\L}$, so that $\G = \L \T \transpose{\T} \transpose{\L}$.
We note that this is particularly useful for sampling from Wisharts with non-integer $\nu$, where it does not make sense to sample a non-integer number of columns $\f_\lambda$.

\section{Deriving Jacobians for matrix transformations}\label{sec:app:jac:deriving}
We now turn to deriving the Jacobians to be used in deriving the densities our generalized Wishart distributions.
Throughout this text, note that we use the term ``Jacobian'' to refer to the ``Jacobian determinant.''
Following the approach in \citet{mathai1997jacobians,mathai2008special}, we define the Jacobian of a function from $x$ to $y$ as the ratio of volume elements:
\begin{align*}
  \text{jacobian} &= \frac{dy_1 dy_2 \dotsm dy_N}{dx_1 dx_2 \dotsm dx_N}.
\end{align*}
Importantly, in this notation $dx_i$ and $dy_i$ are basis vectors, \textit{not} scalars. 
As we are multiplying vectors, not scalars, we need to take care with our choice of multiplication operation. 
The correct choice \citep{mathai1997jacobians} in our context is an antisymmetric exterior product, representing a directed area or volume element, such that
\begin{align*}
  dx_i dx_j &= - dx_j dx_i.
\end{align*}
As the product is antisymmetric, the product of a basis vector with itself is zero:
\begin{align*}
  dx_i dx_i &= - dx_i dx_i = 0,
\end{align*}
which makes sense because the product represents an area, and the area is zero if the two vectors are aligned.
To confirm that this matches usual expressions for Jacobians, consider a $2 \times 2$ matrix-vector multiplication, $\y = \A \x$:
\begin{align*}
  \begin{pmatrix} dy_1 \\ dy_2 \end{pmatrix} &=
  \begin{pmatrix} A_{11} & A_{12} \\ A_{21} & A_{22} \end{pmatrix}
  \begin{pmatrix} dx_1 \\ dx_2 \end{pmatrix} = 
  \begin{pmatrix} A_{11} dx_1 + A_{12} dx_2 \\ A_{21} dx_1 + A_{22} dx_2 \end{pmatrix}.
\end{align*}
Therefore,
\begin{align*}
  dy_1 dy_2 &= \b{A_{11} dx_1 + A_{12} dx_2} \b{A_{21} dx_1 + A_{22} dx_2}\\
   &= A_{11} A_{21} dx_1^2 + A_{11} A_{22} dx_1 dx_2 + A_{12} A_{21} dx_2 dx_1 + A_{12} A_{22} dx_2^2.
\end{align*}
As $dx_1^2 = dx_2^2 = 0$, and $dx_1 dx_2 = -dx_1 dx_2$, we have
\begin{align*}
  dy_1 dy_2 &= \b{A_{11} A_{22} - A_{12} A_{21}} dx_1 dx_2\\
  &= \abs{\A} dx_1 dx_2,
\end{align*}
so that the Jacobian obtained computed using the ratio of volume elements is indeed correct.
Note that as we are dealing with probability density functions, we will typically ignore the sign of the Jacobians that we compute. 

To derive Jacobians for matrix-matrix transformations, it is sometimes helpful to consider an equivalent formulation whereby we vectorize the matrices we are interested in, and take the determinant of the resulting transformation matrix.
As a concrete example, consider
\begin{align*}
    \underbrace{\begin{pmatrix} Y_{11} & Y_{12} \\ Y_{21} & Y_{22} \end{pmatrix}}_{\Y} &=
    \underbrace{\begin{pmatrix} A_{11} & A_{12} \\ A_{21} & A_{22} \end{pmatrix}}_{\mathbf{A}}
    \underbrace{\begin{pmatrix} X_{11} & X_{12} \\ X_{21} & X_{22} \end{pmatrix}}_{\X}.
\end{align*}
We vectorize $\Y$ and $\X$ to obtain
\begin{align*}
    \begin{pmatrix}
      Y_{11} \\
      Y_{21} \\
      Y_{12} \\
      Y_{22}
    \end{pmatrix}
    &=
    \underbrace{\begin{pmatrix}
      A_{11} & A_{12} & 0 & 0 \\
      A_{21} & A_{22} & 0 & 0 \\
      0 & 0 & A_{11} & A_{12} \\
      0 & 0 & A_{21} & A_{22}
    \end{pmatrix}}_{\mathbf{A}^*}
    \begin{pmatrix}
      X_{11} \\
      X_{21} \\
      X_{12} \\
      X_{22}
    \end{pmatrix}.
\end{align*}
The Jacobian of this transformation is clearly $|\mathbf{A}|^2$.
To verify that this is indeed correct, we note that 
\begin{align*}
    d\X = dX_{11}dX_{21}dX_{12}dX_{22} = d\X_{:, 1}d\X_{:, 2}, \\
    d\Y = dY_{11}dY_{21}dY_{12}dY_{22} = d\Y_{:, 1}d\Y_{:, 2}.
\end{align*}
Since
\begin{align*}
    d\Y_{:, 1} = |\mathbf{A}| d\X_{:, 1}, \\
    d\Y_{:, 2} = |\mathbf{A}| d\X_{:, 2},
\end{align*}
we have that 
\begin{align*}
    d\Y = |\mathbf{A}|^2 d\X_{:, 1} d\X_{:, 2} = |\mathbf{A}|^2 d\X,
\end{align*}
which verifies our above calculation.

We now briefly consider how to compute Jacobians for matrices with low-rank or other special (e.g., symmetric) structure.
These matrices will have fewer deterministic elements than unstructured full-rank matrices.
For instance, a low-rank $N\times N$ matrix of rank $p < N$ will be uniquely determined by $2Np - p^2$ elements, whereas an $N\times N$ full-rank, symmetric matrix will be uniquely determined by $N(N+1)/2$ elements.
In these cases, we must therefore identify an equal number of \textit{functionally independent} variables in both $\X$ and $\Y$ to evaluate the Jacobian with \citep{mathai1997jacobians, srivastava2003singular}.
For instance, in the low-rank case, we would choose $2Np - p^2$ arbitrary elements of both $\X$ and $\Y$, whereas in the symmetric case we would choose to inspect (without loss of generality) the lower triangular elements of the matrices.

\subsection{Jacobian for the product of a lower triangular matrix with itself}
\label{sec:jac:J_LLT}
In this section, we compute the Jacobian for the transformation from $\La \in \reals^{N \times N}$ to $\G=\La \transpose{\La}$, where $\La$ is lower triangular.
We begin by noting that the top left block of the product of a lower triangular matrix with itself is a product of smaller lower triangular matrices:
\begin{align*}
  \begin{pmatrix}
     \La_{:p, :p} & \0\\
     \La_{p+1:, :p} & \La_{p+1:, p+1:}
  \end{pmatrix}
  \begin{pmatrix}
     \transpose{\La}_{:p, :p} & \transpose{\La}_{p+1:, :p}\\
     \0 & \transpose{\La}_{p+1:, p+1:}
  \end{pmatrix}
  &=
  \begin{pmatrix}
    \La_{:p, :p} \transpose{\La}_{:p, :p}  & \dotsm\\
    \vdots  & \ddots\\
  \end{pmatrix}.
\end{align*}
We first demonstrate on a $2\times 2$ matrix, before proceeding by induction:   
\begin{align*}
  \begin{pmatrix}
    G_{11} & G_{12}\\
    G_{21} & G_{22}
  \end{pmatrix}
  =
  \begin{pmatrix}
     \Lambda_{11} & 0\\
     \Lambda_{21} & \Lambda_{22}
  \end{pmatrix}
  \begin{pmatrix}
     \Lambda_{11} & \Lambda_{21}\\
     0 & \Lambda_{22}
  \end{pmatrix}
  =
  \begin{pmatrix}
    \Lambda_{11}^2 & \Lambda_{21}\Lambda_{11}\\
    \Lambda_{21} \Lambda_{11} & \Lambda_{21}^2 + \Lambda_{22}^2
  \end{pmatrix}.
\end{align*}
We start with the top-left $1 \times 1$ matrix,
\begin{align}
  G_{11} &= \Lambda_{11}^2,\\
  \label{eq:jac:dG11}
  d G_{11} &= 2 \Lambda_{11} d\Lambda_{11}.
\end{align}
Considering the rest of the matrix (noting that, following the above discussion, we only look at the lower triangular part of $\mathbf{G}$):
\begin{align*}
  d G_{21} &= \Lambda_{21} d\Lambda_{11} + \Lambda_{11} d\Lambda_{21},\\
  d G_{22} &= 2 \Lambda_{22} d\Lambda_{22} + 2 \Lambda_{21} d\Lambda_{21}.
  \intertext{Combining $d G_{11}$ and $d G_{21}$ gives}
  d G_{11} d G_{21} &= \b{2 \Lambda_{11} d\Lambda_{11}} \b{\Lambda_{21} d\Lambda_{11} + \Lambda_{11} d\Lambda_{21}}\\
  &= 2 \Lambda_{11}^2 \b{d\Lambda_{11} d\Lambda_{21}},\\
  \intertext{and then combining $d G_{11} dG_{21}$ and $d G_{22}$ gives}
  dG_{11} dG_{21} dG_{22} &= \b{2 \Lambda_{11}^2 \b{d\Lambda_{11} d\Lambda_{21}}} \b{2 \Lambda_{22} d\Lambda_{22} + 2 \Lambda_{21} d\Lambda_{21}}\\
   &= 4 \Lambda_{11}^2 \Lambda_{22} \b{d\Lambda_{11} d\Lambda_{21} d\Lambda_{22}}.
\end{align*}

By following this process, we can prove by induction that the volume element for the top left $p\times p$ block of $\G$, and in addition the first $K < p+1$ off-diagonal elements of the $p+1^\text{th}$ row is
\begin{align*}
  \underbrace{\b{\prod_{i=1}^p \prod_{k=1}^i d G_{ik}}}_\text{vol.\ elem.\ for $\G_{:p, :p}$} \underbrace{\b{\prod_{k=1}^K d G_{p+1,k}}}_\text{vol.\ elem.\ for $\G_{p+1, :K}$} 
  &= 2^p \b{\prod_{i=1}^p \prod_{k=1}^i \Lambda_{kk} d\Lambda_{ik}} \b{\prod_{k=1}^K \Lambda_{kk} d \Lambda_{p+1,k}}.
\end{align*}
The proof consists of three parts: the base case, adding an off-diagonal element, and adding an on-diagonal element.
For the base case, note that the expression is correct for $p=1$ and $K=0$ (Eq.~\ref{eq:jac:dG11}).
Next, we add an off-diagonal element, $G_{p+1, K+1}$, where $K+1 < p+1$.
We begin by computing $dG_{p+1, K+1}$:
\begin{align*}
  G_{p+1, K+1} &= \sum_{j=1}^{K+1} \Lambda_{p+1, j} \Lambda_{K+1, j},\\
  d G_{p+1, K+1} &= \sum_{j=1}^{K+1} \b{\Lambda_{K+1, j} d \Lambda_{p+1, j} + \Lambda_{p+1, j} d \Lambda_{K+1, j}}.
\end{align*}
Note that the sum only goes to $K+1$, because $\Lambda_{K+1, j} = 0$ for $j > (K+1)$.
Remembering that $d\Lambda_{ij}^2 = 0$, the only term that does not cancel when we multiply by the volume element for the previous terms is that for $d\Lambda_{p+1, K+1}$:
\begin{align*}
  \underbrace{\b{\prod_{i=1}^p \prod_{k=1}^i d G_{ik}}}_\text{vol.\ elem.\ for $\G_{:p, :p}$} & \underbrace{\b{\prod_{k=1}^{K+1} d G_{p+1,k}}}_\text{vol.\ elem.\ for $\G_{p+1, :K+1}$} =
  \b{\prod_{i=1}^p \prod_{k=1}^i d G_{ik}} \b{\prod_{k=1}^{K} d G_{p+1,k}} dG_{p+1, K+1} \\
  &= 2^p \b{\prod_{i=1}^p \prod_{k=1}^i \Lambda_{kk} d\Lambda_{ik}} \b{\prod_{k=1}^K \Lambda_{kk} d\Lambda_{p+1,k}} d G_{p+1, K+1}\\
  &= 2^p \b{\prod_{i=1}^p \prod_{k=1}^i \Lambda_{kk} d\Lambda_{ik}} \b{\prod_{k=1}^K \Lambda_{kk} d\Lambda_{p+1,k}} \b{\Lambda_{K+1, K+1} d \Lambda_{p+1, K+1}}\\
  &= 2^p \b{\prod_{i=1}^p \prod_{k=1}^i \Lambda_{kk} d\Lambda_{ik}} \b{\prod_{k=1}^{K+1} \Lambda_{kk} d\Lambda_{p+1,k}}.
\end{align*}
So the expression is consistent when adding an on-diagonal element.
Finally, the volume element for $G_{p+1, p+1}$ is given by
\begin{align*}
  G_{p+1, p+1} &= \sum_{j=1}^{p+1} \Lambda_{p+1, j}^2, \\
  d G_{p+1, p+1} &= 2 \sum_{j=1}^{p+1} \Lambda_{p+1, j} d \Lambda_{p+1, j}.
\end{align*}
Remembering again that $d\Lambda_{ij}^2 = 0$, the only term that does not cancel when we multiply by the volume element for the previous terms is that for $d\Lambda_{p+1, p+1}$, giving us
\begin{align*}
  \underbrace{\b{\prod_{i=1}^{p+1} \prod_{k=1}^i d G_{ik}}}_\text{vol.\ elem.\ for $\G_{:p+1, :p+1}$} &=
  \underbrace{\b{\prod_{i=1}^p \prod_{k=1}^i d G_{ik}}}_\text{vol.\ elem.\ for $\G_{:p, :p}$} \underbrace{\b{\prod_{k=1}^{p+1} d G_{p+1,k}}}_\text{vol.\ elem.\ for $\G_{p+1, :p+1}$}\\
  &= \b{\prod_{i=1}^p \prod_{k=1}^i d G_{ik}} \b{\prod_{k=1}^{p} d G_{p+1,k}} d G_{p+1,p+1}\\
  &= 2^p \b{\prod_{i=1}^p \prod_{k=1}^i \Lambda_{kk} d\Lambda_{ik}} \b{\prod_{k=1}^p \Lambda_{kk} d \Lambda_{p+1,k}} d G_{p+1, p+1}\\
  &= 2^p \b{\prod_{i=1}^p \prod_{k=1}^i \Lambda_{kk} d\Lambda_{ik}} \b{\prod_{k=1}^p \Lambda_{kk} d \Lambda_{p+1,k}} \b{2 \Lambda_{p+1, p+1} d \Lambda_{p+1, p+1}}\\
  &= 2^{p+1} \b{\prod_{i=1}^p \prod_{k=1}^i \Lambda_{kk} d\Lambda_{ik}} \b{\prod_{k=1}^{p+1} \Lambda_{kk} d \Lambda_{p+1,k}}\\
  &= 2^{p+1} \b{\prod_{i=1}^{p+1} \prod_{k=1}^i \Lambda_{kk} d\Lambda_{ik}}.
\end{align*}
Thus, the final result is:
\begin{align}
  \b{\prod_{i=1}^{N} \prod_{k=1}^i d G_{ik}} &= 
  \b{2^N \prod_{i=1}^{N} \Lambda_{ii}^{N - i + 1}} \b{\prod_{i=1}^{N} \prod_{k=1}^i d \Lambda_{ik}},\\
\end{align}
which we can rewrite as
\begin{align}
  \label{eq:jac:J_LaLaT}
  d\G &= d\La \prod_{i=1}^N 2 \Lambda_{ii}^{N - i + 1}.
\end{align}

\subsubsection{Singular matrices}
\label{sec:jac:J_LaLaT}
The above derivation can be extended to the singular case, where $\La \in \reals^{N \times \nu}$:
\begin{align}
  \label{eq:jac:La}
  \La &= \begin{pmatrix}
    \Lambda_{11}    & \dotsm & 0          \\ 
    \vdots          & \ddots & \vdots     \\ 
    \Lambda_{\nu 1} & \dotsm & \Lambda_{\nu \nu} \\ 
    \vdots          & \vdots & \vdots     \\ 
    \Lambda_{N 1}   & \dotsm & \Lambda_{N \nu}
  \end{pmatrix}.
\end{align}
As discussed above (App.~\ref{sec:app:jac:deriving}), to form a valid Jacobian we need the same number of inputs as outputs.
We therefore consider differences in only the corresponding part of $\G$ (i.e., $G_{i, j\leq \min(i,\nu)}$).
The recursive expression is
\begin{align*}
  \underbrace{\b{\prod_{i=1}^p \prod_{k=1}^{\min(i, \nu)} d G_{ik}}}_\text{vol.\ elem.\ for $\G_{:p, :p}$} \underbrace{\b{\prod_{k=1}^K d G_{p+1,k}}}_\text{vol.\ elem.\ for $\G_{p+1, :K}$} 
  &= 2^{\min(p, \nu)} \b{\prod_{i=1}^p \prod_{k=1}^{\min(i, \nu)} \Lambda_{kk} d\Lambda_{ik}} \b{\prod_{k=1}^K \Lambda_{kk} d\Lambda_{p+1,k}},
\end{align*}
where $K < \min(p, \nu)$.
For $N \leq \nu$, the recursion is exactly as in the full-rank case above.
For $N > \nu$, the key difference is that there are no longer any on-diagonal elements.
As such, for $K=\nu$ we have
\begin{align*}
  \b{\prod_{i=1}^{p+1} \prod_{k=1}^{\min(i, \nu)} d G_{ik}} &= 
  \underbrace{\b{\prod_{i=1}^p \prod_{k=1}^{\min(i, \nu)} d G_{ik}}}_\text{vol.\ elem.\ for $\G_{:p, :p}$} \underbrace{\b{\prod_{k=1}^\nu d G_{p+1,k}}}_\text{vol.\ elem.\ for $\G_{p+1, :\nu}$}\\
  &= 2^{\min(p, \nu)} \b{\prod_{i=1}^p \prod_{k=1}^{\min(i, \nu)} \Lambda_{kk} d\Lambda_{ik}} \b{\prod_{k=1}^K \Lambda_{kk} d\Lambda_{p+1,k}}\\
  &= 2^{\min(p, \nu)} \b{\prod_{i=1}^{p+1} \prod_{k=1}^{\min(i, \nu)} \Lambda_{kk} d\Lambda_{ik}}. 
\end{align*}
The final expression, allowing for the possibility of singular and non-singular matrices, is thus
\begin{align}
  \b{\prod_{i=1}^{N} \prod_{k=1}^{\min(i, \nu)} d G_{ik}} &= 
  \b{\prod_{i=1}^\nt 2 \Lambda_{ii}^{N - i + 1}} \b{\prod_{i=1}^{N} \prod_{k=1}^{\min(i, \nu)} d \Lambda_{ik}}\\
  d\G &= d\La \prod_{i=1}^\nt 2 \Lambda_{ii}^{N - i + 1}.\label{eq:jac:J_LaLaTsing}
\end{align}

\subsection{Jacobian for the product of two different lower triangular matrices}
\subsubsection{From $\T$ to $\La = \L\T$}
\label{sec:jac:J_LA}
In this section, we compute the Jacobian for the transformation from $\T \in \reals^{N \times \nu}$ to $\La = \L \T$, where $\T$ is potentially low rank (i.e., $\nu < N$).
We begin by noting that $\La$ (Eq.~\ref{eq:jac:La}) is a potentially rectangular lower-triangular matrix, with the same structure as $\T$.
Writing this out with a concrete example, where $\T \in \reals^{5 \times 3}$,
\begin{align*}
  \begin{pmatrix}
     \Lambda_{11} & 0      & 0 \\
     \Lambda_{21} & \Lambda_{22} & 0 \\
     \Lambda_{31} & \Lambda_{32} & \Lambda_{33} \\
     \Lambda_{41} & \Lambda_{42} & \Lambda_{43} \\
     \Lambda_{51} & \Lambda_{52} & \Lambda_{53}
  \end{pmatrix}
  &=
  \begin{pmatrix}
     L_{11} & 0      & 0      & 0      & 0\\
     L_{21} & L_{22} & 0      & 0      & 0\\
     L_{31} & L_{32} & L_{33} & 0      & 0\\
     L_{41} & L_{42} & L_{43} & L_{44} & 0\\
     L_{51} & L_{52} & L_{53} & L_{54} & L_{55}
  \end{pmatrix}
  \begin{pmatrix}
     T_{11} & 0      & 0 \\
     T_{21} & T_{22} & 0 \\
     T_{31} & T_{32} & T_{33} \\
     T_{41} & T_{42} & T_{43} \\
     T_{51} & T_{52} & T_{53}
  \end{pmatrix}.
\end{align*}
For the first column,
\begin{align*}
  \begin{pmatrix}
     \Lambda_{11} \\
     \Lambda_{21} \\
     \Lambda_{31} \\
     \Lambda_{41} \\
     \Lambda_{51}
  \end{pmatrix}
  &=
  \begin{pmatrix}
     L_{11} & 0      & 0      & 0      & 0\\
     L_{21} & L_{22} & 0      & 0      & 0\\
     L_{31} & L_{32} & L_{33} & 0      & 0\\
     L_{41} & L_{42} & L_{43} & L_{44} & 0\\
     L_{51} & L_{52} & L_{53} & L_{54} & L_{55}
  \end{pmatrix}
  \begin{pmatrix}
     T_{11} \\
     T_{21} \\
     T_{31} \\
     T_{41} \\
     T_{51}
  \end{pmatrix},
\end{align*}
i.e.
\begin{align*}
  \La_{:, 1} &= \L \T_{:, 1}.
\end{align*}
Continuing for the second column,
\begin{align*}
  \begin{pmatrix}
     0      \\
     \Lambda_{22} \\
     \Lambda_{32} \\
     \Lambda_{42} \\
     \Lambda_{52}
  \end{pmatrix}
  &=
  \begin{pmatrix}
     L_{11} & 0      & 0      & 0      & 0\\
     L_{21} & L_{22} & 0      & 0      & 0\\
     L_{31} & L_{32} & L_{33} & 0      & 0\\
     L_{41} & L_{42} & L_{43} & L_{44} & 0\\
     L_{51} & L_{52} & L_{53} & L_{54} & L_{55}
  \end{pmatrix}
  \begin{pmatrix}
     0      \\
     T_{22} \\
     T_{32} \\
     T_{42} \\
     T_{52}
  \end{pmatrix}.
\end{align*}
We can eliminate the first row and column of $\L$, leading to
\begin{align*}
  \begin{pmatrix}
     \Lambda_{22} \\
     \Lambda_{32} \\
     \Lambda_{42} \\
     \Lambda_{52}
  \end{pmatrix}
  &=
  \begin{pmatrix}
     L_{22} & 0      & 0      & 0\\
     L_{32} & L_{33} & 0      & 0\\
     L_{42} & L_{43} & L_{44} & 0\\
     L_{52} & L_{53} & L_{54} & L_{55}
  \end{pmatrix}
  \begin{pmatrix}
     T_{22} \\
     T_{32} \\
     T_{42} \\
     T_{52}
  \end{pmatrix},
\end{align*}
i.e.
\begin{align*}
  \La_{2:, 2} &= \L_{2:, 2:} \T_{2:, 2}.
\end{align*}
Similarly, for the third column,
\begin{align*}
  \begin{pmatrix}
     0      \\
     0      \\
     \Lambda_{33} \\
     \Lambda_{43} \\
     \Lambda_{53}
  \end{pmatrix}
  &=
  \begin{pmatrix}
     L_{11} & 0      & 0      & 0      & 0\\
     L_{21} & L_{22} & 0      & 0      & 0\\
     L_{31} & L_{32} & L_{33} & 0      & 0\\
     L_{41} & L_{42} & L_{43} & L_{44} & 0\\
     L_{51} & L_{52} & L_{53} & L_{54} & L_{55}
  \end{pmatrix}
  \begin{pmatrix}
     0      \\
     0      \\
     T_{33} \\
     T_{43} \\
     T_{53}
  \end{pmatrix},
\end{align*}
so we can eliminate the first two rows and columns of $\L$:
\begin{align*}
  \begin{pmatrix}
     \Lambda_{33} \\
     \Lambda_{43} \\
     \Lambda_{53}
  \end{pmatrix}
  &=
  \begin{pmatrix}
     L_{33} & 0      & 0\\
     L_{43} & L_{44} & 0\\
     L_{53} & L_{54} & L_{55}
  \end{pmatrix}
  \begin{pmatrix}
     T_{33} \\
     T_{43} \\
     T_{53}
  \end{pmatrix},
\end{align*}
i.e.,\ 
\begin{align*}
  \La_{3:, 3} &= \L_{3:, 3:} \T_{3:, 3}.
\end{align*}

Generalizing this result, the full computation $\La = \L \T$ can be written as a matrix-vector multiplication by rearranging the columns of $\La$ and $\T$ into a single vector:
\begin{align*}
  \begin{pmatrix}
    \La_{1:, 1}\\
    \La_{2:, 2}\\
    \vdots\\
    \La_{\nu:, \nu}
  \end{pmatrix} &=
  \begin{pmatrix}
    \L_{1:, 1:} & \0           & \dotsm & \0\\
    \0          & \L_{2:, 2:} & \dotsm & \0\\
    \vdots      & \vdots      & \ddots & \vdots\\
    \0          & \0          & \dotsm & \L_{\nu:, \nu:}
  \end{pmatrix}
  \begin{pmatrix}
    \T_{1:, 1}\\
    \T_{2:, 2}\\
    \vdots\\
    \T_{\nu:, \nu}
  \end{pmatrix}.
\end{align*}
The Jacobian is given by the determinant of the large square matrix. 
As this matrix is lower-triangular, the determinant can be written in terms of the diagonal elements of $\L$,
\begin{align}
\label{eq:jac:J_LA}
  \b{\prod_{i=1}^N \prod_{k=1}^{\min(i, \nu)} d\Lambda_{ik}} &= \b{\prod_{i=1}^N L_{ii}^{\min(i, \nu)}} \b{\prod_{i=1}^N \prod_{k=1}^{\min(i, \nu)} dT_{ik}}.
\end{align}

\subsubsection{From $\T$ to $\La = \T\B$}
\label{sec:jac:J_AL}
We can also consider the Jacobian from $\T$ to $\La = \T\B$, again where $\T$ is potentially low-rank with $\nu \leq N$ and $\B \in \reals^{\nu \times \nu}$ is lower triangular:
\begin{align}
  \label{eq:jac:AL}
  \begin{pmatrix}
    \Lambda_{11}    & \dotsm & 0          \\ 
    \vdots          & \ddots & \vdots     \\ 
    \Lambda_{\nu 1} & \dotsm & \Lambda_{\nu \nu} \\ 
    \vdots          & \vdots & \vdots     \\ 
    \Lambda_{N 1}   & \dotsm & \Lambda_{N \nu}
  \end{pmatrix} = 
    \begin{pmatrix}
    T_{11}    & \dotsm & 0          \\ 
    \vdots          & \ddots & \vdots     \\ 
    T_{\nu 1} & \dotsm & T_{\nu \nu} \\ 
    \vdots          & \vdots & \vdots     \\ 
    T_{N 1}   & \dotsm & T_{N \nu}
  \end{pmatrix}
  \begin{pmatrix}
  B_{11} & \dotsm & 0 \\
  \vdots & \ddots & \vdots \\
  B_{\nu 1} & \dotsm & B_{\nu\nu}
  \end{pmatrix}.
\end{align}
Instead of considering columns of $\La$ and $\T$, we consider rows.
For the first row, we have
\begin{align*}
    \begin{pmatrix}
      \Lambda_{11} 
    \end{pmatrix}
    = 
    \begin{pmatrix}
      T_{11}
    \end{pmatrix}
    \begin{pmatrix}
      B_{11}
    \end{pmatrix},
\end{align*}
or equivalently,
\begin{align*}
    \La_{1, :1} = \T_{1, :1}\B_{:1, :1}.
\end{align*}
Similarly, for rows up to the $\nu^\text{th}$ row, i.e., for $i \leq \nu$, we have
\begin{align*}
    \begin{pmatrix}
      \Lambda_{i1} & \dotsm & \Lambda_{ii}
    \end{pmatrix}
    =
    \begin{pmatrix}
      T_{i1} & \dotsm & T_{ii}
    \end{pmatrix}
    \begin{pmatrix}
      B_{11} & \dotsm & 0 \\
      \vdots & \ddots & \vdots \\
      B_{i1} & \dotsm & B_{ii}
    \end{pmatrix},
\end{align*}
which can be written as
\begin{align*}
    \La_{i, :i} = \T_{i, :i}\B_{:i, :i}.
\end{align*}
For rows beyond the $\nu^\text{th}$ row, i.e., $i > \nu$, the expression becomes
\begin{align*}
    \begin{pmatrix}
      \Lambda_{i1} & \dotsm & \Lambda_{i\nu}
    \end{pmatrix}
    =
    \begin{pmatrix}
      T_{i1} & \dotsm & T_{i\nu}
    \end{pmatrix}
    \begin{pmatrix}
      B_{11} & \dotsm & 0 \\
      \vdots & \ddots & \vdots \\
      B_{\nu1} & \dotsm & B_{\nu \nu}
    \end{pmatrix},
\end{align*}
which again can be written as
\begin{align*}
    \La_{i, :\nu} = \T_{i, :\nu} \B_{:\nu, :\nu} = \T_{i, :} \B.
\end{align*}
To calculate the Jacobian, we proceed by taking the transpose of each of the rows and stacking them, giving
\begin{align*}
    \begin{pmatrix}
      \La_{1, :1}^\top \\
      \La_{2, :2}^\top \\
      \vdots \\
      \La_{\nu, :\nu}^\top \\
      \La_{\nu + 1, :\nu}^\top \\
      \vdots \\
      \La_{N, :\nu}^\top
    \end{pmatrix}
    =
    \begin{pmatrix}
      \B_{:1, :1}^\top & \0 & \dotsm & \0 & \0 & \dotsm & \0 \\
      \0 & \B_{:2, :2}^\top & \dotsm & \0 & \0 & \dotsm & \0 \\
      \vdots & \vdots & \ddots & \vdots & \vdots & \ddots & \vdots \\
      \0 & \0 & \dotsm & \B^\top & \0 & \dotsm & \0 \\
      \0 & \0 & \dotsm & \0 & \B^\top & \dotsm & \0 \\
      \vdots & \vdots & \ddots & \vdots & \vdots & \ddots & \vdots \\
      \0 & \0 & \dotsm & \0 & \vdots & \dotsm & \B^\top \\
    \end{pmatrix}
    \begin{pmatrix}
      \T_{1, :1}^\top \\
      \T_{2, :2}^\top \\
      \vdots \\
      \T_{\nu, :\nu}^\top \\
      \T_{\nu + 1, :\nu}^\top \\
      \vdots \\
      \T_{N, :\nu}^\top
    \end{pmatrix}.
\end{align*}
As the square matrix is upper triangular, we can simply take the diagonal elements to find the Jacobian, which gives us
\begin{align}
\label{eq:jac:TB}
  \b{\prod_{i=1}^N \prod_{k=1}^{\min(i, \nu)} d\Lambda_{ik}} &= \b{\prod_{i=1}^{\nt} B_{ii}^{N - i + 1}} \b{\prod_{i=1}^N \prod_{k=1}^{\min(i, \nu)} dT_{ik}}.
\end{align}

\subsection{Jacobian from $\C = \La\La^\top$ to $\A\C\transpose{\A}$, where $\A$ is a general (invertible) matrix}
We now consider the Jacobian from $\C = \La\La^\top$, where $\La \in \reals^{N\times \nu}$ has rank $\nu$ as in Eq.~\ref{eq:jac:La}, to $\D = h(\C) = \A\C\A^\top$, where $A$ is any (invertible) $N\times N$ matrix.
This Jacobian is difficult to derive from scratch; however, we can obtain it straightforwardly by using the density of the singular Wishart.
In particular, the probability density function of $\D \sim \mathcal{W}_{1}(\S, \nu)$ is given by
\begin{align*}
    \mathrm{p}_{1}(\D) = \frac{\pi^{\nu(\nt - N)/2}}{2^{\nu N/2}|\S|^{\nu/2}\Gamma_{\nt}\b{\tfrac{\nu}{2}}} |\D_{:\nt, :\nt}|^{(\nu - N - 1)/2}\etr \b{-\S^{-1}\D /2},
\end{align*}
where $\nt = \min \b{\nu, N}$ as before. 
Note that $\D_{:\nt, :\nt}$ is almost surely full rank.
For $\C \sim \mathcal{W}_{2}(\I_N, \nu)$, this simplifies to
\begin{align*}
    \mathrm{p}_{2}(\C) = \frac{\pi^{\nu(\nt-N)/2}}{2^{\nu N/2}\Gamma_{\nt}\b{\tfrac{\nu}{2}}}|\C_{:\nt,:\nt}|^{(\nu-N-1)/2}\etr\b {-\C/2}.
\end{align*}
Using these densities, we can use the identity
\begin{align*}
    \mathrm{p}_1(\D) = \mathrm{p}_2\b{h^{-1}(\D)} \left|\frac{\partial \C}{\partial \D}\right|
\end{align*}
to obtain the desired Jacobian:
\begin{align*}
    J = \left|\frac{\partial \D}{\partial \C}\right| = \mathrm{p}_2\b{h^{-1}(\D)}/\mathrm{p}_1(\D).
\end{align*}
By defining $\S = \A\transpose{\A}$, we obtain
\begin{align}
    \label{eq:jac:J_AXAT}
    J = |\S|^{\nu/2} \frac{|\C_{:\nt,:\nt}|^{(\nu-N-1)/2}}{|\D_{:\nt,:\nt}|^{(\nu - N - 1)/2}}.
\end{align}

\subsubsection{A sanity check}
We now check that this result is consistent with those we obtained for lower-triangular matrices in the previous sections.
In particular, we consider the transformation from $\C = \T\transpose{\T}$, where $\T \in \reals^{N \times \nu}$ is lower triangular with $\nu < N$, to $\D = \L\C\transpose{\L} = \L\T\transpose{(\L\T)}$, with $\L \in \reals^{N \times N}$ being lower triangular and full rank.
Considering first the transformation from $\T$ to $\La = \L\T$, Eq.~\ref{eq:jac:J_LA} gives us
\begin{align*}
    \left|\frac{\partial \La}{\partial \T}\right| = \prod_{i=1}^N L_{ii}^{\min(i, \nu)}.
\end{align*}
Applying Eq.~\ref{eq:jac:J_LaLaTsing} subsequently gives us
\begin{align*}
    \left|\frac{\partial \D}{\partial \La}\right| = \prod_{i=1}^{\nt} 2\Lambda_{ii}^{N - i + 1} = \prod_{i=1}^{\nt} 2(L_{ii}T_{ii})^{N - i + 1}.
\end{align*}
Combining the two (using the chain rule) gives
\begin{align*}
    \left|\frac{\partial \D}{\partial \T}\right| = \left|\frac{\partial \D}{\partial \La}\right|\left|\frac{\partial \La}{\partial \T}\right| = \left(\prod_{i=1}^{\nt} 2(L_{ii}T_{ii})^{N - i + 1}\right)\left(\prod_{i=1}^N L_{ii}^{\min(i, \nu)}\right).
\end{align*}
As we would like the Jacobian from $\C = \T\transpose{\T}$ to $\D$, we must divide out the Jacobian from $\T$ to $\C$, which from Eq.~\ref{eq:jac:J_LaLaTsing}) is
\begin{align*}
    \left|\frac{\partial \C}{\partial \T}\right| = \prod_{i=1}^{\nt} 2T_{ii}^{N - i + 1},
\end{align*}
leading to our final result:
\begin{align}\label{eq:jac:J_LXLT}
    \left|\frac{\partial \D}{\partial \C}\right| = \left(\prod_{i=1}^{\nt} L_{ii}^{N - i + 1}\right) \left(\prod_{i=1}^N L_{ii}^{\min(i, \nu)}\right).
\end{align}
We now compare this to the result of applying Eq.~\ref{eq:jac:J_AXAT}:
\begin{align*}
    J &= |\L\transpose{\L}|^{\nu/2} \frac{|\C_{:\nu, :\nu}|^{(\nu - N -1)/2}}{|\{\L\C\transpose{\L}\}_{:\nu, :\nu}|^{(\nu - N -1)/2}} \\
    &= \frac{|\L|^\nu}{|\L_{:\nu, :\nu}|^{\nu-N-1}}.
\end{align*}
Some algebraic manipulation confirms that the two results are equivalent.

\section{Deriving densities using the above Jacobians}
We now show how the above Jacobians can be used to derive the Wishart densities used in the main text.
We start by demonstrating that our singular Bartlett decomposition from Sec.~\ref{sec:dwp:gen_wishart} indeed leads to the correct probability density function for the standard singular Wishart.

\subsection{Deriving the (singular) Wishart density from the (singular) Bartlett decomposition}
\label{sec:jac:singular_proof}
We wish to derive the density of $\H \sim \Wish{\I_N, \nu}$ from its Bartlett decomposition $\H = \T\transpose{\T}$, where in the singular case
\begin{align*}
    \A = \begin{pmatrix}
      T_{11} & \dotsm & 0 \\
      \vdots & \ddots & \vdots \\
      T_{\nu 1} & \dotsm & T_{\nu\nu} \\
      \vdots & \ddots & \vdots \\
      T_{N 1} & \dotsm & T_{N\nu}
    \end{pmatrix},
\end{align*}
and where 
\begin{align*}
    \p{T_{jj}^2} &= \text{Gamma}\b{T_{jj}^2;\, \alpha=\tfrac{\nu - j + 1}{2}, \beta=\tfrac{1}{2}}, \\
    \p{T_{j>k}} &= \Nc{T_{jk}}{ 0, 1}.
\end{align*}
We first need to change variables to $T_{jj}$ rather than $T_{jj}^2$:
\begin{align*}
  \p{T_{jj}} &= \p{T_{jj}^2} \abs{\dd[T_{jj}^2]{T_{jj}}},\\
  &= \text{Gamma}\b{T_{jj}^2;\, \tfrac{\nu-j+1}{2}, \tfrac{1}{2}} 2 T_{jj},\\
  &= \frac{\b{T_{jj}^2}^{\b{\nu-j+1}/2 - 1} e^{-T_{jj}^2/2}}{2^{\b{\nu-j+1}/2} \Gamma\b{{\tfrac{\nu-j+1}{2}}}} 2 T_{jj},\\
  &= \frac{T_{jj}^{\nu-j} e^{-T_{jj}^2/2}}{2^{\b{\nu-j-1}/2} \Gamma\b{{\tfrac{\nu-j+1}{2}}}}.
\end{align*}
Thus, the probability density for $\T$ under the Bartlett sampling operation is
\begin{align}
  \label{eq:jac:bartlett}
  \p{\T} &= \underbrace{\prod_{j=1}^{\nt} \frac{T_{jj}^{\nu-j} e^{-T^2_{jj}/2}} {2^{\tfrac{\nu-j-1}{2}} \Gamma\b{\tfrac{\nu-j+1}{2}}}}_\text{on-diagonals}
  \underbrace{\prod_{i=j+1}^N \frac{1}{\sqrt{2 \pi}} e^{-T_{ij}^2/2}}_\text{off-diagonals},
\end{align}
where $\nt = \min(\nu, N)$, so that we handle the non-singular case as well. 
To convert this to a density for $\H$, we need the volume element for the transformation from $\T$ to $\H = \T \transpose{\T}$, which is given by Eq.~\ref{eq:jac:J_LaLaTsing}:
\begin{align*} 
  d\H &= d\T \prod_{j=1}^{\nt} 2 T_{jj}^{N-j+1}.
\end{align*}
Thus
\begin{align*}
  \p{\H} &= \p{\T} \b{\prod_{j=1}^{\nt} \frac{1}{2} T_{jj}^{-(N-j+1)}}\\
  &= \prod_{j=1}^{\nt} \frac{T_{jj}^{\nu-N-1} e^{-T^2_{jj}/2}} {2^{(\nu-j+1)/2} \Gamma\b{\tfrac{\nu-j+1}{2}}}
  \prod_{i=j+1}^N \frac{1}{\sqrt{2 \pi}} e^{-T_{ij}^2/2}.
\end{align*}
To obtain the standard form of the p.d.f. for the Wishart, we break this expression down into separate components.
First, we manipulate a product over the diagonal elements of $\T$ to obtain the determinant of $\H$:
\begin{align*} 
  \prod_{j=1}^{\nt} T_{jj}^{\nu-N-1} &= \b{\prod_{j=1}^{\nt} T_{jj}}^{\nu-N-1} = \abs{\T_{:\nt, :} \transpose{\T}_{:\nt, :}}^{\b{\nu-N-1}/2} = \abs{\H_{:\nt, :\nt}}^{\b{\nu-N-1}/2}.
\end{align*}
Next, we manipulate the exponential terms to form an exponentiated trace.  
By combining on- and off-diagonal terms, and noting that $T_{ij} = 0$ for $i<j$, we can extend the product
\begin{align*}
  \prod_{j=1}^\nt e^{-T^2_{jj}/2} \prod_{i=j+1}^N e^{-T_{ij}^2/2}
  &= \prod_{j=1}^\nt \prod_{i=j}^N e^{-T_{ij}^2/2}
  = \prod_{j=1}^\nt \prod_{i=1}^N e^{-T_{ij}^2/2}.\\
  \intertext{Then, taking the product inside the exponential and recalling that $\H = \T \transpose{\T}$, we can write the product in terms of the trace of $\H$:}
  \prod_{j=1}^\nt e^{-T^2_{jj}/2} \prod_{i=j+1}^N e^{-T_{ij}^2/2}
  &= e^{\sum_{j=1}^\nt \sum_{i=1}^N -T_{ij}^2/2}
  = \etr\b{-\H/2}.
\end{align*}
Next, we consider the powers of $2$. We begin by computing the number of $1/\sqrt{2}$ terms, arising from the off-diagonal elements:
\begin{align*}
  \prod_{j=1}^{\nt} \prod_{i=j+1}^N \frac{1}{\sqrt{2}} &=
  \b{\frac{1}{\sqrt{2}}}^{\nu(N-\nt) + \nt (\nt-1)/2}.
\end{align*}
Note that the $\nt (\nt-1)/2$ term corresponds to the off-diagonal terms in the square block $\T_{:\nt, :}$, and the $\nu(N-\nt)$ term corresponds to the terms from $\T_{\nt:, :}$.
Next we consider the on-diagonal terms:
\begin{align*}
  \prod_{j=1}^{\nt} \frac{1}{2^{\b{\nu-j+1}/2}}= 
  \b{\frac{1}{\sqrt{2}}}^{\nt(\nu+1)} \prod_{j=1}^{\nt} \b{\frac{1}{\sqrt{2}}}^{-j} = 
  \b{\frac{1}{\sqrt{2}}}^{\nt(\nu+1) - \nt (\nt+1)/2}.
\end{align*}
Combining the on and off-diagonal terms gives
\begin{align*}
  \prod_{j=1}^{\nt} \frac{1}{2^{\b{\nu-j+1}/2}}
  \prod_{i=j+1}^N \frac{1}{\sqrt{2}}
  &= \b{\frac{1}{\sqrt{2}}}^{\nu(N-\nt) + \nt (\nt-1)/2 + \nt(\nu+1) - \nt (\nt+1)/2}\\
  &= \b{\frac{1}{\sqrt{2}}}^{(\nu N- \nu \nt) + (\nt^2/2 - \nt/2) + (\nt \nu + \nt) + (-\nt^2/2 - \nt/2)}\\
  &= \b{\frac{1}{\sqrt{2}}}^{\nu N}.
\end{align*}
Finally, using the definition of the multivariate Gamma function,
\begin{align*}
  \prod_{j=1}^{\nt} \Gamma\b{\tfrac{\nu-j+1}{2}} \prod_{i=j+1}^N \sqrt{\pi} &= \pi^{\nu (N-\nt)/2} \underbrace{\pi^{\nt(\nt-1)/4} \prod_{j=1}^{\nt} \Gamma\b{\tfrac{\nu-j+1}{2}}}_{=\Gamma_\nt\b{\tfrac{\nu}{2}}}\\
  &= \pi^{\nu (N-\nt)/2} \Gamma_\nt\b{\tfrac{\nu}{2}}.
\end{align*}
We thereby obtain the probability density for the standard singular Wishart distribution,
\begin{align*}
  \p{\H} &= \frac{\pi^{\nu (\nt-N)/2}}{2^{\nu N/2} \Gamma_\nt\b{\tfrac{\nu}{2}}} \abs{\H_{:\nt, :\nt}}^{\b{\nu-N-1}/2} \etr \b{-\H/2}.
\end{align*}
For $\nt=\nu$, this matches Eq.~3.2 in \citet{srivastava2003singular}, and for $\nt=N$ it matches the standard full-rank Wishart probability density function.

\subsection{The generalized singular Wishart}
We now use the above Jacobians to derive the densities of the singular Wishart distributions from the main text, starting with the generalized singular Wishart from Sec.~\ref{sec:dwp:gen_wishart}.
Recall that we have defined the generalized singular Wishart with scale matrix $\S = \L\transpose{\L}$, degrees of freedom $\nu$, and Bartlett-generalizing parameters $\boldsymbol{\alpha},\,\boldsymbol{\beta},\,\boldsymbol{\mu},\,\boldsymbol{\sigma}$ as a distribution over $\G \in \reals^{N \times N}$, where $\G = \L \T \transpose{\T} \transpose{\L}$, with 
\begin{align*}
    \T &= \begin{pmatrix}
      T_{11} & \dotsm & 0 \\
      \vdots & \ddots & \vdots \\
      T_{\nu 1} & \dotsm & T_{\nu \nu} \\
      \vdots & \ddots & \vdots \\
      T_{N 1} & \dotsm & T_{NN}
    \end{pmatrix},\\
    \q{T_{jj}^2} &= \text{Gamma}\b{T_{jj}^2; \, \alpha_j, \beta_j} && \text{for } j \leq \nu, \\
    \q{T_{i > j}} &= \Nc{T_{ij}}{\mu_{ij}, \sigma_{ij}^2} && \text{for } j \leq \nu.
\end{align*}
We now show how to obtain the density of $\G$ using the above Jacobians.
Note that we leave open the possibility that $\nu > N$, and so we again make use of $\nt = \min \b{\nu, N}$.
First, we obtain the density of the generalized Bartlett factor by change-of-variables from $T_{jj}^2$ to $T_{jj}$:
\begin{align*}
    \q{\T} &= \prod_{j}^\nt\text{Gamma}\b{T_{jj}^2;\, \alpha_j, \beta_j}\abs{\dd[T_{jj}^2]{T_{jj}}} \prod_{i=j+1}^N \Nc{T_{ij}}{\mu_{ij}, \sigma_{ij}^2}\\
    &= 2^\nt \prod_{j}^\nt T_{jj}\text{Gamma}\b{T_{jj}^2;\, \alpha_j, \beta_j} \prod_{i=j+1}^N \Nc{T_{ij}}{\mu_{ij}, \sigma_{ij}^2},
\end{align*}
since $\abs{\partial T_{jj}^2/\partial T_{jj}} = 2T_{jj}$. 
We can now obtain the density function of $\G$ by using the chain rule:
\begin{align*}
    \q{\G} = \q{\T} \abs{\dd[\T]{\G}} = \q{\T} \underbrace{\abs{\dd[\T]{\La}}}_{(a)}\underbrace{\abs{\dd[\La]{\G}}}_{(b)},
\end{align*}
where we have defined $\La = \L \T$.
The first of these terms, (a), can be obtained from Eq.~\ref{eq:jac:J_LA} as
\begin{align*}
    \abs{\dd[\T]{\La}} = \prod_{i=1}^N \frac{1}{L_{ii}^{\min \b{i, \nu}}}
\end{align*},
whereas (b) can be obtained by applying Eq.~\ref{eq:jac:J_LaLaTsing}:
\begin{align*}
    \abs{\dd[\La]{\G}} = \prod_{i=1}^\nt \frac{1}{2 \Lambda_{ii}^{N - i + 1}} = \frac{1}{2^\nt}\prod_{i=1}^\nt \frac{1}{\b{L_{ii}T_{ii}}^{N - i + 1}}.
\end{align*}
Combining these terms and performing a few cancellations, we arrive at the required density,
\begin{align*}
    \q{\G} = \b{\prod_{j=1}^N \frac{1}{L_{jj}^{\min \b{j, \nu}}}}\prod_{j=1}^\nt \frac{\textup{Gamma}\b{T_{jj}^2;\,\alpha_j, \beta_j}}{T_{jj}^{N-j}L_{jj}^{N-j+1}}\prod_{i=j+1}^N \Nc{T_{ij}}{\mu_{ij}, \sigma_{ij}^2}.
\end{align*}

\subsection{The A-generalized singular Wishart}
The A-generalized singular Wishart is identical to the above, with the addition of a general invertible matrix $\A \in \reals^{N \times N}$ so that $\G = \L \A \T \transpose{\b{\L \A \T}}$.
However, our path to the density of the A-generalized singular Wishart is quite different.
We define three terms to find the needed density, starting with $\C = \T \transpose{\T}$.
We then consider $\D = \A \C \transpose{\A}$, followed by our result $\G = \L \D \transpose{\L}$, so that
\begin{align*}
    \q{\G} = \q{\T}\abs{\dd[\T]{\C}}\abs{\dd[\C]{\D}}\abs{\dd[\D]{\G}},
\end{align*}
where $\q{\T}$ is identical to above.
We find $\abs{\partial \T/\partial \C}$ by applying Eq.~\ref{eq:jac:J_LaLaTsing}, which gives us
\begin{align*}
    \abs{\dd[\T]{\C}} = \frac{1}{2^\nt}\prod_{i=1}^\nt\frac{1}{T_{ii}^{N - i + 1}}.
\end{align*}
The second of these Jacobians is given by Eq.~\ref{eq:jac:J_AXAT}:
\begin{align*}
    \abs{\dd[\C]{\D}} &= \frac{\abs{\D_{:\nt, :\nt}}^{(\nu - N - 1)/2}}{|\A|^\nu\abs{\C_{:\nt, :\nt}}^{(\nu - N - 1)/2}}.
\end{align*}
While we might hope to be able to simplify this expression, it is unfortunately not the case in general, as $\D_{:\nt, :\nt} = \cb{\A\C\transpose{\A}}_{:\nt, :\nt}$ does not factorize straightforwardly when $\A$ has no special structure.
Computing the density for the A-generalized Wishart therefore necessitates storing these determinants when $\C$ and $\D$ are computed.
We now turn to the last Jacobian, $\abs{\partial \D/\partial \G}$.
This term can be found by applying Eq.~\ref{eq:jac:J_LXLT},
\begin{align*}
    \abs{\dd[\D]{\G}} = \b{\prod_{i=1}^\nt \frac{1}{L_{ii}^{N - i + 1}}}\b{\prod_{i=1}^N \frac{1}{L_{ii}^{\min \b{i, \nu}}}}.
\end{align*}
Combining these terms, we arrive at the final density,
\begin{multline*}
    \q{\G} = \b{\prod_{j=1}^N \frac{1}{L_{jj}^{\min \b{j, \nu}}}}\frac{\abs{\D_{:\nt, :\nt}}^{(\nu - N - 1)/2}}{|\A|^\nu\abs{\C_{:\nt, :\nt}}^{(\nu - N - 1)/2}} \times \\
    \prod_{j=1}^\nt \frac{\textup{Gamma}\b{T_{jj}^2;\,\alpha_j, \beta_j}}{T_{jj}^{N-j}L_{jj}^{N-j+1}}\prod_{i=j+1}^N \Nc{T_{ij}}{\mu_{ij}, \sigma_{ij}^2}.
\end{multline*}

\subsection{The AB-generalized singular Wishart}
The derivation for the AB-generalized singular Wishart is similar to that of the A-generalized Wishart, with the addition of one extra step.
Namely, as the AB-generalized Wishart defines $\G = \L\A\T\B\transpose{\b{\L\A\T\B}}$, we define $\La = \T\B$ and $\C = \La\transpose{\La}$, so that 
\begin{align*}
    \q{\G} = \q{\T}\abs{\dd[\T]{\La}}\abs{\dd[\La]{\C}}\abs{\dd[\C]{\D}}\abs{\dd[\D]{\G}}.
\end{align*}
The first term is found directly from Eq.~\ref{eq:jac:TB}, 
\begin{align*}
    \abs{\dd[\T]{\La}} = \prod_{i=1}^\nt \frac{1}{B_{ii}^{N - i + 1}},
\end{align*}
whereas 
\begin{align*}
    \abs{\dd[\La]{\C}} &= \frac{1}{2^\nt}\prod_{i=1}^{\nt} \frac{1}{\Lambda_{ii}^{N - i + 1}}\\
    &= \frac{1}{2^\nt}\prod_{i=1}^{\nt} \frac{1}{T_{ii}^{N - i + 1}B_{ii}^{N - i + 1}}.
\end{align*}
arises from Eq.~\ref{eq:jac:J_LaLaTsing}.
The remaining Jacobians remain unchanged in form, so that our final density is given by
\begin{multline*}
    \q{\G} = \b{\prod_{j=1}^N \frac{1}{L_{jj}^{\min \b{j, \nu}}}}\frac{\abs{\D_{:\nt, :\nt}}^{(\nu - N - 1)/2}}{|\A|^\nu\abs{\C_{:\nt, :\nt}}^{(\nu - N - 1)/2}} \times \\
    \prod_{j=1}^\nt \frac{\textup{Gamma}\b{T_{jj}^2;\,\alpha_j, \beta_j}}{T_{jj}^{N-j}L_{jj}^{N-j+1}B_{jj}^{2(N-j+1)}}\prod_{i=j+1}^N \Nc{T_{ij}}{\mu_{ij}, \sigma_{ij}^2}.
\end{multline*}

\chapter{Appendix to Chapter 3}



\section{Experimental details}\label{app:dkl:exp-deets}
All experiments on real datasets (MNIST, UCI, CIFAR-10, UTKFace) were written in TensorFlow 2 \citep{tensorflow2015-whitepaper}, using GPflow \citep{GPflow2017} to implement the DKL models.
We use jug \citep{coelho2017jug} to easily run the experiments.
The experiments were run on single GPUs using both NVIDIA Tesla P100-PCIE-16GB GPUs and NVIDIA GeForce RTX 2080 Ti GPUs.

\subsection{Datasets}\label{app:exp-datasets}
We describe the datasets used as well as the splits and preprocessing.
\paragraph{Toy dataset} The toy dataset is that as introduced in \citet{snelson2006sparse}. The dataset comprises 200 input-output pairs and can be found at \url{http://www.gatsby.ucl.ac.uk/~snelson/}. We normalize both inputs and outputs for training and plot the unnormalized values and predictions.

\paragraph{MNIST} We take the first 5,000 datapoints from the standard MNIST dataset \citep{lecun2010mnist}, and use the standard 10,000 point test set for evaluation.
We preprocess the images by dividing the pixel values by 255.

\paragraph{UCI} We use a slightly modified version of Bayesian Benchmarks (\url{https://github.com/hughsalimbeni/bayesian_benchmarks}) to obtain the UCI datasets we use. 
The modification is to rectify minor data leakage in the normalization code: they normalize using the statistics from the entire dataset before dividing into train/test splits, instead of normalizing using only the train split statistics. 
We perform cross-validation using 20 90\%/10\% train/test splits, and report means and standard errors for each metric.
Note that we report metrics on the normalized datasets to lead to more interpretable results: namely, an RMSE of 1 corresponds to predicting 0 for each test point.

\paragraph{CIFAR-10} We use the standard CIFAR-10 dataset \citep{krizhevsky2009learning}, with the standard train-validation split of 50,000 and 10,000 images, respectively, using the validation split as the test set, as is common practice. 
We preprocess the images by simply dividing the pixel values by 255, so that each value lies between 0 and 1. 

\paragraph{UTKFace} The UTKFace dataset \citep{zhifei2017cvpr} is a large face dataset consisting of 23,708 images of faces, annotated with age, gender, and ethnicity. 
The faces have an age range from 0 to 116.
We use the aligned and cropped version to limit the amount of preprocessing necessary, available at \url{https://susanqq.github.io/UTKFace/}.
These cropped images have sizes $200\times 200\times 3$.
We choose 20,000 images to be in the train dataset, with the remaining being used for testing.
We again perform preprocessing by dividing the pixel values by 255, and we additionally normalize the age values.
The metrics we report all use the normalized values, as with the UCI datasets.

\subsection{Models}\label{app:exp-models}
We describe the models used for the experiments. 
To ensure that the comparisons between neural networks and DKL models are as fair as possible, we ensure that each model used in direct comparison has the same number of layers: for the DKL models, we remove the last fully-connected layer of the neural network and replace it with the ARD SE GP. 
All neural networks use ReLU activations.
For all SVDKL models, the inducing points live in the neural network feature space at the input to the GP.

\paragraph{Toy dataset} We use an architecture of [100, 50, 2] for the hidden-layer widths for the neural network.
For DKL, we use the pre-activation features of the final hidden layer for the input to the GP.

\paragraph{MNIST} We use a large fully-connected ReLU architecture of [1,000, 500, 500, 100, 100, 50, 50, 10].
For the DKL model, we use the same feature extractor with an ARD SE kernel, and 5,000 inducing points to minimize the bias from the variational approximation.
The inducing points are initialized using the RobustGP method from \citet{burt2020convergence}.
We use the softmax likelihood for all models.

\paragraph{UCI} For \textsc{Boston, Energy}, we use a ReLU architecture of [50, 50], and a ReLU architecture [1,000, 500, 50] for \textsc{Kin40K, Power}, and \textsc{Protein}.
We note that these architectures are smaller than the ones proposed by \citet{wilson2016deep}.
For the SVGP baseline, we use an ARD SE kernel, with 100 inducing points for the small datasets (\textsc{Boston}, \textsc{Energy}) and 1,000 inducing points for the larger ones.
For the DKL models, we use the post-activation features from the final hidden layer as inputs to the GPs, which use ARD SE kernels.
For SVDKL, we initialize the inducing points using the k-means algorithm on a subset of the training set.
We use 100 inducing points for the smaller datasets (\textsc{Boston, Energy}) and 1,000 on the larger ones.
The method for initializing the inducing points, and the number of inducing points, is the same for the SVGP baseline model, which uses a standard ARD SE kernel.

\paragraph{CIFAR-10} We use a modified ResNet18 \citep{he2016deep} architecture as the baseline neural network architecture; the main modification is that we have added another fully-connected layer at the output to ensure that the neural network and DKL models are comparable in depth.
Therefore, instead of the standard single fully-connected layer after a global average pooling layer, we have two fully-connected layers.
While we could take the output of the global average pooling layer, this is typically very high-dimensional and thus potentially unsuitable as an input to a GP.
For most experiments, we fix the width of the last hidden layer (the final feature width) to 10, although we do consider changing that in App.~\ref{app:dkl:add_exp}.
We additionally add batchnorm layers \citep{ioffe2015batch} before the ReLU activations in the residual blocks.
For SVDKL, we use 1,000 inducing points initialized with k-means on a subset of the training set.
As with UCI, the features at the input to the GP are post-activation features.
For all classification models, we use softmax activation to obtain probabilities for the cross-entropy loss, and for the SVDKL models we use 10 samples from the latent function posterior to compute the log likelihood term of the ELBO.

\paragraph{UTKFace} As with CIFAR-10, we again use a modified ResNet18 \citep{he2016deep} architecture with an additional fully-connected layer at the output.
For SVDKL, we again use 1,000 inducing points.

\subsection{Implementation details}\label{app:dkl:imp-details}
All models are optimized using Adam \citep{kingma2014adam}.
Throughout, we try to ensure that we train each model for comparable numbers of gradient steps and learning rates.

\paragraph{Toy dataset} We train both the NN and DKL models for 10,000 gradient steps using learning rates of 0.001. No weight decay was used. For the HMC experiments, we use a step size of 0.005, 20 leapfrog steps, and a prior variance of 1 on the network weights. We burn in for 10,000 samples, then use 1,000 iterations to sample, thinned by a factor of 10.

\paragraph{MNIST} We train all models using full batch training, i.e., a batch size of 5,000. 
For the pretraining of the feature extractor, we use 96,000 gradient steps (corresponding to 160 epochs of training full MNIST with batch size 100), with an initial learning rate of 1e-3 and no weight decay.
We incorporate learning rate steps at halfway and three quarters through the training, stepping down by a factor of 10 each time.
For the neural network after pretraining and for fDKL, we use the same procedure.
For the DKL model, we first train only the variational and ARD SE parameters for 9,600 gradient steps (corresponding to 16 epochs of training full MNIST with batch size 100), with no learning rate schedule.
We then train everything jointly for 86,400 gradient steps (corresponding to 144 epochs), starting again at a learning rate of 1e-3 and decreasing the learning rate by a factor of 10 at the halfway and three quarters mark.

\paragraph{UCI} For each model, we train with an initial learning rate of 0.001. For \textsc{Boston} and \textsc{Energy}, we use a batch size of 32 and train the minibatched algorithms for a total of 400 epochs, and use a learning rate scheduler that decreases the learning rate by a factor of 10 after 200 and 300 epochs. For \textsc{Kin40K}, \textsc{Power}, and \textsc{Protein} we use a batch size of 100, training for 160 epochs with the same learning rate schedule that triggers at 80 and 120 epochs. For the full-batch methods, we ensure that they are trained for the lesser of the same number of gradient steps or 8,000 gradient steps (due to limited computational budget), with the learning rate schedule set to trigger at the corresponding number of gradient steps as the batched methods. This ensures a fair comparison when claiming that the full-batch methods overfit in comparison to the stochastic versions. For the deep models, we use a weight decay of 1e-4 on the neural network weights. We do not use any pretraining for the DKL models, as we did not find it necessary for these datasets. We initialize the log noise variance to -4 for the DKL models. We train the neural network models using mean squared error loss, and use the maximum likelihood noise estimate after training to compute train and test log likelihoods.

\paragraph{CIFAR-10} We describe the details for batch size 100; for batch size 500, we ensure that we use the same number of gradient steps. We do not use weight decay as we found that it hurt test accuracy. For the NN and SVDKL models, we train for 160 epochs total: we decrease the learning rate from the initial 1e-3 by a factor of 10 at 80 and then 120 epochs. For pNN, we train for an additional 160 epochs in the same way (restarting the learning rate at 1e-3). For pSVDKL, we start by training with the neural network parameters fixed for 80 epochs, with learning rate decreases at 40 and 60 epochs. We then reset the learning rate to 1e-3, and train for an additional 80 epochs with the same learning rate schedule. For the experiments with data augmentation, we use random horizontal flipping and randomly crop $32\times32\times3$ images from the original images zero-padded up to $40\times40\times3$.

We use the same losses as the potentials for SGLD. We use the trained NNs to initialize the weights to reasonable values, and set the batch size to 100. We initialize the learning rate to 1e-3 (which we then scale down by the dataset size to account for the scale of the potential), and decay the learning rate at each epoch by a factor of $1/(1 + 0.4\times \text{epoch})$ to satisfy Robbins-Monro. For the NN, we burn in for 100 epochs, and then sample every other epoch for 100 epochs, leading to 50 samples. For SVDKL, we follow the approach in \citet{hensman2015mcmc}, and learn the variational parameters (i.e. 1,000 inducing points) and GP hyperparameters with the fixed, pretrained NN weights, using the same hyperparameters as for pSVDKL. We then follow the SGLD approach we took for the NN, with 100 epochs of burn in and 100 epochs of sampling, starting with a learning rate of 1e-3.

\paragraph{UTKFace} We follow the same approach as for CIFAR-10. We list the minor differences. We use a small weight decay of 1e-4. For the SVDKL models, we initialize the log noise variance to -4. We use mean squared error loss for the NNs; however, for SGLD we use a Gaussian likelihood with log noise variance initialized to -4. For the SGLD experiments, we initialize the learning rate to 1e-5. For data augmentation, we again use random horizontal flipping as well as randomly cropping $200\times200\times3$ images from the original images zero-padded up to $240\times240\times3$.

\section{Additional experimental results}\label{app:dkl:add_exp}
Here we briefly present some additional experimental results.

\subsection{Toy}\label{app:add_toy}
\begin{figure*}[t]
     \centering
     \begin{subfigure}[b]{0.32\textwidth}
         \centering
         \includegraphics[width=\textwidth]{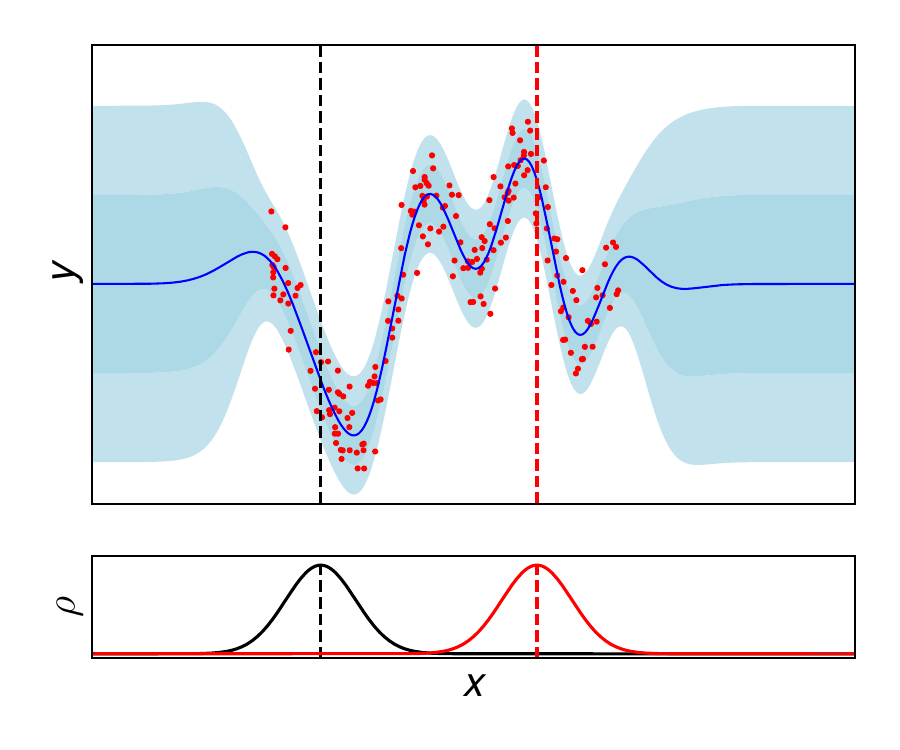}
         \caption{SE kernel. $\mathrm{LML}=-89.3$}
         \label{fig:dkl:app-RBF_200}
     \end{subfigure}
     \hfill
     \begin{subfigure}[b]{0.32\textwidth}
         \centering
         \includegraphics[width=\textwidth]{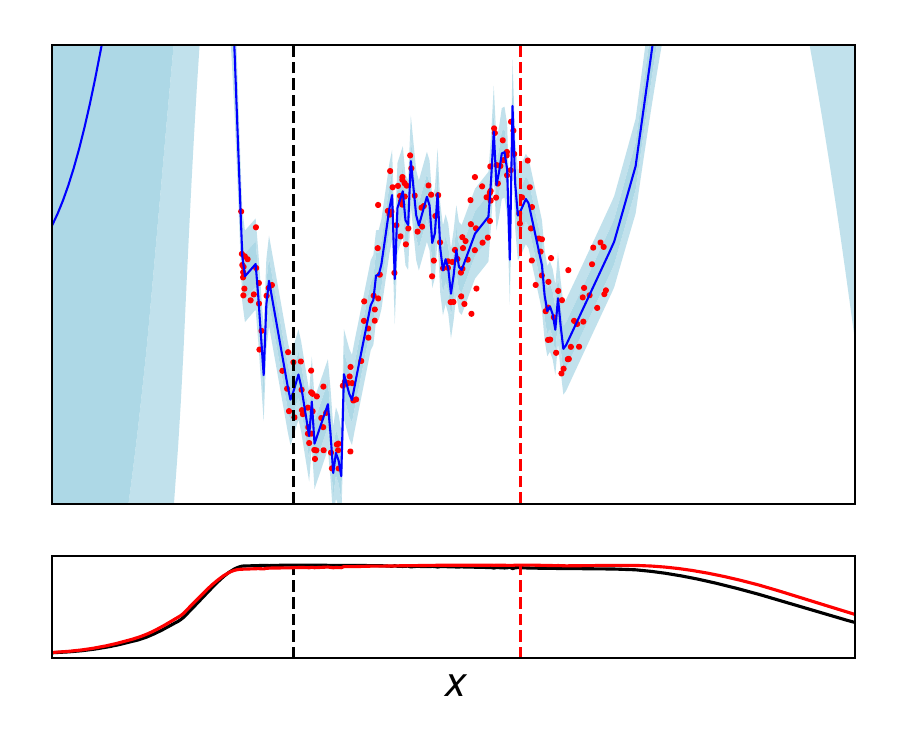}
         \caption{Example DKL. $\mathrm{LML} = -23.4$}
         \label{fig:dkl:app-DKL_200_0}
     \end{subfigure}
     \hfill
     \begin{subfigure}[b]{0.32\textwidth}
         \centering
         \includegraphics[width=\textwidth]{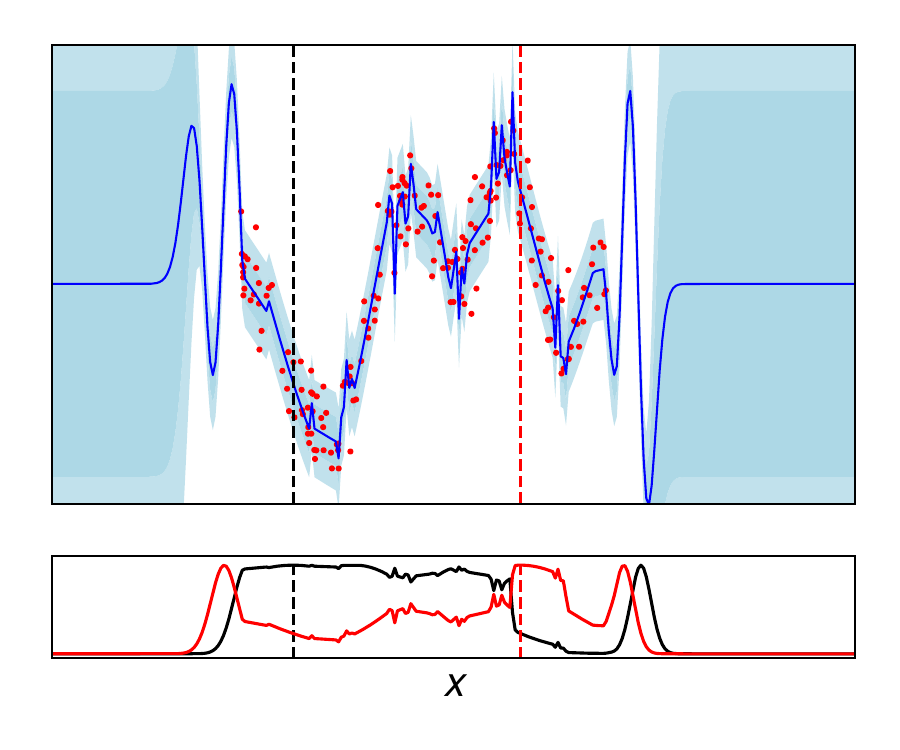}
         \caption{Example DKL. $\mathrm{LML} = -45.7$}
         \label{fig:dkl:app-DKL_200_1}
     \end{subfigure}

     \begin{subfigure}[b]{0.32\textwidth}
         \centering
         \includegraphics[width=\textwidth]{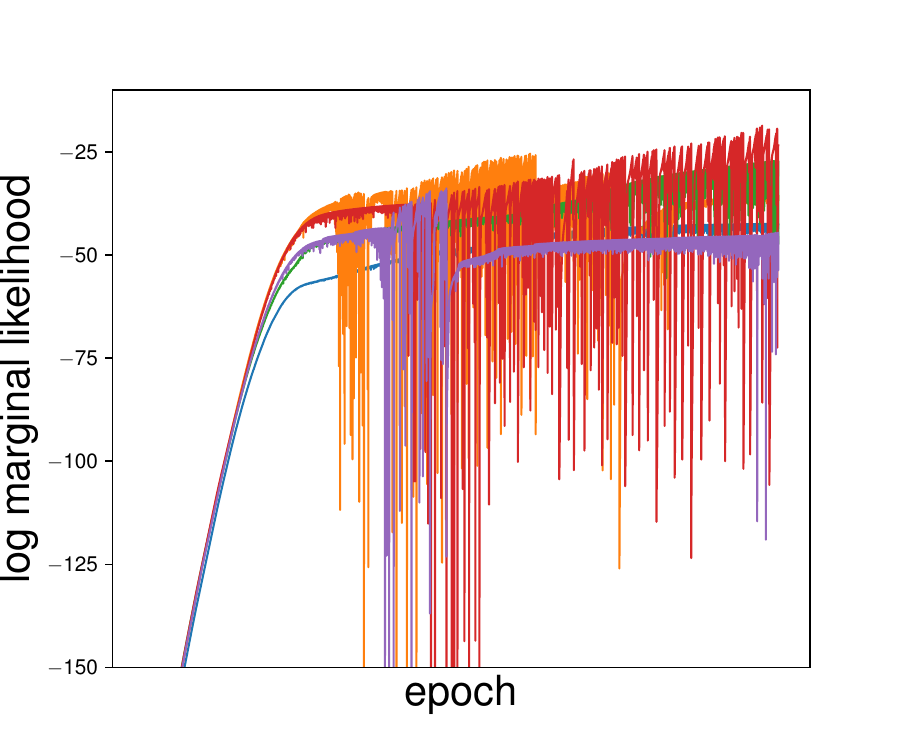}
         \caption{Training curves for 5 different initializations for DKL on the full toy dataset}
         \label{fig:dkl:app-training_curves}
     \end{subfigure}
     \hfill
     \begin{subfigure}[b]{0.32\textwidth}
         \centering
         \includegraphics[width=\textwidth]{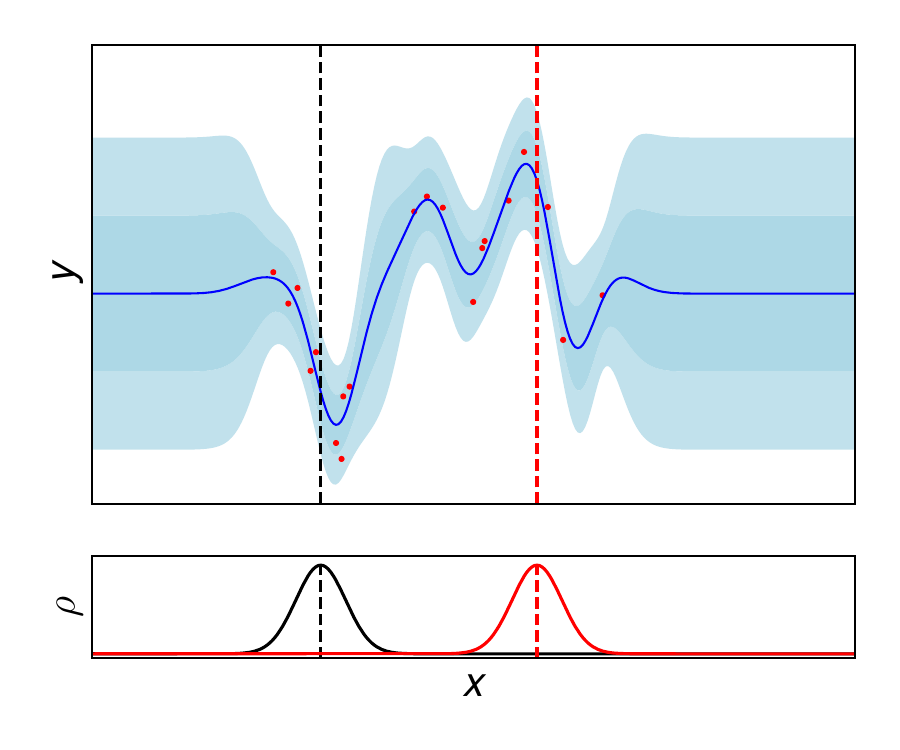}
         \caption{SE kernel. $\mathrm{LML} = -17.5$}
         \label{fig:dkl:app-RBF_20}
     \end{subfigure}
     \hfill
     \begin{subfigure}[b]{0.32\textwidth}
         \centering
         \includegraphics[width=\textwidth]{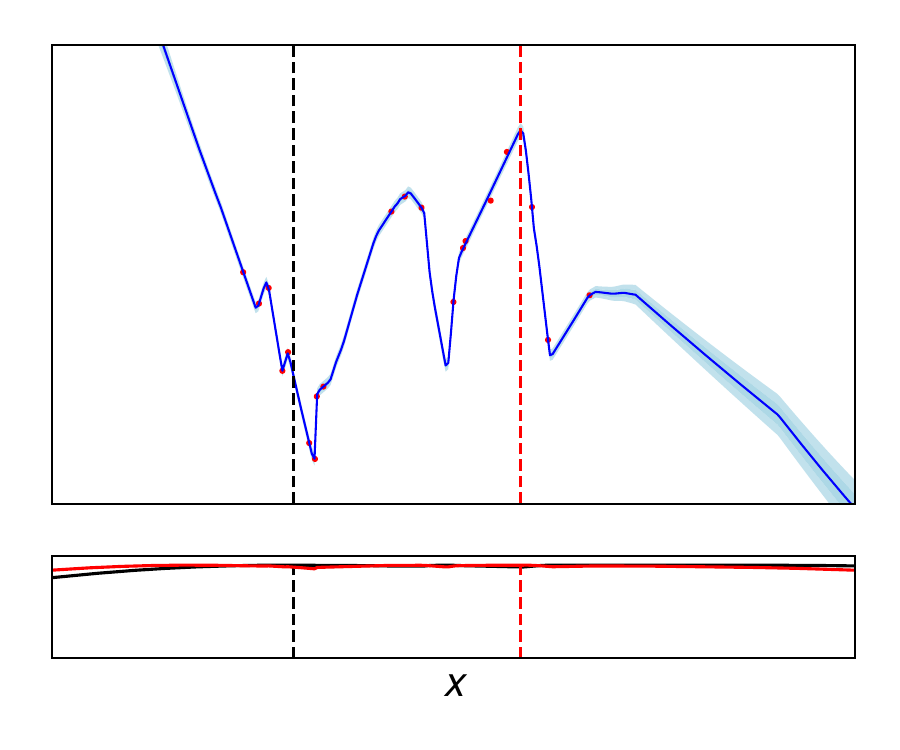}
         \caption{Example DKL. $\mathrm{LML}=28.3$}
         \label{fig:dkl:app-DKL_20}
     \end{subfigure}
        \caption{Plots of fits and training curves using standard SE kernel and DKL. Below each fit we plot two correlation functions $\rho_{x'}(x) = k(x, x')/\sigma_f^2$ induced by each kernel, where the location of $x'$ is given by the dashed vertical lines. (a)-(d) show fits and training curves for the full toy dataset introduced in \citet{snelson2006sparse}, whereas (e)-(f) show fits on the subsampled version from \citet{titsias2009variational}.}
        \label{fig:dkl:app-toy}
\end{figure*}
We show additional plots of fits and training curves for the toy problem from \citet{snelson2006sparse} in Fig.~\ref{fig:dkl:app-toy}. Below the fits, we again show the kernel correlation at two different points $x'$, marked by the vertical dashed lines. In Figure~\ref{fig:dkl:app-RBF_200}, we show the fit using the standard squared exponential kernel, followed by two fits using DKL in Figures~\ref{fig:dkl:app-DKL_200_0} and~\ref{fig:dkl:app-DKL_200_1}. In Fig.~\ref{fig:dkl:app-training_curves} we show training curves for 5 different initializations; note that unlike in Fig.~\ref{fig:dkl:toy_DKL_traincurves} we use a learning rate of 1e-4 here and so require more training iterations to converge. Finally, we consider plots of fits on the subsampled version of the dataset as in \citet{titsias2009variational}; we show fits using the SE kernel and DKL in Figures~\ref{fig:dkl:app-RBF_20} and~\ref{fig:dkl:app-DKL_20}, respectively. For each fit, we also show the log marginal likelihood in the caption.

We make a few observations. First, note that different initializations can lead to very different fits and LMLs. Moreover, as predicted by our theory, the highest log marginal likelihoods are obtained when the prior attempts to correlate all the points in the input domain: Fig.~\ref{fig:dkl:app-DKL_200_0} obtains a higher LML than Fig.~\ref{fig:dkl:app-DKL_200_1}. However, instability in training often leads to worse LMLs than could be obtained (Fig.~\ref{fig:dkl:app-training_curves}). Finally, we note that the overfitting is substantially worse on the subsampled version of the dataset: we also see that the prior is more correlated than previously (Fig.~\ref{fig:dkl:app-DKL_20}).

\subsection{Changing the feature dimension}\label{app:dkl:fd}
We perform experiments changing the feature dimension $Q$ for the UTKFace and CIFAR-10 datasets. We present the results in Tables~\ref{tab:dkl:fd_age} and~\ref{tab:dkl:fd_cifar}, where each model name is followed by the feature space dimension. For UTKFace, it is clear that 2 neurons is not sufficient to fit the data. Beyond 2, we see only minor changes in performance. For CIFAR-10, we find that we need at least 10 neurons to fit well, but beyond 10 there are again only minor differences. We choose 10 neurons for both experiments out of convenience.

\begin{table}
    \centering
    \small
    \caption{Results for UTKFace.}\label{tab:dkl:fd_age}
    \begin{tabular}{rccccc}
\toprule 
& ELBO &  Train RMSE & Test RMSE & Train LL & Test LL \\ 
\midrule 
NN-2 & - & 0.37 $\pm$ 0.26 & 0.61 $\pm$ 0.16 & 0.58 $\pm$ 0.82 & -23.14 $\pm$ 10.94 \\ 
SVDKL-2& -0.22 $\pm$ 1.00 & 0.39 $\pm$ 0.25 & 0.61 $\pm$ 0.16 & 0.14 $\pm$ 0.68 & -3.60 $\pm$ 1.90 \\ 
pNN-2 & - & 0.36 $\pm$ 0.26 & 0.61 $\pm$ 0.16 & 0.74 $\pm$0.88 & -35.66 $\pm$ 14.00 \\ 
fSVDKL-2& -0.32 $\pm$ 4.00 & 0.45 $\pm$ 0.22 & 0.45 $\pm$0.23 & -0.26 $\pm$ 0.50 & -0.23 $\pm$0.52 \\ 
pSVDKL-2& 0.36 $\pm$3.00 & 0.36 $\pm$0.26 & 0.53 $\pm$ 0.20 & 0.47 $\pm$ 0.77 & -3.24 $\pm$ 0.76 \\ 
\midrule 
NN-5 & - & 0.04 $\pm$0.00 & 0.41 $\pm$ 0.01 & 1.72 $\pm$ 0.05 & -42.86 $\pm$ 2.79 \\ 
SVDKL-5& -0.17 $\pm$ 0.59 & 0.11 $\pm$ 0.01 & 0.47 $\pm$ 0.01 & 0.40 $\pm$ 0.11 & -1.41 $\pm$ 0.16 \\ 
pNN-5 & - & 0.04 $\pm$0.00 & 0.41 $\pm$ 0.00 & 1.79 $\pm$0.01 & -48.92 $\pm$ 0.62 \\ 
fSVDKL-5& 0.31 $\pm$ 0.48 & 0.17 $\pm$0.00 & 0.17 $\pm$0.00 & 0.38 $\pm$ 0.02 & 0.37 $\pm$0.03 \\ 
pSVDKL-5& 0.99 $\pm$0.73 & 0.04 $\pm$0.00 & 0.32 $\pm$ 0.01 & 1.18 $\pm$ 0.06 & -2.61 $\pm$ 0.26 \\ 
\midrule 
NN-10 & - & 0.04 $\pm$0.00 & 0.40 $\pm$ 0.00 & 1.81 $\pm$ 0.01 & -48.73 $\pm$ 1.64 \\ 
SVDKL-10& 0.92 $\pm$ 0.15 & 0.04 $\pm$0.00 & 0.40 $\pm$ 0.01 & 1.30 $\pm$ 0.01 & -6.88 $\pm$ 0.38 \\ 
pNN-10 & - & 0.04 $\pm$0.00 & 0.41 $\pm$ 0.00 & 1.83 $\pm$0.01 & -53.72 $\pm$ 1.71 \\ 
fSVDKL-10& 1.05 $\pm$0.02 & 0.08 $\pm$ 0.03 & 0.31 $\pm$0.07 & 1.16 $\pm$ 0.31 & -7.55 $\pm$ 3.42 \\ 
pSVDKL-10& 1.03 $\pm$ 0.07 & 0.04 $\pm$0.00 & 0.38 $\pm$ 0.02 & 1.20 $\pm$ 0.08 & -4.74 $\pm$1.35 \\ 
\midrule 
NN-20 & - & 0.04 $\pm$0.00 & 0.40 $\pm$ 0.00 & 1.78 $\pm$ 0.01 & -46.77 $\pm$ 1.72 \\ 
SVDKL-20& 0.22 $\pm$ 0.01 & 0.08 $\pm$ 0.02 & 0.43 $\pm$ 0.01 & 0.78 $\pm$ 0.22 & -3.42 $\pm$1.52 \\ 
pNN-20 & - & 0.04 $\pm$0.00 & 0.41 $\pm$ 0.00 & 1.80 $\pm$0.01 & -50.51 $\pm$ 0.47 \\ 
fSVDKL-20& 0.71 $\pm$ 0.30 & 0.12 $\pm$ 0.03 & 0.24 $\pm$0.06 & 0.83 $\pm$ 0.34 & -5.06 $\pm$ 4.46 \\ 
pSVDKL-20& 1.15 $\pm$0.10 & 0.04 $\pm$0.00 & 0.34 $\pm$ 0.03 & 1.33 $\pm$ 0.04 & -4.16 $\pm$ 0.42 \\ 
\midrule 
NN-50 & - & 0.04 $\pm$0.00 & 0.40 $\pm$ 0.00 & 1.79 $\pm$ 0.01 & -47.72 $\pm$ 0.68 \\ 
SVDKL-50& 0.92 $\pm$ 0.27 & 0.04 $\pm$0.00 & 0.40 $\pm$ 0.00 & 1.33 $\pm$ 0.03 & -7.35 $\pm$ 0.45 \\ 
pNN-50 & - & 0.04 $\pm$0.00 & 0.41 $\pm$ 0.01 & 1.81 $\pm$0.00 & -51.14 $\pm$ 1.33 \\ 
fSVDKL-50& 1.14 $\pm$ 0.31 & 0.08 $\pm$ 0.03 & 0.32 $\pm$0.06 & 1.29 $\pm$ 0.35 & -11.60 $\pm$ 4.91 \\ 
pSVDKL-50& 1.21 $\pm$0.03 & 0.04 $\pm$0.00 & 0.37 $\pm$ 0.02 & 1.37 $\pm$ 0.02 & -5.71 $\pm$0.55 \\ 
\bottomrule
    \end{tabular}
\end{table}

\begin{landscape}
\begin{table}
    \centering
    \small
    \caption{Results for CIFAR-10.}\label{tab:dkl:fd_cifar}
    \begin{tabular}{rccccccc}
\toprule 
& ELBO &  Train Acc. & Test Acc. & Train LL & Test LL & Inc. Test LL & ECE \\ 
\midrule 
NN-2 & - & 0.69 $\pm$ 0.24 & 0.51 $\pm$ 0.17 & -0.81 $\pm$ 0.61 & -6.42 $\pm$ 2.23 & -6.40 $\pm$ 1.78 & 0.14 $\pm$ 0.06 \\ 
SVDKL-2& -1.38 $\pm$ 0.38 & 0.52 $\pm$ 0.17 & 0.46 $\pm$ 0.15 & -1.34 $\pm$ 0.39 & -1.57 $\pm$0.30 & -2.61 $\pm$0.13 & 0.04 $\pm$0.02 \\ 
pNN-2 & - & 0.70 $\pm$0.24 & 0.53 $\pm$0.17 & -0.77 $\pm$0.63 & -5.25 $\pm$ 1.36 & -7.28 $\pm$ 2.03 & 0.15 $\pm$ 0.06 \\ 
fSVDKL-2& -0.85 $\pm$ 0.59 & 0.69 $\pm$ 0.24 & 0.51 $\pm$ 0.17 & -0.81 $\pm$ 0.61 & -1.83 $\pm$ 0.20 & -4.35 $\pm$ 0.84 & 0.14 $\pm$ 0.06 \\ 
pSVDKL-2& -0.78 $\pm$ 0.62 & 0.70 $\pm$0.24 & 0.53 $\pm$0.17 & -0.78 $\pm$ 0.62 & -1.76 $\pm$ 0.23 & -4.51 $\pm$ 0.90 & 0.13 $\pm$ 0.05 \\ 
\midrule 
NN-5 & - & 0.70 $\pm$0.25 & 0.55 $\pm$0.18 & -0.77 $\pm$0.63 & -3.03 $\pm$ 0.58 & -7.00 $\pm$ 1.92 & 0.13 $\pm$ 0.05 \\ 
SVDKL-5& -1.66 $\pm$ 0.26 & 0.40 $\pm$ 0.12 & 0.39 $\pm$ 0.12 & -1.63 $\pm$ 0.28 & -1.68 $\pm$ 0.25 & -2.30 $\pm$ 0.00 & 0.02 $\pm$ 0.01 \\ 
pNN-5 & - & 0.70 $\pm$ 0.25 & 0.55 $\pm$ 0.18 & -0.77 $\pm$ 0.63 & -3.22 $\pm$ 0.68 & -7.34 $\pm$ 2.06 & 0.13 $\pm$ 0.05 \\ 
fSVDKL-5& -0.79 $\pm$ 0.62 & 0.70 $\pm$ 0.24 & 0.55 $\pm$ 0.18 & -0.77 $\pm$ 0.63 & -1.61 $\pm$ 0.29 & -4.25 $\pm$ 0.80 & 0.09 $\pm$ 0.04 \\ 
pSVDKL-5& -0.77 $\pm$ 0.62 & 0.70 $\pm$ 0.24 & 0.55 $\pm$ 0.19 & -0.77 $\pm$ 0.63 & -1.64 $\pm$ 0.27 & -4.66 $\pm$ 0.96 & 0.11 $\pm$ 0.04 \\ 
\midrule 
NN-10 & - & 1.00 $\pm$ 0.00 & 0.79 $\pm$ 0.00 & -0.00 $\pm$ 0.00 & -2.05 $\pm$ 0.03 & -8.87 $\pm$ 0.10 & 0.18 $\pm$ 0.00 \\ 
SVDKL-10& -0.76 $\pm$ 0.28 & 0.76 $\pm$ 0.09 & 0.63 $\pm$ 0.03 & -0.71 $\pm$ 0.28 & -1.37 $\pm$ 0.10 & -3.38 $\pm$ 0.77 & 0.10 $\pm$ 0.05 \\ 
pNN-10 & - & 1.00 $\pm$ 0.00 & 0.79 $\pm$ 0.00 & -0.00 $\pm$ 0.00 & -2.30 $\pm$ 0.11 & -9.48 $\pm$ 0.30 & 0.19 $\pm$ 0.00 \\ 
fSVDKL-10& -0.02 $\pm$ 0.00 & 1.00 $\pm$ 0.00 & 0.78 $\pm$ 0.00 & -0.01 $\pm$ 0.00 & -1.14 $\pm$ 0.00 & -5.10 $\pm$ 0.01 & 0.14 $\pm$ 0.00 \\ 
pSVDKL-10& -0.00 $\pm$ 0.00 & 1.00 $\pm$ 0.00 & 0.79 $\pm$ 0.00 & -0.00 $\pm$ 0.00 & -1.13 $\pm$ 0.01 & -5.24 $\pm$ 0.05 & 0.15 $\pm$ 0.00 \\ 
\midrule 
NN-20 & - & 1.00 $\pm$ 0.00 & 0.79 $\pm$ 0.00 & -0.00 $\pm$ 0.00 & -2.06 $\pm$ 0.02 & -8.91 $\pm$ 0.14 & 0.18 $\pm$ 0.00 \\ 
SVDKL-20& -0.30 $\pm$ 0.20 & 0.91 $\pm$ 0.06 & 0.70 $\pm$ 0.02 & -0.26 $\pm$ 0.19 & -1.46 $\pm$ 0.16 & -4.72 $\pm$ 0.88 & 0.16 $\pm$ 0.04 \\ 
pNN-20 & - & 1.00 $\pm$ 0.00 & 0.79 $\pm$ 0.00 & -0.00 $\pm$ 0.00 & -2.24 $\pm$ 0.01 & -9.37 $\pm$ 0.08 & 0.19 $\pm$ 0.00 \\ 
fSVDKL-20& -0.02 $\pm$ 0.00 & 1.00 $\pm$ 0.00 & 0.79 $\pm$ 0.00 & -0.01 $\pm$ 0.00 & -1.14 $\pm$ 0.02 & -5.16 $\pm$ 0.10 & 0.13 $\pm$ 0.00 \\ 
pSVDKL-20& -0.00 $\pm$ 0.00 & 1.00 $\pm$ 0.00 & 0.79 $\pm$ 0.00 & -0.00 $\pm$ 0.00 & -1.09 $\pm$ 0.01 & -5.12 $\pm$ 0.05 & 0.15 $\pm$ 0.00 \\ 
\midrule 
NN-50 & - & 1.00 $\pm$ 0.00 & 0.79 $\pm$ 0.00 & -0.00 $\pm$ 0.00 & -2.18 $\pm$ 0.01 & -9.23 $\pm$ 0.04 & 0.19 $\pm$ 0.00 \\ 
SVDKL-50& -2.30 $\pm$ 0.00 & 0.10 $\pm$ 0.00 & 0.10 $\pm$ 0.00 & -2.30 $\pm$ 0.00 & -2.30 $\pm$ 0.00 & -2.30 $\pm$ 0.00 & 0.00 $\pm$ 0.00 \\ 
pNN-50 & - & 1.00 $\pm$ 0.00 & 0.79 $\pm$ 0.00 & -0.00 $\pm$ 0.00 & -2.38 $\pm$ 0.06 & -9.73 $\pm$ 0.15 & 0.19 $\pm$ 0.00 \\ 
fSVDKL-50& -0.02 $\pm$ 0.00 & 1.00 $\pm$ 0.00 & 0.79 $\pm$ 0.00 & -0.00 $\pm$ 0.00 & -1.22 $\pm$ 0.01 & -5.48 $\pm$ 0.05 & 0.14 $\pm$ 0.00 \\ 
pSVDKL-50& -0.00 $\pm$ 0.00 & 1.00 $\pm$ 0.00 & 0.79 $\pm$ 0.00 & -0.00 $\pm$ 0.00 & -1.11 $\pm$ 0.01 & -5.13 $\pm$ 0.04 & 0.15 $\pm$ 0.00 \\ 
\bottomrule
    \end{tabular}
\end{table}
\end{landscape}

\newpage
\section{Tabulated UCI results}\label{app:dkl:uci_tables}
Here we tabulate the results for the UCI datasets.

\begin{table}[ht]
    \centering
    \small
    \caption{Results for \textsc{Boston}. We report means plus or minus one standard error averaged over the splits.}\label{tab:dkl:boston}
    \begin{tabular}{rccccc}
\toprule 
& loss & train RMSE & test RMSE & train LL & test LL \\ 
\midrule 
SVGP & 1.66 $\pm$ 0.06 & 0.39 $\pm$ 0.01 & 0.37 $\pm$ 0.02 & -0.34 $\pm$ 0.01 & -0.33 $\pm$ 0.05 \\ 
fNN & 0.01 $\pm$ 0.00 & 0.02 $\pm$ 0.00 & 0.39 $\pm$ 0.03 & 2.28 $\pm$ 0.03 & -132.41 $\pm$ 22.39 \\ 
sNN & 0.01 $\pm$ 0.00 & 0.10 $\pm$ 0.00 & 0.34 $\pm$ 0.02 & 0.93 $\pm$ 0.02 & -5.61 $\pm$ 1.03 \\ 
DKL & -2.47 $\pm$ 0.00 & 0.00 $\pm$ 0.00 & 0.41 $\pm$ 0.02 & 2.72 $\pm$ 0.00 & -67.55 $\pm$ 3.97 \\ 
SVDKL & -0.47 $\pm$ 0.01 & 0.13 $\pm$ 0.00 & 0.35 $\pm$ 0.02 & 0.57 $\pm$ 0.01 & -1.12 $\pm$ 0.24 \\ 
      \bottomrule 
    \end{tabular}
\end{table}

\begin{table}[ht]
    \centering
    \caption{Results for \textsc{Energy}.}\label{tab:dkl:energy}
    \begin{tabular}{rccccc}
\toprule 
& loss & train RMSE & test RMSE & train LL & test LL \\ 
\midrule 
SVGP & 0.07 $\pm$ 0.01 & 0.19 $\pm$ 0.00 & 0.20 $\pm$ 0.00 & 0.19 $\pm$ 0.01 & 0.15 $\pm$ 0.02 \\ 
fNN & 0.00 $\pm$ 0.00 & 0.02 $\pm$ 0.00 & 0.04 $\pm$ 0.00 & 2.55 $\pm$ 0.02 & -0.04 $\pm$ 0.38 \\ 
sNN & 0.00 $\pm$ 0.00 & 0.02 $\pm$ 0.00 & 0.05 $\pm$ 0.00 & 2.31 $\pm$ 0.02 & 0.62 $\pm$ 0.19 \\ 
DKL & -3.01 $\pm$ 0.02 & 0.01 $\pm$ 0.00 & 0.05 $\pm$ 0.00 & 3.15 $\pm$ 0.02 & -2.63 $\pm$ 0.49 \\ 
SVDKL & -1.21 $\pm$ 0.00 & 0.03 $\pm$ 0.00 & 0.04 $\pm$ 0.00 & 1.26 $\pm$ 0.00 & 1.22 $\pm$ 0.01 \\ 
      \bottomrule 
    \end{tabular}
\end{table}

\begin{table}[ht]
    \centering
    \caption{Results for \textsc{Kin40K}.}\label{tab:dkl:kin40k}
    \begin{tabular}{rccccc}
\toprule 
& loss & train RMSE & test RMSE & train LL & test LL \\ 
\midrule 
SVGP & -0.14 $\pm$ 0.00 & 0.16 $\pm$ 0.00 & 0.17 $\pm$ 0.00 & 0.36 $\pm$ 0.00 & 0.33 $\pm$ 0.00 \\ 
fNN & 0.01 $\pm$ 0.00 & 0.03 $\pm$ 0.00 & 0.05 $\pm$ 0.00 & 2.18 $\pm$ 0.00 & 1.17 $\pm$ 0.02 \\ 
sNN & 0.01 $\pm$ 0.00 & 0.03 $\pm$ 0.00 & 0.05 $\pm$ 0.00 & 2.03 $\pm$ 0.00 & 1.51 $\pm$ 0.01 \\ 
VDKL & -1.41 $\pm$ 0.00 & 0.02 $\pm$ 0.00 & 0.05 $\pm$ 0.00 & 1.44 $\pm$ 0.00 & 1.33 $\pm$ 0.00 \\ 
SVDKL & -2.62 $\pm$ 0.00 & 0.01 $\pm$ 0.00 & 0.03 $\pm$ 0.00 & 2.68 $\pm$ 0.00 & 1.73 $\pm$ 0.02 \\ 
      \bottomrule 
    \end{tabular}
\end{table}

\begin{table}[ht]
    \centering
    \caption{Results for \textsc{Power}.}\label{tab:dkl:power}
    \begin{tabular}{rccccc}
\toprule 
& loss & train RMSE & test RMSE & train LL & test LL \\ 
\midrule 
SVGP & -0.01 $\pm$ 0.00 & 0.23 $\pm$ 0.00 & 0.23 $\pm$ 0.00 & 0.06 $\pm$ 0.00 & 0.07 $\pm$ 0.01 \\ 
fNN & 0.04 $\pm$ 0.00 & 0.17 $\pm$ 0.00 & 0.21 $\pm$ 0.00 & 0.37 $\pm$ 0.00 & 0.11 $\pm$ 0.02 \\ 
sNN & 0.05 $\pm$ 0.00 & 0.21 $\pm$ 0.00 & 0.22 $\pm$ 0.00 & 0.14 $\pm$ 0.00 & 0.11 $\pm$ 0.01 \\ 
VDKL & -0.57 $\pm$ 0.00 & 0.13 $\pm$ 0.00 & 0.21 $\pm$ 0.00 & 0.62 $\pm$ 0.00 & -0.02 $\pm$ 0.02 \\ 
SVDKL & -0.25 $\pm$ 0.00 & 0.18 $\pm$ 0.00 & 0.21 $\pm$ 0.00 & 0.28 $\pm$ 0.00 & 0.16 $\pm$ 0.01 \\ 
      \bottomrule 
    \end{tabular}
\end{table}

\begin{table}[ht]
    \centering
    \caption{Results for \textsc{Protein}.}\label{tab:dkl:protein}
    \begin{tabular}{rccccc}
\toprule 
& loss & train RMSE & test RMSE & train LL & test LL \\ 
\midrule 
SVGP & 1.06 $\pm$ 0.00 & 0.64 $\pm$ 0.00 & 0.66 $\pm$ 0.00 & -0.98 $\pm$ 0.00 & -1.00 $\pm$ 0.00 \\ 
fNN & 0.19 $\pm$ 0.00 & 0.39 $\pm$ 0.00 & 0.58 $\pm$ 0.00 & -0.46 $\pm$ 0.00 & -1.09 $\pm$ 0.01 \\ 
sNN & 0.17 $\pm$ 0.00 & 0.35 $\pm$ 0.00 & 0.55 $\pm$ 0.00 & -0.36 $\pm$ 0.00 & -1.14 $\pm$ 0.01 \\ 
VDKL & 0.32 $\pm$ 0.01 & 0.30 $\pm$ 0.00 & 0.59 $\pm$ 0.00 & -0.23 $\pm$ 0.01 & -1.86 $\pm$ 0.01 \\ 
SVDKL & 0.35 $\pm$ 0.00 & 0.31 $\pm$ 0.00 & 0.57 $\pm$ 0.00 & -0.26 $\pm$ 0.00 & -1.29 $\pm$ 0.01 \\ 

      \bottomrule 
    \end{tabular}
\end{table}

\chapter{Appendix to Chapter 4}
\ifpdf
    \graphicspath{{GI/Figs/Raster/}{GI/Figs/PDF/}{GI/Figs/}}
\else
    \graphicspath{{GI/Figs/Vector/}{GI/Figs/}}
\fi

\section{Reparameterized variational inference}
\label{app:reparam}
In variational inference for Bayesian neural networks, the ELBO takes the form
\begin{align}
  \mathcal{L}(\phi) &= \E[{\qsub[\phi]{\w}}]{\log \pc{\Y}{\X, \w} + \log \frac{\p{\w}}{\qsub[\phi]{\w}}},
\end{align}
where $\w$ is a vector containing all of the elements of the weight matrices in the full network, $\slpo{\W_\ell}$, and $\phi=(\Z_0, \slpo{\V_\ell, \La_\ell})$ are the parameters of the approximate posterior.
This objective is difficult to differentiate with respect to $\phi$, because $\phi$ parameterises the distribution over which the expectation is taken.
Following \citet{kingma2013auto} and \citet{rezende2014stochastic}, we sample $\epsilon$ from a simple, fixed distribution (e.g. a standard normal), and transform them to give samples from $\q{w}$:
\begin{align}
  \w(\epsilon; \phi) \sim \qsub[\phi]{\w}.
\end{align}
Thus the ELBO can be written as
\begin{align}
  \mathcal{L}(\phi) &= \E[\epsilon]{\log \pc{\Y}{\X, \w(\epsilon; \phi)} + \log \frac{\p{\w(\epsilon; \phi)}}{\qsub[\phi]{\w(\epsilon; \phi)}}}.
\end{align}
As the distribution over which the expectation is taken is now independent of $\phi$, we can form unbiased estimates of the gradient of $\mathcal{L}(\phi)$ by drawing one or a few samples of $\epsilon$.
Variational inference in deep Gaussian processes can be handled similarly.

\section{Motivating the approximate posterior for deep GPs}
\label{sec:gi:app:gpmotivation}
Our original motivation for the approximate posterior was for the BNN case, which we then extended to deep GPs. Here, we show how the same approximate posterior can be motivated from a deep GP perspective. As with the BNN case, we first derive the form of the optimal approximate posterior for the last layer, in the regression case. Without inducing points, the ELBO becomes
\begin{align}
    \label{eq:gi:gp-ELBO-fonly}
    \mathcal{L} &= \E[\q{\slpo{\F_\ell}}]
    {\log \frac{\p{\Y, \slpo{\F_\ell}}}{\q{\slpo{\F_\ell}}}},
\end{align}
where we have defined a generic variational posterior $\q{\slpo{\F_\ell}}$. Since we are interested in the form of $\qc{\F_{L+1}}{\slp{\F_\ell}}$, we rearrange the ELBO so that all terms that do not depend on $\F_{L+1}$ are absorbed into a constant, $c$:
\begin{align}
    \label{eq:gi:gp-ELBO-constant}
    \mathcal{L} &= \E[\q{\slpo{\F_\ell}}]{\log \pc{\Y, \F_{L+1}}{\slp{\F_\ell}} - \log \q{\F_{L+1}} + c}.
\end{align}
Some straightforward rearrangements lead to a similar form to before,
\begin{align}
    \mathcal{L} = \E[\q{\slp{\F_\ell}}]{-\KL{\qc{\F_{L+1}}{\slp{\F_\ell}}}{\pc{\F_{L+1}}{\Y, \slp{\F_\ell}}} + c},
\end{align}
from which we see that the optimal conditional posterior is given by $\qc{\F_{L+1}}{\slp{\F_\ell}} = \qc{\F_{L+1}}{\F_L} = \pc{\F_{L+1}}{\Y,\F_L}$, which has a closed form for regression: it is simply the standard GP posterior given by training data $\Y$ at inputs $\F_L$. In particular, for a Gaussian likelihood
\begin{align}
    \pc{\Y}{\F_{L+1}} &= \prodlnlp \Nc{\y_\lambda^{L+1}}{\f_\lambda^{L+1}, \La_{L+1}^{-1}},
\end{align}
where $\La_{L+1}$ is the precision,
\begin{align}
    \qc{\F_{L+1}}{\F_L} &= \prodlnlp \Nc{\f_\lambda^{L+1}}{\S_{L+1}^{\f}\La_{L+1}\y_\lambda^{L+1}, \S_{L+1}^\f}, \\
    \S_{L+1}^\f &= (\Kfl{\F_L}^{-1} + \La_{L+1})^{-1}.
\end{align}
This can be understood as kernelized Bayesian linear regression conditioned on the features from the previous layers.
Finally, as is usual in GPs \citep{rasmussen2006gaussian}, the predictive distribution for test points can be obtained by conditioning using this approximate posterior.

\section{Parameter scaling for Adam}
\label{sec:gi:app:param-scaling}
The standard optimiser for variational BNNs and DGPs is Adam \citep{kingma2014adam}, which we also use.
Considering similar RMSprop updates for simplicity \citep{tieleman2012lecture},
\begin{align}
  \Delta w &= \eta \frac{g}{\sqrt{\E{g^2}}},
\end{align}
where the expectation over $g^2$ is approximated using a moving-average of past gradients.
Thus, absolute parameter changes are going to be of order $\eta$.
This is fine if all the parameters have roughly the same order of magnitude, but becomes a serious problem if some of the parameters are very large and others are very small.
For instance, if a parameter is around $10^{-4}$ and $\eta=10^{-4}$, then a single Adam step can easily double the parameter estimate, or change it from positive to negative.
In contrast, if a parameter is around $1$, then Adam, with $\eta=10^{-4}$ can make proportionally much smaller changes to this parameter, (around $0.01\%$).
Thus, we need to ensure that all of our parameters have the same scale, especially as we mix methods, such as combining factorised and global inducing points.
We thus design all our new approximate posteriors (i.e., the inducing inputs and outputs) such that the parameters have a scale of around $1$.
The key issue is that the mean weights in factorised methods tend to be quite small --- they have scale around $1/\sqrt{\text{fan-in}}$.  
To resolve this issue, we store scaled weights, and we divide these stored, scaled mean parameters by the fan-in as part of the forward pass,
\begin{align}
  \text{weights} &= \frac{\text{scaled weights}}{\sqrt{\text{fan-in}}}.
\end{align}
This scaling allows us to use larger learning rates than are typically used.





\section{Understanding compositional uncertainty}
\label{sec:gi:app:compositional}
In this section, we take inspiration from the experiments of \citet{ustyuzhaninov2019compositional}, which investigate the compositional uncertainty obtained by different approximate posteriors for DGPs. They noted that methods which factorise over layers have a tendency to cause the posterior distribution for each layer to collapse to a (nearly) deterministic function, resulting in worse uncertainty quantification within layers and worse ELBOs. In contrast, they found that allowing the approximate posterior to have correlations between layers allows those layers to capture more uncertainty, resulting in better ELBOs and therefore a closer approximation to the true posterior. They argue that this then allows the model to better discover compositional structure in the data. 
\begin{figure}[!t]
    \centering 
    \includegraphics{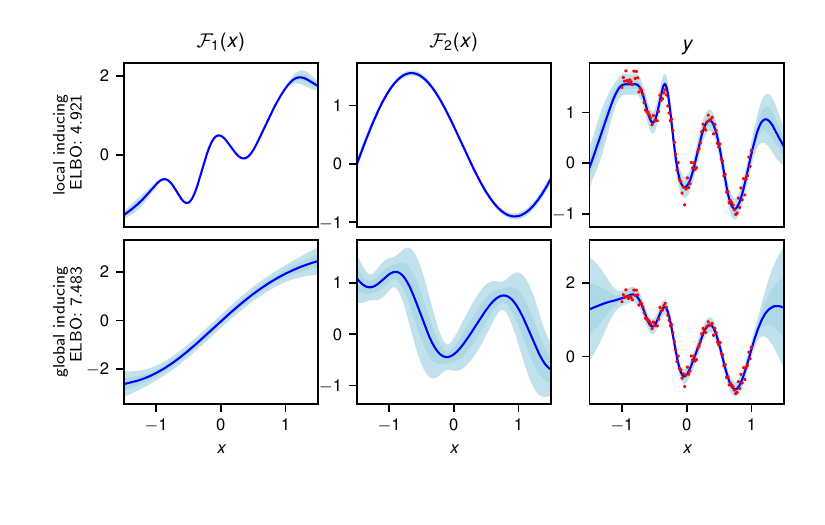}
    \caption{Posterior distributions for 2-layer DGPs with local inducing and global inducing. The first two columns show the predictive distributions for each layer taken individually, while the last column shows the predictive distribution of the output $y$.}
    \label{fig:gi:dgp-uncertainty}
\end{figure}
\begin{table}
    \centering
    \caption{ELBOs and variances of the intermediate functions for a BNN fit to the toy data of Fig.~\ref{fig:gi:toy}.}
    \label{tab:gi:comp-uncertainty}
    \begin{tabular}{lcccc}
        \toprule
         & ELBO & $\mathbb{V}[\mathcal{F}_1]$ & $\mathbb{V}[\mathcal{F}_2]$ & $\mathbb{V}[\mathcal{F}_3]$  \\
         \midrule
         factorised & -4.585 & 0.0728 & 0.4765 & 0.1926 \\
         local inducing & -5.469 & 0.0763 & 0.4473 & 0.0643 \\
         global inducing & 2.236 & 0.4820 & 0.4877 & 1.0820 \\
         \bottomrule
     \end{tabular}
\end{table}

We first consider a toy problem consisting of 100 datapoints generated by sampling from a two-layer DGP of width one, with squared-exponential kernels in each layer. We then fit two two-layer DGPs to this data - one using local inducing, the other using global inducing. The results of this experiment can be seen in Fig.~\ref{fig:gi:dgp-uncertainty}, which show the final fit, along with the learned posteriors over intermediate functions $\mathcal{F}_1$ and $\mathcal{F}_2$. These results mirror those observed by \citet{ustyuzhaninov2019compositional} on a similar experiment (c.f. Figures 5 and 6): local inducing, which factorises over layers, collapses to a nearly deterministic posterior over the intermediate functions, whereas global inducing provides a much broader distribution over functions for the two layers. Therefore, global inducing leads to a wider range of plausible functions that could explain the data via composition, which can be important in understanding the data. We observe that this behaviour directly leads to better uncertainty quantification for the out-of-distribution region, as well as better ELBOs.

To illustrate a similar phenomenon in BNNs, we reconsider the toy problem of Sec.~\ref{sec:gi:toy}. As it is not meaningful to consider neural networks with only one hidden unit per layer, instead of plotting intermediate functions we instead look at the mean variance of the functions at random input points, following roughly the experiment of \citet{dutordoir2020bayesian} in Table 1. For each layer, we consider the quantity
\begin{equation}
    \mathbb{E}_x\left[\frac{1}{\nu_\ell}\sum_{\lambda=1}^{\nu_\ell}\mathbb{V}[f^l_\lambda(x)]\right],
\end{equation}
where the expectation is over random input points, which we sample from a standard normal. We expect that for methods which introduce correlations across layers, this quantity will be higher, as there will be a wider range of intermediate functions that could plausibly explain the data. We confirm this in Table~\ref{tab:gi:comp-uncertainty}, which indicates that global inducing leads to many more compositions of functions being considered as plausible explanations of the data. This is additionally reflected in the ELBO, which is far better for global inducing than the other, factorised methods. However, we note that the variances are far closer than we might otherwise expect for the second layer. We hypothesise that this is due to the pruning effects described in \citet{trippe2018overpruning}, where a layer has many weights that are close to the prior that are then pruned out by the following layer by having the outgoing weights collapse to zero. In fact, we note that the variances in the last layer are small for the factorised methods, which supports this hypothesis. By contrast, global inducing leads to high variances across all layers.

We believe that understanding the role of compositional uncertainty in variational inference for deep Bayesian models can lead to important conclusions about both the models being used and the compositional structure underlying the data being modelled, and is therefore an important direction for future work to consider.

\section{UCI results with Bayesian neural networks}
\label{sec:gi:app:ucibnn}
\begin{figure}[!t]
    \centering
    \includegraphics{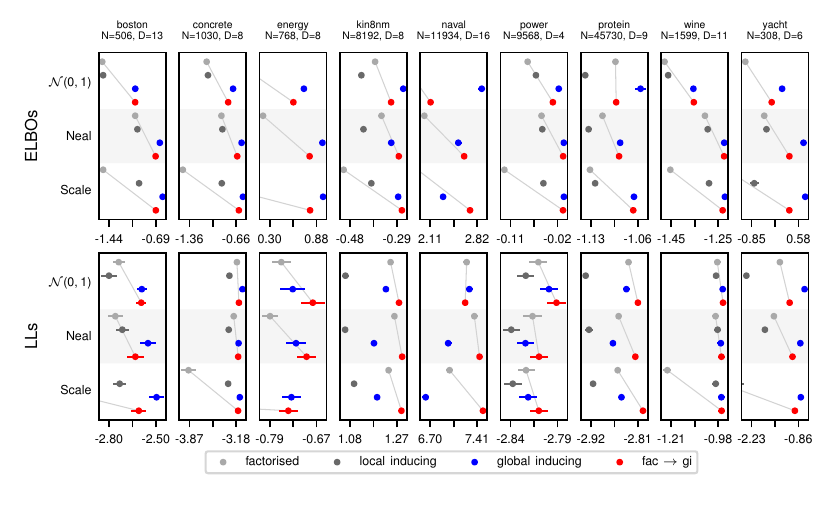}
    \caption{ELBOs per datapoint and average test log likelihoods for BNNs on UCI datasets.
    \label{fig:gi:uci_rmse_elbo}}
\end{figure}
For this Appendix, we consider all of the UCI datasets from \cite{hernandez2015probabilistic}, along with four approximation families: factorised (i.e. mean-field), local inducing, global inducing, and fac$\rightarrow$gi, which may offer some computational advantages to global inducing. We also considered three priors: the standard $\mathcal{N}(0, 1)$ prior, NealPrior, and ScalePrior. The test LLs and ELBOs for BNNs applied to UCI datasets are given in Fig.~\ref{fig:gi:uci_rmse_elbo}.
Note that the ELBOs for the global inducing methods (both global inducing and fac$\rightarrow$gi) are almost always better than those for baseline methods, often by a very large margin. However, as noted earlier, this does not necessarily correspond to better test log likelihoods due to model misspecification: there is not a straightforward relationship between the ELBO and the predictive performance, and so it is possible to obtain better test log likelihoods with worse inference. We present all the results, including for the test RMSEs, in tabulated form in Appendix~\ref{sec:gi:app:tables}.

\subsection{Experimental details}
The architecture we considered for all BNN UCI experiments were fully-connected ReLU networks with 2 hidden layers of 50 hidden units each. We performed a grid search to select the learning rate and minibatch size. For the fully factorised approximation, we selected the learning rate from $\{3\textrm{e-}4, 1\textrm{e-}3, 3\textrm{e-}3, 1\textrm{e-}2\}$ and the minibatch size from $\{32, 100, 500\}$, optimising for 25000 gradient steps; for the other methods we selected the learning rate from $\{3\textrm{e-}3, 1\textrm{e-}2\}$ and fixed the minibatch size to 10000 \citep[as in][]{salimbeni2017doubly}, optimising for 10000 gradient steps. For all methods we selected the hyperparameters that gave the best ELBO. 
We trained the models using 10 samples from the approximate posterior, while using 100 for evaluation. For the inducing point methods, we used the selected batch size for the number of inducing points per layer. For all methods, we initialised the log noise variance at -3, but use the scaling trick in Appendix~\ref{sec:gi:app:param-scaling} to accelerate convergence, scaling by a factor of 10. Note that for the fully factorised method we used the local reparameterisation trick \citep{kingma2015variational}; however, for fac $\rightarrow$ gi we cannot do so because the inducing point methods require that covariances be propagated through the network correctly.
For the inducing point methods, we additionally use output channel-specific precisions, $\La_\lambda^\ell$, which effectively allows the network to prune unnecessary neurons if that benefits the ELBO.
However, we only parameterise the diagonal of these precision matrices to save on computational and memory cost.

\section{Uncertainty calibration \& out-of-distribution detection for CIFAR-10}
\label{sec:gi:app:uncertainty}
To assess how well our methods capture uncertainty, we consider calibration, as well as the predictive entropy for out-of-distribution data.
Calibration is assessed by comparing the model's probabilistic assessment of its confidence with its accuracy --- the proportion of the time that it is actually correct.
For instance, gathering model predictions with some confidence (e.g. softmax probabilities in the range 0.9 to 0.95), and looking at the accuracy of these predictions, we would expect the model to be correct with probability 0.925; a higher or lower value would represent miscalibration.

\begin{figure}
    \centering
    \includegraphics{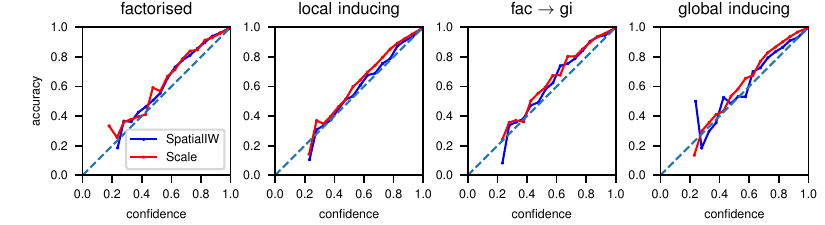}
    \caption{Calibration curves for CIFAR-10}
    \label{fig:gi:calibration}
\end{figure}

We begin by plotting calibration curves (for the small `ResNet' model) in Fig.~\ref{fig:gi:calibration}, obtained by binning the predictions in 20 equal bins and assessing the mean accuracy of the binned predictions.
For well-calibrated models, we expect the line to lie on the diagonal.
A line above the diagonal indicates the model is underconfident (the model is performing better than it expects), whereas a line below the diagonal indicates it is overconfident (it is performing worse than it expects).
While it is difficult to draw strong conclusions from these plots, it appears generally that factorised is poorly calibrated for both priors, that SpatialIWPrior generally improves calibration over ScalePrior, and that local inducing with SpatialIWPrior performs very well.

To come to more quantitative conclusions, we use expected calibration error (ECE; \citep{naeini2015obtaining, guo2017calibration}), which measures the expected absolute difference between the model's confidence and accuracy.
Confirming the results from the plots (Fig.~\ref{fig:gi:calibration}), we find that using the more sophisticated SpatialIWPrior gave considerable improvements in calibration.
While, as expected, we find that our most accurate prior, SpatialIWPrior, in combination with global inducing points did very well (ECE of 0.021), the model with the best ECE is actually local inducing with SpatialIWPrior, albeit by a very small margin.
We leave investigation of exactly why this is to future work.
Finally, note our final ECE value of 0.021 is a considerable improvement over those for uncalibrated models in \citet{guo2017calibration} (Table 1), which are in the region of 0.03-0.045 (although considerably better calibration can be achieved by post-hoc scaling of the model's confidence).
\begin{table}
    \centering
    \caption{Expected calibration error for CIFAR-10}
    \label{tab:gi:ece-table}
    \begin{tabular}{lcccc}
        \toprule
         & factorised & local inducing & fac $\rightarrow$ gi & global inducing  \\
         \midrule
         ScalePrior & 0.053 & 0.040 & 0.049 & \textbf{0.038} \\
         SpatialIWPrior & 0.045 & \textbf{0.018} & 0.036 & 0.021 \\
         \bottomrule
     \end{tabular}
\end{table}

In addition to calibration, we consider out-of-distribution performance.
Given out-of-distribution data, we would hope that the network would give high output entropies (i.e. low confidence in the predictions), whereas for in-distribution data, we would hope for low entropy predictions (i.e. high confidence in the predictions).
To evaluate this, we consider the mean predictive entropies for both the CIFAR-10 test set and the SVHN \citep{netzer2011reading} test set, when the network has been trained on CIFAR-10.
We compare the ratio (CIFAR-10 entropy/ SVHN entropy) of these mean predictive entropies for each model in Table~\ref{tab:gi:entropy-table}; a lower ratio indicates that the model is doing a better job of differentiating the datasets.
We see that for both priors we considered, global inducing performs the best.

\begin{table}[t]
    \centering
    \caption{Predictive entropy ratios for CIFAR-10 \& SVHN}
    \label{tab:gi:entropy-table}
    \begin{tabular}{lcccc}
        \toprule
         & factorised & local inducing & fac $\rightarrow$ gi & global inducing  \\
         \midrule
         ScalePrior & 0.506 & 0.534 & 0.462 & \textbf{0.352} \\
         SpatialIWPrior & 0.461 & 0.480 & 0.412 & \textbf{0.342} \\
         \bottomrule
     \end{tabular}
\end{table}

\section{UCI results with deep Gaussian processes}
\label{sec:gi:app:ucidgp}
In this appendix, we again consider all of the UCI datasets from \citet{hernandez2015probabilistic} for DGPs with depths ranging from two to five layers.
We compare DSVI \citep{salimbeni2017doubly}, local inducing, and global inducing.
While local inducing uses the same inducing-point architecture as \citet{salimbeni2017doubly}, the actual implementation and parameterisation is very different.
As such, we do expect to see differences between local inducing and \citet{salimbeni2017doubly}.

\begin{figure}[!t]
    \centering
    \includegraphics{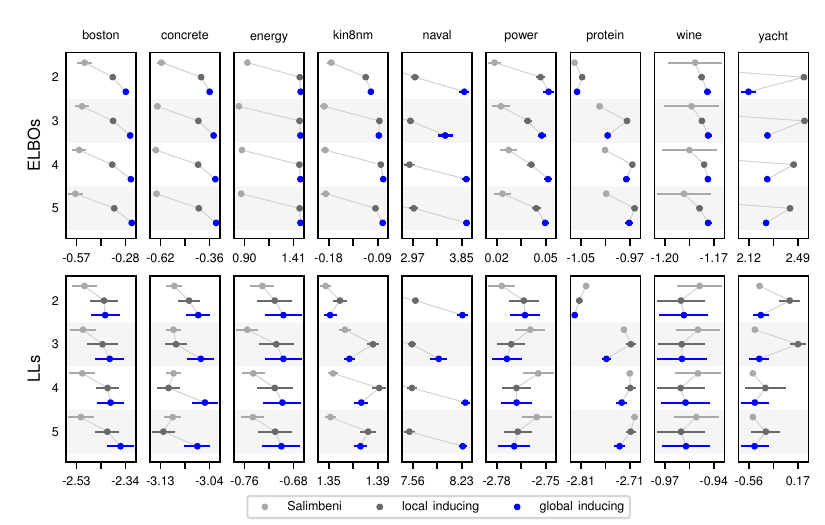}
    \caption{ELBOs per datapoint and average test log likelihoods for DGPs on UCI datasets. The numbers indicate the depths of the models.
    \label{fig:gi:uciresgp}}
\end{figure}

We show our results in Fig.~\ref{fig:gi:uciresgp}. Here, the results are not as clear-cut as in the BNN case. 
For the smaller datasets (i.e. boston, concrete, energy, wine, but with the notable exception of yacht), global inducing generally outperforms both local inducing and DSVI, as noted in the main text, especially when considering the ELBOs.
We do however observe that for power, protein, yacht, and one model for kin8nm, the local approaches sometimes outperform global inducing, even for the ELBOs.
We believe this is due to the fact that relatively few inducing points were used (100), in combination with the fact that global inducing has far fewer variational parameters than the local approaches.
This may make optimisation harder in the global inducing case, especially for 
larger datasets where the model uncertainty does not matter as much as the posterior concentration will be stronger.
Importantly, however, our results on CIFAR-10 indicate that these issues do not arise in very large-scale, high-dimensional datasets, which are of most interest for future work.
Surprisingly, local inducing generally significantly outperforms DSVI, even though they are different parameterisations of the same approximate posterior.
We leave consideration of why this is for future work.

We provide tabulated results, including for RMSEs, in Appendix~\ref{sec:gi:app:tables}.

\subsection{Experimental details}
Here, we matched the experimental setup in \citet{salimbeni2017doubly} as closely as possible.
In particular, we used 100 inducing points, and full-covariance observation noise.
However, our parameterisation is still somewhat different from theirs, in part because our approximate posterior is defined in terms of noisy function-values, while their approximate posterior was defined in terms of the function-values themselves.

As the original results in \citet{salimbeni2017doubly} used different UCI splits, and did not provide the ELBO, we reran their code,\footnote{\texttt{https://github.com/ICL-SML/Doubly-Stochastic-DGP}} changing the number of epochs and noise variance to reflect the values in their paper. 
This gave very similar log likelihoods to those reported in the paper.

\section{MNIST 500}
\label{sec:gi:app:mnist500}
For MNIST, we considered a LeNet-inspired model consisting of two conv2d-relu-maxpool blocks, followed by conv2d-relu-linear, where the convolutions all have $3\times3$ kernels with 64 channels.
We trained all models using a learning rate of $10^{-3}$.

When training on very small datasets, such as the first 500 training examples in MNIST, we can see a variety of pathologies emerge with standard methods.
To help build intuition for these pathologies, we introduce a sanity check for the ELBO.
In particular, we could imagine a model that sets the distribution over all lower-layer parameters equal to the prior, and sets the top-layer parameters so as to ensure that the predictions are uniform.
With $10$ classes, this results in an average test log likelihood of $-2.30 \approx \log (1/10)$, and an ELBO (per datapoint) of approximately $-2.30$.
We found that many combinations of the approximate posterior/prior converged to ELBOs near this baseline.
Indeed, the only approximate posterior to escape this baseline for ScalePrior and SpatialIWPrior is global inducing points.
This is because ScalePrior and SpatialIWPrior both offer the flexibility to shrink the prior variance, and hence shrink the weights towards zero, giving uniform predictions, and potentially zero KL divergence.
In contrast, NealPrior and StandardPrior do not offer this flexibility: you always have to pay something in KL divergence in order to give uniform predictions.
We believe that this is the reason that factorised performs better than expected with NealPrior, despite having an ELBO that is close to the baseline.
Furthermore, it is unclear why local inducing gives very test log likelihood and performance, despite having an ELBO that is similar to factorised.
For StandardPrior, all the ELBOs are far lower than the baseline, and far lower than for any other priors.
Despite this, factorised and fac $\rightarrow$ gi in combination with StandardPrior appear to transiently perform better in terms of predictive accuracy than any other method.
These results should sound a note of caution whenever we try to use factorised approximate posteriors with fixed prior covariances \citep[e.g.][]{blundell2015weight,farquhar2020liberty}.
We leave a full investigation of these effects for future work.

\begin{figure}
    \centering
    \includegraphics{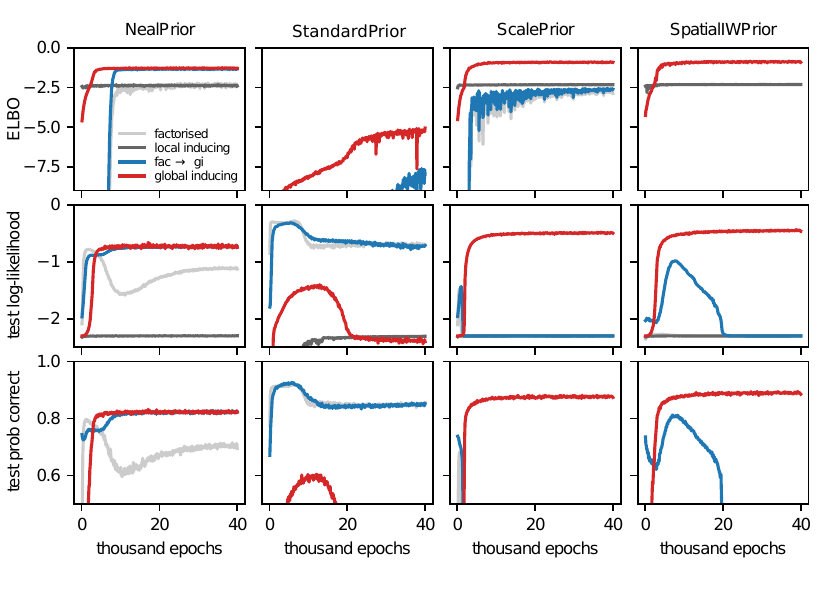}
    \caption{The ELBO, test log likelihoods and classification accuracy with different priors and approximate posteriors on a reduced MNIST dataset consisting of only the first 500 training examples.
    \label{fig:gi:mnist_500}}
\end{figure}

\section{Additional experimental details}
\label{sec:gi:app:deets}
All the methods were implemented in PyTorch. We ran the toy experiments on CPU, with the UCI experiments being run on a mixture of CPU and GPU. The remaining experiments -- linear, CIFAR-10, and MNIST 500 -- were run on various GPUs. For CIFAR-10, the most intensive of our experiments, we trained the models on one NVIDIA Tesla P100-PCIE-16GB. We optimised using Adam \citep{kingma2014adam} throughout.

\paragraph{Factorised} We initialise the posterior weight means to follow the Neal scaling (i.e. drawn from NealPrior); however, we use the scaling described in Appendix~\ref{sec:gi:app:param-scaling} to accelerate training. We initialise the weight variances to $1\textrm{e-}3/\sqrt{\nu_{\ell-1}}$ for each layer.

\paragraph{Inducing point methods} For global inducing, we initialise the inducing inputs, $\Z_0$, and pseudo-outputs for the last layer, $\V_{L+1}$, using the first batch of data, except for the toy experiment, where we initialise using samples from $\N{0, 1}$ (since we used more inducing points than datapoints). For the remaining layers, we initialise the pseudo-outputs by sampling from $\N{0, 1}$. We initialise the log precision to $-4$, except for the last layer, where we initialise it to 0. We additionally use a scaling factor of 3 as described in Appendix~\ref{sec:gi:app:param-scaling}. For local inducing, the initialisation is largely the same, except we initialise the pseudo-outputs for every layer by sampling from $\N{0, 1}$. We additionally sample the inducing inputs for every layer from $\N{0, 1}$.

\paragraph{Toy experiment} For each variational method, we optimise the ELBO over 5000 epochs, using full batches for the gradient descent. We use a learning rate of $1\textrm{e-}2$. We fix the noise variance at its true value, to help assess the differences between each method more clearly. We use 10 samples from the variational posterior for training, using 100 for testing. For HMC, we use 10000 samples to burn in, and 10000 samples for evaluation, which we subsequently thin by a factor of 10. We initialise the samples from a standard normal distribution, and use 20 leapfrog steps for each sample. We hand-tune the leapfrog step sizes to be 0.0007 and 0.003 for the burn-in and sampling phases, respectively.

\paragraph{Deep linear network} We use 10 inducing points for the inducing point methods. We use 1 sample from the approximate posterior for training and 10 for testing, training for 40 periods of 1000 gradient steps, using full batches for each step, with a learning rate of $1\textrm{e-}2$.

\paragraph{UCI experiments} The splits that we used for the UCI datasets can be found at \url{https://github.com/yaringal/DropoutUncertaintyExps}.

\paragraph{CIFAR-10} The CIFAR-10 dataset (\url{https://www.cs.toronto.edu/~kriz/cifar.html}; \citep{krizhevsky2009learning}) is a 10-class dataset comprising RGB, $32\times 32$ images. It is divided in two sets: a training set of 50,000 examples, and a validation set of 10,000 examples. For the purposes of this paper, we use the validation set as our test set and refer to it as such, as is commonly done in the literature. We use a batch size of 500, with one sample from the approximate posterior for training and 10 for testing. For pre-processing, we normalise the data using the training dataset's mean and standard deviation. Finally, we train for 1000 epochs with a learning rate of 1e-2 (see App.~\ref{sec:gi:app:param-scaling} for an explanation of why our learning rate is higher than might be expected), and we use a tempering scheme for the first 100 epochs, slowly increasing the influence of the KL divergence to the prior by multiplying it by a factor that increases from 0 to 1. In our scheme, we increase the factor in a step-wise manner, meaning that for the first ten epochs it is 0, then 0.1 for the next ten, 0.2 for the following ten, and so on. Importantly, we still have 900 epochs of training where the standard, untempered ELBO is used, meaning that our results reflect that ELBO. Finally, we note that we share the precisions $\La_\ell$ within layers instead of using a separate precision for each output channel as was done in the UCI case. This saves memory and computational cost although possibly at the expense of predictive performance. For the full ResNet-18 experiments, we use the same training procedure. We do not use batch normalization \citep{ioffe2015batch} for the fully factorised case, since batch normalization is difficult to interpret in a Bayesian manner as it treats training and testing points differently. However, for the inducing point methods, we use a modified version of batch normalization that computes batch statistics from the inducing data, which do not change between training and testing. For the random cropping, we use zero padding of four pixels on each edge (resulting in $40\times 40$ images), and randomly crop out a $32\times 32$ image.

\paragraph{MNIST 500} The MNIST dataset (\url{http://yann.lecun.com/exdb/mnist/}) is a dataset of grayscale handwritten digits, each $28\times 28$ pixels, with 10 classes. It comprises 60,000 training images and 10,000 test images. For the MNIST 500 experiments, we trained using the first 500 images from the training dataset and discarded the rest. We normalised the images using the full training dataset's statistics.

\paragraph{Deep GPs} As mentioned, we largely follow the approach of \citet{salimbeni2017doubly} for hyperparameters. For global inducing, we initialise the inducing inputs to the first batch of training inputs, and we initialise the pseudo-outputs for the last layer to the respective training outputs. For the remaining layers, we initialise the pseudo-outputs by sampling from a standard normal distribution. We initialise the precision matrix to be diagonal with log precision zero for the output layer, and log precision -4 for the remaining layers. For local inducing, we initialise inducing inputs and pseudo-outputs by sampling from a standard normal for every layer, and initialise the precision matrices to be diagonal with log precision zero.

\newpage
\section{Tables of UCI Results}

\label{sec:gi:app:tables}
We provide tables of test log likelihoods, root mean square error, and ELBOs for all our experiments.
We highlight the best result for each configuration across all methods in bold.
For the ELBOs, the highlighted results take the error bars into account, while we did not do this for the test metrics.
This is because we found that most of the variation in the test metrics across splits was due to the splitting procedure, rather than from the method's actual performance, meaning that one method would perform consistently better across all (or nearly all) splits.

\begin{table}[ht]
\small
  \caption{Average test log likelihoods in nats for BNNs on UCI datasets (errors are $\pm$ 1 standard error)}
  \label{tab:gi:uci_lls}
  \centering
  \begin{tabular}{rccccccccc}
  \toprule 
& factorised & local inducing & global inducing & fac $\rightarrow$ global \\ 
\midrule 
boston - $\mathcal{N}(0, 1)$ & -2.74 $\pm$ 0.03 & -2.80 $\pm$ 0.04 & $\mathbf{-2.59 \pm 0.03}$ & -2.60 $\pm$ 0.02 \\ 
NealPrior & -2.76 $\pm$ 0.04 & -2.71 $\pm$ 0.04 & $\mathbf{-2.55 \pm 0.05}$ & -2.63 $\pm$ 0.05 \\ 
ScalePrior& -3.63 $\pm$ 0.03 & -2.73 $\pm$ 0.03 & $\mathbf{-2.50 \pm 0.04}$ & -2.61 $\pm$ 0.04 \\ 
\midrule 
concrete - $\mathcal{N}(0, 1)$ & -3.17 $\pm$ 0.02 & -3.28 $\pm$ 0.01 & $\mathbf{-3.08 \pm 0.01}$ & -3.14 $\pm$ 0.01 \\ 
NealPrior & -3.21 $\pm$ 0.01 & -3.29 $\pm$ 0.01 & $\mathbf{-3.14 \pm 0.01}$ & -3.15 $\pm$ 0.02 \\ 
ScalePrior& -3.89 $\pm$ 0.09 & -3.30 $\pm$ 0.01 & $\mathbf{-3.12 \pm 0.01}$ & -3.15 $\pm$ 0.01 \\ 
\midrule 
energy - $\mathcal{N}(0, 1)$ & -0.76 $\pm$ 0.02 & -1.75 $\pm$ 0.01 & -0.73 $\pm$ 0.03 & $\mathbf{-0.68 \pm 0.03}$ \\ 
NealPrior & -0.79 $\pm$ 0.02 & -2.06 $\pm$ 0.09 & -0.72 $\pm$ 0.02 & $\mathbf{-0.70 \pm 0.02}$ \\ 
ScalePrior& -2.55 $\pm$ 0.01 & -2.42 $\pm$ 0.02 & $\mathbf{-0.73 \pm 0.02}$ & -0.74 $\pm$ 0.02 \\ 
\midrule 
kin8nm - $\mathcal{N}(0, 1)$ & 1.24 $\pm$ 0.01 & 1.06 $\pm$ 0.01 & 1.22 $\pm$ 0.01 & $\mathbf{1.28 \pm 0.01}$ \\ 
NealPrior & 1.26 $\pm$ 0.01 & 1.06 $\pm$ 0.01 & 1.18 $\pm$ 0.01 & $\mathbf{1.29 \pm 0.01}$ \\ 
ScalePrior& 1.23 $\pm$ 0.01 & 1.10 $\pm$ 0.01 & 1.19 $\pm$ 0.01 & $\mathbf{1.29 \pm 0.00}$ \\ 
\midrule 
naval - $\mathcal{N}(0, 1)$ & 7.25 $\pm$ 0.04 & 6.06 $\pm$ 0.10 & $\mathbf{7.29 \pm 0.04}$ & 7.23 $\pm$ 0.02 \\ 
NealPrior & 7.37 $\pm$ 0.03 & 4.28 $\pm$ 0.37 & 6.97 $\pm$ 0.04 & $\mathbf{7.45 \pm 0.02}$ \\ 
ScalePrior& 6.99 $\pm$ 0.03 & 2.80 $\pm$ 0.00 & 6.63 $\pm$ 0.04 & $\mathbf{7.50 \pm 0.02}$ \\ 
\midrule 
power - $\mathcal{N}(0, 1)$ & -2.81 $\pm$ 0.01 & -2.82 $\pm$ 0.01 & -2.80 $\pm$ 0.01 & $\mathbf{-2.79 \pm 0.01}$ \\ 
NealPrior & $\mathbf{-2.81 \pm 0.01}$ & -2.84 $\pm$ 0.01 & -2.82 $\pm$ 0.01 & $\mathbf{-2.81 \pm 0.01}$ \\ 
ScalePrior& -2.82 $\pm$ 0.01 & -2.84 $\pm$ 0.01 & -2.82 $\pm$ 0.01 & $\mathbf{-2.81 \pm 0.01}$ \\ 
\midrule 
protein - $\mathcal{N}(0, 1)$ & -2.83 $\pm$ 0.00 & -2.93 $\pm$ 0.00 & -2.84 $\pm$ 0.00 & $\mathbf{-2.81 \pm 0.00}$ \\ 
NealPrior & -2.86 $\pm$ 0.00 & -2.92 $\pm$ 0.01 & -2.87 $\pm$ 0.00 & $\mathbf{-2.82 \pm 0.00}$ \\ 
ScalePrior& -2.86 $\pm$ 0.00 & -2.91 $\pm$ 0.00 & -2.85 $\pm$ 0.00 & $\mathbf{-2.80 \pm 0.00}$ \\ 
\midrule 
wine - $\mathcal{N}(0, 1)$ & -0.98 $\pm$ 0.01 & -0.99 $\pm$ 0.01 & $\mathbf{-0.96 \pm 0.01}$ & $\mathbf{-0.96 \pm 0.01}$ \\ 
NealPrior & -0.99 $\pm$ 0.01 & -0.98 $\pm$ 0.01 & -0.97 $\pm$ 0.01 & $\mathbf{-0.96 \pm 0.01}$ \\ 
ScalePrior& -1.22 $\pm$ 0.01 & -0.99 $\pm$ 0.01 & $\mathbf{-0.96 \pm 0.01}$ & $\mathbf{-0.96 \pm 0.01}$ \\ 
\midrule 
yacht - $\mathcal{N}(0, 1)$ & -1.41 $\pm$ 0.05 & -2.39 $\pm$ 0.05 & $\mathbf{-0.68 \pm 0.03}$ & -1.12 $\pm$ 0.02 \\ 
NealPrior & -1.58 $\pm$ 0.04 & -1.84 $\pm$ 0.05 & $\mathbf{-0.81 \pm 0.03}$ & -1.04 $\pm$ 0.01 \\ 
ScalePrior& -4.12 $\pm$ 0.03 & -2.71 $\pm$ 0.22 & $\mathbf{-0.79 \pm 0.02}$ & -0.97 $\pm$ 0.05 \\ 
\bottomrule 
  \end{tabular}
\end{table}

\begin{table}[ht]
\small
  \caption{Test RMSEs for BNNs on UCI datasets (errors are $\pm$ 1 standard error)}
  \label{tab:gi:uci_rmse}
  \centering
  \begin{tabular}{rccccccccc}
\toprule 
& factorised & local inducing & global inducing & fac $\rightarrow$ global \\ 
\midrule 
boston - $\mathcal{N}(0, 1)$ & 3.60 $\pm$ 0.21 & 3.85 $\pm$ 0.26 & $\mathbf{3.13 \pm 0.20}$ & 3.14 $\pm$ 0.20 \\ 
NealPrior & 3.64 $\pm$ 0.24 & 3.55 $\pm$ 0.23 & $\mathbf{3.14 \pm 0.18}$ & 3.33 $\pm$ 0.21 \\ 
ScalePrior& 9.03 $\pm$ 0.26 & 3.57 $\pm$ 0.20 & $\mathbf{2.97 \pm 0.19}$ & 3.27 $\pm$ 0.20 \\ 
\midrule 
concrete - $\mathcal{N}(0, 1)$ & 5.73 $\pm$ 0.11 & 6.34 $\pm$ 0.11 & $\mathbf{5.39 \pm 0.09}$ & 5.55 $\pm$ 0.10 \\ 
NealPrior & 5.96 $\pm$ 0.11 & 6.35 $\pm$ 0.12 & $\mathbf{5.64 \pm 0.10}$ & 5.70 $\pm$ 0.11 \\ 
ScalePrior& 12.66 $\pm$ 1.04 & 6.48 $\pm$ 0.12 & $\mathbf{5.56 \pm 0.10}$ & 5.68 $\pm$ 0.09 \\ 
\midrule 
energy - $\mathcal{N}(0, 1)$ & 0.51 $\pm$ 0.01 & 1.35 $\pm$ 0.02 & 0.50 $\pm$ 0.01 & $\mathbf{0.47 \pm 0.02}$ \\ 
NealPrior & 0.51 $\pm$ 0.01 & 1.95 $\pm$ 0.14 & 0.49 $\pm$ 0.01 & $\mathbf{0.47 \pm 0.01}$ \\ 
ScalePrior& 3.02 $\pm$ 0.05 & 2.67 $\pm$ 0.06 & 0.50 $\pm$ 0.01 & $\mathbf{0.49 \pm 0.01}$ \\ 
\midrule 
kin8nm - $\mathcal{N}(0, 1)$ & $\mathbf{0.07 \pm 0.00}$ & 0.08 $\pm$ 0.00 & $\mathbf{0.07 \pm 0.00}$ & $\mathbf{0.07 \pm 0.00}$ \\ 
NealPrior & $\mathbf{0.07 \pm 0.00}$ & 0.08 $\pm$ 0.00 & $\mathbf{0.07 \pm 0.00}$ & $\mathbf{0.07 \pm 0.00}$ \\ 
ScalePrior& $\mathbf{0.07 \pm 0.00}$ & 0.08 $\pm$ 0.00 & $\mathbf{0.07 \pm 0.00}$ & $\mathbf{0.07 \pm 0.00}$ \\ 
\midrule 
naval - $\mathcal{N}(0, 1)$ & $\mathbf{0.00 \pm 0.00}$ & $\mathbf{0.00 \pm 0.00}$ & $\mathbf{0.00 \pm 0.00}$ & $\mathbf{0.00 \pm 0.00}$ \\ 
NealPrior & $\mathbf{0.00 \pm 0.00}$ & 0.01 $\pm$ 0.00 & $\mathbf{0.00 \pm 0.00}$ & $\mathbf{0.00 \pm 0.00}$ \\ 
ScalePrior& $\mathbf{0.00 \pm 0.00}$ & 0.01 $\pm$ 0.00 & $\mathbf{0.00 \pm 0.00}$ & $\mathbf{0.00 \pm 0.00}$ \\ 
\midrule 
power - $\mathcal{N}(0, 1)$ & 4.00 $\pm$ 0.03 & 4.06 $\pm$ 0.03 & 3.96 $\pm$ 0.04 & $\mathbf{3.93 \pm 0.04}$ \\ 
NealPrior & 4.03 $\pm$ 0.04 & 4.13 $\pm$ 0.03 & 4.06 $\pm$ 0.03 & $\mathbf{4.00 \pm 0.04}$ \\ 
ScalePrior& 4.06 $\pm$ 0.04 & 4.12 $\pm$ 0.04 & 4.05 $\pm$ 0.04 & $\mathbf{4.01 \pm 0.04}$ \\ 
\midrule 
protein - $\mathcal{N}(0, 1)$ & 4.12 $\pm$ 0.02 & 4.54 $\pm$ 0.02 & 4.14 $\pm$ 0.02 & $\mathbf{4.04 \pm 0.01}$ \\ 
NealPrior & 4.21 $\pm$ 0.01 & 4.50 $\pm$ 0.03 & 4.27 $\pm$ 0.02 & $\mathbf{4.06 \pm 0.01}$ \\ 
ScalePrior& 4.22 $\pm$ 0.02 & 4.46 $\pm$ 0.01 & 4.19 $\pm$ 0.02 & $\mathbf{4.00 \pm 0.02}$ \\ 
\midrule 
wine - $\mathcal{N}(0, 1)$ & 0.65 $\pm$ 0.01 & 0.65 $\pm$ 0.01 & $\mathbf{0.63 \pm 0.01}$ & $\mathbf{0.63 \pm 0.01}$ \\ 
NealPrior & 0.66 $\pm$ 0.01 & 0.65 $\pm$ 0.01 & $\mathbf{0.64 \pm 0.01}$ & $\mathbf{0.64 \pm 0.01}$ \\ 
ScalePrior& 0.82 $\pm$ 0.01 & 0.65 $\pm$ 0.01 & $\mathbf{0.64 \pm 0.01}$ & $\mathbf{0.64 \pm 0.01}$ \\ 
\midrule 
yacht - $\mathcal{N}(0, 1)$ & 0.98 $\pm$ 0.07 & 2.35 $\pm$ 0.13 & 0.56 $\pm$ 0.04 & $\mathbf{0.50 \pm 0.04}$ \\ 
NealPrior & 1.15 $\pm$ 0.07 & 1.37 $\pm$ 0.11 & $\mathbf{0.57 \pm 0.04}$ & 0.63 $\pm$ 0.05 \\ 
ScalePrior& 14.55 $\pm$ 0.59 & 5.75 $\pm$ 1.34 & $\mathbf{0.56 \pm 0.04}$ & 0.63 $\pm$ 0.04 \\ 
\bottomrule 
  \end{tabular}
\end{table}

\begin{table}[ht]
\small
  \caption{ELBOs per datapoint in nats for BNNs on UCI datasets (errors are $\pm$ 1 standard error)}
  \label{tab:gi:uci_elbos}
  \centering
  \begin{tabular}{rccccccccc}
\toprule 
& factorised & local inducing & global inducing & fac $\rightarrow$ global \\ 
\midrule 
boston - $\mathcal{N}(0, 1)$ & -1.55 $\pm$ 0.00 & -1.54 $\pm$ 0.00 & $\mathbf{-1.02 \pm 0.01}$ & -1.03 $\pm$ 0.00 \\ 
NealPrior & -1.03 $\pm$ 0.00 & -0.99 $\pm$ 0.00 & $\mathbf{-0.63 \pm 0.00}$ & -0.70 $\pm$ 0.00 \\ 
ScalePrior& -1.54 $\pm$ 0.00 & -0.96 $\pm$ 0.00 & $\mathbf{-0.59 \pm 0.00}$ & -0.70 $\pm$ 0.00 \\ 
\midrule 
concrete - $\mathcal{N}(0, 1)$ & -1.10 $\pm$ 0.00 & -1.08 $\pm$ 0.00 & $\mathbf{-0.71 \pm 0.00}$ & -0.78 $\pm$ 0.00 \\ 
NealPrior & -0.88 $\pm$ 0.00 & -0.87 $\pm$ 0.00 & $\mathbf{-0.59 \pm 0.00}$ & -0.65 $\pm$ 0.00 \\ 
ScalePrior& -1.45 $\pm$ 0.01 & -0.88 $\pm$ 0.00 & $\mathbf{-0.57 \pm 0.00}$ & -0.63 $\pm$ 0.00 \\ 
\midrule 
energy - $\mathcal{N}(0, 1)$ & -0.13 $\pm$ 0.02 & -0.53 $\pm$ 0.01 & $\mathbf{0.72 \pm 0.00}$ & 0.59 $\pm$ 0.00 \\ 
NealPrior & 0.21 $\pm$ 0.00 & -0.33 $\pm$ 0.04 & $\mathbf{0.95 \pm 0.00}$ & 0.79 $\pm$ 0.01 \\ 
ScalePrior& -1.12 $\pm$ 0.00 & -0.47 $\pm$ 0.01 & $\mathbf{0.96 \pm 0.01}$ & 0.80 $\pm$ 0.01 \\ 
\midrule 
kin8nm - $\mathcal{N}(0, 1)$ & -0.38 $\pm$ 0.00 & -0.43 $\pm$ 0.00 & $\mathbf{-0.26 \pm 0.00}$ & -0.31 $\pm$ 0.00 \\ 
NealPrior & -0.35 $\pm$ 0.00 & -0.43 $\pm$ 0.00 & -0.31 $\pm$ 0.00 & $\mathbf{-0.28 \pm 0.00}$ \\ 
ScalePrior& -0.51 $\pm$ 0.00 & -0.39 $\pm$ 0.01 & -0.29 $\pm$ 0.00 & $\mathbf{-0.27 \pm 0.00}$ \\ 
\midrule 
naval - $\mathcal{N}(0, 1)$ & 1.68 $\pm$ 0.01 & 1.65 $\pm$ 0.09 & $\mathbf{2.89 \pm 0.04}$ & 2.11 $\pm$ 0.02 \\ 
NealPrior & 2.02 $\pm$ 0.02 & -0.09 $\pm$ 0.34 & 2.53 $\pm$ 0.04 & $\mathbf{2.62 \pm 0.02}$ \\ 
ScalePrior& 1.91 $\pm$ 0.03 & -1.42 $\pm$ 0.00 & 2.30 $\pm$ 0.03 & $\mathbf{2.71 \pm 0.02}$ \\ 
\midrule 
power - $\mathcal{N}(0, 1)$ & -0.08 $\pm$ 0.00 & -0.06 $\pm$ 0.00 & $\mathbf{-0.02 \pm 0.00}$ & -0.03 $\pm$ 0.00 \\ 
NealPrior & -0.05 $\pm$ 0.00 & -0.05 $\pm$ 0.00 & $\mathbf{-0.01 \pm 0.00}$ & $\mathbf{-0.01 \pm 0.00}$ \\ 
ScalePrior& -0.13 $\pm$ 0.00 & -0.05 $\pm$ 0.00 & $\mathbf{-0.01 \pm 0.00}$ & $\mathbf{-0.01 \pm 0.00}$ \\ 
\midrule 
protein - $\mathcal{N}(0, 1)$ & -1.09 $\pm$ 0.00 & -1.14 $\pm$ 0.00 & $\mathbf{-1.06 \pm 0.01}$ & -1.09 $\pm$ 0.00 \\ 
NealPrior & -1.11 $\pm$ 0.00 & -1.13 $\pm$ 0.00 & $\mathbf{-1.09 \pm 0.00}$ & $\mathbf{-1.09 \pm 0.00}$ \\ 
ScalePrior& -1.13 $\pm$ 0.00 & -1.12 $\pm$ 0.00 & $\mathbf{-1.07 \pm 0.00}$ & $\mathbf{-1.07 \pm 0.00}$ \\ 
\midrule 
wine - $\mathcal{N}(0, 1)$ & -1.48 $\pm$ 0.00 & -1.47 $\pm$ 0.00 & $\mathbf{-1.36 \pm 0.00}$ & $\mathbf{-1.36 \pm 0.00}$ \\ 
NealPrior & -1.31 $\pm$ 0.00 & -1.30 $\pm$ 0.00 & $\mathbf{-1.22 \pm 0.00}$ & -1.23 $\pm$ 0.00 \\ 
ScalePrior& -1.46 $\pm$ 0.00 & -1.29 $\pm$ 0.00 & $\mathbf{-1.22 \pm 0.00}$ & -1.23 $\pm$ 0.00 \\ 
\midrule 
yacht - $\mathcal{N}(0, 1)$ & -1.04 $\pm$ 0.02 & -1.30 $\pm$ 0.02 & $\mathbf{0.08 \pm 0.01}$ & -0.23 $\pm$ 0.01 \\ 
NealPrior & -0.46 $\pm$ 0.02 & -0.39 $\pm$ 0.01 & $\mathbf{0.74 \pm 0.01}$ & 0.31 $\pm$ 0.01 \\ 
ScalePrior& -1.61 $\pm$ 0.00 & -0.77 $\pm$ 0.10 & $\mathbf{0.79 \pm 0.01}$ & 0.30 $\pm$ 0.01 \\ 
\bottomrule 
  \end{tabular}
\end{table}

\begin{table}[ht]
\small
  \caption{Average test log likelihoods for our rerun of \citet{salimbeni2017doubly}, and our implementations of local and global inducing points for deep GPs of various depths.}
  \label{tab:gi:uci_dsvi2_lls}
  \centering
  \begin{tabular}{rccccccc}
    \toprule
    \{dataset\} - \{depth\} & DSVI & local inducing & global inducing \\
\midrule 
boston - 2& -2.50 $\pm$ 0.05 & $\mathbf{-2.42 \pm 0.05}$ & $\mathbf{-2.42 \pm 0.05}$ \\ 
3& -2.51 $\pm$ 0.05 & -2.43 $\pm$ 0.06 & $\mathbf{-2.40 \pm 0.05}$ \\ 
4& -2.51 $\pm$ 0.05 & -2.41 $\pm$ 0.04 & $\mathbf{-2.40 \pm 0.05}$  \\ 
5& -2.51 $\pm$ 0.05 & -2.41 $\pm$ 0.04 & $\mathbf{-2.36 \pm 0.05}$  \\ 
\midrule 
concrete - 2& -3.11 $\pm$ 0.01 & -3.08 $\pm$ 0.02 & $\mathbf{-3.06 \pm 0.02}$ \\ 
3& -3.11 $\pm$ 0.01 & -3.10 $\pm$ 0.02 & $\mathbf{-3.06 \pm 0.02}$ \\ 
4& -3.11 $\pm$ 0.01 & -3.12 $\pm$ 0.02 & $\mathbf{-3.05 \pm 0.02}$ \\ 
5& -3.11 $\pm$ 0.01 & -3.13 $\pm$ 0.02 & $\mathbf{-3.06 \pm 0.02}$ \\ 
\midrule 
energy - 2& -0.73 $\pm$ 0.02 & -0.71 $\pm$ 0.03 & $\mathbf{-0.70 \pm 0.03}$ \\ 
3& -0.76 $\pm$ 0.02 & -0.71 $\pm$ 0.03 & $\mathbf{-0.70 \pm 0.03}$ \\ 
4& -0.75 $\pm$ 0.02 & -0.71 $\pm$ 0.03 & $\mathbf{-0.70 \pm 0.03}$ \\ 
5& -0.75 $\pm$ 0.02 & -0.71 $\pm$ 0.03 & $\mathbf{-0.70 \pm 0.03}$ \\ 
\midrule 
kin8nm - 2& 1.34 $\pm$ 0.00 & $\mathbf{1.36 \pm 0.01}$ & 1.35 $\pm$ 0.00 \\ 
3& 1.36 $\pm$ 0.00 & $\mathbf{1.38 \pm 0.00}$ & 1.36 $\pm$ 0.00 \\ 
4& 1.35 $\pm$ 0.00 & $\mathbf{1.39 \pm 0.01}$ & 1.37 $\pm$ 0.01 \\ 
5& 1.35 $\pm$ 0.00 & $\mathbf{1.38 \pm 0.01}$ & 1.37 $\pm$ 0.00 \\ 
\midrule 
naval - 2& 6.77 $\pm$ 0.07 & 7.59 $\pm$ 0.04 & $\mathbf{8.24 \pm 0.07}$ \\ 
3& 6.61 $\pm$ 0.07 & 7.54 $\pm$ 0.04 & $\mathbf{7.91 \pm 0.10}$ \\ 
4& 6.54 $\pm$ 0.14 & 7.54 $\pm$ 0.06 & $\mathbf{8.28 \pm 0.05}$ \\ 
5& 5.02 $\pm$ 0.41 & 7.51 $\pm$ 0.06 & $\mathbf{8.24 \pm 0.04}$ \\ 
\midrule 
power - 2& -2.78 $\pm$ 0.01 & $\mathbf{-2.76 \pm 0.01}$ & $\mathbf{-2.76 \pm 0.01}$ \\ 
3& $\mathbf{-2.76 \pm 0.01}$ & -2.77 $\pm$ 0.01 & -2.77 $\pm$ 0.01 \\ 
4& $\mathbf{-2.75 \pm 0.01}$ & -2.77 $\pm$ 0.01 & -2.77 $\pm$ 0.01 \\ 
5& $\mathbf{-2.75 \pm 0.01}$ & -2.77 $\pm$ 0.01 & -2.77 $\pm$ 0.01 \\ 
\midrule 
protein - 2& $\mathbf{-2.80 \pm 0.00}$ & -2.82 $\pm$ 0.00 & -2.83 $\pm$ 0.00 \\ 
3& -2.73 $\pm$ 0.00 & $\mathbf{-2.71 \pm 0.01}$ & -2.76 $\pm$ 0.01 \\ 
4& $\mathbf{-2.71 \pm 0.01}$ & -2.71 $\pm$ 0.01 & -2.73 $\pm$ 0.01 \\ 
5& $\mathbf{-2.70 \pm 0.01}$ & -2.71 $\pm$ 0.01 & -2.74 $\pm$ 0.01 \\ 
\midrule 
wine - 2& $\mathbf{-0.95 \pm 0.01}$ & -0.96 $\pm$ 0.01 & -0.96 $\pm$ 0.01 \\ 
3& $\mathbf{-0.95 \pm 0.01}$ & -0.96 $\pm$ 0.01 & -0.96 $\pm$ 0.01 \\ 
4& $\mathbf{-0.95 \pm 0.01}$ & -0.96 $\pm$ 0.01 & -0.96 $\pm$ 0.01 \\ 
5& $\mathbf{-0.95 \pm 0.01}$ & -0.96 $\pm$ 0.01 & -0.96 $\pm$ 0.01 \\ 
\midrule 
yacht - 2& -0.40 $\pm$ 0.03 & $\mathbf{0.05 \pm 0.14}$ & -0.38 $\pm$ 0.10 \\ 
3& -0.47 $\pm$ 0.02 & $\mathbf{0.17 \pm 0.11}$ & -0.41 $\pm$ 0.13 \\ 
4& -0.50 $\pm$ 0.02 & $\mathbf{-0.31 \pm 0.29}$ & -0.47 $\pm$ 0.19 \\ 
5& -0.50 $\pm$ 0.02 & $\mathbf{-0.31 \pm 0.20}$ & -0.48 $\pm$ 0.19 \\ 
    \bottomrule
  \end{tabular}
\end{table}

\begin{table}[ht]
\small
  \caption{Test RMSEs for our rerun of \citet{salimbeni2017doubly}, and our implementations of local and global inducing points for deep GPs of various depths.}
  \label{tab:gi:uci_dsvi2_rmse}
  \centering
  \begin{tabular}{rccccccc}
    \toprule
\{dataset\} - \{depth\} & DSVI & local inducing & global inducing \\
\midrule 
boston - 2& 2.95 $\pm$ 0.18 & $\mathbf{2.78 \pm 0.15}$ & 2.82 $\pm$ 0.14 \\ 
3& 2.98 $\pm$ 0.18 & $\mathbf{2.78 \pm 0.14}$ & 2.79 $\pm$ 0.14 \\ 
4& 2.99 $\pm$ 0.18 & $\mathbf{2.75 \pm 0.12}$ & 2.80 $\pm$ 0.15 \\ 
5& 3.01 $\pm$ 0.19 & 2.78 $\pm$ 0.13 & $\mathbf{2.73 \pm 0.13}$ \\ 
\midrule 
concrete - 2& 5.51 $\pm$ 0.10 & 5.24 $\pm$ 0.11 & $\mathbf{5.21 \pm 0.12}$ \\ 
3& 5.53 $\pm$ 0.10 & 5.38 $\pm$ 0.11 & $\mathbf{5.18 \pm 0.12}$ \\ 
4& 5.50 $\pm$ 0.09 & 5.47 $\pm$ 0.11 & $\mathbf{5.16 \pm 0.13}$ \\ 
5& 5.53 $\pm$ 0.11 & 5.50 $\pm$ 0.10 & $\mathbf{5.23 \pm 0.13}$ \\ 
\midrule 
energy - 2& 0.50 $\pm$ 0.01 & 0.49 $\pm$ 0.01 & $\mathbf{0.48 \pm 0.01}$ \\ 
3& 0.50 $\pm$ 0.01 & 0.49 $\pm$ 0.01 & $\mathbf{0.48 \pm 0.01}$ \\ 
4& 0.50 $\pm$ 0.01 & 0.49 $\pm$ 0.01 & $\mathbf{0.48 \pm 0.01}$ \\ 
5& 0.50 $\pm$ 0.01 & 0.49 $\pm$ 0.01 & $\mathbf{0.48 \pm 0.01}$ \\ 
\midrule 
kin8nm - 2& $\mathbf{0.06 \pm 0.00}$ & $\mathbf{0.06 \pm 0.00}$ & $\mathbf{0.06 \pm 0.00}$ \\ 
3& $\mathbf{0.06 \pm 0.00}$ & $\mathbf{0.06 \pm 0.00}$ & $\mathbf{0.06 \pm 0.00}$ \\ 
4& $\mathbf{0.06 \pm 0.00}$ & $\mathbf{0.06 \pm 0.00}$ & $\mathbf{0.06 \pm 0.00}$ \\ 
5& $\mathbf{0.06 \pm 0.00}$ & $\mathbf{0.06 \pm 0.00}$ & $\mathbf{0.06 \pm 0.00}$ \\ 
\midrule 
naval - 2& $\mathbf{0.00 \pm 0.00}$ & $\mathbf{0.00 \pm 0.00}$ & $\mathbf{0.00 \pm 0.00}$ \\ 
3& $\mathbf{0.00 \pm 0.00}$ & $\mathbf{0.00 \pm 0.00}$ & $\mathbf{0.00 \pm 0.00}$ \\ 
4& $\mathbf{0.00 \pm 0.00}$ & $\mathbf{0.00 \pm 0.00}$ & $\mathbf{0.00 \pm 0.00}$ \\ 
5& 0.01 $\pm$ 0.00 & $\mathbf{0.00 \pm 0.00}$ & $\mathbf{0.00 \pm 0.00}$ \\ 
\midrule 
power - 2& 3.88 $\pm$ 0.03 & 3.82 $\pm$ 0.04 & $\mathbf{3.81 \pm 0.04}$ \\ 
3& $\mathbf{3.80 \pm 0.04}$ & 3.85 $\pm$ 0.04 & 3.87 $\pm$ 0.04 \\ 
4& $\mathbf{3.78 \pm 0.04}$ & 3.84 $\pm$ 0.04 & 3.84 $\pm$ 0.04 \\ 
5& $\mathbf{3.78 \pm 0.04}$ & 3.83 $\pm$ 0.04 & 3.84 $\pm$ 0.04 \\ 
\midrule 
protein - 2& $\mathbf{4.01 \pm 0.01}$ & 4.07 $\pm$ 0.02 & 4.11 $\pm$ 0.01 \\ 
3& 3.75 $\pm$ 0.01 & $\mathbf{3.73 \pm 0.03}$ & 3.88 $\pm$ 0.03 \\ 
4& $\mathbf{3.73 \pm 0.01}$ & $\mathbf{3.73 \pm 0.03}$ & 3.77 $\pm$ 0.03 \\ 
5& $\mathbf{3.70 \pm 0.02}$ & 3.73 $\pm$ 0.03 & 3.80 $\pm$ 0.03 \\ 
\midrule 
wine - 2& $\mathbf{0.06 \pm 0.00}$ & $\mathbf{0.06 \pm 0.00}$ & $\mathbf{0.06 \pm 0.00}$ \\ 
3& $\mathbf{0.06 \pm 0.00}$ & $\mathbf{0.06 \pm 0.00}$ & $\mathbf{0.06 \pm 0.00}$ \\ 
4& $\mathbf{0.06 \pm 0.00}$ & $\mathbf{0.06 \pm 0.00}$ & $\mathbf{0.06 \pm 0.00}$ \\ 
5& $\mathbf{0.06 \pm 0.00}$ & $\mathbf{0.06 \pm 0.00}$ & $\mathbf{0.06 \pm 0.00}$ \\ 
\midrule 
yacht - 2& 0.40 $\pm$ 0.03 & $\mathbf{0.36 \pm 0.03}$ & 0.41 $\pm$ 0.03 \\ 
3& 0.42 $\pm$ 0.03 & 0.37 $\pm$ 0.03 & $\mathbf{0.36 \pm 0.03}$ \\ 
4& 0.44 $\pm$ 0.03 & 0.37 $\pm$ 0.03 & $\mathbf{0.36 \pm 0.03}$ \\ 
5& 0.44 $\pm$ 0.03 & 0.40 $\pm$ 0.03 & $\mathbf{0.35 \pm 0.03}$ \\ 
    \bottomrule
  \end{tabular}
\end{table}

\begin{table}[ht]
\small
  \caption{ELBOs per datapoint for our rerun of \citet{salimbeni2017doubly}, and our implementations of local and global inducing points for deep GPs of various depths.}
  \label{tab:gi:uci_dsvi2_elbos}
  \centering
  \begin{tabular}{rccccccc}
    \toprule
\{dataset\} - \{depth\} & DSVI & local inducing & global inducing \\
\midrule 
boston - 2& -0.52 $\pm$ 0.04 & -0.35 $\pm$ 0.00 & $\mathbf{-0.28 \pm 0.01}$ \\ 
3& -0.54 $\pm$ 0.04 & -0.35 $\pm$ 0.01 & $\mathbf{-0.25 \pm 0.01}$ \\ 
4& -0.55 $\pm$ 0.04 & -0.36 $\pm$ 0.00 & $\mathbf{-0.25 \pm 0.01}$ \\ 
5& -0.58 $\pm$ 0.04 & -0.35 $\pm$ 0.01 & $\mathbf{-0.24 \pm 0.01}$ \\ 
\midrule 
concrete - 2& -0.61 $\pm$ 0.02 & -0.41 $\pm$ 0.00 & $\mathbf{-0.36 \pm 0.00}$ \\ 
3& -0.63 $\pm$ 0.01 & -0.42 $\pm$ 0.00 & $\mathbf{-0.34 \pm 0.00}$ \\ 
4& -0.64 $\pm$ 0.01 & -0.42 $\pm$ 0.00 & $\mathbf{-0.33 \pm 0.00}$ \\ 
5& -0.64 $\pm$ 0.01 & -0.42 $\pm$ 0.00 & $\mathbf{-0.33 \pm 0.00}$ \\ 
\midrule 
energy - 2& 0.93 $\pm$ 0.02 & $\mathbf{1.48 \pm 0.00}$ & $\mathbf{1.48 \pm 0.00}$ \\ 
3& 0.84 $\pm$ 0.02 & 1.47 $\pm$ 0.00 & $\mathbf{1.48 \pm 0.00}$ \\ 
4& 0.87 $\pm$ 0.01 & 1.47 $\pm$ 0.00 & $\mathbf{1.48 \pm 0.00}$ \\ 
5& 0.86 $\pm$ 0.01 & 1.47 $\pm$ 0.00 & $\mathbf{1.48 \pm 0.00}$ \\ 
\midrule 
kin8nm - 2& -0.18 $\pm$ 0.01 & -0.11 $\pm$ 0.00 & $\mathbf{-0.10 \pm 0.00}$ \\ 
3& -0.19 $\pm$ 0.01 & $\mathbf{-0.09 \pm 0.00}$ & $\mathbf{-0.09 \pm 0.00}$ \\ 
4& -0.19 $\pm$ 0.01 & $\mathbf{-0.08 \pm 0.00}$ & $\mathbf{-0.08 \pm 0.00}$ \\ 
5& -0.19 $\pm$ 0.01 & -0.09 $\pm$ 0.00 & $\mathbf{-0.08 \pm 0.00}$ \\ 
\midrule 
naval - 2& 2.29 $\pm$ 0.08 & 3.01 $\pm$ 0.05 & $\mathbf{3.89 \pm 0.07}$ \\ 
3& 2.07 $\pm$ 0.10 & 2.93 $\pm$ 0.05 & $\mathbf{3.55 \pm 0.12}$ \\ 
4& 1.90 $\pm$ 0.25 & 2.92 $\pm$ 0.07 & $\mathbf{3.93 \pm 0.05}$ \\ 
5& 0.61 $\pm$ 0.37 & 2.99 $\pm$ 0.06 & $\mathbf{3.93 \pm 0.04}$ \\ 
\midrule 
power - 2& 0.02 $\pm$ 0.00 & 0.04 $\pm$ 0.00 & $\mathbf{0.05 \pm 0.00}$ \\ 
3& 0.02 $\pm$ 0.00 & $\mathbf{0.04 \pm 0.00}$ & $\mathbf{0.04 \pm 0.00}$ \\ 
4& 0.03 $\pm$ 0.00 & 0.04 $\pm$ 0.00 & $\mathbf{0.05 \pm 0.00}$ \\ 
5& 0.02 $\pm$ 0.00 & $\mathbf{0.04 \pm 0.00}$ & $\mathbf{0.04 \pm 0.00}$ \\ 
\midrule 
protein - 2& -1.06 $\pm$ 0.00 & $\mathbf{-1.05 \pm 0.00}$ & -1.06 $\pm$ 0.00 \\ 
3& -1.02 $\pm$ 0.00 & $\mathbf{-0.98 \pm 0.00}$ & -1.01 $\pm$ 0.00 \\ 
4& -1.01 $\pm$ 0.00 & $\mathbf{-0.97 \pm 0.00}$ & -0.98 $\pm$ 0.00 \\ 
5& -1.01 $\pm$ 0.00 & $\mathbf{-0.97 \pm 0.00}$ & $\mathbf{-0.97 \pm 0.00}$ \\ 
\midrule 
wine - 2& -1.18 $\pm$ 0.02 & $\mathbf{-1.17 \pm 0.00}$ & $\mathbf{-1.17 \pm 0.00}$ \\ 
3& -1.18 $\pm$ 0.02 & $\mathbf{-1.17 \pm 0.00}$ & $\mathbf{-1.17 \pm 0.00}$ \\ 
4& -1.18 $\pm$ 0.02 & $\mathbf{-1.17 \pm 0.00}$ & $\mathbf{-1.17 \pm 0.00}$ \\ 
5& -1.18 $\pm$ 0.02 & -1.18 $\pm$ 0.00 & $\mathbf{-1.17 \pm 0.00}$ \\ 
\midrule 
yacht - 2& 1.05 $\pm$ 0.06 & $\mathbf{2.53 \pm 0.01}$ & 2.12 $\pm$ 0.05 \\ 
3& 1.00 $\pm$ 0.06 & $\mathbf{2.54 \pm 0.01}$ & 2.26 $\pm$ 0.01 \\ 
4& 0.97 $\pm$ 0.06 & $\mathbf{2.46 \pm 0.02}$ & 2.26 $\pm$ 0.01 \\ 
5& 0.95 $\pm$ 0.06 & $\mathbf{2.43 \pm 0.01}$ & 2.25 $\pm$ 0.01 \\ 
    \bottomrule
  \end{tabular}
\end{table}
\chapter{Appendix to Chapter 5}

\section{Choice of $\tF_\textrm{i}$}
\label{sec:dwp:app:F}

Here, we establish that the distribution over $\tF_\text{t} \transpose{\tF}_\text{i}$ and $\tF_\text{t} \transpose{\tF}_\text{t}$ resulting from conditioning according to Eq.~\ref{eq:dwp:Ft|Fi} does not depend on the choice of $\tF_\text{i}$, as long as $\tF_\text{i}\transpose{\tF}_\text{i} = \G_\text{ii}$.\footnote{In this section we omit labeling with $\ell$ for clarity.}
With this conditioning rule, and by the definition of the matrix-variate Gaussian, we can write
\begin{align*}
  \tF_\text{t} &= \S_\text{ti} \S_\text{ii}^{-1} \tF_\text{i} + \S_{\text{tt}\cdot \text{i}}^{1/2} \mathbf{\Xi},
\end{align*}
where $\mathbf{\Xi}$ is a matrix with IID standard Gaussian elements.
Thus,
\begin{align*}
  \tF_\text{t} \transpose{\tF}_\text{i} &= \S_\text{ti} \S_\text{ii}^{-1} \tF_\text{i} \transpose{\tF}_\text{i} + \S_{\text{tt}\cdot \text{i}}^{1/2} \mathbf{\Xi} \transpose{\tF}_\text{i},\\
  \tF_\text{t} \transpose{\tF}_\text{i} &\sim \MN{\S_\text{ti} \S_\text{ii}^{-1} \G_\text{ii}, \S_{\text{tt}\cdot \text{i}}, \G_\text{ii}},
\end{align*}
where the latter line again comes from the definition of the matrix-variate Gaussian.
We can do the same for $\tF_\text{t} \transpose{\tF}_\text{t}$:
\begin{align*}
  \tF_\text{t} \transpose{\tF}_\text{t} &= 
  \S_\text{ti} \S_\text{ii}^{-1} \tF_\text{i} \transpose{\tF}_\text{i} \S_\text{ii}^{-1} \transpose{\S}_\text{ti}
  +\S_\text{ti} \S_\text{ii}^{-1} \tF_\text{i} \transpose{\mathbf{\Xi}} \S_{\text{tt}\cdot \text{i}}^{1/2}\\
  &+\S_{\text{tt}\cdot \text{i}}^{1/2} \mathbf{\Xi} \transpose{\tF}_\text{i} \S_\text{ii}^{-1} \transpose{\S}_\text{ti}
  +\S_{\text{tt}\cdot \text{i}}^{1/2} \mathbf{\Xi} \transpose{\mathbf{\Xi}} \S_{\text{tt}\cdot \text{i}}^{1/2}.
\end{align*}
The first term is independent of the choice of of $\tF_\text{i}$ because $\G_\text{ii} = \tF_\text{i} \transpose{\tF}_\text{i}$.
Meanwhile, the final term does not depend on $\tF_\text{i}$ at all.
Finally, the two terms in the middle are each other's transposes.
Considering only the first, we observe that it is Gaussian with a covariance that depends on $\G_\text{ii}$, but not on the specific choice of $\tF_\text{i}$:
\begin{align*}
  \S_\text{ti} \S_\text{ii}^{-1} \tF_\text{i} \transpose{\mathbf{\Xi}} \S_{\text{tt}\cdot \text{i}}^{1/2} \sim \MN{\0, \S_\text{ti} \S_\text{ii}^{-1} \G_\text{ii} \S_\text{ii}^{-1} \transpose{\S}_\text{ti},  \S_{\text{tt}\cdot \text{i}}},
\end{align*}
Thus, $\G_\text{ti} = \tF_\text{t} \transpose{\tF}_\text{i}$ and $\G_\text{tt} = \tF_\text{t} \transpose{\F}_\text{t}$ depend (in distribution) on $\G_\text{ii}$ but not on the specific choice of $\tF_\text{i}$.
Therefore, we are free to use any $\tF_\text{i}$ as long as $\G_\text{ii} = \tF_\text{i} \transpose{\tF}_\text{i}$.

\section{Experimental details}
\label{app:dwp:exp-details}

\paragraph{Datasets} All experiments were performed using the UCI splits from \citet{Gal2015DropoutB}, available at \url{https://github.com/yaringal/DropoutUncertaintyExps/tree/master/UCI_Datasets}. For each dataset there are twenty splits, with the exception of \textsc{Protein}, which only has five. We report mean plus or minus one standard error over the splits.

\paragraph{Model details} As standard, we set $\nu$ (the `width' of each layer) to be equal to the dimensionality of the input space. We use the squared exponential kernel, with automatic relevance determination (ARD) in the first layer, but without for the intermediate layers as ARD relies on explicit features existing. However, we found in practice that using ARD for intermediate layers in a DGP did not hugely affect the results, as each output GP in a layer shares the same prior and hence output prior variance. For the final GP layer of the DWP model we use a global inducing approximate posterior \citep{ober2020global}, as done for the entirety of the DGP. We leave the particular implementation details for the code, but we note that we use the `sticking the landing' gradient estimator \citep{roeder2017sticking} for the $\sl{\boldsymbol{\alpha}_\ell, \boldsymbol{\beta}_\ell, \boldsymbol{\mu}_\ell, \boldsymbol{\sigma}_\ell}$ approximate posterior parameters of the DWP (using it for the other parameters, as well as for the DGP parameters, is difficult as the parameters of one layer will affect the KL estimate of the following layers).

\paragraph{Training details} We train all models using the same training scheme. We use 20,000 gradient steps to train each model with the Adam optimizer \citep{kingma2014adam} with an initial learning rate of 1e-2. We anneal the KL using a factor increasing linearly from 0 to 1 over the first 1,000 gradient steps, and step the learning rate down to 1e-3 after 10,000 gradient steps. We use 10 samples from the approximate posterior for training, and 100 for testing. Experiments were performed using an internal cluster of machines with NVIDIA GeForce 2080 Ti GPUs, although we used CPU (Intel Core i9-10900X) for the smaller datasets (\textsc{Boston}, \textsc{Concrete}, \textsc{Energy}, \textsc{Wine}, \textsc{Yacht}).



\newpage

\section{Tables}
We provide tables of ELBOs, test log likelihoods, and root mean square error for all our experiments.
We highlight the best result for each configuration across all methods in bold, while italicizing the best result between the DGP and DWP (with the simpler posterior).
For the ELBOs, the highlighted results take the error bars into account, while we did not do this for the test metrics.
This is because we found that most of the variation in the test metrics across splits was due to the splitting procedure, rather than from the method's actual performance, meaning that one method would perform consistently better across all (or nearly all) splits.

In Table~\ref{tab:dwp:uci_lls}, we include a comparison to the reported results for the 3-layer DIWP with squared exponential kernel from \citet{aitchison2020deep}; they did not provide ELBOs or RMSEs. 
However, it should be noted that the specific implementation and architectural details differ significantly from those presented in this paper, and so these results are not directly comparable. 
Additionally, \citet{aitchison2020deep} bases its error bars on paired comparisons to the other methods instead of the standard error bars we use here; we therefore omit the error bars completely.

Finally, we provide a table of runtime per training epoch for each method on the $\textsc{Boston}$ and $\textsc{Protein}$ datasets in Table~\ref{tab:dwp:runtime}.

\label{sec:dwp:tables}
\begin{table}[ht]
\footnotesize
  \caption{ELBOs per datapoint. We report mean plus or minus one standard error over the splits. Bold numbers correspond to the best over all models, whereas the italicized models only compare DGP and DWP.}
  \label{tab:dwp:uci_elbos}
  \centering
  \begin{tabular}{rcccc}
  \toprule
\{dataset\} - \{depth\} & DGP & DWP & DWP-A & DWP-AB \\ 
\midrule 
\textsc{Boston} - 2 & -0.38 $\pm$ 0.01 & \textit{-0.33 $\pm$ 0.00} & \textbf{-0.32 $\pm$ 0.01} & \textbf{-0.32 $\pm$ 0.00} \\ 
3 & -0.40 $\pm$ 0.00 & \textit{-0.34 $\pm$ 0.01} & \textbf{-0.33 $\pm$ 0.00} & \textbf{-0.33 $\pm$ 0.01} \\ 
4 & -0.43 $\pm$ 0.00 & \textit{\textbf{-0.35 $\pm$ 0.00}} & \textbf{-0.34 $\pm$ 0.01} & \textbf{-0.34 $\pm$ 0.01} \\ 
5 & -0.45 $\pm$ 0.00 & \textit{\textbf{-0.37 $\pm$ 0.01}} & \textbf{-0.36 $\pm$ 0.00} & \textbf{-0.36 $\pm$ 0.00} \\ 
\midrule 
\textsc{Concrete} - 2 & -0.45 $\pm$ 0.00 & \textit{-0.42 $\pm$ 0.00} & -0.40 $\pm$ 0.00 & \textbf{-0.39 $\pm$ 0.00} \\ 
3 & -0.47 $\pm$ 0.00 & \textit{-0.43 $\pm$ 0.00} & \textbf{-0.41 $\pm$ 0.00} & \textbf{-0.41 $\pm$ 0.00} \\ 
4 & -0.49 $\pm$ 0.00 & \textit{-0.46 $\pm$ 0.00} & \textbf{-0.43 $\pm$ 0.00} & \textbf{-0.43 $\pm$ 0.00} \\ 
5 & -0.50 $\pm$ 0.00 & \textit{-0.49 $\pm$ 0.00} & \textbf{-0.45 $\pm$ 0.00} & \textbf{-0.45 $\pm$ 0.00} \\ 
\midrule 
\textsc{Energy} - 2 & 1.43 $\pm$ 0.00 & \textit{\textbf{1.46 $\pm$ 0.00}} & \textbf{1.46 $\pm$ 0.00} & \textbf{1.46 $\pm$ 0.00} \\ 
3 & 1.42 $\pm$ 0.00 & \textit{1.44 $\pm$ 0.00} & \textbf{1.45 $\pm$ 0.00} & \textbf{1.45 $\pm$ 0.00} \\ 
4 & 1.40 $\pm$ 0.00 & \textit{1.42 $\pm$ 0.00} & \textbf{1.43 $\pm$ 0.00} & \textbf{1.43 $\pm$ 0.00} \\ 
5 & 1.38 $\pm$ 0.00 & \textit{1.40 $\pm$ 0.00} & \textbf{1.42 $\pm$ 0.00} & 1.41 $\pm$ 0.00 \\ 
\midrule 
\textsc{Kin8nm} - 2 & \textit{-0.15 $\pm$ 0.00} & -0.16 $\pm$ 0.00 & \textbf{-0.14 $\pm$ 0.00} & \textbf{-0.14 $\pm$ 0.00} \\ 
3 & \textit{-0.14 $\pm$ 0.00} & -0.15 $\pm$ 0.00 & \textbf{-0.13 $\pm$ 0.00} & \textbf{-0.13 $\pm$ 0.00} \\ 
4 & -0.14 $\pm$ 0.00 & -0.14 $\pm$ 0.00 & \textbf{-0.11 $\pm$ 0.00} & \textbf{-0.11 $\pm$ 0.00} \\ 
5 & -0.14 $\pm$ 0.00 & -0.14 $\pm$ 0.00 & \textbf{-0.11 $\pm$ 0.00} & \textbf{-0.11 $\pm$ 0.00} \\ 
\midrule 
\textsc{Naval} - 2 & 3.93 $\pm$ 0.05 & 3.82 $\pm$ 0.09 & 3.80 $\pm$ 0.13 & 3.84 $\pm$ 0.10 \\ 
3 & 3.83 $\pm$ 0.06 & 3.71 $\pm$ 0.12 & 3.86 $\pm$ 0.06 & \textbf{3.99 $\pm$ 0.04} \\ 
4 & \textit{\textbf{3.91 $\pm$ 0.05}} & 3.66 $\pm$ 0.13 & \textbf{3.75 $\pm$ 0.11} & \textbf{3.85 $\pm$ 0.09} \\ 
5 & \textit{\textbf{3.92 $\pm$ 0.04}} & 3.59 $\pm$ 0.12 & \textbf{3.97 $\pm$ 0.02} & 3.63 $\pm$ 0.22 \\ 
\midrule 
\textsc{Power} - 2 & 0.03 $\pm$ 0.00 & 0.03 $\pm$ 0.00 & \textbf{0.04 $\pm$ 0.00} & \textbf{0.04 $\pm$ 0.00} \\ 
3 & 0.03 $\pm$ 0.00 & 0.03 $\pm$ 0.00 & 0.03 $\pm$ 0.00 & 0.03 $\pm$ 0.00 \\ 
4 & 0.03 $\pm$ 0.00 & 0.03 $\pm$ 0.00 & 0.03 $\pm$ 0.00 & 0.03 $\pm$ 0.00 \\ 
5 & \textit{\textbf{0.03 $\pm$ 0.00}} & 0.02 $\pm$ 0.00 & \textbf{0.03 $\pm$ 0.00} & \textbf{0.03 $\pm$ 0.00} \\ 
\midrule 
\textsc{Protein} - 2 & \textit{\textbf{-1.06 $\pm$ 0.00}} & -1.07 $\pm$ 0.00 & \textbf{-1.06 $\pm$ 0.00} & \textbf{-1.06 $\pm$ 0.00} \\ 
3 & -1.04 $\pm$ 0.00 & -1.04 $\pm$ 0.00 & \textbf{-1.03 $\pm$ 0.00} & \textbf{-1.03 $\pm$ 0.00} \\ 
4 & -1.02 $\pm$ 0.00 & -1.02 $\pm$ 0.00 & \textbf{-1.00 $\pm$ 0.00} & -1.01 $\pm$ 0.00 \\ 
5 & \textit{\textbf{-1.00 $\pm$ 0.00}} & -1.01 $\pm$ 0.00 & \textbf{-1.00 $\pm$ 0.00} & \textbf{-1.00 $\pm$ 0.00} \\ 
\midrule 
\textsc{Wine} - 2 & -1.18 $\pm$ 0.00 & -1.18 $\pm$ 0.00 & \textbf{-1.18 $\pm$ 0.00} & \textbf{-1.18 $\pm$ 0.00} \\ 
3 & -1.19 $\pm$ 0.00 & \textit{\textbf{-1.18 $\pm$ 0.00}} & \textbf{-1.18 $\pm$ 0.00} & \textbf{-1.18 $\pm$ 0.00} \\ 
4 & -1.19 $\pm$ 0.00 & \textit{\textbf{-1.18 $\pm$ 0.00}} & \textbf{-1.18 $\pm$ 0.00} & \textbf{-1.18 $\pm$ 0.00} \\ 
5 & -1.19 $\pm$ 0.00 & -1.19 $\pm$ 0.00 & -1.19 $\pm$ 0.00 & -1.19 $\pm$ 0.00 \\ 
\midrule 
\textsc{Yacht} - 2 & 1.88 $\pm$ 0.03 & \textit{2.02 $\pm$ 0.01} & \textbf{2.07 $\pm$ 0.01} & \textbf{2.07 $\pm$ 0.01} \\ 
3 & 1.62 $\pm$ 0.01 & \textit{1.86 $\pm$ 0.02} & \textbf{2.02 $\pm$ 0.01} & \textbf{2.03 $\pm$ 0.01} \\ 
4 & 1.47 $\pm$ 0.02 & \textit{1.73 $\pm$ 0.02} & \textbf{1.93 $\pm$ 0.01} & 1.91 $\pm$ 0.01 \\ 
5 & 1.46 $\pm$ 0.02 & \textit{1.59 $\pm$ 0.02} & \textbf{1.79 $\pm$ 0.02} & \textbf{1.79 $\pm$ 0.02} \\ 
\bottomrule
  \end{tabular}
\end{table}

\begin{table}[ht]
\footnotesize
  \caption{Average test log likelihoods. We report mean plus or minus one standard error over the splits, along with quoted results for the DIWP model from \citet{aitchison2020deep}. We only directly compare between DWP and DGP models and do not quote error bars for the DIWP due to the differences noted above. Bold numbers correspond to the best over all models, whereas the italicized models only compare DGP and DWP.}
  \label{tab:dwp:uci_lls}
  \centering
  \begin{tabular}{rcccc|c}
  \toprule
\{dataset\} - \{depth\} & DGP & DWP & DWP-A & DWP-AB & DIWP \\ 
\midrule 
\textsc{Boston} - 2 & -2.43 $\pm$ 0.05 & \textit{-2.40 $\pm$ 0.05} & \textbf{-2.37 $\pm$ 0.05} & \textbf{-2.37 $\pm$ 0.05} & - \\ 
3 & -2.39 $\pm$ 0.04 & \textit{-2.38 $\pm$ 0.05} & \textbf{-2.35 $\pm$ 0.04} & \textbf{-2.35 $\pm$ 0.04} & -2.40 \\ 
4 & -2.41 $\pm$ 0.04 & \textit{-2.38 $\pm$ 0.04} & \textbf{-2.37 $\pm$ 0.04} & \textbf{-2.37 $\pm$ 0.04} & - \\ 
5 & -2.43 $\pm$ 0.04 & \textit{-2.38 $\pm$ 0.04} & -2.39 $\pm$ 0.05 & \textbf{-2.38 $\pm$ 0.04} & - \\ 
\midrule 
\textsc{Concrete} - 2 & \textit{-3.10 $\pm$ 0.02} & -3.12 $\pm$ 0.02 & \textbf{-3.08 $\pm$ 0.02} & \textbf{-3.08 $\pm$ 0.02} & - \\ 
3 & \textit{-3.08 $\pm$ 0.02} & -3.10 $\pm$ 0.02 & \textbf{-3.06 $\pm$ 0.02} & -3.07 $\pm$ 0.02 & -3.08 \\ 
4 & -3.13 $\pm$ 0.02 & \textit{-3.12 $\pm$ 0.02} & \textbf{-3.07 $\pm$ 0.02} & \textbf{-3.07 $\pm$ 0.02} & - \\ 
5 & -3.13 $\pm$ 0.02 & -3.13 $\pm$ 0.02 & \textbf{-3.07 $\pm$ 0.02} & -3.08 $\pm$ 0.02 & - \\ 
\midrule 
\textsc{Energy} - 2 & -0.70 $\pm$ 0.03 & -0.70 $\pm$ 0.03 & -0.70 $\pm$ 0.03 & -0.70 $\pm$ 0.03 & - \\ 
3 & -0.70 $\pm$ 0.03 & -0.70 $\pm$ 0.03 & -0.70 $\pm$ 0.03 & -0.70 $\pm$ 0.03 & -0.70 \\ 
4 & -0.70 $\pm$ 0.03 & -0.70 $\pm$ 0.03 & -0.70 $\pm$ 0.03 & -0.70 $\pm$ 0.03 & - \\ 
5 & -0.71 $\pm$ 0.03 & -0.71 $\pm$ 0.03 & \textbf{-0.70 $\pm$ 0.03} & \textbf{-0.70 $\pm$ 0.03} & - \\ 
\midrule 
\textsc{Kin8nm} - 2 & 1.35 $\pm$ 0.00 & 1.35 $\pm$ 0.00 & \textbf{1.36 $\pm$ 0.00} & \textbf{1.36 $\pm$ 0.00} & - \\ 
3 & 1.37 $\pm$ 0.00 & 1.37 $\pm$ 0.00 & \textbf{1.38 $\pm$ 0.00} & \textbf{1.38 $\pm$ 0.00} & 1.01 \\ 
4 & 1.38 $\pm$ 0.00 & \textit{1.39 $\pm$ 0.01} & \textbf{1.40 $\pm$ 0.00} & \textbf{1.40 $\pm$ 0.00} & - \\ 
5 & 1.38 $\pm$ 0.00 & \textit{1.40 $\pm$ 0.01} & \textbf{1.41 $\pm$ 0.01} & \textbf{1.41 $\pm$ 0.01} & - \\ 
\midrule 
\textsc{Naval} - 2 & \textit{\textbf{8.24 $\pm$ 0.06}} & 8.23 $\pm$ 0.08 & 8.18 $\pm$ 0.11 & 8.18 $\pm$ 0.13 & - \\ 
3 & 8.15 $\pm$ 0.06 & \textit{8.18 $\pm$ 0.07} & 8.27 $\pm$ 0.05 & \textbf{8.38 $\pm$ 0.03} & 5.92 \\ 
4 & \textit{8.28 $\pm$ 0.04} & 8.17 $\pm$ 0.11 & 8.14 $\pm$ 0.13 & \textbf{8.32 $\pm$ 0.06} & - \\ 
5 & \textit{8.28 $\pm$ 0.04} & 8.17 $\pm$ 0.07 & \textbf{8.40 $\pm$ 0.02} & 8.10 $\pm$ 0.19 & - \\ 
\midrule 
\textsc{Power} - 2 & -2.78 $\pm$ 0.01 & \textit{-2.77 $\pm$ 0.01} & \textbf{-2.76 $\pm$ 0.01} & \textbf{-2.76 $\pm$ 0.01} & - \\ 
3 & -2.77 $\pm$ 0.01 & \textit{\textbf{-2.76 $\pm$ 0.01}} & \textbf{-2.76 $\pm$ 0.01} & \textbf{-2.76 $\pm$ 0.01} & -2.78 \\ 
4 & -2.78 $\pm$ 0.01 & \textit{-2.77 $\pm$ 0.01} & \textbf{-2.75 $\pm$ 0.01} & \textbf{-2.75 $\pm$ 0.01} & - \\ 
5 & -2.78 $\pm$ 0.01 & \textit{-2.77 $\pm$ 0.01} & \textbf{-2.76 $\pm$ 0.01} & \textbf{-2.76 $\pm$ 0.01} & - \\ 
\midrule 
\textsc{Protein} - 2 & -2.82 $\pm$ 0.00 & \textit{\textbf{-2.81 $\pm$ 0.00}} & \textbf{-2.81 $\pm$ 0.00} & \textbf{-2.81 $\pm$ 0.00} & - \\ 
3 & -2.78 $\pm$ 0.00 & \textit{-2.77 $\pm$ 0.00} & \textbf{-2.76 $\pm$ 0.00} & \textbf{-2.76 $\pm$ 0.00} & -2.74 \\ 
4 & -2.75 $\pm$ 0.00 & \textit{-2.73 $\pm$ 0.00} & \textbf{-2.72 $\pm$ 0.00} & -2.73 $\pm$ 0.01 & - \\ 
5 & -2.73 $\pm$ 0.01 & \textit{-2.72 $\pm$ 0.01} & -2.71 $\pm$ 0.01 & \textbf{-2.70 $\pm$ 0.00} & - \\ 
\midrule 
\textsc{Wine} - 2 & -0.96 $\pm$ 0.01 & -0.96 $\pm$ 0.01 & -0.96 $\pm$ 0.01 & -0.96 $\pm$ 0.01 & - \\ 
3 & -0.96 $\pm$ 0.01 & -0.96 $\pm$ 0.01 & -0.96 $\pm$ 0.01 & -0.96 $\pm$ 0.01 & -1.00 \\ 
4 & -0.96 $\pm$ 0.01 & -0.96 $\pm$ 0.01 & -0.96 $\pm$ 0.01 & -0.96 $\pm$ 0.01 & - \\ 
5 & -0.96 $\pm$ 0.01 & -0.96 $\pm$ 0.01 & -0.96 $\pm$ 0.01 & -0.96 $\pm$ 0.01 & - \\ 
\midrule 
\textsc{Yacht} - 2 & -0.29 $\pm$ 0.12 & \textit{\textbf{-0.04 $\pm$ 0.10}} & \textbf{-0.04 $\pm$ 0.08} & -0.08 $\pm$ 0.10 & - \\ 
3 & -0.63 $\pm$ 0.04 & \textit{-0.13 $\pm$ 0.07} & 0.12 $\pm$ 0.07 & \textbf{0.14 $\pm$ 0.06} & -0.39 \\ 
4 & -0.77 $\pm$ 0.07 & \textit{-0.26 $\pm$ 0.07} & \textbf{-0.04 $\pm$ 0.09} & \textbf{-0.04 $\pm$ 0.09} & - \\ 
5 & -0.73 $\pm$ 0.07 & \textit{-0.58 $\pm$ 0.06} & -0.22 $\pm$ 0.09 & \textbf{-0.18 $\pm$ 0.07} & - \\ 
\bottomrule
  \end{tabular}
\end{table}

\begin{table}[ht]
\footnotesize
  \caption{Root mean square error. We report mean plus or minus one standard error over the splits. Bold numbers correspond to the best over all models, whereas the italicized models only compare DGP and DWP.}
  \label{tab:dwp:uci_rmses}
  \centering
  \begin{tabular}{rcccc}
  \toprule
\{dataset\} - \{depth\} & DGP & DWP & DWP-A & DWP-AB \\ 
\midrule 
\textsc{Boston} - 2 & 2.72 $\pm$ 0.14 & \textit{2.67 $\pm$ 0.14} & 2.60 $\pm$ 0.12 & \textbf{2.59 $\pm$ 0.13} \\ 
3 & 2.73 $\pm$ 0.14 & \textit{2.66 $\pm$ 0.13} & \textbf{2.62 $\pm$ 0.13} & 2.63 $\pm$ 0.13 \\ 
4 & 2.76 $\pm$ 0.14 & \textit{2.74 $\pm$ 0.15} & 2.71 $\pm$ 0.14 & \textbf{2.68 $\pm$ 0.14} \\ 
5 & \textit{2.81 $\pm$ 0.14} & 2.82 $\pm$ 0.17 & \textbf{2.77 $\pm$ 0.16} & 2.81 $\pm$ 0.17 \\ 
\midrule 
\textsc{Concrete} - 2 & \textit{5.41 $\pm$ 0.10} & 5.50 $\pm$ 0.12 & \textbf{5.29 $\pm$ 0.12} & 5.30 $\pm$ 0.12 \\ 
3 & \textit{5.31 $\pm$ 0.11} & 5.32 $\pm$ 0.10 & \textbf{5.22 $\pm$ 0.12} & 5.23 $\pm$ 0.12 \\ 
4 & 5.54 $\pm$ 0.10 & \textit{5.43 $\pm$ 0.11} & 5.24 $\pm$ 0.13 & \textbf{5.22 $\pm$ 0.13} \\ 
5 & \textit{5.49 $\pm$ 0.10} & 5.53 $\pm$ 0.10 & 5.26 $\pm$ 0.11 & \textbf{5.24 $\pm$ 0.11} \\ 
\midrule 
\textsc{Energy} - 2 & 0.48 $\pm$ 0.01 & 0.48 $\pm$ 0.01 & 0.48 $\pm$ 0.01 & 0.48 $\pm$ 0.01 \\ 
3 & 0.48 $\pm$ 0.01 & 0.48 $\pm$ 0.01 & 0.48 $\pm$ 0.01 & 0.48 $\pm$ 0.01 \\ 
4 & 0.48 $\pm$ 0.01 & 0.48 $\pm$ 0.01 & 0.48 $\pm$ 0.01 & 0.48 $\pm$ 0.01 \\ 
5 & 0.49 $\pm$ 0.01 & \textit{\textbf{0.48 $\pm$ 0.01}} & \textbf{0.48 $\pm$ 0.01} & \textbf{0.48 $\pm$ 0.01} \\ 
\midrule 
\textsc{Kin8nm} - 2 & 0.06 $\pm$ 0.01 & 0.06 $\pm$ 0.01 & 0.06 $\pm$ 0.00 & 0.06 $\pm$ 0.00 \\ 
3 & 0.06 $\pm$ 0.01 & 0.06 $\pm$ 0.01 & 0.06 $\pm$ 0.00 & 0.06 $\pm$ 0.00 \\ 
4 & 0.06 $\pm$ 0.01 & 0.06 $\pm$ 0.01 & 0.06 $\pm$ 0.00 & 0.06 $\pm$ 0.00 \\ 
5 & 0.06 $\pm$ 0.01 & 0.06 $\pm$ 0.01 & 0.06 $\pm$ 0.00 & 0.06 $\pm$ 0.00 \\ 
\midrule 
\textsc{Naval} - 2 & 0.00 $\pm$ 0.00 & 0.00 $\pm$ 0.00 & 0.00 $\pm$ 0.00 & 0.00 $\pm$ 0.00 \\ 
3 & 0.00 $\pm$ 0.00 & 0.00 $\pm$ 0.00 & 0.00 $\pm$ 0.00 & 0.00 $\pm$ 0.00 \\ 
4 & 0.00 $\pm$ 0.00 & 0.00 $\pm$ 0.00 & 0.00 $\pm$ 0.00 & 0.00 $\pm$ 0.00 \\ 
5 & 0.00 $\pm$ 0.00 & 0.00 $\pm$ 0.00 & 0.00 $\pm$ 0.00 & 0.00 $\pm$ 0.00 \\ 
\midrule 
\textsc{Power} - 2 & 3.87 $\pm$ 0.04 & \textit{3.83 $\pm$ 0.04} & 3.82 $\pm$ 0.04 & \textbf{3.81 $\pm$ 0.04} \\ 
3 & 3.87 $\pm$ 0.03 & \textit{3.82 $\pm$ 0.04} & \textbf{3.81 $\pm$ 0.04} & \textbf{3.81 $\pm$ 0.04} \\ 
4 & 3.89 $\pm$ 0.04 & \textit{3.84 $\pm$ 0.04} & \textbf{3.78 $\pm$ 0.04} & \textbf{3.78 $\pm$ 0.04} \\ 
5 & 3.88 $\pm$ 0.04 & \textit{3.84 $\pm$ 0.04} & \textbf{3.80 $\pm$ 0.04} & \textbf{3.80 $\pm$ 0.04} \\ 
\midrule 
\textsc{Protein} - 2 & 4.08 $\pm$ 0.01 & \textit{4.06 $\pm$ 0.01} & \textbf{4.05 $\pm$ 0.02} & \textbf{4.05 $\pm$ 0.01} \\ 
3 & 3.92 $\pm$ 0.02 & \textit{3.90 $\pm$ 0.01} & 3.88 $\pm$ 0.01 & \textbf{3.87 $\pm$ 0.01} \\ 
4 & 3.82 $\pm$ 0.01 & \textit{3.79 $\pm$ 0.01} & \textbf{3.75 $\pm$ 0.01} & 3.79 $\pm$ 0.02 \\ 
5 & 3.77 $\pm$ 0.02 & \textit{3.76 $\pm$ 0.02} & 3.73 $\pm$ 0.02 & \textbf{3.70 $\pm$ 0.01} \\ 
\midrule 
\textsc{Wine} - 2 & 0.63 $\pm$ 0.01 & 0.63 $\pm$ 0.01 & 0.63 $\pm$ 0.01 & 0.63 $\pm$ 0.01 \\ 
3 & 0.63 $\pm$ 0.01 & 0.63 $\pm$ 0.01 & 0.63 $\pm$ 0.01 & 0.63 $\pm$ 0.01 \\ 
4 & 0.63 $\pm$ 0.01 & 0.63 $\pm$ 0.01 & 0.63 $\pm$ 0.01 & 0.63 $\pm$ 0.01 \\ 
5 & 0.63 $\pm$ 0.01 & 0.63 $\pm$ 0.01 & 0.63 $\pm$ 0.01 & 0.63 $\pm$ 0.01 \\ 
\midrule 
\textsc{Yacht} - 2 & 0.41 $\pm$ 0.04 & \textit{\textbf{0.33 $\pm$ 0.03}} & \textbf{0.33 $\pm$ 0.03} & \textbf{0.33 $\pm$ 0.03} \\ 
3 & 0.53 $\pm$ 0.03 & \textit{0.35 $\pm$ 0.03} & 0.31 $\pm$ 0.03 & \textbf{0.30 $\pm$ 0.03} \\ 
4 & 0.58 $\pm$ 0.05 & \textit{0.41 $\pm$ 0.04} & \textbf{0.33 $\pm$ 0.03} & \textbf{0.33 $\pm$ 0.03} \\ 
5 & 0.57 $\pm$ 0.05 & \textit{0.50 $\pm$ 0.04} & \textbf{0.37 $\pm$ 0.03} & 0.38 $\pm$ 0.03 \\ 
\bottomrule
  \end{tabular}
\end{table}

\begin{table}[ht]
  \caption{Average runtime (seconds) for an epoch of \textsc{Boston} and \textsc{Protein}. Error bars are negligible.}
  \label{tab:dwp:runtime}
  \centering
  \begin{tabular}{rcccc}
    \toprule
\{dataset\} - \{depth\} & DGP & DWP & DWP-A & DWP-AB\\
\midrule 
\textsc{Boston} - 2 & 0.463 & 0.200 & 0.203 & 0.202 \\
5 & 1.292 & 0.358 & 0.373 & 0.370 \\
\midrule 
\textsc{Protein} - 2 & 0.903 & 0.843 & 0.854 & 0.869 \\
5 & 2.012 & 1.806 & 1.846 & 1.839 \\
\bottomrule
  \end{tabular}
\end{table}

\end{appendices}

\printthesisindex 

\end{document}